\documentclass[onefignum,onetabnum]{siamonline250211}

\usepackage{lipsum}
\usepackage{amsfonts}
\usepackage{graphicx}
\usepackage{epstopdf}
\usepackage{algorithmic}
\ifpdf
  \DeclareGraphicsExtensions{.eps,.pdf,.png,.jpg}
\else
  \DeclareGraphicsExtensions{.eps}
\fi
\usepackage{xr} 
\usepackage{enumitem}
\setlist[enumerate]{leftmargin=.5in}
\setlist[itemize]{leftmargin=.5in}

\newsiamremark{remark}{Remark}
\newsiamremark{hypothesis}{Hypothesis}
\crefname{hypothesis}{Hypothesis}{Hypotheses}
\newsiamthm{claim}{Claim}
\newsiamremark{fact}{Fact}
\crefname{fact}{Fact}{Facts}

\headers{Bayesian Dimension Reduction for GP Modeling}{E. H. Gyamfi, E. L. Kang, B. A. Konomi, and G. Lin}

\title{A Bayesian Framework for Built-in Input Dimension Reduction for Gaussian Process Modeling\thanks{Submitted to the editors 12/2025.
\funding{This work was funded by NSF.}}}

\author{Eric Herrison Gyamfi\thanks{Division of Statistics and Data Science, Department of Mathematical Sciences, University of Cincinnati, Cincinnati, OH (\email{gyamfien@mail.uc.edu}, \email{kangel@ucmail.uc.edu}, \email{konomibr@ucmail.uc.edu}).}
\and Emily L. Kang\footnotemark[2]
\and Bledar A. Konomi\footnotemark[2]
\and Guang Lin\footnotemark[3]\thanks{Department of Mathematics, Purdue University, West Lafayette, IN (\email{guanglin@purdue.edu}).}}

\usepackage{amsopn}

\ifpdf
\hypersetup{
  pdftitle={A Bayesian Framework for Built-in Input Dimension Reduction for Gaussian Process Modeling},
  pdfauthor={E. H. Gyamfi, E. L. Kang, B. A. Konomi, and G. Lin}
}
\fi

\usepackage{subcaption}
\usepackage{hyperref}

\def\ba{\mathbf{a}}

\def\bx{\mathbf{x}}

\def\bz{\mathbf{z}}
\def\bA{\mathbf{A}}

\def\bI{\mathbf{I}}

\def\bW{\mathbf{W}}

\begin{document}

\maketitle


\begin{abstract}
Gaussian process (GP) modeling is widely used in computational science and engineering. However, fitting a GP to high-dimensional inputs remains challenging due to the curse of dimensionality. While various methods have been proposed to reduce input dimensionality, they typically follow a two-stage approach, performing dimension reduction and GP fitting separately. We introduce a Bayesian framework that seamlessly integrates dimensionality reduction with GP modeling and inference. Our approach, built on a hierarchical Bayesian model with priors on the Stiefel manifold, enforces orthonormality on the projection matrix and enables posterior inference via Hamiltonian Monte Carlo with geodesic flow. Additionally, we extend this framework by incorporating Deep Gaussian Processes (DGP) with built-in dimension reduction, providing a more flexible and powerful tool for complex datasets. Through extensive numerical studies, we demonstrate that while the proposed Bayesian method incurs higher computational costs, it improves predictive performance and uncertainty quantification, providing a principled and robust alternative to existing methods.
\end{abstract}




\section{Introduction}\label{sec:intro}


Computer experiments have become indispensable in scientific discovery and engineering design, particularly when direct physical measurements are expensive, time-consuming, or infeasible. Modern simulators approximate complex physical, biological, and environmental processes, ranging from the spread of infectious diseases \cite{Fadikar2017CalibratingAS, Charles2023, Andrianakis2015BayesianHM, Plumlee2014BuildingAE, Hooten2010StatisticalAM, Andrianakis2017EfficientHM, Ahmed2024}, to earthquake early-warning and hazard mitigation \cite{LI2024, SAMADIAN2024, Fei_2024}, to large-scale environmental and climate dynamics \cite{Li2024SurrogateMF, Kennedy2001BayesianCO, McNeall2019, Donnelly2024, Marmin2022, Mehta2014}. Although computational platforms continue to advance rapidly, improvements in the predictive fidelity of physics-based models have not kept pace. Increasing model complexity introduces substantial computational cost, often forcing compromises in fidelity, study region, or other model approximations. In addition, these physics-based models or simulators typically involve many parameters that require calibration, making uncertainty quantification (UQ) \cite{Smith2024} both essential and challenging—particularly in high-dimensional settings.


Surrogate models, or statistical emulators, alleviate this computational burden by providing fast approximations to expensive simulators. \textcolor{black}{Among these surrogates, Gaussian process (GP) regression \cite{gramacy2020surrogates, Santner2003, Rasmussen2005} has become a cornerstone tool due to its flexibility, and natural quantification of uncertainty}. Its nonparametric formulation enables adaptation to diverse functional relationships, making it particularly appealing in computer experiments.

However, standard GP modeling deteriorates significantly as the input dimension increases. Predictive accuracy declines and computation becomes prohibitive---the classical curse of dimensionality \cite{Richard56}. At high dimensions, Euclidean distances become uninformative, and an exponentially growing number of samples is required to adequately explore the input space \cite{Bengio2005}. Consequently, classic GP regression is generally infeasible for modern high-dimensional computer experiments.

Much of the recent research on scalable surrogate modeling therefore focuses on identifying and exploiting low-dimensional structure embedded within high-dimensional inputs. Traditional approaches include input screening or ranking methods such as sensitivity indices \cite{Saltelli2008} and automatic relevance determination \cite{Neal2012}. These approaches can be effective but degrade when inputs are numerous, strongly correlated, or jointly influential.


\textcolor{black}{A more systematic class of approaches applies a dimensionality reduction prior to GP modeling}. Examples include truncated Karhunen--Lo\`eve expansions \cite{Ghanem2003}, principal component analysis (PCA) \cite{Tao2020ApplicationOA}, kernel PCA \cite{Kapsoulis2016TheUO, Zhou2020KernelPC}, partial least squares \cite{Bouhlel2016AnIA}, active subspaces \cite{as, Constantine2014, Constantine2015, Mufti2022}, and gradient kernel dimension reduction \cite{fukumizu2011, gkdr_juq}. While effective at reducing the input dimension, these methods operate in a two-stage manner: dimension reduction is performed first, followed by GP modeling on the reduced inputs. This separation ignores uncertainty in the projection step and may distort the relationship between inputs and outputs, leading to compromised prediction and degraded UQ.

More recently, gradient-free methods have embedded the dimension-reduction step directly into the GP model. Projection matrices are optimized on the Stiefel manifold using Cayley or Householder transforms \cite{TRIPATHY2016191, gautier2021fullybayesian}. These approaches are attractive for black-box simulators that lack gradient information. However, existing methods are either fully optimization-based \cite{TRIPATHY2016191} or only partially Bayesian \cite{gautier2021fullybayesian}, and therefore cannot propagate uncertainty in the dimension-reduction step. This limitation is problematic in UQ applications, where uncertainty quantification is often as important as point prediction.

Even with dimension reduction, classic GPs with stationary kernels may be too restrictive to capture complex, heterogeneous, or nonstationary relationships. Deep Gaussian processes (DGPs) \cite{Damianou2013, Sauer02012023} address this limitation by composing multiple GP layers to construct flexible hierarchical mappings. Prior work has shown that DGPs can capture nonstationarity and multi-scale structure \cite{Havasi2018, Bui2016, Dunlop2017, Majdi2020, Salimbeni2017}. However, inference in DGPs remains challenging. Variational approaches scale well but may sacrifice posterior accuracy \cite{Damianou2013, Salimbeni2017}, whereas sampling-based methods such as stochastic gradient  Hamiltonian Monte Carlo (HMC) \cite{Havasi2018} are more accurate but often mismatched to the small-sample regimes typical of computer experiments. Recent work has explored Elliptical Slice Sampling (ESS) \cite{Murray2010, Sauer02012023}, which is better suited to these settings.

Parallel developments have incorporated HMC on the Stiefel manifold for sampling projection matrices, for example in polynomial chaos expansions \cite{chaos2018} and multifidelity surrogate modeling \cite{TSILIFIS2021114147}, using matrix Langevin priors \cite{Hoff01012009, CHIKUSE2003375}. These works emphasize the importance of formally accounting for uncertainty in the projection matrix when modeling high-dimensional inputs.


This paper advances surrogate modeling and dimension reduction for Gaussian processes through three main contributions:

\begin{enumerate}
    \item \textbf{A unified Bayesian framework for dimension reduction and GP modeling.}  
    We introduce a hierarchical model that integrates dimension reduction and GP regression within a single probabilistic formulation. Unlike two-stage methods, our framework fully propagates posterior uncertainty from the projection step to the surrogate predictions.

    \item \textbf{Matrix Langevin priors and HMC for orthonormal projections.}  
    The projection matrix is modeled using a matrix Langevin prior supported on the Stiefel manifold, enforcing orthonormality by construction. Posterior sampling is performed via Hamiltonian Monte Carlo with geodesic flows, providing a principled and efficient alternative to optimization-based manifold methods.

    \item \textbf{Extension to deep Gaussian processes with built-in dimension reduction.}  
    The proposed framework naturally extends beyond shallow GPs to two- and three-layer DGPs. This enables flexible, nonlinear, and nonstationary mappings while maintaining coherent uncertainty quantification across all latent layers.

\end{enumerate}

\textcolor{black}{The framework directly addresses the curse of dimensionality while preserving rigorous Bayesian inference. It is particularly suited for small-data, high-dimensional regimes typical in computer experiments, offering improved predictive performance and more reliable uncertainty quantification}. Together, these contributions provide the first framework to learn a low-dimensional input projection jointly with the Gaussian process surrogate through Bayesian inference, and it is naturally extensible to deep architectures. The resulting methodology is flexible, uncertainty-aware, and well aligned with the demands of modern UQ. 

\textcolor{black}{Our work is most closely related to \cite{Sauer02012023}, who develop Bayesian deep GP surrogates for nonstationary computer experiments and use active learning for sequential design. We build upon this line of work while pursuing a distinct primary objective: rather than using DGPs only to warp the input space, we additionally learn an orthonormal projection of the high-dimensional inputs \emph{jointly} with the GP or DGP surrogate, place a Matrix Langevin prior on this projection, and sample it via HMC with geodesic flow on the Stiefel manifold, propagating the projection uncertainty through to prediction.}


The remainder of the paper is organized as follows. \Cref{sec:method} introduces the model formulation, prior specification, and inference algorithms. \Cref{sec:Numerical_studies} presents numerical studies, including three synthetic benchmarks and two real-world applications. \Cref{sec:Con_Discus} concludes with a discussion of the main findings and future research directions.

\section{Methodology}
\label{sec:method}

This section presents a Bayesian framework for Gaussian process (GP) surrogate modeling with \emph{built-in dimension reduction} (BDR). The approach integrates an orthonormal projection of high-dimensional inputs with shallow and deep Gaussian processes, enabling flexible modeling of nonlinear and nonstationary computer model outputs while quantifying uncertainty in both the projection and the surrogate. \Cref{subsec:dimred} introduces the dimension reduction formulation on the Stiefel manifold. \Cref{subsec:gp_bdr} develops the standard GP with BDR. \Cref{subsec:dgp_bdr} extends the construction to deep Gaussian processes (DGPs). Prior specification and Bayesian inference are detailed in \Cref{subsec:priors}--\Cref{subsec:posterior}. Computational complexity is summarized in \Cref{subsec:complexity}.

\subsection{Dimension Reduction Framework}
\label{subsec:dimred}

Let $\mathbf{x} \in \mathbb{R}^{p}$ denote a high-dimensional input and $y = f(\mathbf{x}) \in \mathbb{R}$ a scalar simulator output. We assume that the variation in $f(\mathbf{x})$ concentrates on a $D$-dimensional linear subspace $\mathbb{R}^{D}$ with $D \ll p$. Let 
\[
\mathbf{z} = W^\top \mathbf{x} \in \mathbb{R}^{D},
\quad W \in \mathcal{V}_{p,D},
\]
where $\mathcal{V}_{p,D} = \{W \in \mathbb{R}^{p\times D} : W^\top W = I_D\}$ is the Stiefel manifold of orthonormal $p\times D$ projection matrices. Further background on notation used in the Supplement is summarized in \Cref{sec:SM_notation}. The low-dimensional representation $\mathbf{z}$ induces a surrogate functional relationship
\[
f(\mathbf{x}) \approx \eta(\mathbf{z}),
\qquad \eta : \mathbb{R}^{D} \to \mathbb{R}.
\]
This formulation is consistent with sufficient dimension reduction (SDR), which assumes 
\[
y \;\perp\; \mathbf{x} \mid W^\top \mathbf{x} ,
\]
so that the subspace spanned by the columns of $W$ is sufficient for predicting $y$.

For an input matrix $X_n \in \mathbb{R}^{n \times p}$ with rows $\mathbf{x}_1,\dots,\mathbf{x}_n$, the reduced inputs are
\[
Z_n = X_n W \in \mathbb{R}^{n\times D}.
\]

\begin{figure}[t]
    \centering
    \includegraphics[scale=0.75]{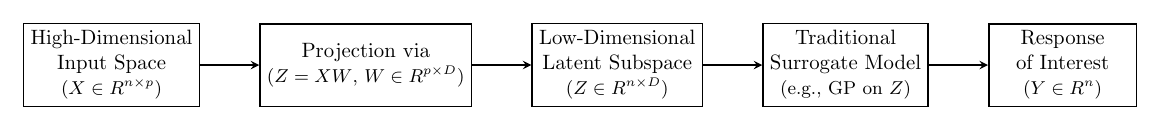}
    \caption{Schematic illustration of built-in dimension reduction: high-dimensional inputs $X_n$ are projected onto a $D$-dimensional subspace via an orthonormal matrix $W$, and the surrogate is defined on the reduced coordinates $Z_n = X_n W$.}
    \label{fig:schmaticDR}
\end{figure}

\textcolor{black}{The goal is to infer $W$, and the kernel hyperparameters jointly conditioned on fixed $D$, so that uncertainty in the projection and surrogate are coherently propagated. In this work, $D$ is treated as a model choice rather than a random variable. For each candidate $D$, we perform fully Bayesian inference over the $W$ and all surrogate parameters. Model comparison across different $D$ values is conducted using predictive criteria. The detailed choices of $D$ are described in Section \ref{sec:Numerical_studies}. A fully Bayesian treatment of $D$ could be obtained by placing a discrete prior on D and performing reversible-jump MCMC or marginal likelihood-based model averaging. However, such trans-dimensional sampling introduces substantial computational overhead and is beyond the scope of this work.}



\subsection{Gaussian Process Regression with Built-in Dimension Reduction}
\label{subsec:gp_bdr}

Given reduced inputs $Z_n = X_n W$, the GP prior for the latent function $g$ is
\[
\eta(\cdot) \sim \mathrm{GP}\!\left(0,\; C(\cdot,\cdot\,; \theta_y)\right),
\]
where $C(\cdot,\cdot\,;\theta_y)$ is a covariance kernel parameterized by the
lengthscale $\theta_y$ controlling smoothness. In this work we use the squared
exponential kernel
\[
C(\mathbf{z}_i,\mathbf{z}_j;\theta_y)
=
\exp\!\left(
-\frac{1}{2\theta_y}\|\mathbf{z}_i-\mathbf{z}_j\|^2
\right).
\]

Evaluating this kernel at the projected inputs $Z_n=(\mathbf{z}_1,\dots,\mathbf{z}_n)$
yields the $n\times n$ correlation matrix $C_y$ with entries
\[
[C_y]_{ij}=C(\mathbf{z}_i,\mathbf{z}_j;\theta_y).
\]

The corresponding likelihood for $Y_n=(y_1,\dots,y_n)^\top$ is
\begin{equation}
Y_n \mid X_n, W, \theta_y, \tau^2
\sim \mathcal{N}_n\!\left(0,\;
\Sigma_y(Z_n)\right),
\qquad
\Sigma_y(Z_n)=\tau^2\big(C_y + g I_n\big),
\qquad 
C_y=C(Z_n,Z_n'\,; W, \theta_y),
\label{eq:gp_cov_top}
\end{equation}
$\tau^2$ is a scale parameter, and $g > 0$ is a nugget that stabilizes the covariance and accounts for small numerical noise.

The log-likelihood is 
\begin{equation}
\log \mathcal{L}(Y_n \mid X_n, \theta_y, \tau^2, g, W)
=
-\frac{1}{2}\left[
\log |\Sigma_y|
+ Y_n^\top \Sigma_y^{-1} Y_n
+ n\log(2\pi)
\right].
\label{eq:gp_likelihood}
\end{equation}
\subsubsection*{Projection-dependent kernel}

The kernel depends on $W$ through the projected distance
\[
\|\mathbf{z}_i - \mathbf{z}_j\|^2
= \|W^\top(\mathbf{x}_i - \mathbf{x}_j)\|^2
= (\mathbf{x}_i - \mathbf{x}_j)^\top W W^\top (\mathbf{x}_i - \mathbf{x}_j).
\]
Differentiating the kernel with respect to $W$ gives the gradient
\begin{equation}
\frac{\partial C_y}{\partial W}
=
-\frac{1}{\theta_y}
\left(
C_y \odot D
\right)X^\top (XW),
\tag{2.3}
\label{eq:kernel_grad_W}
\end{equation}
where  
(i) $\odot$ denotes the Hadamard product,  and 
(ii) $D_{ij} = \frac{1}{2}\|\mathbf{x}_i - \mathbf{x}_j\|^2$. 

\subsubsection*{Identifiability of $W$}
\textcolor{black}{
For isotropic kernel defined on the reduced space, the likelihood depends on $W$ only through the projected inputs, $Z_n$. As a result, $W$ is identifiable only up to right multiplication by an orthogonal matrix $G \in \mathcal{V}_{D,D}$. In particular, $W$ and $WG$ span the same $D$-dimensional subspace and give the same likelihood. Hence, statistical identifiability pertains only to the $D$-dimensional column space of $W$, not to a unique orthonormal representation of that space. Imposing orthonormality constraints on \(W\) makes computation more stable and removes some unnecessary redundancy in the parameterization but it does not remove all non-identifiability, because even with orthonormal columns, \(W\) and \(WG\) for any orthogonal \(G\) still represent the same subspace. Accordingly, the paper focused on interpretation and inference on the rotation of the projection matrix, $WW^T$, which is invariant under rotations within the reduced space. 
}

\subsubsection*{Predictive distribution}

For test inputs $\mathcal{X}$, let $\mathcal{Z}= \mathcal{X} W$ and define   
\[
C_* = C(\mathcal{Z}, Z_n ;\theta_y), 
\qquad
C_{**} = C(\mathcal{Z}, \mathcal{Z};\theta_y).
\]
Under the GP with BDR model,
\begin{equation}
\mathcal{Y} \mid Y_n, X_n, W, \theta_y, \tau^2, g
\sim
\mathcal{N}\!\left(
\mu_*,\;
\Sigma_*
\right),
\label{eq:gp_pred}
\end{equation}
where
\[
\mu_* = C_* \Sigma_y^{-1} Y_n,
\qquad
\Sigma_* = C_{**} - C_* \Sigma_y^{-1} C_*^\top.
\]

\begin{figure}[t]
    \centering
    \includegraphics[width=0.48\linewidth]{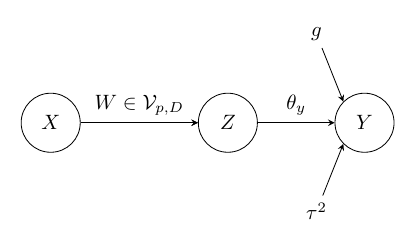}
    \caption{Model structure of a standard Gaussian process with built-in dimension reduction. The projection matrix $W\in \mathcal{V}_{p,D}$ maps $X_n$ to $Z_n$, and a GP prior is placed on the latent function $g(\cdot)$.}
    \label{fig:1lDGP}
\end{figure}

The standard GP with BDR provides a flexible surrogate while learning $W$, but its stationarity may limit performance when the input–output relationship varies across the domain. \Cref{subsec:dgp_bdr} introduces a DGP extension that captures nonstationary features more effectively.

\subsection{Deep Gaussian Processes with Built-In Dimension Reduction}
\label{subsec:dgp_bdr}

Although the standard GP with BDR captures global smooth trends, many computer models exhibit input–output behavior that changes across the domain, producing heterogeneity, varying smoothness, or localized nonlinearities. Deep Gaussian processes (DGPs) address these limitations by composing multiple GP layers, each transforming its input before passing it to the next. In our setting, the first transformation is always the linear projection $Z_n = X_n W$, after which the hierarchy proceeds nonlinearly.

Let $D$ denote the reduced dimension of $Z_n$. A two-layer DGP introduces an intermediate latent variable 
\[
Q = (Q_1,\dots,Q_D)\in\mathbb{R}^{n\times D},
\]
where each column $Q_j$ is a GP mapping of $Z_n$.

\subsubsection{Two-layer DGP with BDR}
The model is
\begin{equation}
\begin{aligned}
    Y_n \mid Q &\sim \mathcal{N}\!\left(0,\,
       \tau^2\big(C(Q,Q';\theta_y)+ g I_n\big)\right), \\
    Q_j &\sim \mathcal{N}\!\left(0,\, C(Z_n,Z_n';\theta_q[j],W)\right),  && j=1,\dots,D,
\end{aligned}
\label{eq:two_layer_def}
\end{equation}
where  
(i) each $Q_j$ is conditionally independent given $Z_n$,  
(ii) each latent node has its own lengthscale $\theta_q[j]$, and  
(iii) the top-level GP uses the latent matrix $Q$ as its input.

The induced marginal likelihood integrates over both $Q$ and $W$:
\[
\mathcal{L}(Y_n \mid X_n)\propto 
\int \mathcal{L}(Y_n\mid Q,\theta_y,g,\tau^2)\,
     \mathcal{L}(Q\mid Z_n,\theta_q)\,
     \pi(W)\, dQ\, dW.
\]

\begin{figure}[t]
\centering
\includegraphics[width=0.80\linewidth]{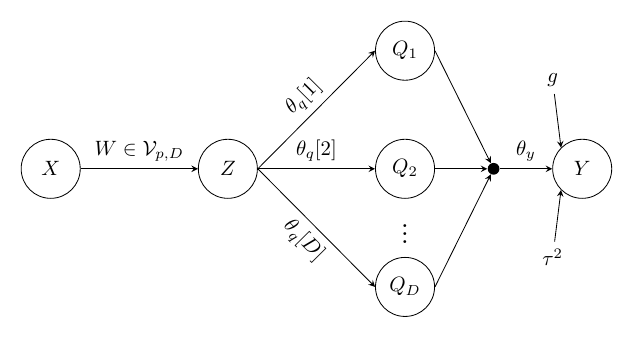}
\caption{Two-layer DGP with built-in dimension reduction. The projection $W$ first maps $X_n$ to $Z_n$, which drives $D$ latent GPs $\{Q_j\}_{j=1}^D$. These propagate through a second GP layer that models $Y_n$.}
\label{fig:2lDGP}
\end{figure}

This structure provides a nonstationary mapping $X_n \to Z_n \to Q \to Y_n$; the intermediate GPs distort the geometry of $Z_n$ and allow the covariance in the output layer to vary across regions of the input space.

\subsubsection{Three-layer DGP with BDR}

A deeper representation introduces an additional nonlinear transformation. Let
\[
R = (R_1,\dots,R_D) \in \mathbb{R}^{n\times D}
\]
be a first latent layer mapping $Z_n$, with the second layer mapping $R$ to $Q$. The model is

\begin{equation}
\begin{aligned}
Y_n \mid Q &\sim \mathcal{N}\!\left(0,\, \tau^2(C(Q,Q';\theta_y)+ g I_n)\right)\\
Q_j \mid R &\sim \mathcal{N}\!\left(0,\, C(R,R';\theta_q[j])\right), && j=1,\dots,D,\\
R_k \mid Z_n &\sim \mathcal{N}\!\left(0,\, C(Z_n,Z_n';\theta_r[k],W)\right), && k=1,\dots,D.
\end{aligned}
\label{eq:three_layer_def}
\end{equation}

The full likelihood is
\[
\mathcal{L}(Y_n\mid X_n)\propto 
\int \mathcal{L}(Y_n\mid Q)
     \mathcal{L}(Q\mid R)
     \mathcal{L}(R\mid Z_n)
     \pi(W)\, dQ\, dR\, dW,
\]
which is analytically intractable and motivates the MCMC scheme in \Cref{subsec:posterior}.

\begin{figure}[t]
\centering
\includegraphics[width=0.80\linewidth]{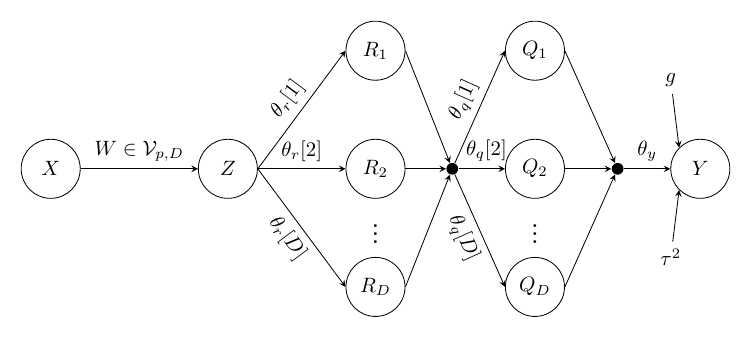}
\caption{Three-layer DGP with built-in dimension reduction. The projection $W$ produces $Z_n$, which enters a first nonlinear layer producing $R$, then a second layer producing $Q$, which drives the observation layer.}
\label{fig:3lDGP}
\end{figure}

\paragraph{Identifiability and regularization}
Following \cite{Sauer02012023}, we impose  
(i) independent isotropic kernels at each latent node;  
(ii) no nugget in the latent layers;  
(iii) a single nugget and variance parameter only at the top layer;  
(iv) equal latent dimensionality across all layers.  
These restrictions improve identifiability and stabilize posterior sampling, though they do not guarantee strict statistical identifiability; rotational invariances within the reduced subspace remain.

\subsection{Prior Specification}
\label{subsec:priors}

The hierarchical Bayesian model assigns priors to the projection matrix $W$, the kernel hyperparameters across all layers, and the noise and scale parameters. The priors reflect two key structural elements: orthogonality of $W$ and nested smoothness assumptions across layers.

\subsubsection{Matrix Langevin prior for $W$}

Because $W$ lives on the Stiefel manifold $\mathcal{V}_{p,D}$, standard Euclidean priors are inappropriate. We use a Matrix Langevin (\(\mathcal{ML}\)) prior:
\begin{equation}
W \sim \mathcal{ML}(F), 
\qquad
\pi_{\mathcal{ML}}(W;F)
= c(F)^{-1} \exp\!\left(\mathrm{tr}(F^\top W)\right),
\label{eq:ml_prior}
\end{equation}
with normalizing constant  
\[
c(F) = {}_0\mathcal{F}_1\!\left(\tfrac{p}{2},\tfrac{1}{4}F^\top F\right),
\]
the hypergeometric function of matrix argument. The parameter matrix $F$ determines the concentration and preferred orientation of $W$. We parameterize $F$ using its singular value decomposition
\[
F = M \Lambda V^\top,
\]
where  
(i) $M\in\tilde{\mathcal{V}}_{p,D}$ and $V\in\mathcal{V}_{D,D}$ are orthonormal matrices,  
(ii) $\Lambda=\mathrm{diag}(\lambda_1,\dots,\lambda_D)$ with  
\[
\infty > \lambda_1 > \cdots > \lambda_D > 0.
\]
This representation separates orientation $(M,V)$ and concentration $(\Lambda)$ and facilitates posterior updates through manifold-aware sampling.

Priors for the SVD components follow
\[
M \sim \mathcal{ML}(F_M), 
\qquad
V \sim \mathcal{ML}(F_V), 
\qquad
\lambda_k \stackrel{i.i.d}{\sim} \Gamma(b_1,b_2),
\]
with diffuse $F_M$ and $F_V$ producing weak directional preferences.

\subsubsection{Priors for kernel hyperparameters and noise}

Let  
\[
\Theta = \{\theta_y,\theta_q,\theta_r,g,\tau^2\}.
\]
We use the standard weakly-informative Gamma priors for lengthscales and nugget:
\[
\theta_y,\;\theta_q[j],\;\theta_r[j],\; g
\stackrel{i.i.d}{\sim}
\Gamma\!\left(\tfrac{3}{2},\, b_{(\cdot)}\right),
\]
with the ordering
\[
b_{\theta_y} > b_{\theta_q} > b_{\theta_r},
\]
which encodes the assumption that deeper layers vary more rapidly than upper layers.

The variance scale parameter has an inverse-gamma prior:
\[
\tau^2 \sim \mathrm{IG}(\alpha_1,\alpha_2).
\]

Together, these priors induce a structured hierarchy in which the projection $W$ is regularized on the manifold, deeper layers are less smooth, and the predictive layer remains globally smooth to prevent overfitting.

\subsection{Posterior Inference}\label{subsec:posterior} 

Posterior inference is challenging due to (i) the presence of the projection matrix
$W\in\mathcal{V}_{p,D}$, (ii) latent variables $(Q,R)$ in deep architectures, and (iii) non-conjugate kernel hyperparameters. We perform Bayesian inference
via a hybrid MCMC scheme that combines Metropolis--Hastings (MH), elliptical slice sampling (ESS), and Hamiltonian Monte Carlo (HMC) on the Stiefel manifold.

Let
\[
\Xi = \{W,M,V,\Lambda,\theta_y,\theta_q,\theta_r,g,\tau^2,Q,R\}
\]
denote the full parameter set (with $R$ present only for three layers).  
Bayes' theorem yields the posterior
\begin{equation}
\pi(\Xi \mid Y_n,X_n)\; \propto\;
\mathcal{L}(Y_n\mid Q,\theta_y,g,\tau^2)\,
\mathcal{L}(Q\mid R, Z_n,\theta_q)\,
\mathcal{L}(R\mid Z_n,\theta_r)\,
\pi(W,M,V,\Lambda)\,\pi(\theta_y,\theta_q,\theta_r,g,\tau^2).
\label{eq:full_joint_post}
\end{equation}
\textcolor{black}{Each likelihood term depends directly only on its immediate parent layer. The projected inputs $Z_n$ enter the model through the first layer, $R$, and influence deeper layers only via propagation through the hierarchy.
The posterior distribution in \eqref{eq:full_joint_post} is proportional to the product of the likelihood terms and prior distributions. The likelihood is decomposed according to the hierarchical structure of the model: the observed response $Y_n$ depends on the latent layer $Q$ through the output GP with parameters $(\theta_y,g,\tau^2)$; the latent layer $Q$ depends on the intermediate layer $R$ and the projected inputs $Z_n$; and the intermediate layer $R$ depends on the projected inputs through its GP prior. The remaining terms correspond to priors for the projection parameters $(W,M,V,\Lambda)$ and the GP hyperparameters $(\theta_y,\theta_q,\theta_r,g,\tau^2)$. Together, these components define the full joint posterior over all unknown quantities $\Xi$.}
Because direct sampling is intractable, we update each block of parameters in sequence, conditioning on all others. The details of each step are given below.

\subsubsection{Updates for variance and nugget parameters}

The conditional posterior of $\tau^2$ remains inverse-gamma under Gaussian
likelihood:
\begin{equation}
\tau^{2} \mid \cdot
\sim
\mathrm{IG}\!\left(
\alpha_1 + \tfrac{n}{2},\;
\alpha_2 + \tfrac{1}{2} Y_n^\top \Sigma_y^{-1} Y_n
\right), 
\qquad 
\Sigma_y = \tau^2\!\big(C(Q,Q';\theta_y) + g I_n\big).
\end{equation}
The update is inexpensive provided the Cholesky factor of $\Sigma_y$ has already
been computed for other updates.

The nugget $g$ and lengthscale $\theta_y$ (and latent lengthscales $\theta_q[j]$,
$\theta_r[j]$) are updated using MH random-walk proposals on the logarithmic
scale. For a generic hyperparameter $\phi$,
\[
\phi^\ast \sim \mathrm{Unif}\!\left(\tfrac{l}{u}\phi^{(t-1)},\, \tfrac{u}{l}\phi^{(t-1)}\right),
\]
which yields the acceptance probability
\[
\alpha_\phi = 
\min\!\left( 1,\;
\frac{\mathcal{L}(Y_n\mid \phi^\ast)\,\pi(\phi^\ast)}%
     {\mathcal{L}(Y_n\mid \phi^{(t-1)})\,\pi(\phi^{(t-1)})}\,
\frac{\phi^{(t-1)}}{\phi^\ast}
\right).
\]
This approach is standard for GP hyperparameters and performs robustly in
moderate dimensions.

\subsubsection{Updates for the hyperparameters: $\mathbf{M}$ and $\mathbf{V}$}  \textcolor{black}{The conditional posterior of \(M\) and \(V\)
remains Matrix Langevin distribution
\begin{equation*}
  M|W,\Lambda,V \sim \mathcal{ML}(W\Lambda V + F_M); \quad
 V | W, \Lambda , M \sim \mathcal{ML}(W^\top  M \Lambda + F_V)
\end{equation*} 
where $F_M$ and $F_V$ are treated as constants. In our implementation, the posterior updates for $M$ and $V$ are obtained using the Gibbs sampling scheme of \cite{Hoff01012009}, developed for simulation from matrix von Mises-Fisher and matrix Bingham-von Mises-Fisher distributions. Since the matrix Langevin distribution is the linear-term case of this family, that Gibbs sampler applies directly here. For general background on matrix Langevin distributions and Stiefel manifolds, see \cite{CHIKUSE2003375}.}

\subsubsection{ESS updates for latent layers}
\textcolor{black}{ESS provides a gradient-free, tuning-free sampler for the latent variables with Gaussian priors. In our setting, this applies to latent variables \(Q_j\) and \(R_j\), which are assigned Gaussian process priors}. For a latent variable such as $Q_j$ in a two-layer DGP, \textcolor{black}{let $Q_j^{\mathrm{cur}}$ denotes the previously accepted sample}, ESS draws a prior sample
\[
Q_j^{\mathrm{prior}} \sim \mathcal{N}\!\left(0,\, C(Z_n,Z_n';\theta_q[j],W)\right),
\]
and proposes an elliptical rotation
\[
Q_j^{\mathrm{prop}} = Q_j^{\mathrm{cur}}\cos\phi + Q_j^{\mathrm{prior}}\sin\phi,
\qquad \phi\sim \mathrm{Unif}(0,2\pi),
\]
(which preserves the Gaussian prior measure) and accepts if
\[
\log\mathcal{L}(Y_n\mid Q^{\mathrm{prop}}) 
\ge 
\log u + \log\mathcal{L}(Y_n\mid Q^{\mathrm{cur}}),
\qquad u\sim\mathrm{Unif}(0,1).
\] 
\textcolor{black}{$\phi$ denotes the rotation angle defining the elliptical proposal, and that the angular bracket refers to the interval of allowable angles that iteratively shrunk when a proposal is rejected.} If rejected, the angular bracket is shrunk and a new angle is drawn; see the  Supplement for details. For a three-layer DGP, an analogous ESS update is applied to $R_j$. 


\subsubsection{Updates for $\lambda$}\textcolor{black}{
We update \(\lambda\) using a slice-type sampling (SS) scheme motivated by ESS. Specifically, the algorithm employs the elliptical proposal construction and angular bracket-shrinking strategy of ESS, but is adapted to the present setting in which \(\lambda\) is assigned a Gamma prior and is restricted to the positive constrained parameter space. Therefore, this update is not a standard Gaussian-prior ESS step.
}

\subsubsection{Manifold HMC updates for the projection matrix $W$}

Updating $W$ is the most computationally involved step. Since $W$ must remain on 
$\mathcal{V}_{p,D}$, we use HMC with geodesic flows on the Stiefel manifold. Let
the negative log-posterior be the potential energy,
\[
U(W) = - \log\mathcal{L}(Y_n\mid Q,\theta_y,g,\tau^2)
       - \log\mathcal{L}(Q\mid R,\theta_q)
       - \log\mathcal{L}(R\mid Z_n,\theta_r)
       - \log\pi_{\mathrm{ML}}(W;F),
\]
{\color{black} and introduce momentum $P$ in the tangent space
\[
T_W\mathcal{V}_{p,D} = 
\big\{ P\in\mathbb{R}^{p\times D}: W^\top P + P^\top W = 0 \big\}.
\]
For completeness, the projection of any Euclidean matrix $A\in\mathbb{R}^{p\times D}$ onto the tangent space $T_W\mathcal{V}_{p,D}$ is
\[
\Pi_W(A)
=
A-W\,\mathrm{sym}(W^\top A),
\qquad
\mathrm{sym}(B)=\frac12(B+B^\top).
\]
The corresponding Hamiltonian is
\[
H(W,P)
=
U(W)+\frac12\langle P,P\rangle,
\]
where $U(W)$ is the potential energy and $P\in T_W\mathcal{V}_{p,D}$ is the tangent-space momentum. In practice, an unconstrained Gaussian momentum is sampled and then projected onto $T_W\mathcal{V}_{p,D}$ before applying the geodesic leapfrog update. This ensures that the proposal respects the Stiefel constraint $W^\top W=I_D$.
The geodesic part of the update evolves according to
\[
\dot W=W\,\mathrm{sym}(P^\top P),
\]
between projected gradient momentum updates. Under the canonical metric, this geodesic flow admits a closed-form matrix-exponential representation, so the proposal moves exactly along the Stiefel manifold during each geodesic step. Therefore, no ad hoc re-orthogonalization of $W$ is required. The required Euclidean gradient is converted into a valid manifold direction through the tangent projection. In particular, since
\[
\nabla_W U(W)
=
-\frac{\partial}{\partial W}\log\pi(W\mid\cdot),
\]
the gradient information enters the leapfrog scheme through projected momentum updates of the form
\[
P
\leftarrow
P+\frac{\epsilon}{2}\Pi_W\!\left(\nabla_W\log\pi(W\mid\cdot)\right),
\]
where $\Pi_W$ is the tangent-space projection. Thus, a geodesic integrator updates $(W,P)$ by alternating (i) movement along the manifold geodesic and (ii) momentum adjustments using the projected gradient. \textcolor{black}{Although $U(W)=-\log\pi(W\mid \cdot)$ is written using the full negative log-posterior for completeness, only the terms involving $Z_n$ and the prior $\pi_{\mathrm{ML}}(W;F)$ depend explicitly on $W$; the remaining likelihood terms are constant with respect to $W$ and therefore do not contribute to the gradient in the HMC update.} The gradient contribution from the Matrix Langevin prior is simply $V\Lambda M^\top$. The gradient from the GP terms
requires derivatives of the covariance matrix with respect to $W$, which for the
squared-exponential kernel has the compact form
\begin{equation}
\frac{\partial C_y}{\partial W}
= -\frac{C_y}{\theta}\,(X_n-X_n') (X_n-X_n')^\top W,
\label{eq:dcdw_compact}
\end{equation}
with complete derivations provided in the Supplement. 
The proposed $(W',P')$ is
accepted or rejected with the usual HMC acceptance probability. Additionally, the summary of the Geodesic flow is detailed in \cref{alg:hmc_algorithm}.}

\begin{algorithm}[H]
\caption{Geodesic Monte Carlo for $W$ on the Stiefel Manifold $\mathcal{V}_{p,D}$}
\label{alg:hmc_algorithm}
\begin{algorithmic}[1]  
\STATE{\textbf{Input:} integration period $T$, step size $\varepsilon$, initial $W^{(0)}\in\mathcal{V}_{p,D}$, $\Theta=\{\tau^2,g,\theta_y\}, F=ML V^T$}
\STATE{\textbf{Output:} a sample $W \sim \pi(W\mid X_n,Y_n,\Theta, F)$}

\STATE{\textbf{Initialization:}}
\STATE{Sample $P^{(0)} \sim \mathcal{N}(0,I_{p\times D})$ and project onto the tangent space}
\begin{equation*}P^{(0)} \gets \Pi_{W^{(0)}}(P^{(0)}) =  P^{(0)} - \tfrac{1}{2}W^{(0)}\!\big(W^{(0)\top}P^{(0)} + P^{(0)\top}W^{(0)}\big)
\end{equation*}
\STATE{Compute  initial Hamiltonian: }
\begin{equation*}H(W^{(0)},P^{(0)}) = -\log \pi(W^{(0)}\mid X_n,Y_n,\Theta, F) + \tfrac{1}{2}P^{(0)\top}P^{(0)}
\end{equation*}

\FOR{$m=1,\ldots,T$}
  \STATE{\textbf{Half-step Momentum Update:}}
  \begin{equation*}
  \begin{aligned}
      P^{(m-1)} & \gets P^{(m-1)} + \tfrac{\varepsilon}{2}\nabla_{W^{(m-1)}}\log \pi\big(W^{(m-1)}\mid X_n,Y_n,\Theta, F\big)\\
     P^{(m-1)} & \gets \Pi_{W^{(m-1)}}(P^{(m-1)}) =  P^{(m-1)} - \tfrac{1}{2}W^{(m-1)}\!\big(W^{(m-1)\top}P^{(m-1)} + P^{(m-1)\top}W^{(m-1)}\big)
   \end{aligned}
  \end{equation*}

  \STATE{\textbf{Position and Momentum Update (Geodesic Flow):}}
  \IF{$D=1$}
    \STATE \[
    \begin{aligned}
        [W^{(m)},P^{(m)}] = [W^{(m-1)},P^{(m-1)}]
        \begin{bmatrix} 1 & 0 \\ 0 & \nu^{-1} \end{bmatrix}
        \begin{bmatrix} \cos(\nu\varepsilon) & -\sin(\nu\varepsilon) \\ \sin(\nu\varepsilon) & \cos(\nu\varepsilon) \end{bmatrix}
        \begin{bmatrix} 1 & 0 \\ 0 & \nu \end{bmatrix}, \nu = \|P^{(m-1)}\|
    \end{aligned}
    \]  
    \ELSE
        \STATE Set
    \[
      A^{(m-1)} = W^{(m-1)\top}P^{(m-1)}, \qquad
      S^{(m-1)} = P^{(m-1)\top}P^{(m-1)}.
    \]
    \STATE \[
    \big[W^{(m)},P^{(m)}\big]
    \gets
    \big[W^{(m-1)},P^{(m-1)}\big]\,
      \exp\!\left(
        \varepsilon
        \begin{bmatrix}
           A^{(m-1)} & -S^{(m-1)} \\
           I & A^{(m-1)}
        \end{bmatrix}
      \right)
      \begin{bmatrix}
        e^{-\varepsilon A^{(m-1)}} & 0 \\[4pt]
        0 & e^{-\varepsilon A^{(m-1)}}
      \end{bmatrix}
    \]
  \ENDIF 
  
  \STATE{\textbf{Half-step Momentum Update:}}
  \begin{equation*}
  \begin{aligned}
  P^{(m)} & \gets P^{(m)} + \tfrac{\varepsilon}{2}\nabla_{W^{(m)}}\log \pi\big(W^{(m)}\mid X_n,Y_n,\Theta, F\big)\\
  P^{(m)} & \gets \Pi_{W^{(m)}}(P^{(m)}) =  P^{(m)} - \tfrac{1}{2}W^{(m)}\!\big(W^{(m)\top}P^{(m)} + P^{(m)\top}W^{(m)}\big)
  \end{aligned}
  \end{equation*}
\ENDFOR

\STATE{\textbf{Accept/Reject Step:}}
\STATE{Compute Hamiltonian: }
 \begin{equation*}
     H(W^{(m)},P^{(m)})= -\log \pi(W^{(m)}|X_n,Y_n,\Theta) + \tfrac{1}{2}P^{(m)\top}P^{(m)}
 \end{equation*}
\STATE{Compute acceptance probability:}
\begin{equation*}
    \alpha = \min\{1,\exp\big(-H(W^{(m)},P^{(m)}) + H(W^{(c-1)},P^{(c-1)})\big)\}
\end{equation*}
\STATE{With probability $\alpha$ accept $W^{(m)}$ if \( \alpha > r=unif(0,1) \), else set $W^{(m)} \gets W^{(c-1)}$}
\RETURN \(W^{(c)} = W^{(m)}\)
\end{algorithmic}
\end{algorithm}

\subsubsection{Full Gibbs--MH--HMC--ESS Sampler}
\label{sec:SM_sampler}

This section provides the complete sampling algorithm for standard GP, two- and three-layer DGP with BDR. Posterior inference is carried out using a hybrid Gibbs--MH--HMC--ESS sampler that jointly updates all the unknown parameters (\(\theta _y, g, \tau ^2, \theta _r, \theta _q, W\)) and latent variables (\(R \text{ and } Q\)). The orthogonal orientation matrices \(M\) and \(V\) are updated from their matrix Langevin full conditional distributions, the diagonal concentration parameters in \(\Lambda\) are sampled via SS, the projection matrix \(W\) is updated using HMC on the Stiefel manifold, and latent Gaussian (\(R \text{ and } Q\)) process variables are updated using elliptical slice sampling. The \(\tau ^2\) is sampled via inverse gamma full conditional distribution and remaining kernel parameters (\(\theta _y, g,\theta _r, \text{ and } \theta _q\)) are sampled using MH steps, yielding a unified posterior computation strategy for the standard GP, two- and three-layer DGP models with built-in dimension reduction.  A complete Gibbs–MH–HMC–ESS sampler for the standard GP, two- and three-layer GP model with BDR is provided in \Cref{alg:Gibbs_MH_HMC_ESS_layer1}, \Cref{alg:two_layer_sampler} and \Cref{alg:three_layer_sampler} respectively.

\begin{algorithm}[H]
\caption{: Gibbs-MH-HMC-ESS Sampler for Standard GP with BDR}
\label{alg:Gibbs_MH_HMC_ESS_layer1}
\begin{algorithmic}[1]
\STATE{\textbf{Input:} Data \((Y_n, X_n)\), prior distributions for parameters, number of iterations \(n_{\text{iter}}\)}
\STATE{\textbf{Output:} Samples of \(\tau^2, \theta_y, g, W, M, V, \text{ and } \Lambda\)}
\STATE{\textbf{Step 1: Initialization}}
\STATE{Initialize parameters from their prior distributions:}
\[
\tau^2_0 \sim \text{InverseGamma}(\alpha_1, \alpha_2), \quad g_0 \sim \text{Gamma}(\beta_1, \beta_2),
\]
\[
\theta_{y,0} \sim \text{i.i.d. Gamma}(\gamma_1, \gamma_2), \quad M_0 \sim \text{mL}(F_M), \quad V_0 \sim \text{mL}(F_V),
\]
\[
\lambda_{k,0} \sim \text{i.i.d. Gamma}(b_1, b_2), \; k = 1, \dots, D, \quad W_0 \sim \text{mL}(M_0 \Lambda_0 V_0).
\]

\STATE{\textbf{Step 2: Iterative Updates}}
\FOR{$c = 2, \dots, n_{\text{iter}}$}

    \STATE{\textbf{Update \(M\):}}
    \[
    M^{(c)} \sim p\big(M^{(c)} |W^{(c-1)}, \Lambda^{(c-1)}, V^{(c-1)}, M^{(c-1)} \big) \makebox[0.15\linewidth][r]{// } \mathbf{\mathcal{ML}}
    \]

    \STATE{\textbf{Update \(V\):}}
    \[
    V^{(c)} \sim p\big(V^{(c)} |W^{(c-1)}, \Lambda^{(c-1)}, M^{(c)}, V^{(c-1)} \big) \makebox[0.15\linewidth][r]{// }\mathbf{\mathcal{ML}}
    \]

    \FOR{$j = 1, \dots, D$}
        \STATE{\textbf{Update \(\lambda[j]\):}}
        \[
        \lambda[j]^{(c)} \sim p\big(\lambda[j]^{(c)} |W^{(c-1)}, V^{(c)}, M^{(c)}, \lambda[j]^{(c-1)} \big) \makebox[0.20\linewidth][r]{//\textbf{ SS } via } \mathcal{L}(W|M, \Lambda,  V)
        \]
       
    \ENDFOR

    \STATE{\textbf{Update \(W\):}}
    \[
    \hspace{2cm} W^{(c)} \sim  p\big(W^{(c)} | Y_n, X_n, W^{(c-1)}, \Lambda^{(c)}, M^{(c)}, V^{(c)}, g^{(c)}, \tau^{2(c)}, \theta_y^{(c)} \big) \makebox[0.15\linewidth][r]{//\textbf{HMC}}
    \]

    \STATE{\textbf{Update \(\tau^2\):}}
    \[
    \tau^{2(c)} \sim  p \big(\tau^{2(c)} | Y_n, X_n, W^{(c)}, \theta_y^{(c-1)}, g^{(c-1)} \big) \makebox[0.20\linewidth][r]{// \textbf{IG}}
    \]

    \STATE{\textbf{Update \(g\):}}
    \[
    \quad \quad g^{(c)} \sim p\big(g^{(c)} | Y_n, X_n, W^{(c)}, \theta_y^{(c-1)}, \tau^{2(c)}, g^{(c-1)} \big) \makebox[0.20\linewidth][r]{//\textbf{MH} \text{ via }}\mathcal{L}(Y_n|X_n, W, \tau ^2,\theta _y, g)
    \]

    \STATE{\textbf{Update \(\theta_y^{(c)}\):}}
    \[
    \quad \qquad \theta_y^{(c)} \sim p\big(\theta_y^{(c)} | Y_n, X_n, W^{(c)}, g^{(c)}, \tau^{2(c)}, \theta_y^{(c-1)} \big) \makebox[0.20\linewidth][r]{//\textbf{MH} \text{ via }}\mathcal{L}(Y_n|X_n, W, \tau ^2, \theta _y, g)
    \]
       
\ENDFOR
\end{algorithmic}
\end{algorithm}

\begin{algorithm}[H]
\caption{Gibbs--MH--HMC--ESS Sampler for Two-Layer DGP with BDR}
\label{alg:two_layer_sampler}
\begin{algorithmic}[1]

\STATE{\textbf{Initialize:}}
\[
\begin{aligned}
\tau^2 &\sim \mathrm{IG}(\alpha_1,\alpha_2), &
g &\sim\mathrm{Gamma}(\beta_1,\beta_2), \\
\theta_y,\theta_q,\theta_r &\overset{\text{i.i.d.}}{\sim}\mathrm{Gamma}(\gamma_1,\gamma_2), &
M,V &\sim \mathrm{mL},\qquad
\lambda[j]\sim\mathrm{Gamma}(b_1,b_2).
\end{aligned}
\]

\FOR{$c=1$ \textbf{to} $n_{\text{iter}}$}

\STATE{\textbf{Update $M$ and $V$ (matrix Langevin):}}
\[
M^{(c)} \sim p(M\mid W^{(c-1)},\Lambda^{(c-1)},V^{(c-1)}),\qquad
V^{(c)} \sim p(V\mid W^{(c-1)},\Lambda^{(c-1)},M^{(c)}).
\]

\FOR{$j=1$ \textbf{to} $D$}
\STATE{\textbf{Update $\lambda[j]$ (SS):}}
\[
\lambda[j]^{(c)} \sim p(\lambda[j]\mid W^{(c-1)},M^{(c)},V^{(c)}).
\]
\ENDFOR

\STATE{\textbf{Update $W$ (HMC):}}
\[
W^{(c)} \sim p(W\mid Y,X,Q^{(c-1)}, M^{(c)},V^{(c)},\Lambda^{(c)},\theta_y^{(c)},g^{(c)},\tau^{2(c)}).
\]

\FOR{$j=1$ \textbf{to} $D$}
\STATE{\textbf{Update $\theta_{q[j]}$ (MH):}}
\[
\theta_{q[j]}^{(c)} \sim p(\theta_{q[j]}\mid Q_j^{(c-1)},X, W^{(c)}, \theta_{q[j]}^{(c-1)}).
\]
\STATE{\textbf{Update $Q_j$ (ESS):}}
\[
Q_j^{(c)} \sim p(Q_j\mid Y,X, W^{(c)},\theta_y^{(c)},\theta_{q[j]}^{(c)}).
\]
\ENDFOR

\STATE{\textbf{Update $\tau^2$ (Gibbs):}}
\[
\tau^{2(c)} \sim p(\tau^2\mid Y,Q^{(c)},\theta_y^{(c-1)},g^{(c-1)}).
\]

\STATE{\textbf{Update $g$ (MH):}}
\[
g^{(c)} \sim p\bigl(g\mid Y,Q^{(c)}, g^{(c-1)},\theta_y^{(c-1)},\tau^{2(c)}\bigr).
\]

\STATE{\textbf{Update $\theta_y$ (MH):}}
\[
\theta_y^{(c)} \sim p(\theta_y\mid Y,Q^{(c)}, \theta_y^{(c-1)}, \tau^{2(c)},g^{(c)}).
\]

\ENDFOR
\end{algorithmic}
\end{algorithm}

\begin{algorithm}[H]
\caption{Gibbs--MH--HMC--ESS Sampler for Three-Layer DGP with BDR}
\label{alg:three_layer_sampler}
\begin{algorithmic}[1]
\STATE{\textbf{Initialize:}}
\[
\begin{aligned}
\tau^2 &\sim \mathrm{IG}(\alpha_1,\alpha_2), &
g &\sim\mathrm{Gamma}(\beta_1,\beta_2), \\
\theta_y,\theta_q,\theta_r &\overset{\text{i.i.d.}}{\sim}\mathrm{Gamma}(\gamma_1,\gamma_2), &
M,V &\sim \mathrm{mL},\qquad
\lambda[j]\sim\mathrm{Gamma}(b_1,b_2).
\end{aligned}
\]
\FOR{$c=1$ \textbf{to} $n_{\text{iter}}$}
\STATE{\textbf{Update $M$ and $V$ (matrix Langevin):}}
\[
M^{(c)} \sim p(M\mid W^{(c-1)},\Lambda^{(c-1)},V^{(c-1)}),\qquad
V^{(c)} \sim p(V\mid W^{(c-1)},\Lambda^{(c-1)},M^{(c)}).
\]
\FOR{$j=1$ \textbf{to} $D$}
\STATE{\textbf{Update $\lambda[j]$ (SS):}}
\[
\lambda[j]^{(c)} \sim p(\lambda[j]\mid W^{(c-1)},M^{(c)},V^{(c)}).
\]
\ENDFOR
\STATE{\textbf{Update $W$ (HMC):}}
\[
W^{(c)} \sim p(W\mid Y,X,Q^{(c-1)}, M^{(c)},V^{(c)},\Lambda^{(c)},\theta_y^{(c)},g^{(c)},\tau^{2(c)}).
\]
\FOR{$j=1$ \textbf{to} $D$}
\STATE{\textbf{Update $\theta_{r[j]}$ (MH):}}
\[
\theta_{r[j]}^{(c)} \sim p(\theta_{r[j]}\mid R_j^{(c-1)},X, W^{(c)}, \theta_{r[j]}^{(c-1)}).
\]
\STATE{\textbf{Update $R_j$ (ESS):}}
\[
R_j^{(c)} \sim p(R_j\mid X,W^{(c)},\theta_{r[j]}^{(c)},\theta_{q[j]}^{(c)}).
\]
\ENDFOR
\FOR{$j=1$ \textbf{to} $D$}
\STATE{\textbf{Update $\theta_{q[j]}$ (MH):}}
\[
\theta_{q[j]}^{(c)} \sim p(\theta_{q[j]}\mid Q_j^{(c-1)},R^{(c-1)}, \theta_{q[j]}^{(c-1)}).
\]
\STATE{\textbf{Update $Q_j$ (ESS):}}
\[
Q_j^{(c)} \sim p(Q_j\mid Y,R^{(c-1)},\theta_y^{(c)},\theta_{q[j]}^{(c)}).
\]
\ENDFOR
\STATE{\textbf{Update $\tau^2$ (Gibbs):}}
\[
\tau^{2(c)} \sim p(\tau^2\mid Y,Q^{(c)},\theta_y^{(c-1)},g^{(c-1)}).
\]
\STATE{\textbf{Update $g$ (MH):}}
\[
g^{(c)} \sim p\bigl(g\mid Y,Q^{(c)}, g^{(c-1)},\theta_y^{(c-1)},\tau^{2(c)}\bigr).
\]
\STATE{\textbf{Update $\theta_y$ (MH):}}
\[
\theta_y^{(c)} \sim p(\theta_y\mid Y,Q^{(c)}, \theta_y^{(c-1)}, \tau^{2(c)},g^{(c)}).
\]
\ENDFOR
\end{algorithmic}
\end{algorithm}




\subsection{Posterior Prediction}
\label{subsec:prediction}

\textcolor{black}{The three model variants are nested, and both inference and prediction specialize accordingly. The single-layer GP is recovered by removing the latent intermediate layer and modeling the response directly with a GP on the projected inputs; the two-layer DGP adds one latent GP layer that warps the projected inputs before the output GP; and the three-layer DGP stacks one further latent GP layer. The prior structure is shared across variants, with an additional GP prior introduced for each extra latent layer, and posterior inference follows the same hierarchical sampling strategy with the parameter blocks expanded accordingly. Posterior prediction specializes in the same way: for the GP it reduces to a single GP conditioning step, whereas for the two- and three-layer DGPs uncertainty is propagated recursively through each latent layer before the output layer. Consequently, the two-layer prediction equations below extend directly to the three-layer model by iterating the same conditional GP prediction layer by layer.}

\textcolor{black}{Posterior prediction is performed conditionally on posterior draws. For \(cth\) posterior sample, we define the training inputs as \(Z_n^{(c)}=X_n W^{(c)} \). Let $\mathcal{X} \in \mathbb{R}^{n_{\mathrm{test}}\times p}$ denote the test inputs and define the projected test inputs for \(cth\) posterior sample as
\[
\mathcal{Z}^{(c)} = \mathcal{X} W^{(c)} \in \mathbb{R}^{n_{\mathrm{test}}\times D}.
\] 
\paragraph{ Prediction for Ordinary GP}
For the ordinary GP model, the GP predictive distribution is
\begin{equation}
\begin{aligned}
   &  \mathcal{Y}^{(c)} \mid Y, Z_n^{(c)}, \mathcal{Z}^{(c)}
\sim \mathcal{N}\big(\mu_{\mathcal{Y}}^{(c)}, \Sigma_{\mathcal{Y}}^{(c)} \big) \text{ where } \mu_{\mathcal{Y}}^{(c)}=  C_y^{(c)}(\mathcal{Z}^{(c)},Z_n^{(c)})
  \left(C_y^{(c)}(Z_n^{(c)},Z_n^{(c)}) + g^{(c)} I_n\right)^{-1} Y \text{ and } \\
  & \Sigma_{\mathcal{Y}}^{(c)} = \tau^{2(c)}\!\left[ C_y^{(c)}(\mathcal{Z}^{(c)},\mathcal{Z}^{(c)}) +g^{(c)}I_{n_{\mathrm{test}}} -
  C_y^{(c)}(\mathcal{Z}^{(c)},Z_n^{(c)})
  \left(C_y^{(c)}(Z_n ^{(c)},Z_n ^{(c)})  + g^{(c)} I_n\right)^{-1}
  C_y^{(c)}(Z_n^{(c)},\mathcal{Z}^{(c)})\right]
\end{aligned}
\end{equation} Thus prediction consists of a single GP conditioning step.
\paragraph{Prediction for Two-Layer DGP} 
For the inner layer, each node $Q_j$, $j=1,\dots,D$, the GP predictive distribution is
\begin{equation}
\begin{aligned}
   & \mathcal{Q}_j^{(c)} \mid Q_j^{(c)},\mathcal{Z}^{(c)}
\sim
\mathcal{N}\!\left(\mu_{ \mathcal{Q}_j}^{(c)},\Sigma_{ \mathcal{Q}_j}^{(c)}\right) \text{ where } \mu_{\mathcal{Q}_j}^{(c)}= 
    C_q^{(c)}(\mathcal{Z}^{(c)},Z_n ^{(c)})C_q^{(c)}(Z_n ^{(c)},Z_n^{(c)})^{-1} Q_j^{(c)}  \text{ and } \\
    & \Sigma_{ \mathcal{Q}_j}^{(c)}=
    C_q^{(c)}(\mathcal{Z}^{(c)},\mathcal{Z}^{(c)}) -
    C_q^{(c)}(\mathcal{Z}^{(c)},Z_n^{(c)})\,
    C_q^{(c)}(Z_n^{(c)},Z_n^{(c)})^{-1}
    C_q^{(c)}(Z_n^{(c)},\mathcal{Z}^{(c)}).
\label{eq:latent_pred} 
\end{aligned}
\end{equation}
Stacking the means yields the warped inputs
\[
\mathcal{Q}^{(c)} = \big(\mu_{\mathcal{Q}_1}^{(c)},\dots, \mu_{\mathcal{Q}_D}^{(c)}\big)
\in \mathbb{R}^{n_{\mathrm{test}}\times D}.
\]
The outer layer, $\mathcal{Y}^{(c)}$, predictive distribution is
\begin{equation}
\begin{aligned}
& \mathcal{Y}^{(c)} \mid \mathcal{Q}^{(c)}, Q^{(c)}, Y
\sim
\mathcal{N}\!\left(\mu_{ \mathcal{Y}}^{(c)},\Sigma_{ \mathcal{Y}}^{(c)}\right) \text{ where } \mu_{\mathcal{Y}}^{(c)}= 
    C_y^{(c)}(\mathcal{Q}^{(c)},Q^{(c)})\left(C_y^{(c)}(Q^{(c)},Q^{(c)}) + g^{(c)} I_n\right)^{-1}Y  \text{ and } \\
 & \Sigma_{ \mathcal{Y}}^{(c)}=
   \tau^{2(c)}\!\left[ C_y^{(c)}(\mathcal{Q}^{(c)},\mathcal{Q}^{(c)}) +g^{(c)}I_{n_{\mathrm{test}}} -
    C_y^{(c)}(\mathcal{Q}^{(c)},Q^{(c)})\,
    \left(C_y^{(c)}(Q^{(c)},Q^{(c)})  + g^{(c)} I_n\right) ^{-1}
    C_y^{(c)}(Q^{(c)},\mathcal{Q}^{(c)})\right].
\end{aligned}
\end{equation}Thus prediction proceeds sequentially:
\[
\mathcal{Z}^{(c)} \rightarrow
\mathcal{Q}^{(c)} \rightarrow
\mathcal{Y}^{(c)}.
\]
\paragraph{Prediction for the Three-Layer DGP} For the additional latent layer $R_k$, $k=1,\dots,D$, the GP predictive distribution is 
\begin{equation}
\begin{aligned}
   & \mathcal{R}_k^{(c)} \mid R_k^{(c)},\mathcal{Z}^{(c)}
\sim
\mathcal{N}\!\left(\mu_{ \mathcal{R}_k}^{(c)},\Sigma_{ \mathcal{R}_k}^{(c)}\right) \text{ where } \mu_{\mathcal{R}_k}^{(c)}= 
    C_r^{(c)}(\mathcal{Z}^{(c)},Z_n^{(c)})C_r^{(c)}(Z_n ^{(c)},Z_n^{(c)})^{-1} R_k^{(c)}  \text{ and } \\
    & \Sigma_{ \mathcal{R}_k}^{(c)}=
    C_r^{(c)}(\mathcal{Z}^{(c)},\mathcal{Z}^{(c)}) -
    C_r^{(c)}(\mathcal{Z}^{(c)},Z_n ^{(c)})\,
    C_r^{(c)}(Z_n^{(c)},Z_n^{(c)})^{-1}
    C_r^{(c)}(Z_n ^{(c)},\mathcal{Z}^{(c)}).
\label{eq:latent_pred} 
\end{aligned}
\end{equation}
Stacking the means yields the warped inputs
\[
\mathcal{R}^{(c)} = \big(\mu_{\mathcal{R}_1}^{(c)},\dots, \mu_{\mathcal{R}_D}^{(c)}\big)
\in \mathbb{R}^{n_{\mathrm{test}}\times D}.
\]
For the second layer, each node $Q_k$, the GP predictive distribution is
\begin{equation}
\begin{aligned}
   & \mathcal{Q}_j^{(c)} \mid Q_j^{(c)},\mathcal{R}^{(c)}, R^{(c)}
\sim
\mathcal{N}\!\left(\mu_{ \mathcal{Q}_j}^{(c)},\Sigma_{ \mathcal{Q}_j}^{(c)}\right) \text{ where } \mu_{\mathcal{Q}_j}^{(c)}= 
    C_q^{(c)}(\mathcal{R}^{(c)},R^{(c)})C_q^{(c)}(R^{(c)},R^{(c)})^{-1} Q_j^{(c)}  \text{ and } \\
    & \Sigma_{ \mathcal{Q}_j}^{(c)}=
    C_q^{(c)}(\mathcal{R}^{(c)},\mathcal{R}^{(c)}) -
    C_q^{(c)}(\mathcal{R}^{(c)},R^{(c)})\,
    C_q^{(c)}(R^{(c)},R^{(c)})^{-1}
    C_q^{(c)}(R^{(c)},\mathcal{R}^{(c)}).
\label{eq:latent_pred} 
\end{aligned}
\end{equation}
Stacking the means yields the warped inputs
\[
\mathcal{Q}^{(c)} = \big(\mu_{\mathcal{Q}_1}^{(c)},\dots, \mu_{\mathcal{Q}_D}^{(c)}\big)
\in \mathbb{R}^{n_{\mathrm{test}}\times D}.
\]
The outer layer, $\mathcal{Y}^{(c)}$, the predictive distribution is
\begin{equation}
\begin{aligned}
& \mathcal{Y}^{(c)} \mid \mathcal{Q}^{(c)}, Q^{(c)}, Y
\sim
\mathcal{N}\!\left(\mu_{ \mathcal{Y}}^{(c)},\Sigma_{ \mathcal{Y}}^{(c)}\right) \text{ where } \mu_{\mathcal{Y}}^{(c)}= 
    C_y^{(c)}(\mathcal{Q}^{(c)},Q^{(c)})\left[C_y^{(c)}(Q^{(c)},Q^{(c)}) + g^{(c)} I_n\right]^{-1}Y  \text{ and } \\
 & \Sigma_{ \mathcal{Y}}^{(c)}=
    \tau^{2(c)}\!\left[C_y^{(c)}(\mathcal{Q}^{(c)},\mathcal{Q}^{(c)}) + g^{(c)}I_{n_{\mathrm{test}}}  -
    C_y^{(c)}(\mathcal{Q}^{(c)},Q^{(c)})\,
    \left(C_y^{(c)}(Q^{(c)},Q^{(c)})  + g^{(c)} I_n\right)^{-1}
    C_y^{(c)}(Q^{(c)},\mathcal{Q}^{(c)})\right].
\end{aligned}
\end{equation}
The sequential prediction proceeds as
\[
\mathcal{Z}^{(c)} \rightarrow
\mathcal{R}^{(c)} \rightarrow
\mathcal{Q}^{(c)} \rightarrow
\mathcal{Y}^{(c)}.
\]
}


Posterior averages over $c\in\mathcal{C}$ then yield the predictive mean and
covariance:
\[
\bar{\mu}_\mathcal{Y} = \frac{1}{|\mathcal{C}|} \sum_{c\in\mathcal{C}} \mu_{\mathcal{Y}}^{(c)}, 
\qquad 
\bar{\Sigma}_\mathcal{Y}
= \frac{1}{|\mathcal{C}|} \sum_{c\in\mathcal{C}} \Sigma_{\mathcal{Y}}^{(c)}
  + \frac{1}{|\mathcal{C}|-1} \sum_{c\in\mathcal{C}}
      (\mu_{\mathcal{Y}}^{(c)}-\bar{\mu}_{\mathcal{Y}})(\mu_{\mathcal{Y}}^{(c)}-\bar{\mu}_{\mathcal{Y}})^\top.
\]

\subsection{Computational Complexity}
\label{subsec:complexity}

The dominant costs arise from $n\times n$ GP covariance operations. Cholesky factorizations are $O(n^3)$ and must be performed multiple times per iteration. This difficulty could be alleviated by using methods such as Vecchia approximation or other scalable methods for GP \cite{SauerVecchia2023} as discussed in \Cref{sec:Con_Discus}.

\paragraph{Standard GP with BDR}
Each update of $g$ or $\theta_y$ requires constructing $C(Z_n,Z_n)$ ($O(n^2 D)$)
and a Cholesky factorization ($O(n^3)$).  
HMC updates for $W$ require $T$ leapfrog steps, each costing
\[
O(T(n^3 + n^2 p D)),
\]
due to evaluation of log-likelihoods and kernel gradients.

\paragraph{Two-layer DGP}
For each of $D$ latent nodes, updates of $\theta_q[j]$ and $Q_j$ require an
$O(n^3)$ Cholesky factorization. Thus, latent layers multiply the $n^3$ cost by
approximately $D$. Overall,
\[
O\!\left( 
T(n^3 + n^2 p D) \;+\; D n^3 
\right).
\]

\paragraph{Three-layer DGP}
A second latent layer doubles the latent cost, giving
\[
O\!\left( 
T(n^3 + n^2 p D) \;+\; 2 D n^3
\right).
\]

The updates for $(M,V,\Lambda)$ operate in spaces of size $p\times D$ or
$D\times D$ and cost at most $O(pD^2+D^3)$ per update, negligible compared to
$n^3$. 

\section{Numerical Examples}\label{sec:Numerical_studies}

In this section, we present a series of numerical experiments to evaluate the effectiveness of GP surrogate modeling equipped with various dimensionality reduction strategies. We begin with controlled synthetic examples where the underlying structure is known, enabling a direct assessment of input–subspace recovery, predictive accuracy, and uncertainty quantification. Importantly, our standard GP formulation provides a unified framework in which dimension reduction of the input variables and prediction are learned jointly. We then turn to more complex synthetic benchmark functions that challenge surrogate models, followed by applications to a stochastic partial differential equation and the ONERA–M6 wing design problem. Together, these examples illustrate the performance of the proposed approach in both standard GP and deep GP (DGP) settings.


\textcolor{black}{ESS for the Gaussian latent nodes of the DGP ($Q=\left[Q_1, \cdots, Q_D\right]$ and $R=\left[R_1, \cdots, R_D\right]$), as in \cite{Sauer02012023}, is tuning-free and provides robust mixing. We update \(\lambda\) using a slice sampling scheme motivated by ESS. Specifically, the algorithm employs the elliptical proposal construction and angular bracket-shrinking strategy of ESS, but is adapted to the present setting in which \(\lambda\) is assigned a Gamma prior and is restricted to the positive constrained parameter space which it does not correspond to standard Gaussian-prior ESS. The slice-type sampling and ESS acceptance rate is $100\%$, because the algorithm adaptively shrinks the bracket until it finds a proposal that satisfies the slice condition.} Since no universal rules exist for selecting optimal HMC integration parameters, these were tuned empirically for each problem. After experimentation, we fixed the HMC step size to \(\varepsilon = 0.09\) and the number of leapfrog steps to \(T = 15\) throughout all studies, which offered stable performance, good acceptance rates whiles maintaining good runtime and fast convergence. \textcolor{black}{One important characteristic of the algorithm is that a higher acceptance rate generally requires a larger \(T\)
and a smaller value of \(\varepsilon\), which can significantly slow down the algorithm. Conversely, using a smaller \(T\) and a larger \(\varepsilon\)  may improve computational speed, but often at the expense of a lower acceptance rate. To obtain a satisfactory acceptance rates, we leveraged on the initial values for \(M, \Lambda , V \) and \(W\). We first sampled from a standard normal distribution and then used SVD to obtain 
\(M, \Lambda, \text{ and } V \). We subsequently sampled 
the initial value of \(W\) from a matrix Langevin distribution parameterized by the initial values \(M\Lambda V^T \). That is, we generate
\[
F \in \mathbb{R}^{p \times D}, \qquad F_{ij} \stackrel{\mathrm{iid}}{\sim} \mathcal{N}(0,1),
\]
and compute its SVD
\[
F = M_{\mathrm{full}} L V^\top.
\]
The resulting values are taken as
\[
M_{\mathrm{init}} = M_{\mathrm{full}}[:,1\!:\!D], \qquad
\Lambda_{\mathrm{init}} = \mathrm{diag}(L), \qquad
V_{\mathrm{init}} = V.
\]
$W_{init}$ is sampled from \(\mathcal{ML}(M_{\mathrm{init}} \Lambda_{\mathrm{init}} V_{\mathrm{init}} ^T\)). For $F_M$ and $F_V$ values, \(F_M\) is generated from a matrix Langevin distribution parametrized as \(M_{\mathrm{init}}\) (\(\mathcal{ML} (M_{\mathrm{init}} )\)), and \(F_V\) parameterized as \(V_{\mathrm{init}}\) (\(\mathcal{ML} (V_{\mathrm{init}} )\)). We take \(\Lambda_{\mathrm{init}}\) as the initial value for \(\Lambda\). For the Gamma prior values for the \(\lambda\), we used \(b_1 =5/2\) and \(b_2=10/3\). This initialization strategy and the prior parameterization values were helpful for achieving a good acceptance rate. The bounds for the uniform sliding-window scheme were set to \(l = 1\) and \(u = 2\) across all layers. For the lengthscale and nugget priors \(G(3/2, b_{[\,\cdot\,]})\), the scale parameters were
\begin{equation}
    b_{[\theta_r]} = \frac{3.9}{6}, \qquad 
    b_{[\theta_q]} = \frac{3.9}{3}, \qquad 
    b_{[\theta_y]} = b_{[g]}=3.9.
    \label{eq:lengthcale_prior_values}
\end{equation}. In addition, the prior on \(\tau^2\) uses the hyperparameter values
\[
\alpha_1 = 0.001, \qquad \alpha_2 = 0.001.
\] We set the initial values for the kernel hyperparameters as \[
\theta _{y _{init}} = \theta _{q _{init}}= \theta _{r _{init}}=1, \qquad g_{init} = 9 \times 10^{-5}
\] and measurement noise, \(\tau ^2 = 0.005\)}.

In each experiment, the data are randomly partitioned into \(80\%\) training and \(20\%\) testing sets. For a given split, posterior inference is performed using only the training set. Specifically, for each fitted model, we generated \(2000\) posterior samples with a burn-in of 500 and a thinning interval of \(3\). Each of the 1500 retained posterior draw produces a new set of parameter values, which is then used to generate predictions on the corresponding held-out test set and to compute the associated performance metrics. 
We evaluate model performance using several quantitative metrics: the Root Mean Square Prediction Error (RMSPE), the Nash–Sutcliffe Model Efficiency coefficient (NSME) \cite{gautier2021fullybayesian}, the Continuous Ranked Probability Score (CRPS) \cite{SauerVecchia2023}, the predictive log-likelihood score \cite{gramacy2020surrogates}, the Bayesian Information Criterion (BIC) \cite{TRIPATHY2016191}, and the Mean Log Pointwise Predictive Density (MLPPD) \cite{gautier2021fullybayesian} \textcolor{black}{as well as the empirical coverage probability (CP) and average length of the $95\%$ credible intervals (ALCI) to further assess posterior predictive calibration and interval efficiency.} \textcolor{black}{Following each experiment, a table a given to report the median values of those metrics}. Let $n_{\text{test}}$ denote the number of test inputs,  
$\mathcal{Y}_i$ the held-out response at $\mathcal{X}_i$,  
$\hat{\mathcal{Y}}_i$ the posterior predictive mean,  
and $\sigma_i^2$ the associated predictive variance. The definitions of these metrics are 

\begin{enumerate}
\item \textbf{Root Mean Square Predictive Error (RMSPE)}
\[
\mathrm{RMSPE}
=
\sqrt{
\frac{1}{n_{\text{test}}}
\sum_{i=1}^{n_{\text{test}}}
(\mathcal{Y}_i - \hat{\mathcal{Y}}_i)^2 }.
\]

\item \textbf{Nash--Sutcliffe Model Efficiency (NSME)} 
\cite{gautier2021fullybayesian}
\[
\mathrm{NSME}
=
1 -
\frac{\sum_{i}(\mathcal{Y}_i-\hat{\mathcal{Y}}_i)^2}
{\sum_{i}(\mathcal{Y}_i-\bar{\mathcal{Y}})^2}.
\]

\item \textbf{Continuous Ranked Probability Score (CRPS)}  
(for Gaussian predictive distributions) \cite{SauerVecchia2023}
\[
\mathrm{CRPS}
=
\frac{1}{n_{\text{test}}}
\sum_{i=1}^{n_{\text{test}}}
\sigma_i 
\left(
\frac{1}{\sqrt{\pi}}
 - 2\phi(z_i)
 - z_i (2\Phi(z_i)-1)
\right),
\quad
z_i=\frac{\mathcal{Y}_i-\hat{\mathcal{Y}}_i}{\sigma_i}.
\]

\item \textbf{Predictive Log-Likelihood Score}
\[
\mathrm{Score}
=
-\log|\Sigma(\mathcal{Z})|
-
(\mathcal{Y}-\hat{\mathcal{Y}})^\top
\Sigma(\mathcal{Z})^{-1}
(\mathcal{Y}-\hat{\mathcal{Y}}).
\]

\item \textbf{Bayesian Information Criterion (BIC)}  
\cite{TRIPATHY2016191}
\[
\mathrm{BIC}
=
\mathcal{L}(\boldsymbol{\theta}_D^{*};\mathcal{Z},Y)
-\frac12\,\#(\boldsymbol{\theta}_D^{*})\log(n_{\text{train}}),
\]
with  
$\boldsymbol{\theta}_D^{*}=\{W,\tau^2,g,\theta_y\}$.

\item \textbf{Mean Log Pointwise Predictive Density (MLPPD)}  
\cite{gautier2021fullybayesian}
\[
\mathrm{MLPPD}
=
\frac{1}{n_{\text{test}}}
\sum_{i=1}^{n_{\text{test}}}
\left[
-\frac12\log(2\pi\sigma_i^2)
-
\frac{(\mathcal{Y}_i-\hat{\mathcal{Y}}_i)^2}{2\sigma_i^2}
\right].
\]

\item \textbf{Empirical coverage probability (CP)}: Let \(\{(x_i,y_i)\}_{i=1}^{n_{\mathrm{test}}}\) denote the test set, and let
\[
[\ell_i,u_i]
\]
be the posterior predictive \(95\%\) credible interval for the response at \(x_i\), where \(\ell_i\) and \(u_i\) are the lower and upper predictive bounds, respectively.

The empirical CP is defined as
\[
\mathrm{CP}
=
\frac{1}{n_{\mathrm{test}}}
\sum_{i=1}^{n_{\mathrm{test}}}
\mathbf{1}\!\left(y_i \in [\ell_i,u_i]\right),
\]
where \(\mathbf{1}(\cdot)\) is the indicator function. This measures the proportion of test responses that fall within the nominal \(95\%\) posterior predictive credible intervals. 

\item \textbf{Average length of the credible intervals (ALCI)}: Following empirical CP, ALCI is defined as
\[
\mathrm{ALCI}
=
\frac{1}{n_{\mathrm{test}}}
\sum_{i=1}^{n_{\mathrm{test}}}
(u_i-\ell_i).
\]
This summarizes the average width of the predictive \(95\%\) credible intervals, with smaller values indicating sharper uncertainty quantification, provided coverage remains adequate.

\end{enumerate}

\textcolor{black}{Additionally, boxplots of RMSPE and MLPPD are presented to summarize the distribution of these posterior-sample-based test-set performance metrics.} Together, they offer complementary insight into predictive accuracy, reliability of uncertainty quantification, and overall model quality. 

This paper explores Monte Carlo benchmarking exercises across three simulated examples and two real-world computer experiments. The comparator methods are:

\begin{itemize}
    \item \textbf{GP ($\mathbf{D}$) BDR}: Bayesian standard GP with built-in dimension reduction (BDR) for input subspace dimension $\mathbf{D}$, where $\mathbf{D}=1,2,3$.

    \item \textbf{DGP $\mathbf{A}$-layer ($\mathbf{D}$) BDR}: Bayesian DGP with $\mathbf{A}$ layers and BDR for input subspace $\mathbf{D}$, where $\mathbf{A}=2,3$.

    \item \textbf{DGP $\mathbf{A}$-layer ($\mathbf{D}$) W/o}: DGP with $\mathbf{A}$ layers and input subspace $\mathbf{D}$ \emph{without} BDR—i.e., without estimating \(W\)—implemented following \cite{Sauer02012023}.

    \item \textbf{GP ($\mathbf{D}$) Oracle}: standard GP with dimension reduction using the true subspace \(W\).

    \item \textbf{DGP $\mathbf{A}$-layer ($\mathbf{D}$) Oracle}: DGP with $\mathbf{A}$ layers using the true input subspace \(W\), as in \cite{Sauer02012023}.
\end{itemize}

The oracle variants serve as idealized benchmarks. All DGP models are restricted to at most three layers. While RMSPE focuses on the accuracy of predictive means, CRPS additionally incorporates predictive variances, offering a more complete view of uncertainty quantification. BIC balances goodness-of-fit with model complexity by penalizing the number of parameters and training size. The predictive log-likelihood score evaluates the likelihood of test outputs under the predictive distribution, and NSME measures predictive efficiency relative to the variability in observations. The sampling procedures for the Gibbs–MH–HMC–ESS algorithms for one-, two- and three-layer DGPs are provided in \Cref{alg:Gibbs_MH_HMC_ESS_layer1}, \Cref{alg:two_layer_sampler} and \Cref{alg:three_layer_sampler} respectively. \textcolor{black}{All computations were performed on a MacBook Pro equipped with an M4 Pro chip, a 16-core CPU, 24 GB of RAM, and 512 GB of storage. A full reproducibility repository is provided at \href{https://github.com/UCStat/Bayesian_BDR_GP.git}{GitHub}. The repository includes complete implementations for data generation, Gibbs and multichain samplers, posterior prediction, diagnostic metrics, and plotting for all model layers considered in this paper. It also contains end-to-end scripts/notebooks (run\_multichains.py, run\_multichains.ipynb) and testing/verification utilities (test\_all\_cases.py, verify\_2\_samples.py) to help users reproduce both core results and workflow checks. Detailed setup and execution guidance is documented in README.md and QUICKSTART.md.}


\textcolor{black}{For Case 1 and 2 synthetic examples, the true projection matrix $W$ was generated following the setup in the \cite{TRIPATHY2016191}. The $p \times D$ matrix was first sampled from a standard normal distribution, and a QR decomposition was then used to produce an orthonormal basis, yielding a valid projection matrix on the Stiefel manifold. This random construction is standard in synthetic benchmarking because it provides a neutral and mathematically convenient way to define a low-dimensional subspace without favoring any particular coordinate direction. The purpose is not to replicate a specific real-world mechanism, but rather to assess whether the proposed method can recover an unknown low-dimensional structure and deliver accurate predictions in high-dimensional settings.}

\subsection{Case 1: Synthetic Quadratic Response Surface with Known Structure}
\label{susec:case1}

We begin with a controlled synthetic experiment designed to evaluate the ability of GP- and DGP-based surrogate models to recover a known low-dimensional input structure and to assess predictive accuracy and uncertainty quantification. Let the input vector $\bx \in \mathbb{R}^p$ follow a standard multivariate normal distribution, $\bx \sim \mathcal{N}_p(0,\bI)$ with $p=10$, where the components are independent standard normal variables. The response is generated according to a latent low-dimensional model of the form
\[
f(\bx) \approx \eta (\bz), \qquad \bz = \bW^\top \bx,
\]
where $\bW \in \mathbb{R}^{p \times D}$ denotes the projection matrix onto a $D$-dimensional active subspace. The mapping $\eta: \mathbb{R}^D \to \mathbb{R}$ is specified as a quadratic function,
\[
\eta(\bz) = \mathbf{a_0} + \ba^\top \bz + \bz^\top \bA \bz,
\]
with intercept $\mathbf{a_0}$, linear coefficients $\ba \in \mathbb{R}^D$, and a $D \times D$ quadratic-form matrix $\bA$. The observed output is then
\[
Y = \eta(\bz) + \epsilon,
\]
where $\epsilon \sim \mathcal{N}(0,\sigma^2_\epsilon)$ is Gaussian observational noise. We generate $n$ independent draws of $\bx$, compute the corresponding noiseless latent response $\eta(\bz)$, and then add Gaussian perturbations to form $Y$.

For each $n \in \{350, 600\}$, the data are randomly partitioned into 80\% training and 20\% testing sets, yielding training sizes $n_{\text{train}} \in \{280, 480\}$. All models—standard GP, two-layer DGP, and three-layer DGP—are trained using only the training data, and their predictive performance is evaluated on the held-out test set. This setup allows us to assess both the accuracy of the learned input subspace and the quality of predictions when the true dimensionality is $D=1$ or $D=2$, consistent with previous studies such as \cite{TRIPATHY2016191, chaos2018, TSILIFIS2021114147}.

\subsubsection{1D Input Subspace}
\label{subsubsec:case1_1D}

In the one-dimensional case ($D=1$), the true projection matrix $\bW$ reduces to a $10$-dimensional vector:
\[
\bW = (-0.0091, -0.0579, -0.1877, 0.4774, 0.4559, -0.6714, -0.1264, -0.0082, 0.0724, -0.2308)^\top.
\]
The remaining parameters of $\eta(\cdot)$ specialize to scalars,
\[
\mathbf{a_0} = -0.16113, 
\qquad \ba = (-0.97483), 
\qquad \bA = (-1.66526), 
\qquad \sigma^2_\epsilon = 0.01.
\]

\begin{figure}[h!]
    \centering
    \begin{subfigure}{0.49\textwidth}
        \centering
        \includegraphics[width=\linewidth]{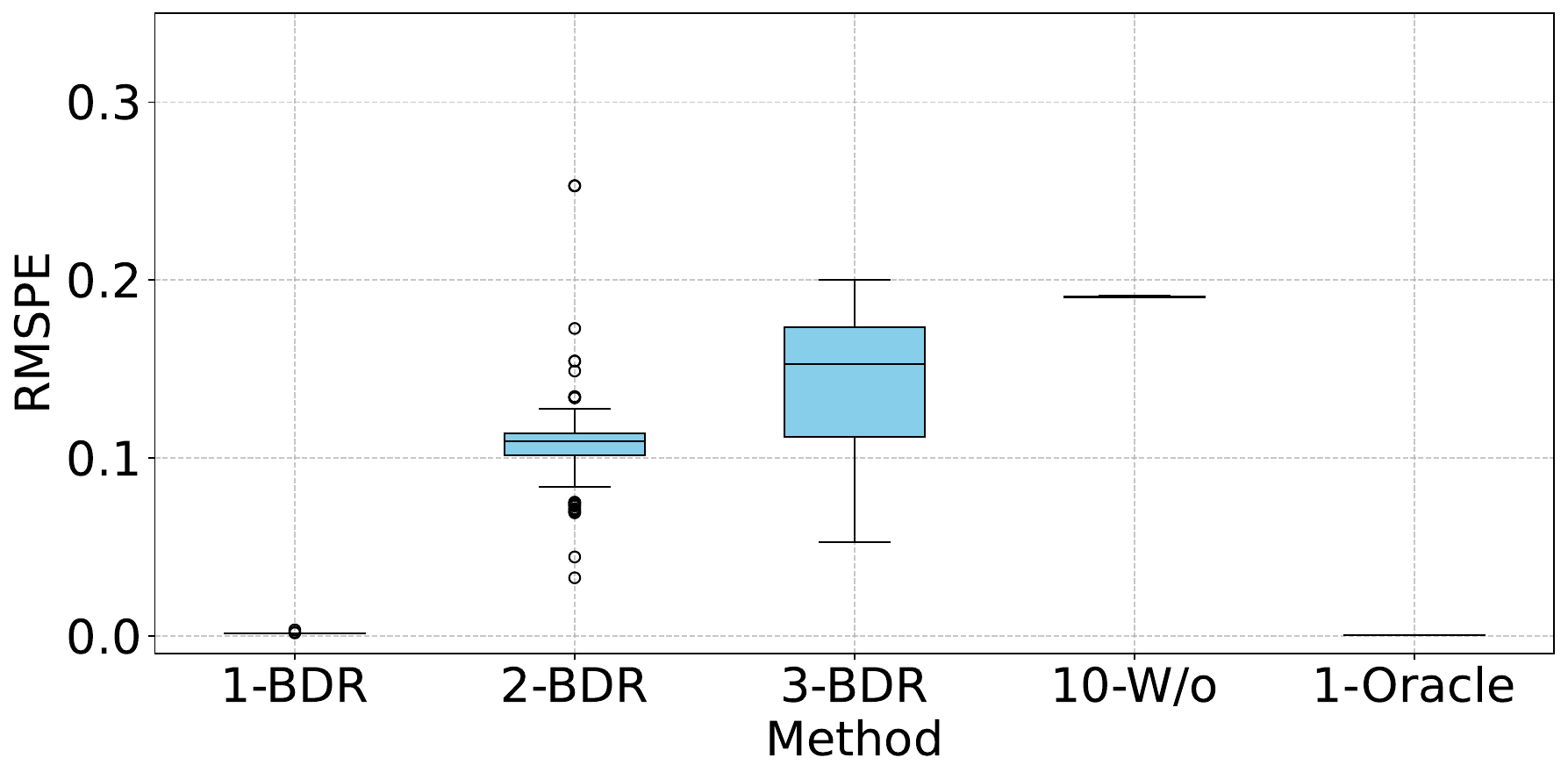}
        \caption{Standard GP at $n_{\text{train}}=280$}
        \label{fig:case1_1d_rmspe_280_l1}
    \end{subfigure}
    \hfill
    \begin{subfigure}{0.49\textwidth}
        \centering
        \includegraphics[width=\linewidth]{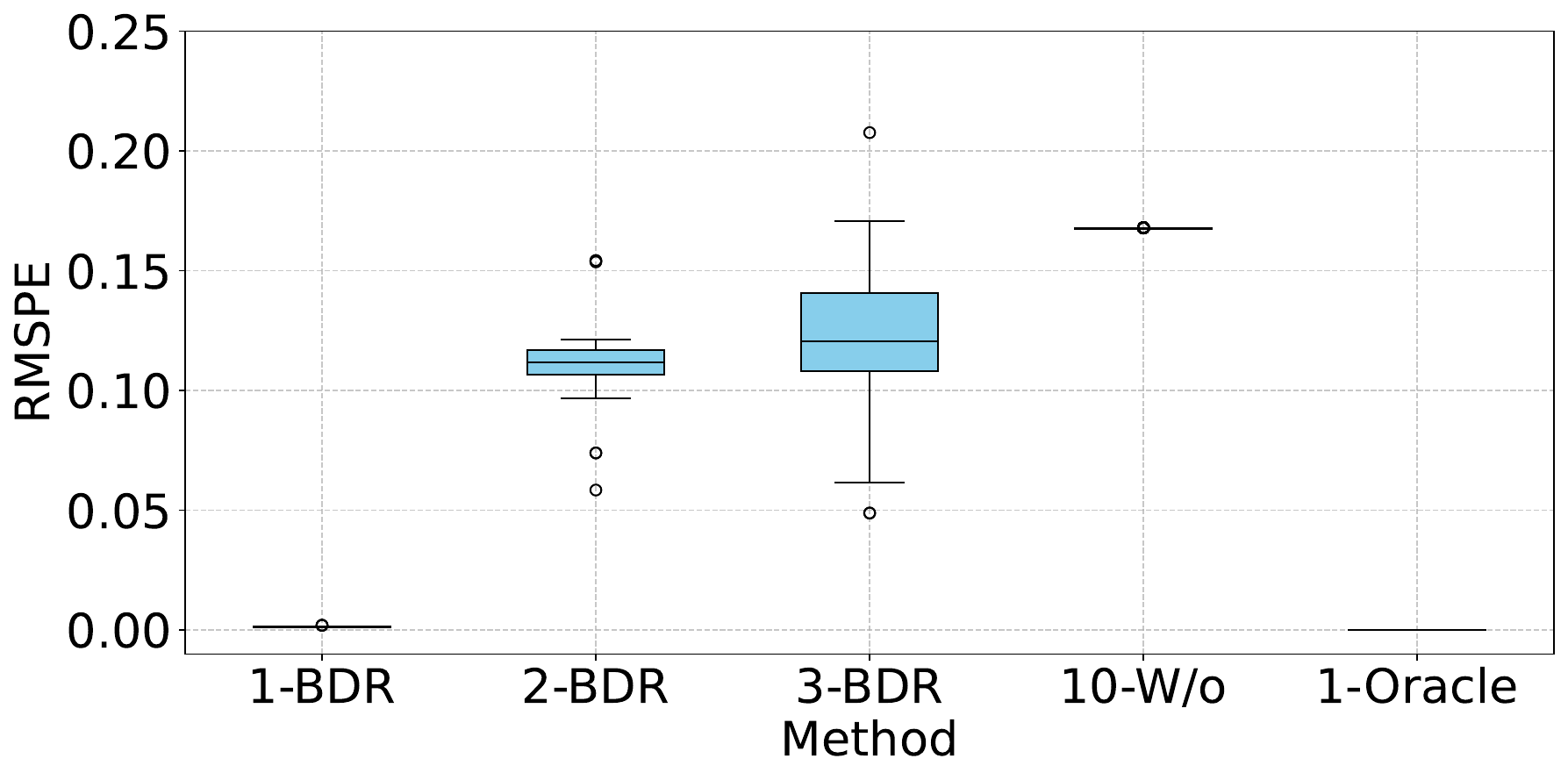}
        \caption{Standard GP at $n_{\text{train}}=480$}
        \label{fig:case1_1d_rmspe_480_l1}
    \end{subfigure}
    \hfill
    \begin{subfigure}{0.49\textwidth}
        \centering
        \includegraphics[width=\linewidth]{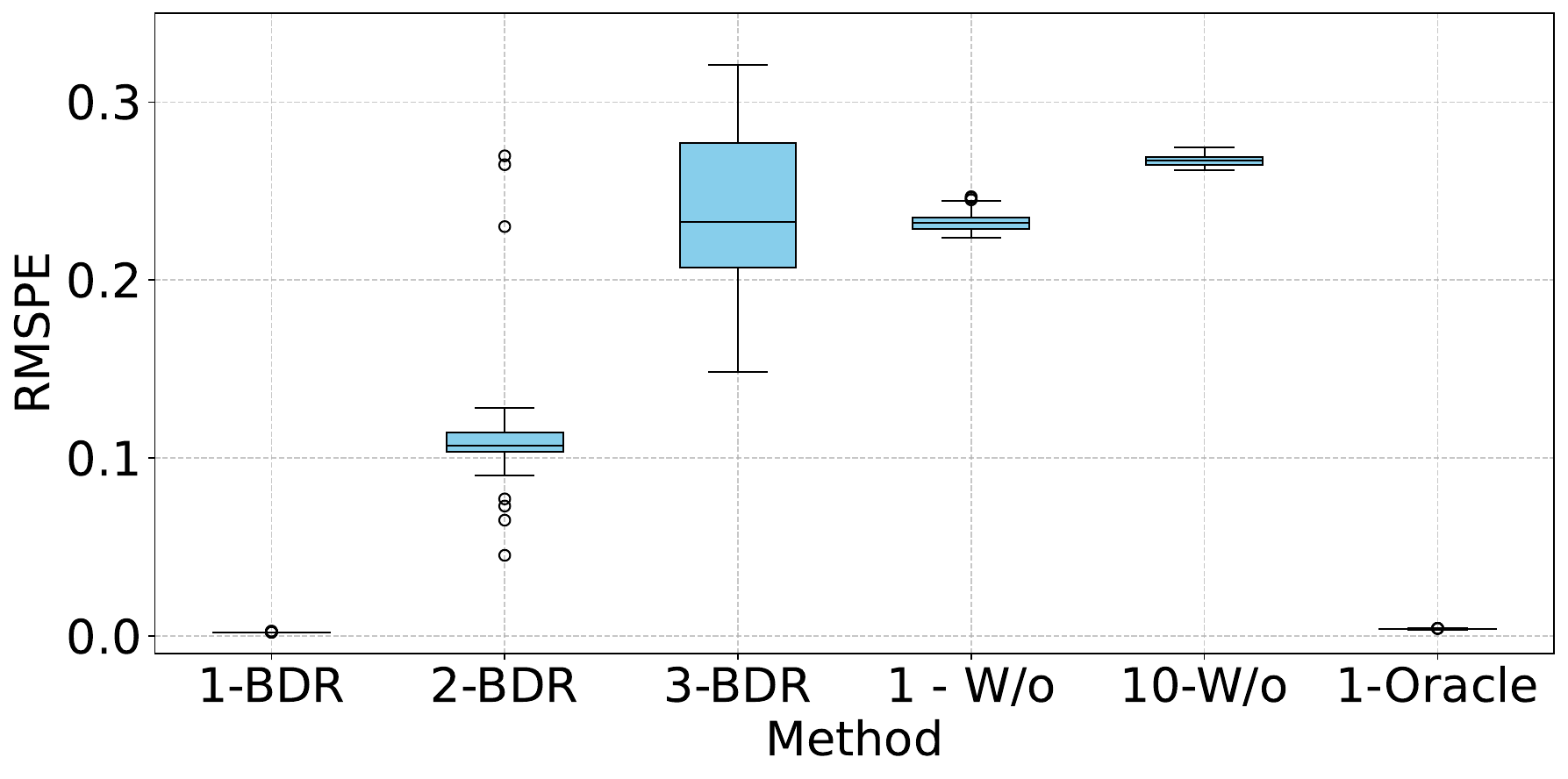}
        \caption{2-layer DGP at $n_{\text{train}}=280$}
        \label{fig:case1_1d_rmspe_280_l2}
    \end{subfigure}
    \begin{subfigure}{0.49\textwidth}
        \centering
        \includegraphics[width=\linewidth]{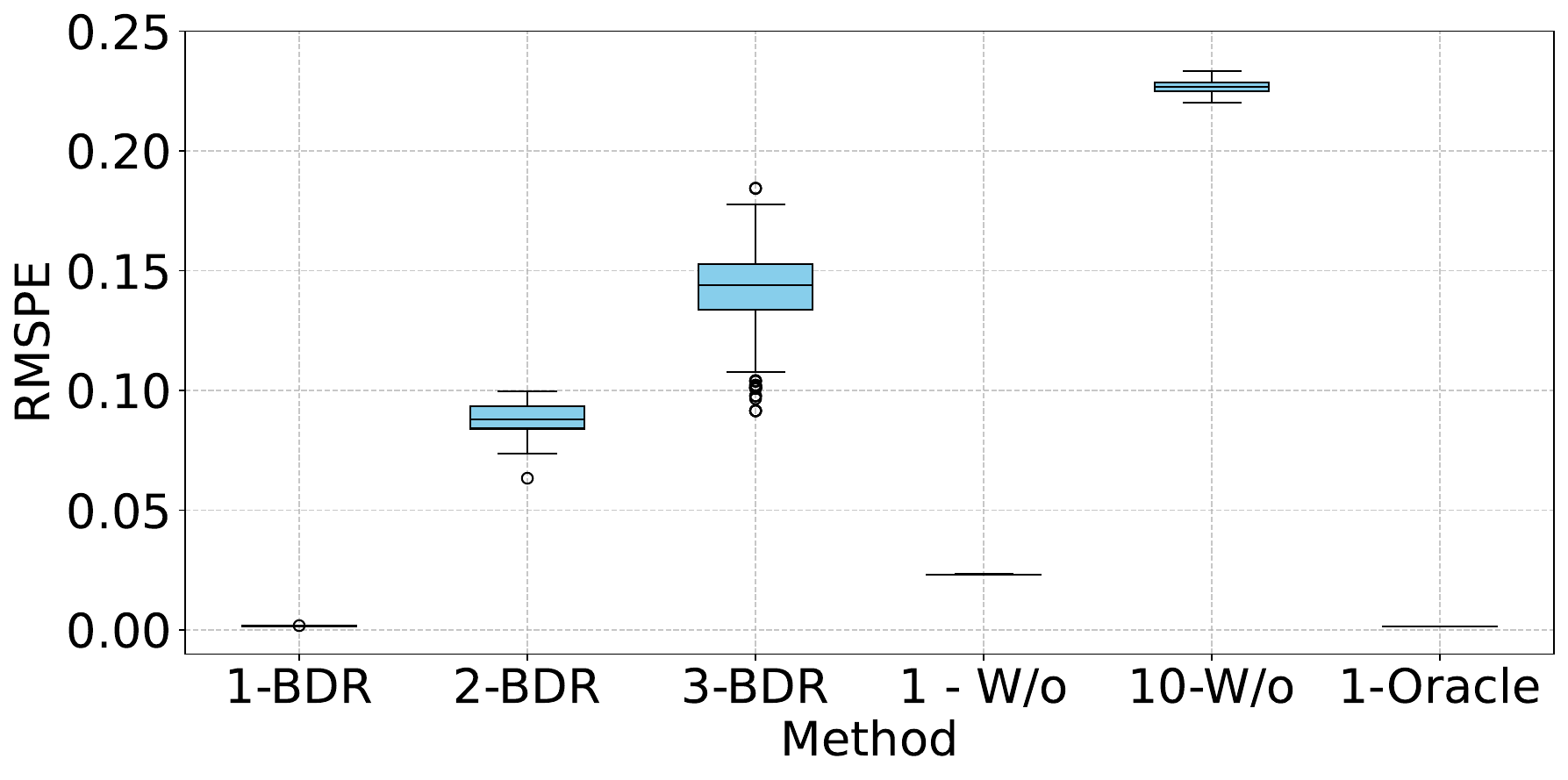}
        \caption{2-layer DGP at $n_{\text{train}}=480$}
        \label{fig:case1_1d_rmspe_480_l2}
    \end{subfigure}
    \hfill
    \begin{subfigure}{0.49\textwidth}
        \centering
        \includegraphics[width=\linewidth]{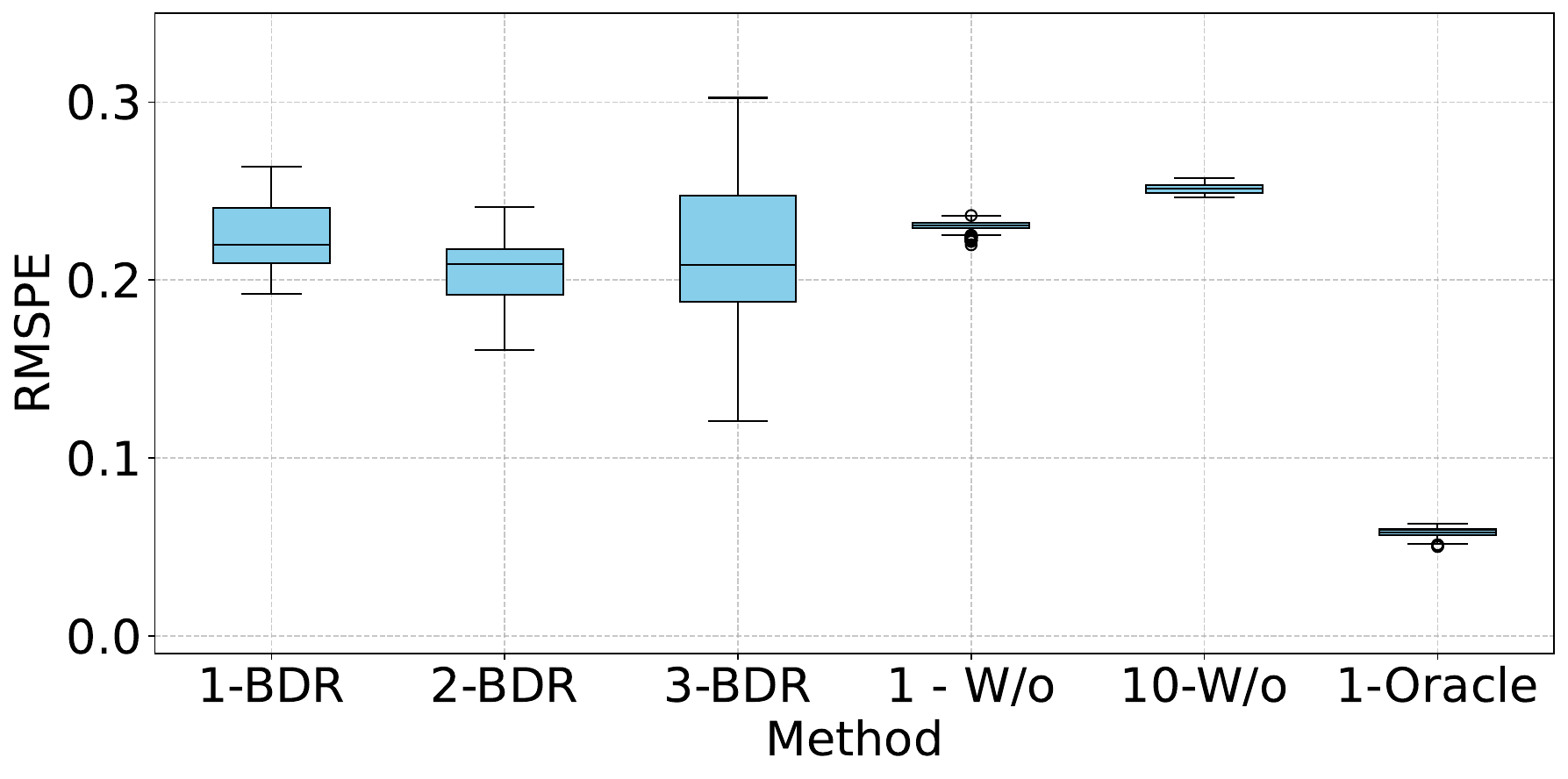}
        \caption{3-layer DGP at $n_{\text{train}}=280$}
        \label{fig:case1_1d_rmspe_280_l3}
    \end{subfigure}
    \hfill
    \begin{subfigure}{0.49\textwidth}
        \centering
        \includegraphics[width=\linewidth]{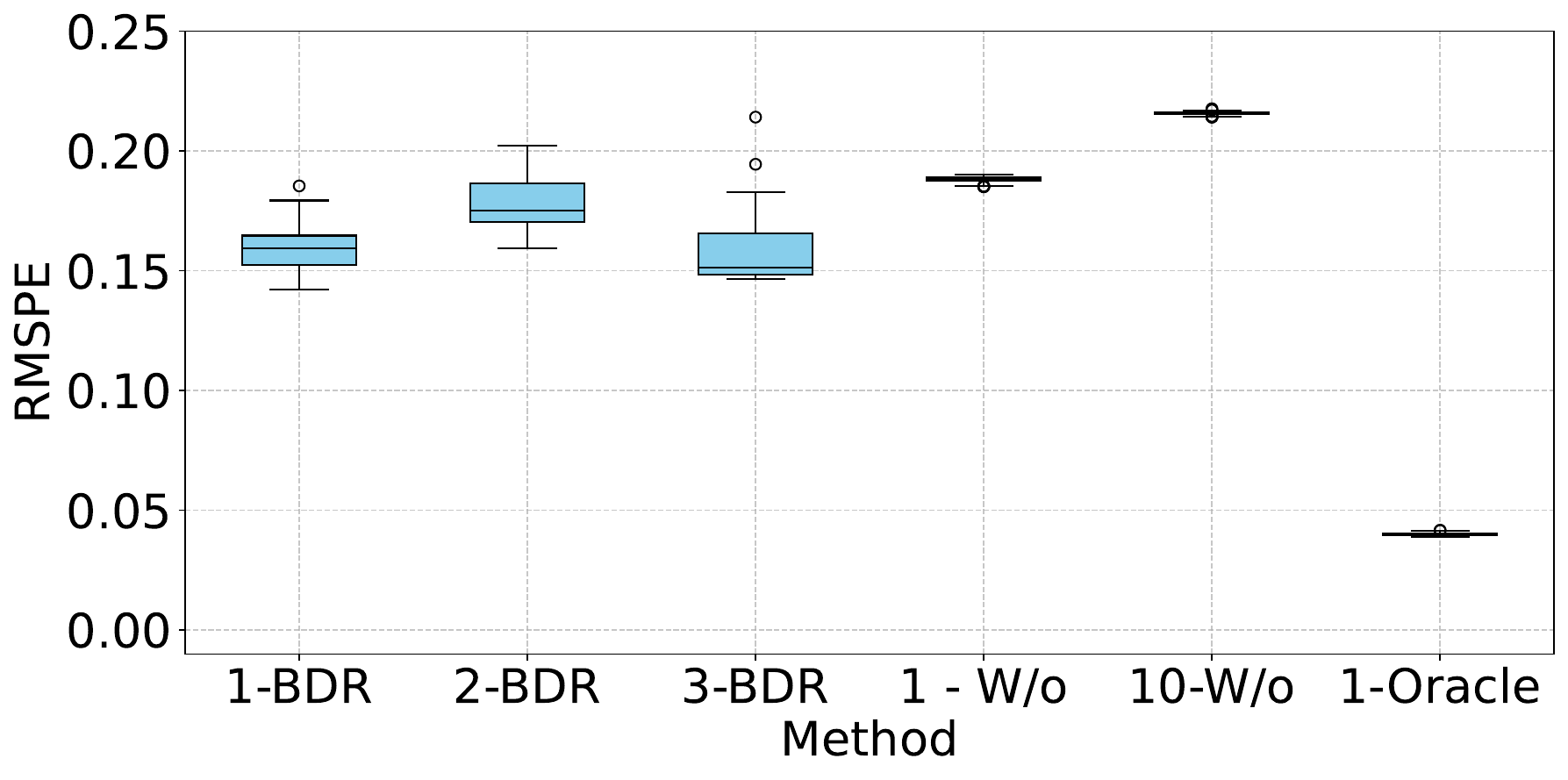}
        \caption{3-layer DGP at $n_{\text{train}}=480$}
        \label{fig:case1_1d_rmspe_480_l3}
    \end{subfigure}
    \caption{RMSPE for 1D input subspace at $n_{\text{train}}=280$ and $480$ based on the response surface of a polynomial function with known structure.}
    \label{fig:case1_1d_rmspe_280_and_480}
\end{figure}

\begin{figure}[h!]
    \centering

    \begin{subfigure}{0.49\textwidth}
        \centering
        \includegraphics[width=\linewidth]{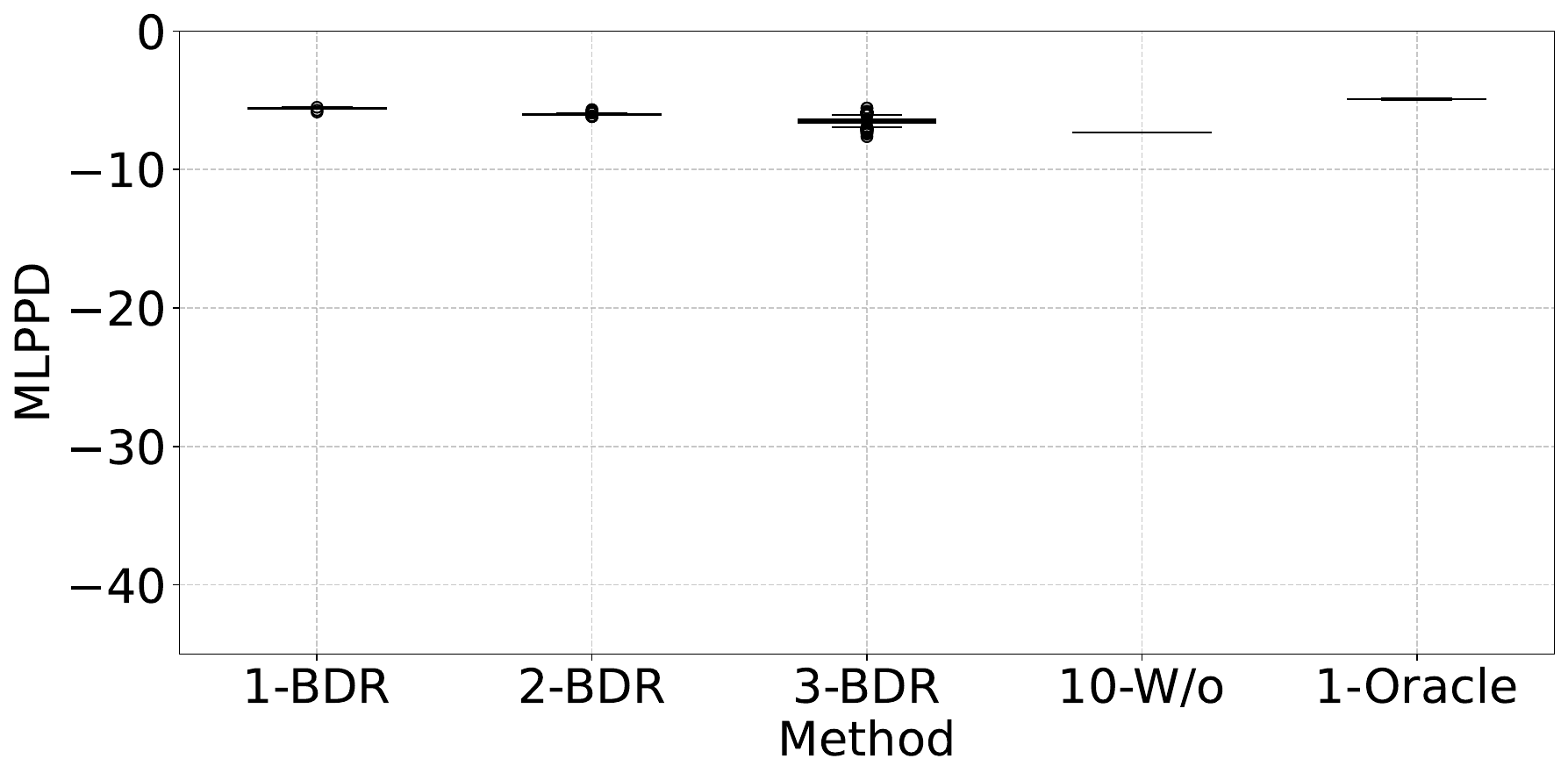}
        \caption{Standard GP at $n_{\text{train}}=280$}
        \label{fig:case1_1d_mlppd_280_l1}
    \end{subfigure}
    \hfill
    \begin{subfigure}{0.49\textwidth}
        \centering
        \includegraphics[width=\linewidth]{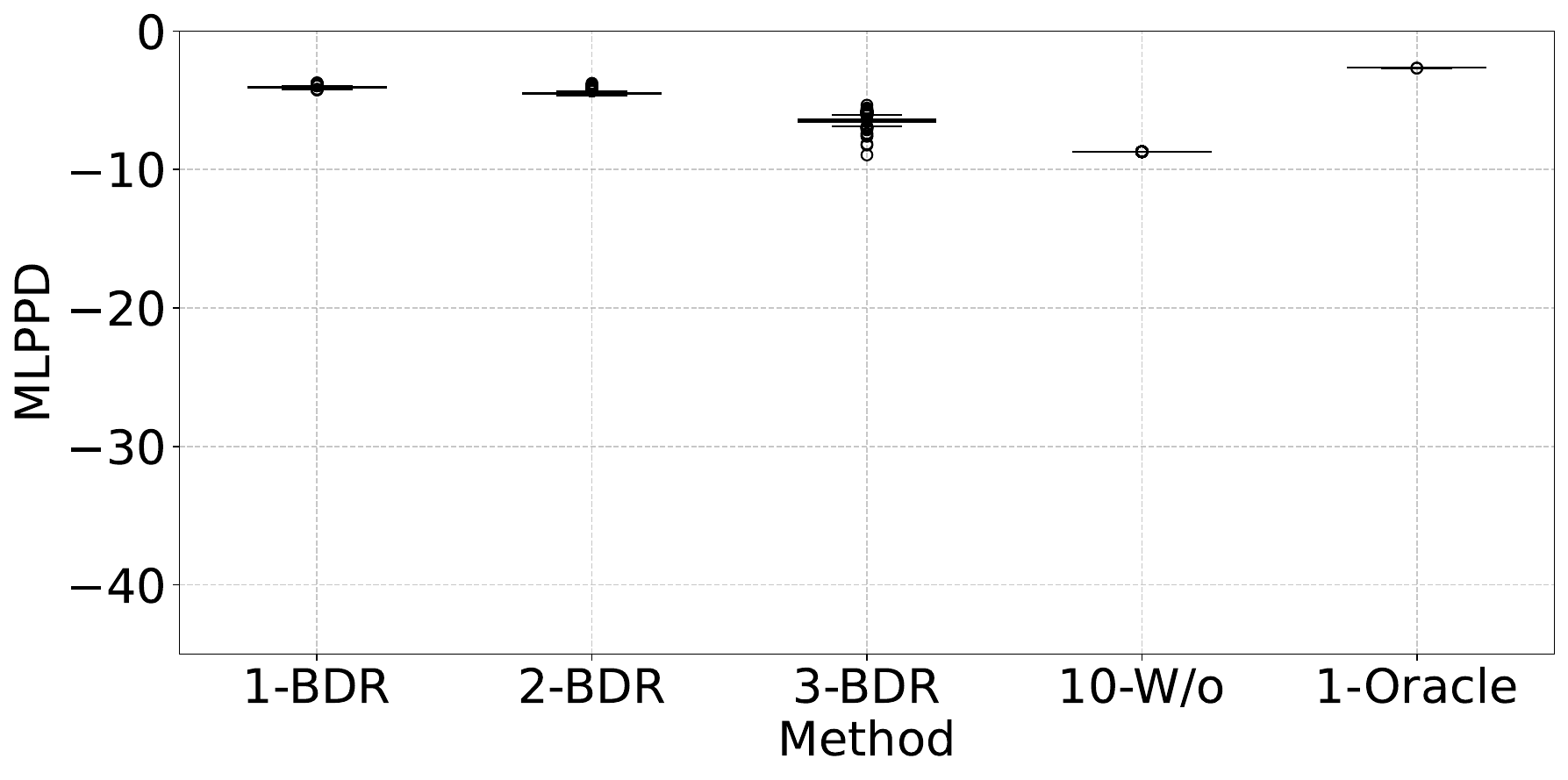}
        \caption{Standard GP at $n_{\text{train}}=480$}
        \label{fig:case1_1d_mlppd_480_l1}
    \end{subfigure}
    \begin{subfigure}{0.49\textwidth}
        \centering
        \includegraphics[width=\linewidth]{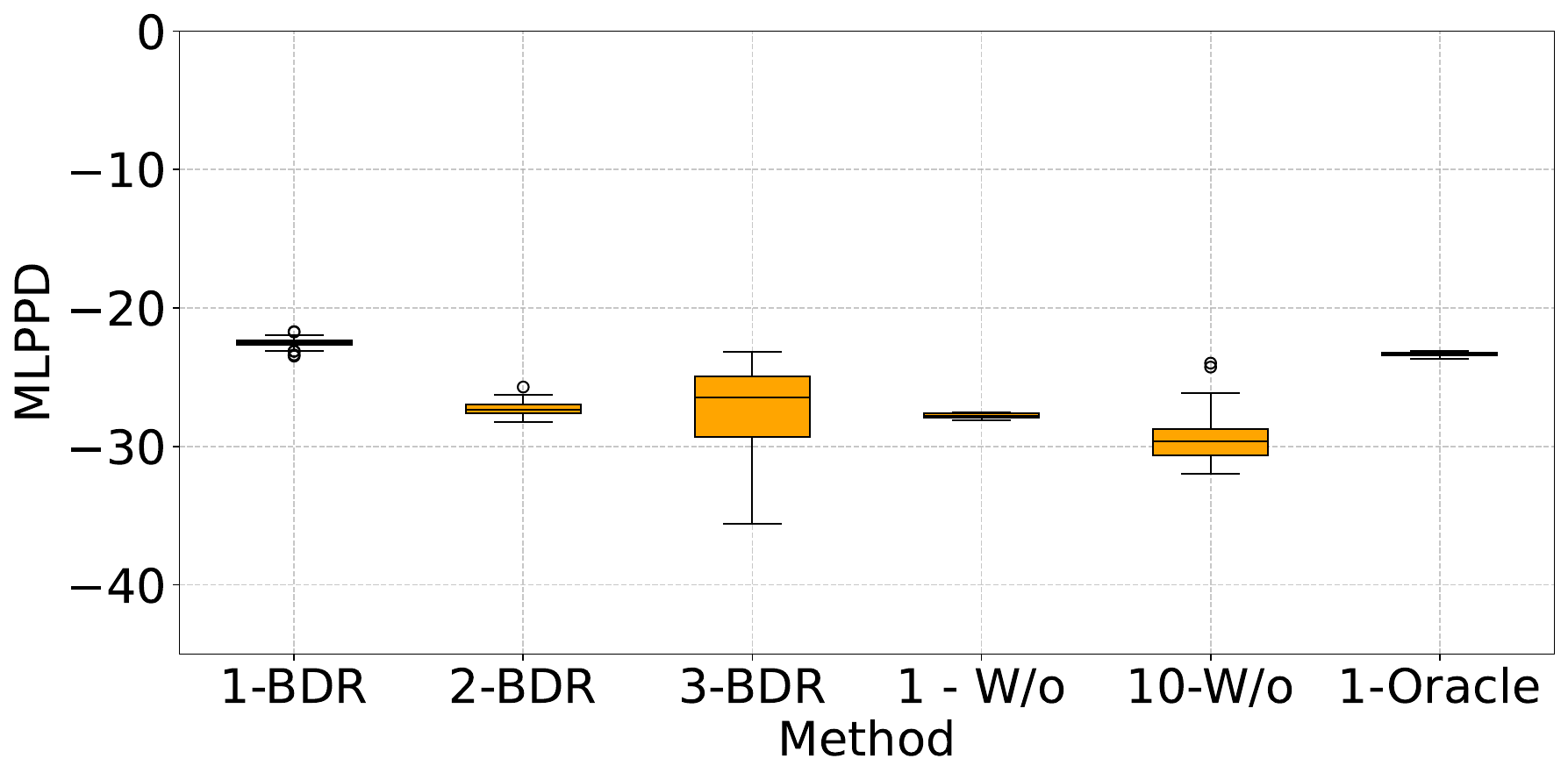}
        \caption{2-layer DGP at $n_{\text{train}}=280$}
        \label{fig:case1_1d_mlppd_280_l2}
    \end{subfigure}
    \hfill
    \begin{subfigure}{0.49\textwidth}
        \centering
        \includegraphics[width=\linewidth]{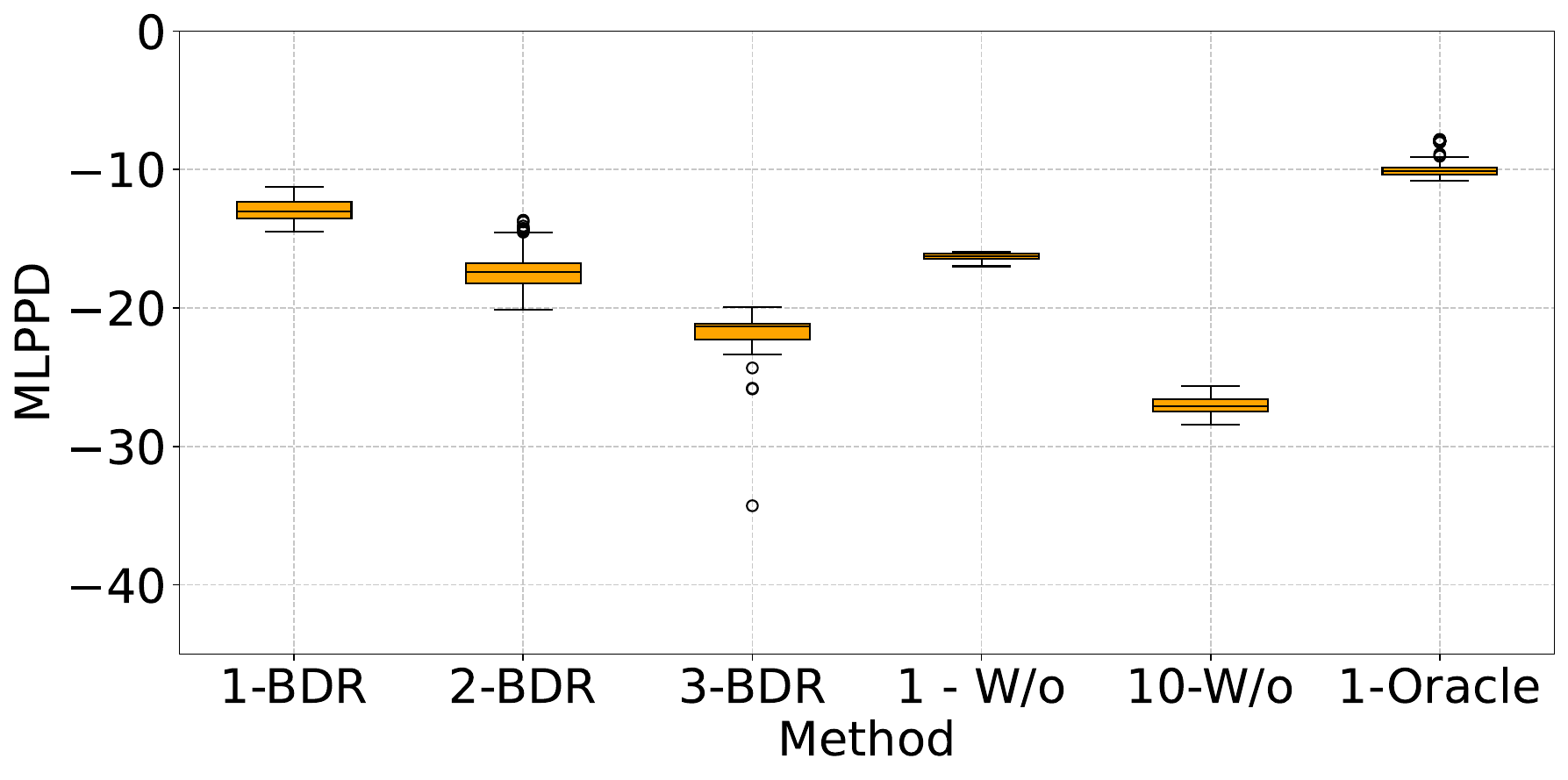}
        \caption{2-layer DGP at $n_{\text{train}}=480$}
        \label{fig:case1_1d_mlppd_480_l2}
    \end{subfigure}
    \begin{subfigure}{0.49\textwidth}
        \centering
        \includegraphics[width=\linewidth]{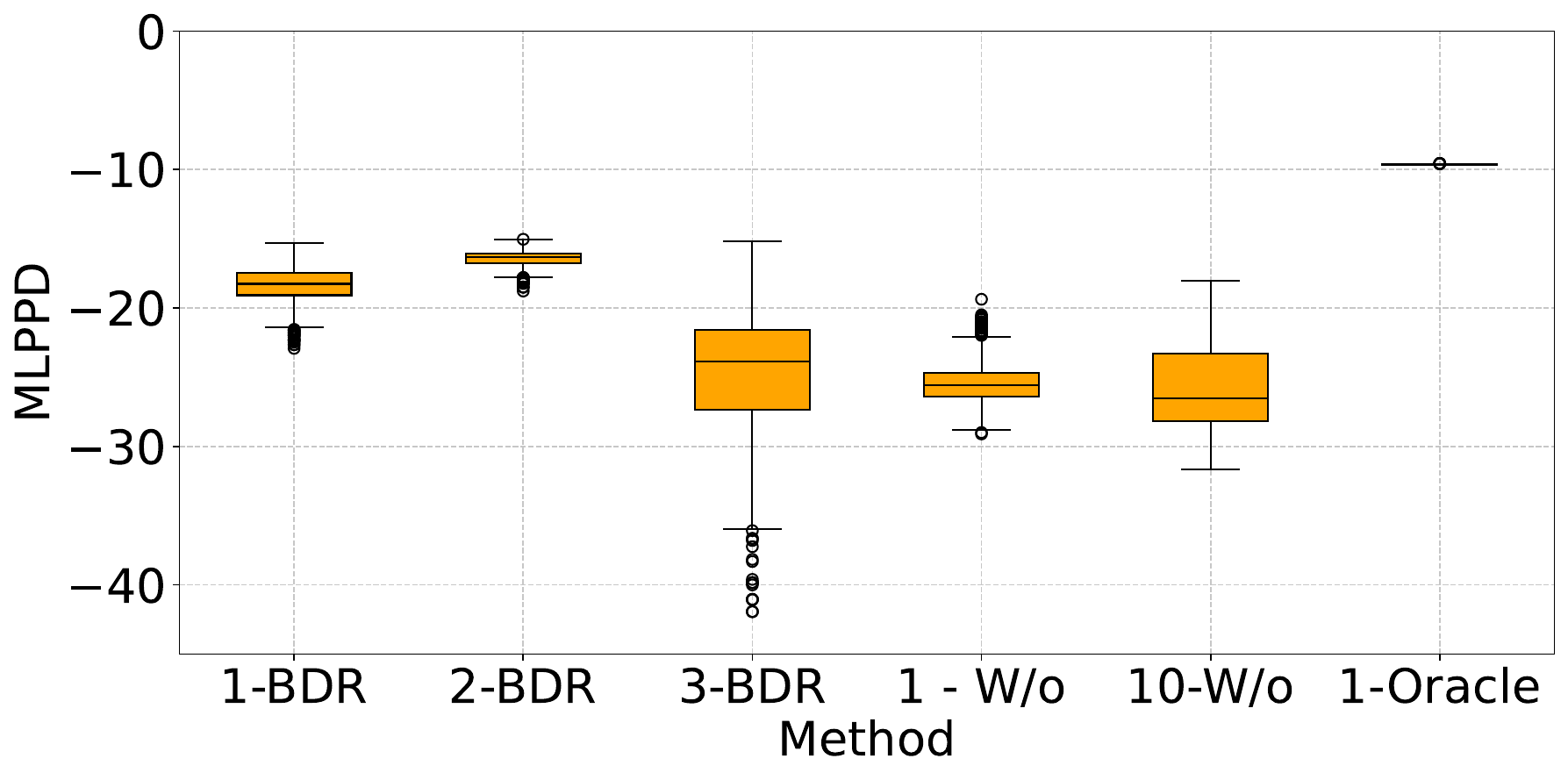}
        \caption{3-layer DGP at $n_{\text{train}}=280$}
        \label{fig:case1_1d_mlppd_280_l3}
    \end{subfigure}
    \begin{subfigure}{0.49\textwidth}
        \centering
        \includegraphics[width=\linewidth]{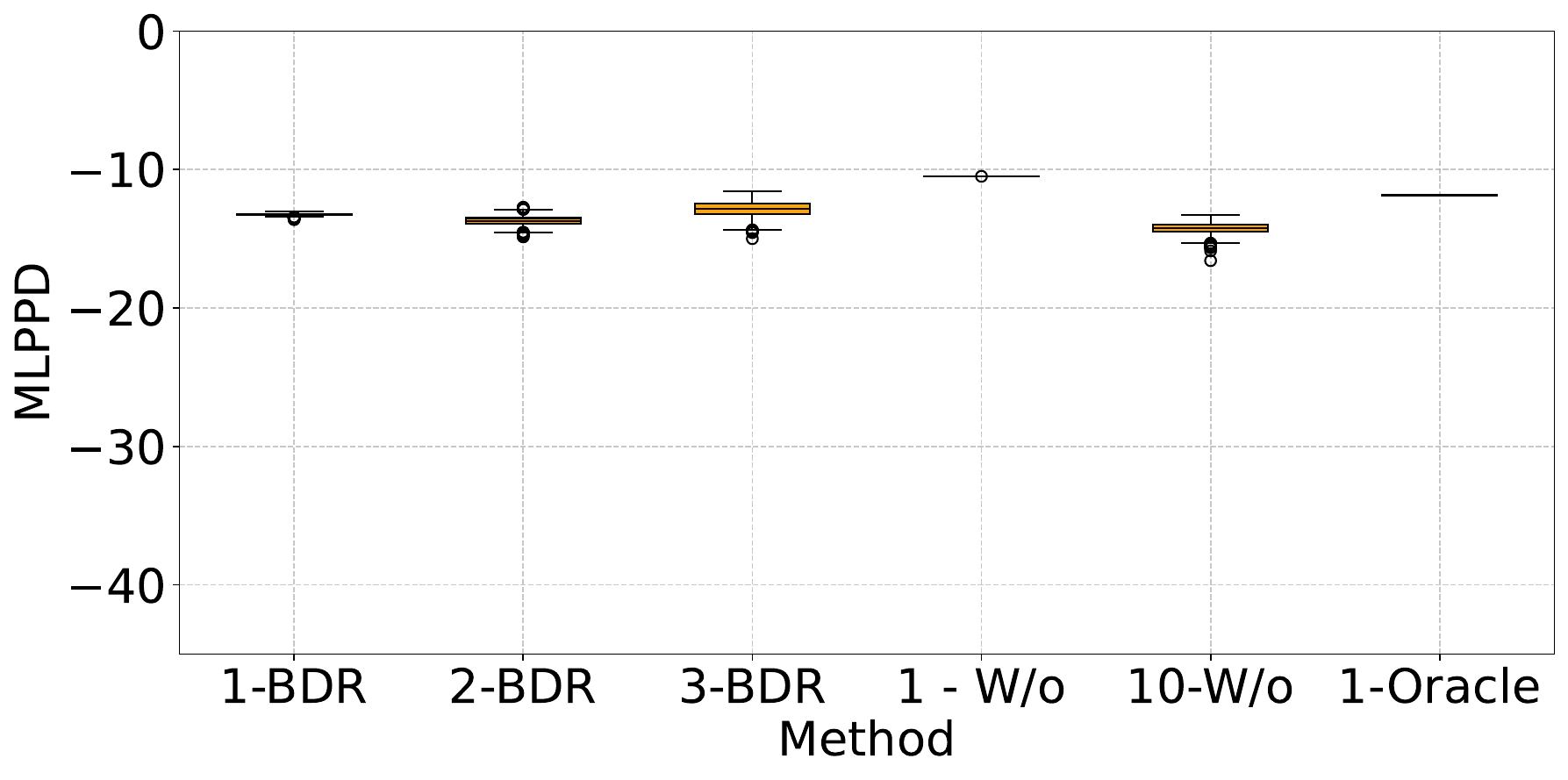}
        \caption{3-layer DGP at $n_{\text{train}}=480$}
        \label{fig:case1_1d_mlppd_480_l3}
    \end{subfigure}
    \caption{MLPPD for 1D input subspace at $n_{\text{train}}=280$ and $480$ based on the response surface of a polynomial function with known structure.}
    \label{fig:case1_1d_mlppd_280_and_480}
\end{figure}

\Cref{tab:case1_1d}, \Cref{fig:case1_1d_rmspe_280_and_480}, and \Cref{fig:case1_1d_mlppd_280_and_480} summarize the performance of the methods. Across both training sizes, the GP(1)–BDR and two-layer DGP(1)–BDR models consistently achieve the best predictive performance. They attain the lowest RMSPE and CRPS values while also producing the highest MLPPD, NSME, and BIC scores, demonstrating that the Bayesian dimension-reduction mechanism successfully identifies the true one-dimensional structure. Increasing the training size from $n_{\text{train}} = 280$ to $480$ further improves accuracy, particularly for the DGP models, which are better able to estimate deeper latent mappings with more data. Models fitted without BDR or with incorrect subspace dimensions ($D>1$) exhibit clear reductions in accuracy and calibration. These performance gaps widen at the larger training size, highlighting the importance of correctly learning $W$ when more information becomes available. Oracle models using the true $W$ provide an upper bound on performance, and the tight agreement between the oracle and BDR-based fits indicates that the proposed method recovers the correct 1D manifold with high fidelity. \textcolor{black}{The CP and ALCI results in \Cref{tab:case1_1dT1} provide additional evidence on uncertainty quantification. For both $n_{\text{train}}=280$ and $480$, the GP(1)-BDR and DGP 2-layer(1)-BDR fits deliver coverage probabilities near the nominal $95\%$ level together with comparatively short credible intervals, showing that they achieve a favorable balance between calibration and interval sharpness. In contrast, models without BDR and models with over-specified subspace dimension tend to exhibit poorer calibration and, in many cases, substantially inflated interval lengths, while the similarity between the BDR and Oracle fits indicates that the learned projection closely matches the true one-dimensional subspace.}


\begin{table}[t]
\centering
\begin{tabular}{|l|c|c|c|c|c|c|c|}
\hline
  \multicolumn{7}{ |c| }{$n=280$ with true W: $10 \times 1$} \\
\hline
Method (D) & TC(mins) &  RMSPE &  NSME & CRPS & Score  & BIC\\
\hline
GP (1) BDR & 534.60 & \textbf{0.0013} & \textit{\textbf{1.0000}}  & 0.1748  & 584.9018 & 160.09\\ 
GP (2) BDR & 700.00  &  0.1055 &  0.9976  &  0.0615  &  192.6912 & 155.03\\
GP (3) BDR & 909.02   &  0.1491 & 0.9944  & 0.6674  & 44.1512 & 152.00 \\
DGP 2-layer (1) BDR & 1529.85 & 0.0019   & \textit{\textbf{1.0000}}   & \textbf{0.0257} & 190.4500 & 159.50\\ 
DGP 2-layer (2) BDR & 2004.60  & 0.1041   & 0.9989 & 0.1882  & \textbf{689.9975} & 156.29\\
DGP 2-layer (3) BDR &  3050.00  &  0.2317 & 0.9894   & 0.5429  & 476.3961 & 150.00 \\
DGP 3-layer (1) BDR & 3300.20   & 0.2133  & 0.9896  & 0.8945  & 138.6220 &149.25\\
DGP 3-layer (2) BDR & 4529.03   & 0.2114  & 0.9899 & 0.6742  &  101.7746 & 150.02 \\
DGP 3-layer (3) BDR & 5900.01   & 0.2116 & 0.9868  & 0.4655 & 656.1235 & 148.27\\
\hline
GP (10) W/o & 630.34   & 0.1907  & 0.9960 & 0.2820   & 138.1613 & 149.84 \\
DGP 2-layer (1) W/o & 1072.08   & 0.2370  & 0.9825 & 0.5852   & 92.2932 & 145.46 \\
DGP 2-layer (10) W/o & 9550.21   & 0.2630  & 0.9738  & 0.9547  & 83.9778 & 147.20\\
DGP 3-layer (1) W/o &  3139.32  & 0.2355  & 0.9820 &  0.6701  & 51.7365 & 143.18 \\
DGP 3-layer (10) W/o &  12800.00  & 0.2483   & 0.9811  & 0.9516  & 58.5189 & 147.86 \\
\hline
GP (1) Oracle & 34.00   & \textit{\textbf{0.0003}}   & \textit{\textbf{1.0000}}  & \textit{\textbf{0.00043}}  & 601.7694 & \textbf{164.37} \\
DGP 2-layer (1) Oracle & 800.26   & 0.0036  & \textit{\textbf{1.0000}} & 0.0253  & 444.8803 & 158.34\\
DGP 3-layer (1) Oracle & 1470.38   & 0.0592   & 0.9438  & 0.6118 & \textbf{638.6674} & 156.80\\
\hline
  \multicolumn{7}{ |c| }{$n=480$ with true W: $10 \times 1$} \\
\hline
GP (1) BDR & 600.04 & \textbf{0.0011}  & \textit{\textbf{1.0000}}  & 0.1628 & 875.7653 & \textbf{195.80}\\
GP (2) BDR & 859.00 & 0.1037  & 0.9979   &  0.0252 & 334.8046 & 193.91\\
GP (3) BDR & 1165.80   & 0.1146  & 0.9963  &  0.6353  & 125.3080 & 190.17  \\
DGP 2-layer (1) BDR & 2390.01  & 0.0017  & 0.9995   & 0.1711  & \textbf{1121.7729} & 194.57 \\ 
DGP 2-layer (2) BDR & 3940.00   & 0.0915  &  0.9981  & \textbf{0.0220}  & 305.4317 & 194.29  \\
DGP 2-layer (3) BDR & 4006.00   & 0.1434  &  0.9957 &  0.5524 & 573.4950 & 185.03\\
DGP 3-layer (1) BDR & 4523.00   & 0.1601  &  0.9781 & 0.8900  &  116.7235 & 180.89\\
DGP 3-layer (2) BDR & 5600.00   & 0.1632  & 0.9767  & 0.6329  & 439.8808 & 184.06\\
DGP 3-layer (3) BDR & 5827.40   & 0.1510  &  0.9803 & 0.5867  & 748.9148  & 183.91 \\
\hline
GP (10) W/o & 745.30   & 0.1790  & 0.9755 & 0.2969  & 370.1816 & 180.18 \\
DGP 2-layer (1) W/o & 2028.26   & 0.0232  & 0.9991 &  0.5030  & 350.2119 &  185.96\\
DGP 2-layer (10) W/o &  11009.20   & 0.2285  & 0.95112  & 0.6179  & 140.4433 & 178.16  \\
DGP 3-layer (1) W/o &  5026.30  & 0.1810  &  0.9730 &  0.4846  & 120.4036 & 173.60 \\
DGP 3-layer (10) W/o &  17370.35  & 0.2167 & 0.9600  & 0.6780  & 155.6789 & 179.91\\
\hline
GP (1) Oracle & 56.03   & \textit{\textbf{0.00019}}  & \textit{\textbf{1.0000}} & \textit{\textbf{0.00045}}  & \textit{\textbf{1102.3938}} & \textit{\textbf{197.45}} \\
DGP 2-layer (1) Oracle &  1000.56  & 0.0015   & 0.9997  & 0.0443  & 1033.1538 & 193.00\\
DGP 3-layer (1) Oracle &  2723.55  & 0.0432  & 0.9989   & 0.8705  & 940.85039 & 186.01\\
\hline
\end{tabular}
\caption{Performance Metrics for 1D input subspace at $n_{train}=280$ and $480$ based on response surface of polynomial function with known structure. Table entries report, for each metric, the median value across posterior samples.}
\label{tab:case1_1d}
\end{table}

\begin{table}[t]
\centering
\begin{tabular}{|l|c|c|c|c|c|c|c|}
\hline
  \multicolumn{3}{ |c| }{$n=280$ with true W: $10 \times 1$} \\
\hline
Method (D) &  CP  &  ALCI (95\%) \\
\hline
GP (1) BDR & 0.9504 & 0.0046\\ 
GP (2) BDR & 0.9676 & 0.1972\\
GP (3) BDR & 0.8429 & 0.2464\\
DGP 2-layer (1) BDR & 0.9550 &  0.0280\\ 
DGP 2-layer (2) BDR & 0.9714 & 0.9759\\
DGP 2-layer (3) BDR & 0.9857 & 2.0619\\
DGP 3-layer (1) BDR & 0.9415 & 0.3357\\
DGP 3-layer (2) BDR & 0.8143 & 1.7848\\
DGP 3-layer (3) BDR & 1.0000 & 3.0183\\
\hline
GP (10) W/o  & 0.2790 & 0.0105 \\
DGP 2-layer (1) W/o & 0.3049 & 0.0245 \\
DGP 2-layer (10) W/o & 0.1000 & 1.2309\\
DGP 3-layer (1) W/o & 0.3810 & 0.9520\\
DGP 3-layer (10) W/o & 0.1285 & 0.8838\\
\hline
GP (1) Oracle & 0.9143 & 0.0017\\
DGP 2-layer (1) Oracle & 0.9557 & 0.0845\\
DGP 3-layer (1) Oracle & 0.9466 & 0.6401\\
\hline
  \multicolumn{3}{ |c| }{$n=480$ with true W: $10 \times 1$} \\
\hline
GP (1) BDR & 0.9529 & 0.0038\\ 
GP (2) BDR & 0.9702 & 0.1563\\
GP (3) BDR & 0.8830 & 0.2022 \\
DGP 2-layer (1) BDR & 0.9883 & 0.0204\\ 
DGP 2-layer (2) BDR & 0.9823 & 0.4737 \\
DGP 2-layer (3) BDR & 0.9333  & 1.5146\\
DGP 3-layer (1) BDR & 0.9750 & 0.3159\\
DGP 3-layer (2) BDR & 0.9414 & 0.9413\\
DGP 3-layer (3) BDR & 0.9083 & 2.0759\\
\hline
GP (10) W/o  & 0.3400 & 0.0121 \\
DGP 2-layer (1) W/o & 0.3240 & 0.0208\\
DGP 2-layer (10) W/o & 0.2705 & 0.4457\\
DGP 3-layer (1) W/o & 0.4022 & 0.6109\\
DGP 3-layer (10) W/o & 0.3050 & 0.3467\\
\hline
GP (1) Oracle & 0.9416 & 0.0014\\
DGP 2-layer (1) Oracle & 0.9580 & 0.0755\\
DGP 3-layer (1) Oracle & 0.9667 & 0.3289\\
\hline
\end{tabular}
\caption{Performance Metrics for 1D input subspace at $n_{train}=280$ and $480$ based on response surface of polynomial function with known structure. Table entries report, for each metric, the median value across posterior samples.}
\label{tab:case1_1dT1}
\end{table}

\subsubsection{2D Input Subspace}
\label{subsubsec:case1_2D}

For the two-dimensional input subspace, the true $10 \times 2$ projection matrix $\bW$ is
{\scriptsize
\[
\bW =
\begin{pmatrix}
0.00840 & -0.18426 \\
0.34300 & -0.05347 \\
0.08108 & 0.06556 \\
-0.41219 & 0.65424 \\
0.48483 & 0.03966 \\
0.06720 & -0.41480 \\
0.48210 & 0.07550 \\
0.21010 & 0.53750 \\
0.07810 & -0.20020 \\
-0.29120 & 0.34800
\end{pmatrix}.
\]
}

The remaining components of $\eta(\cdot)$ are
\[
\mathbf{a_0} = -0.06976, 
\qquad
\ba = (0.4376,\, 0.9870)^\top,
\qquad
\bA =
\begin{bmatrix}
-0.9257 & -0.3840 \\
-0.4174 & -0.6766
\end{bmatrix},
\qquad
\sigma^2_\epsilon = 0.01.
\]

\begin{figure}[h!]
    \centering

    \begin{subfigure}{0.49\textwidth}
        \centering
        \includegraphics[width=\linewidth]{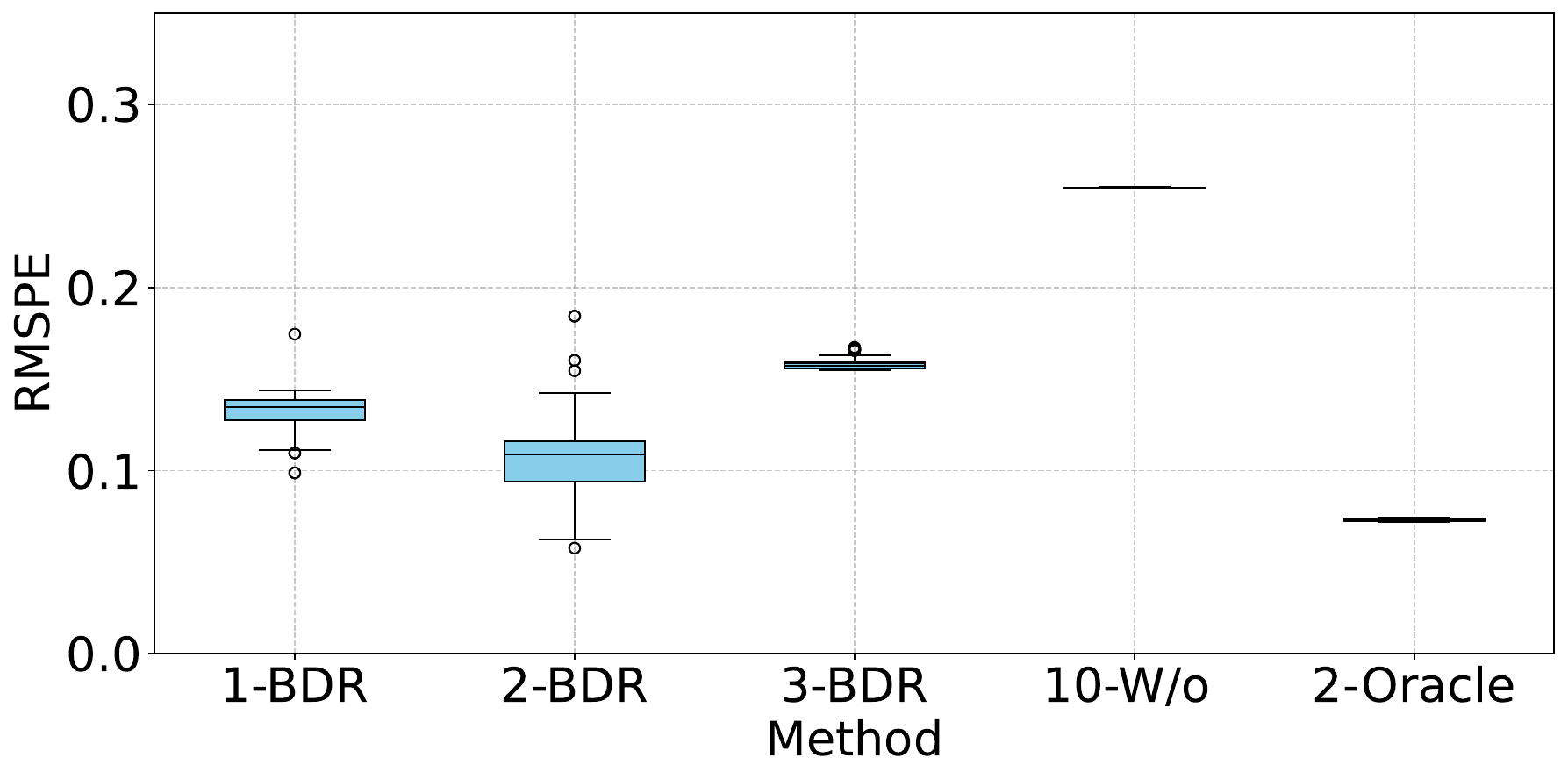}
        \caption{Standard GP at $n_{\text{train}}=280$}
        \label{fig:case1_2d_rmspe_280_l1}
    \end{subfigure}
    \hfill
    \begin{subfigure}{0.49\textwidth}
        \centering
        \includegraphics[width=\linewidth]{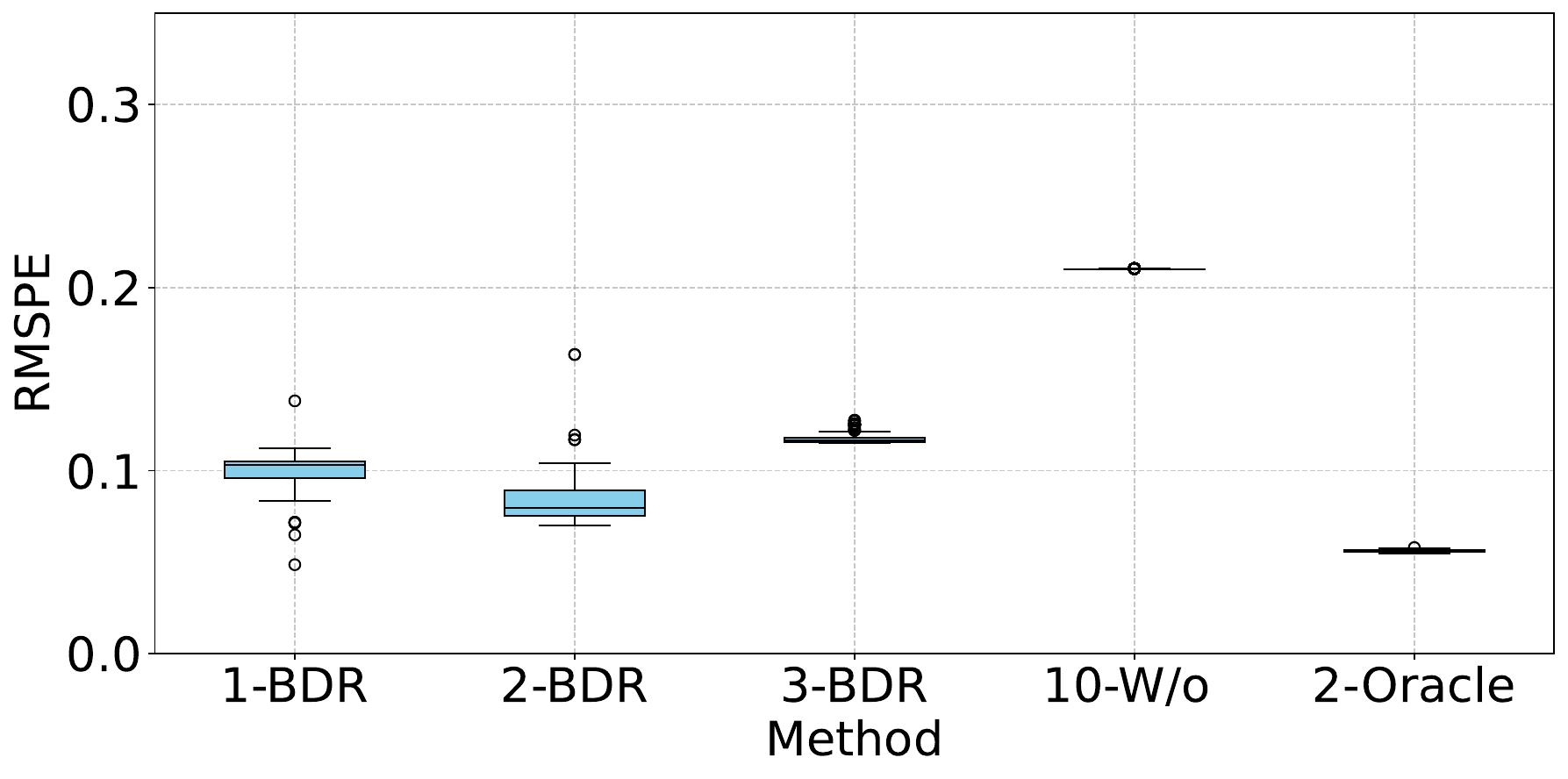}
        \caption{Standard GP at $n_{\text{train}}=480$}
        \label{fig:case1_2d_rmspe_480_l1}
    \end{subfigure}
    \begin{subfigure}{0.49\textwidth}
        \centering
        \includegraphics[width=\linewidth]{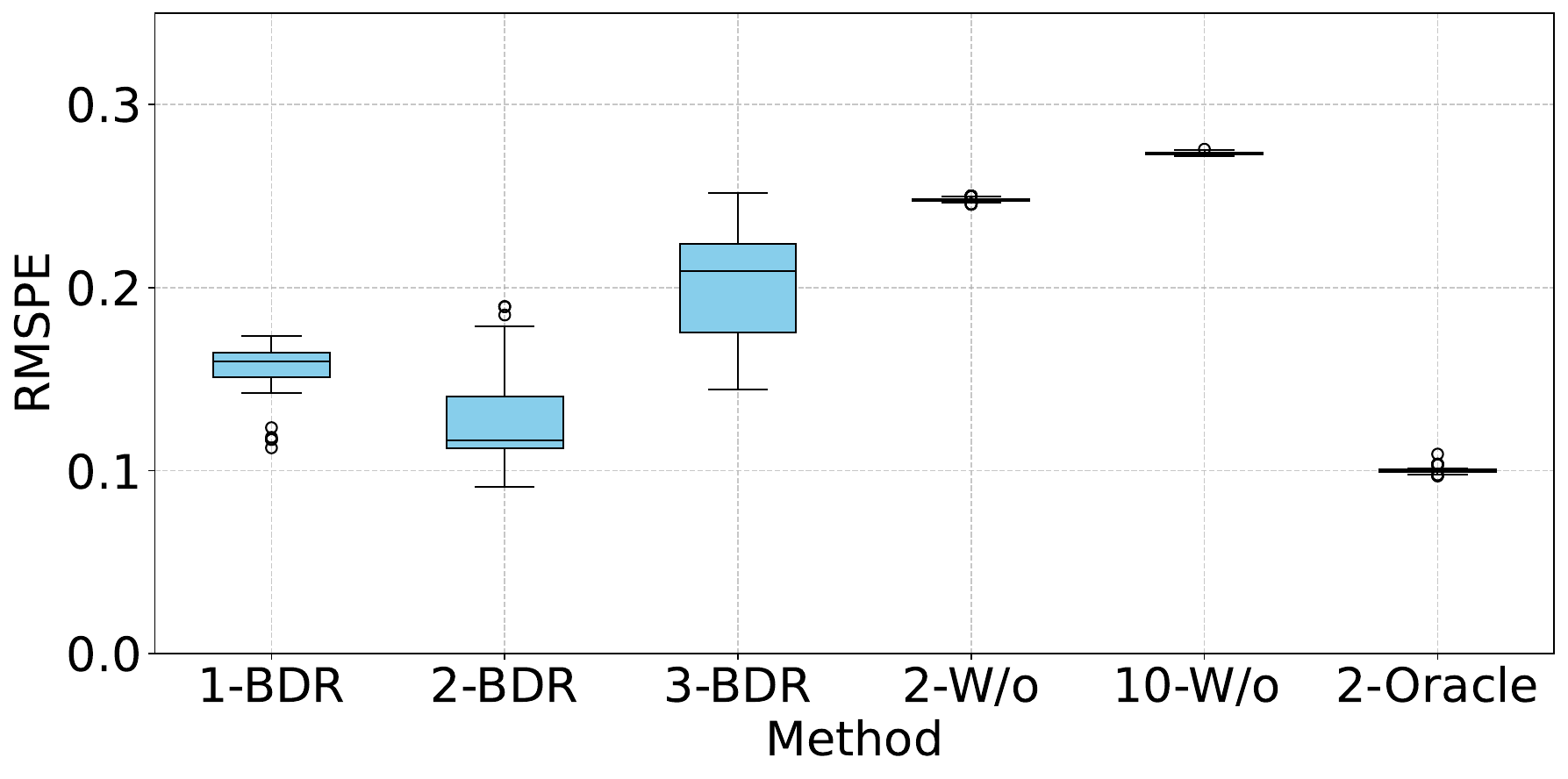}
        \caption{2-layer DGP at $n_{\text{train}}=280$}
        \label{fig:case1_2d_rmspe_280_l2}
    \end{subfigure}
    \hfill
    \begin{subfigure}{0.49\textwidth}
        \centering
        \includegraphics[width=\linewidth]{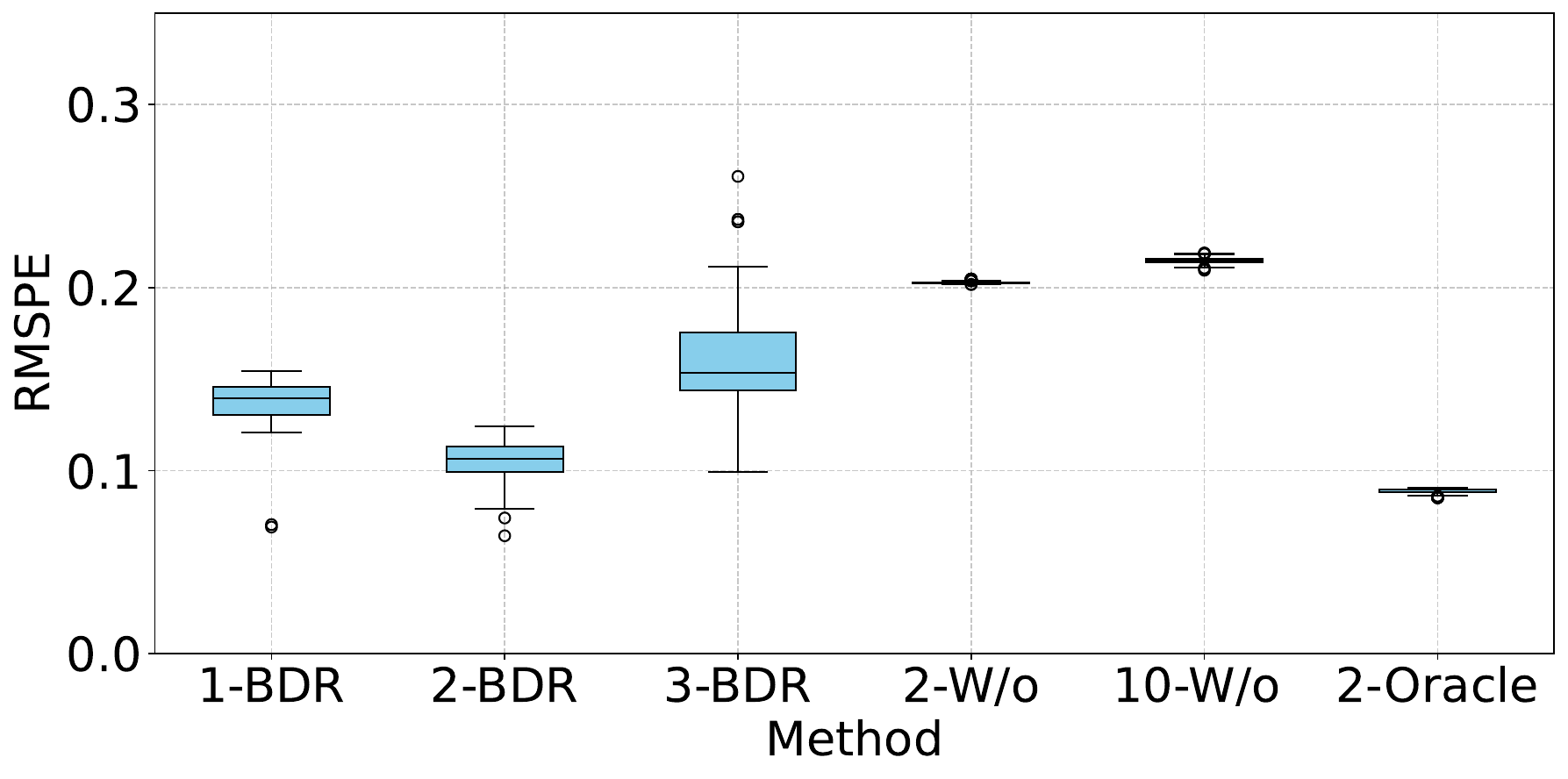}
        \caption{2-layer DGP at $n_{\text{train}}=480$}
        \label{fig:case1_2d_rmspe_480_l2}
    \end{subfigure}
    \begin{subfigure}{0.49\textwidth}
        \centering
        \includegraphics[width=\linewidth]{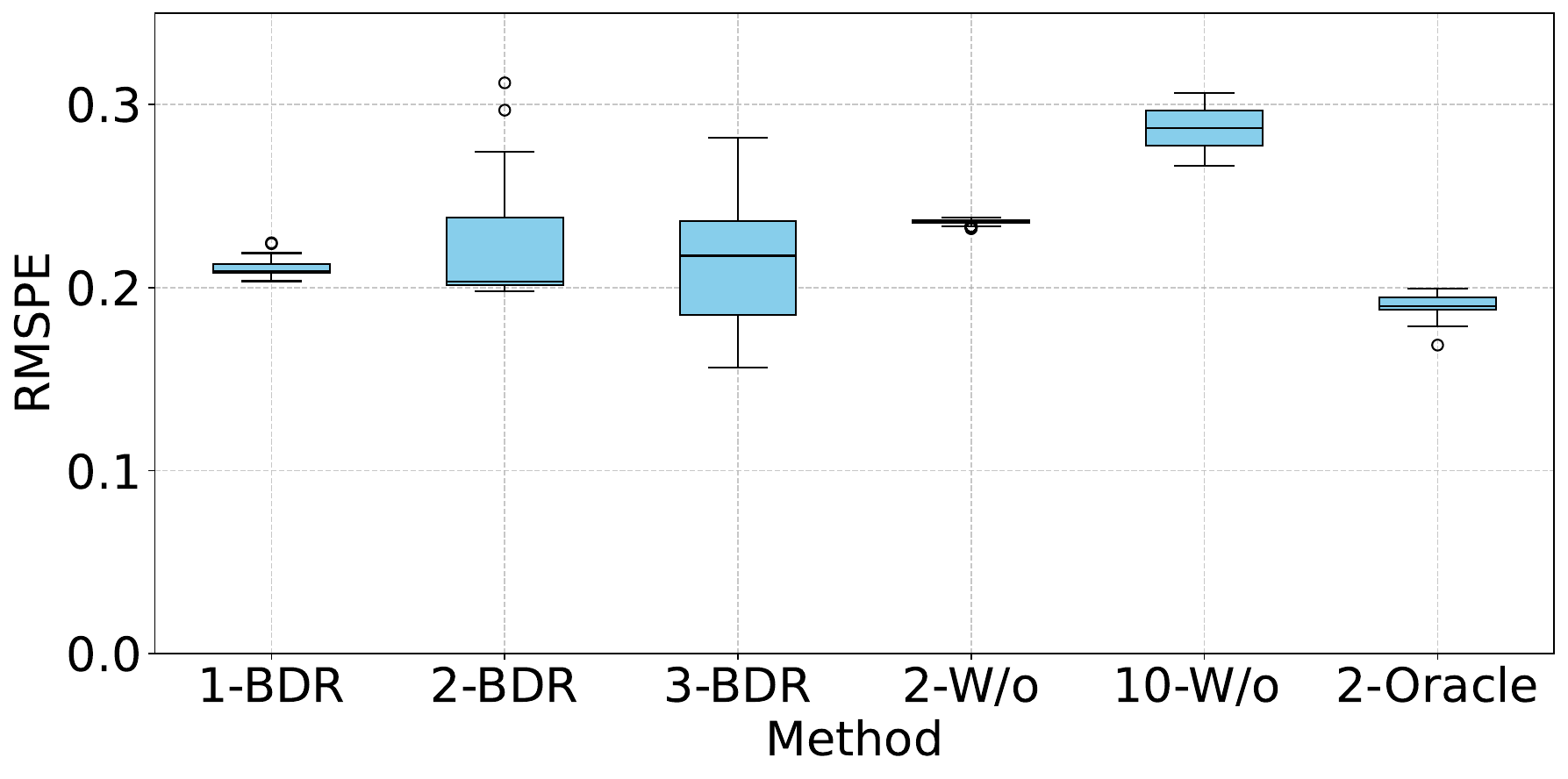}
        \caption{3-layer DGP at $n_{\text{train}}=280$}
        \label{fig:case1_2d_rmspe_280_l3}
    \end{subfigure}
    \begin{subfigure}{0.49\textwidth}
        \centering
        \includegraphics[width=\linewidth]{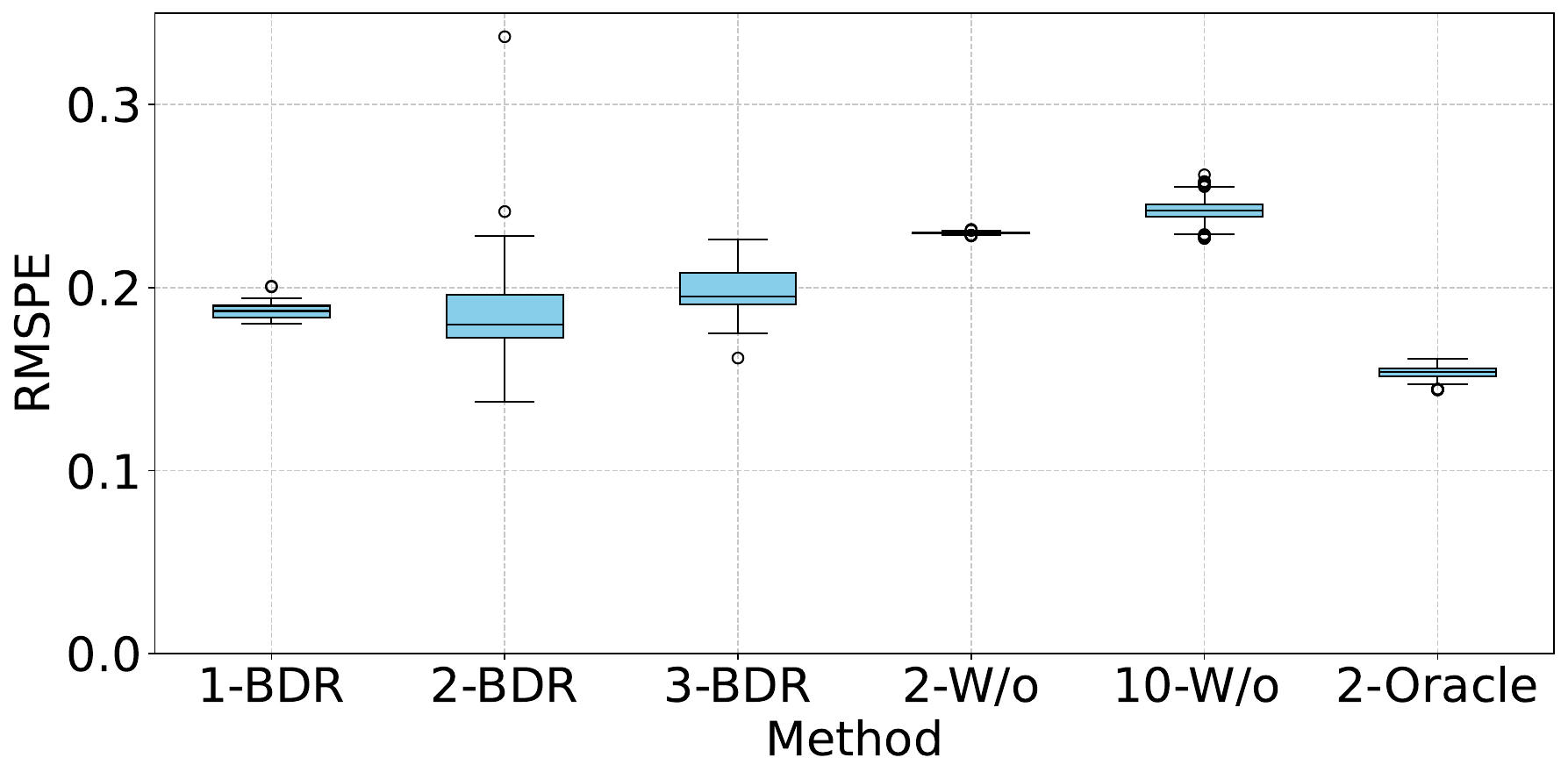}
        \caption{3-layer DGP at $n_{\text{train}}=480$}
        \label{fig:case1_2d_rmspe_480_l3}
    \end{subfigure}
    \caption{RMSPE for 2D input subspace at $n_{\text{train}}=280$ and $480$ based on the response surface of a polynomial function with known structure.}
    \label{fig:case1_2d_rmspe_280_and_480}
\end{figure}

\begin{figure}[h!]
    \centering

    \begin{subfigure}{0.49\textwidth}
        \centering
        \includegraphics[width=\linewidth]{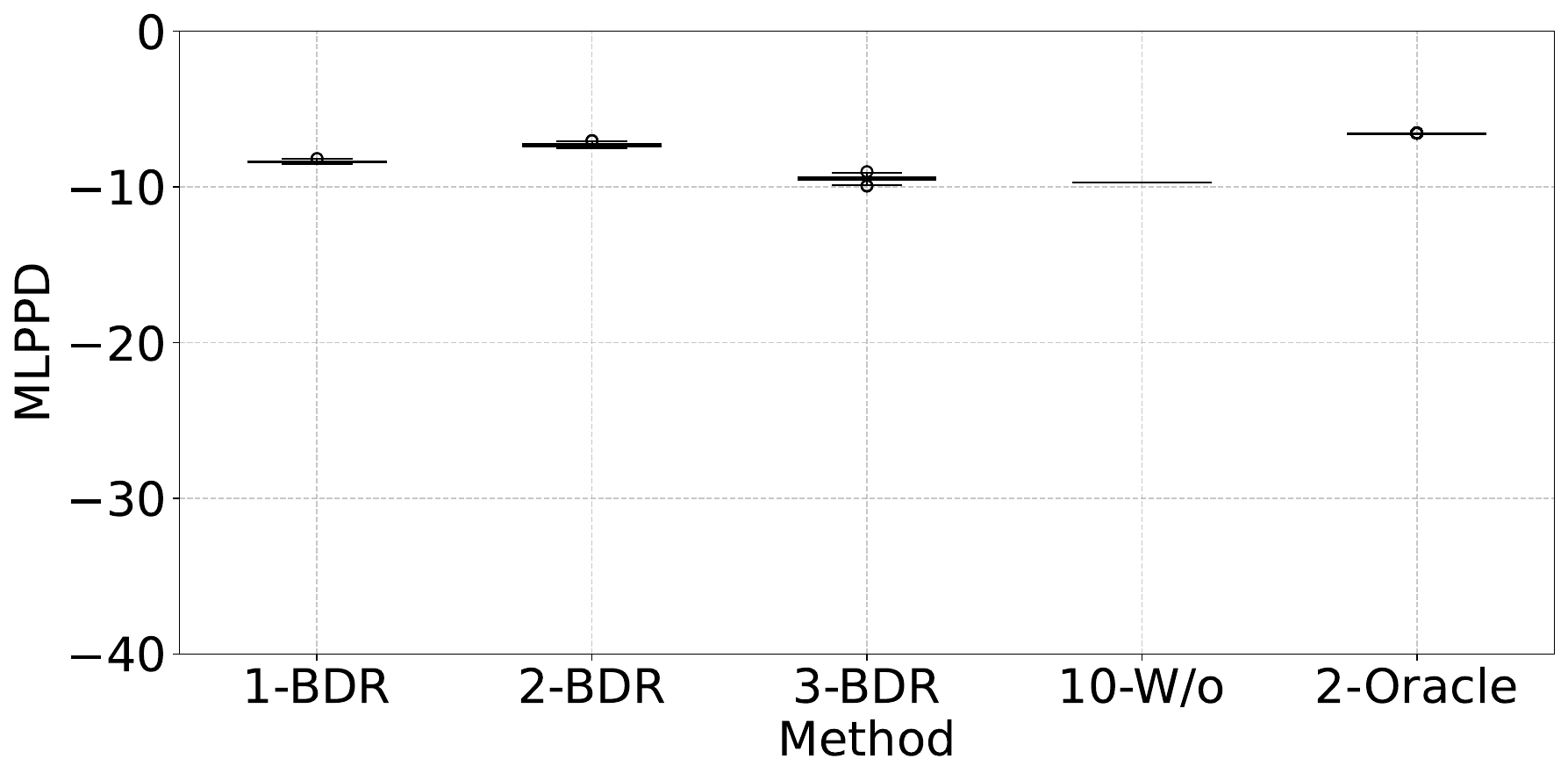}
        \caption{Standard GP}
        \label{fig:case1_2d_mlppd_280_l1}
    \end{subfigure}
    \hfill
    \begin{subfigure}{0.49\textwidth}
        \centering
        \includegraphics[width=\linewidth]{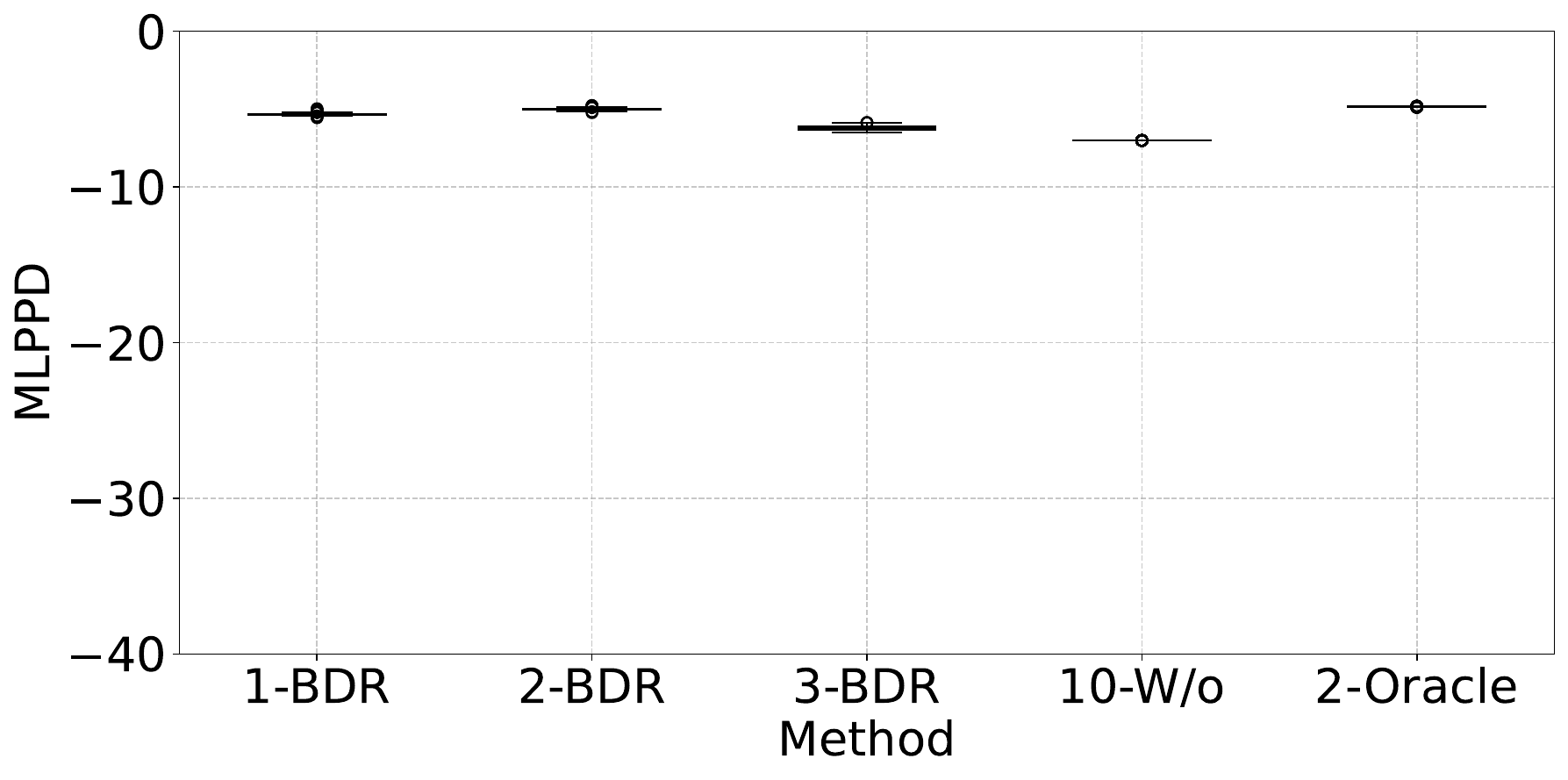}
        \caption{Standard GP}
        \label{fig:case1_2d_mlppd_480_l1}
    \end{subfigure}
    \begin{subfigure}{0.49\textwidth}
        \centering
        \includegraphics[width=\linewidth]{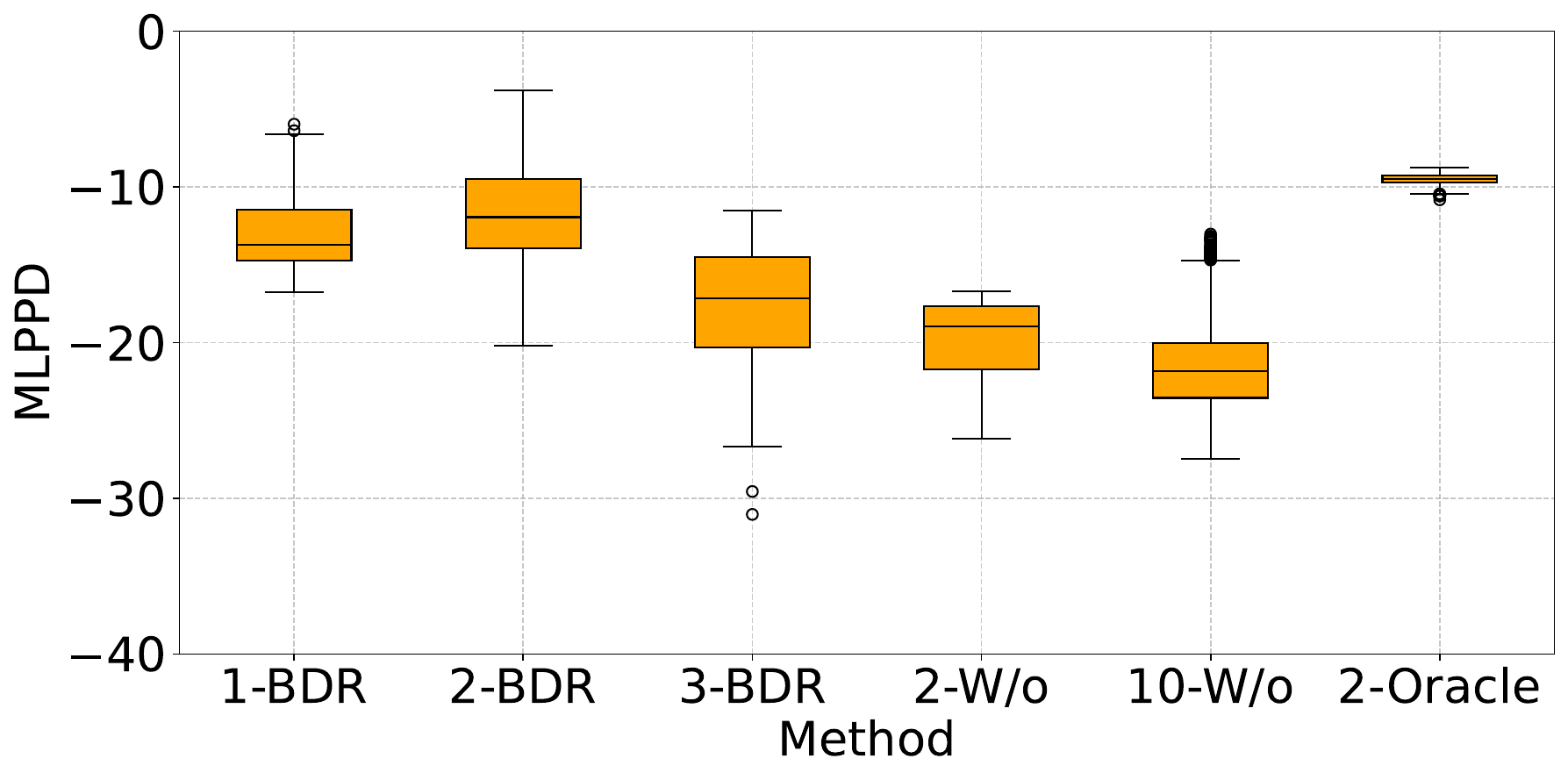}
        \caption{2-layer DGP}
        \label{fig:case1_2d_mlppd_280_l2}
    \end{subfigure}
    \hfill
    \begin{subfigure}{0.49\textwidth}
        \centering
        \includegraphics[width=\linewidth]{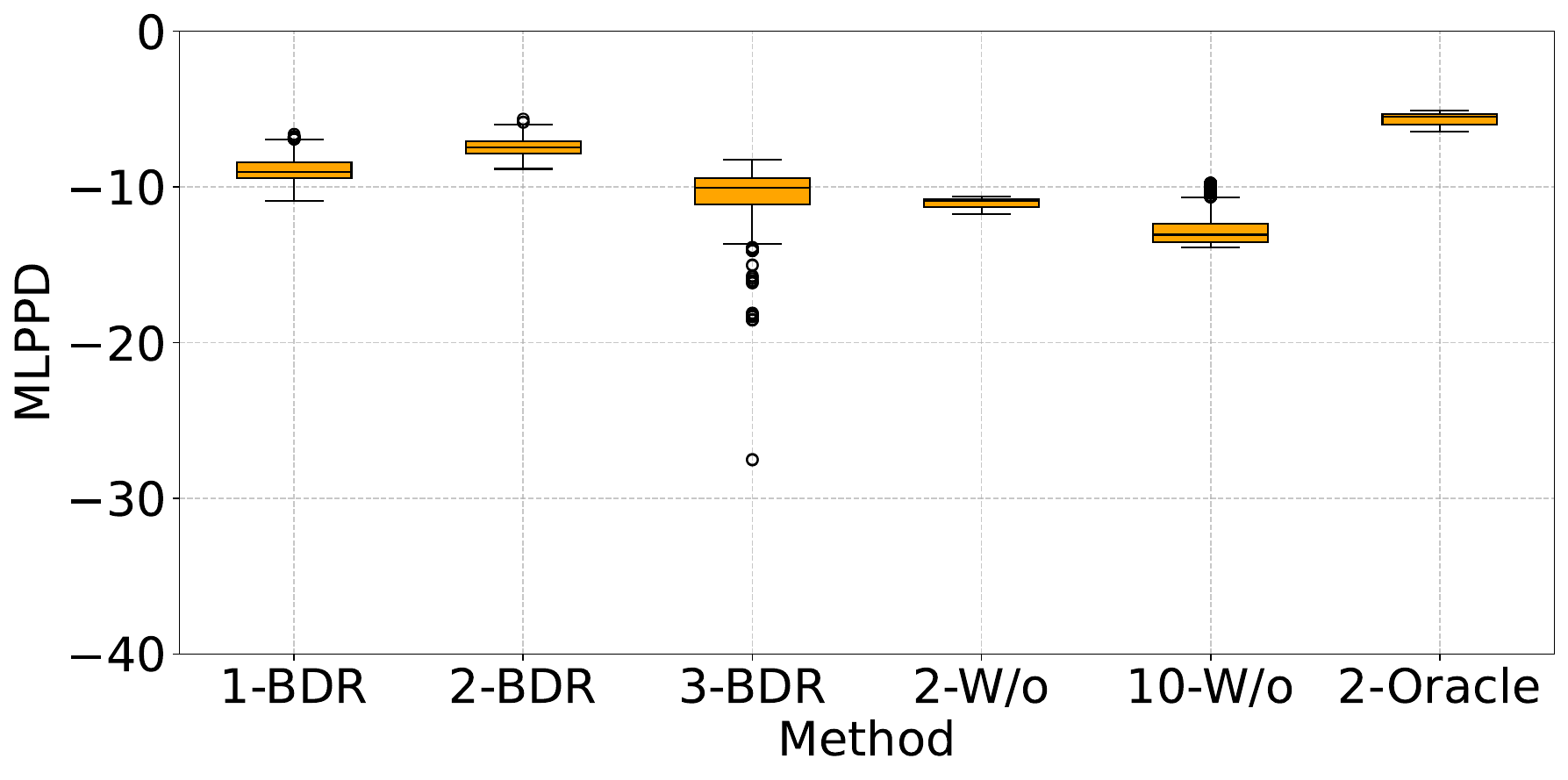}
        \caption{2-layer DGP}
        \label{fig:case1_2d_mlppd_480_l2}
    \end{subfigure}
    \begin{subfigure}{0.49\textwidth}
        \centering
        \includegraphics[width=\linewidth]{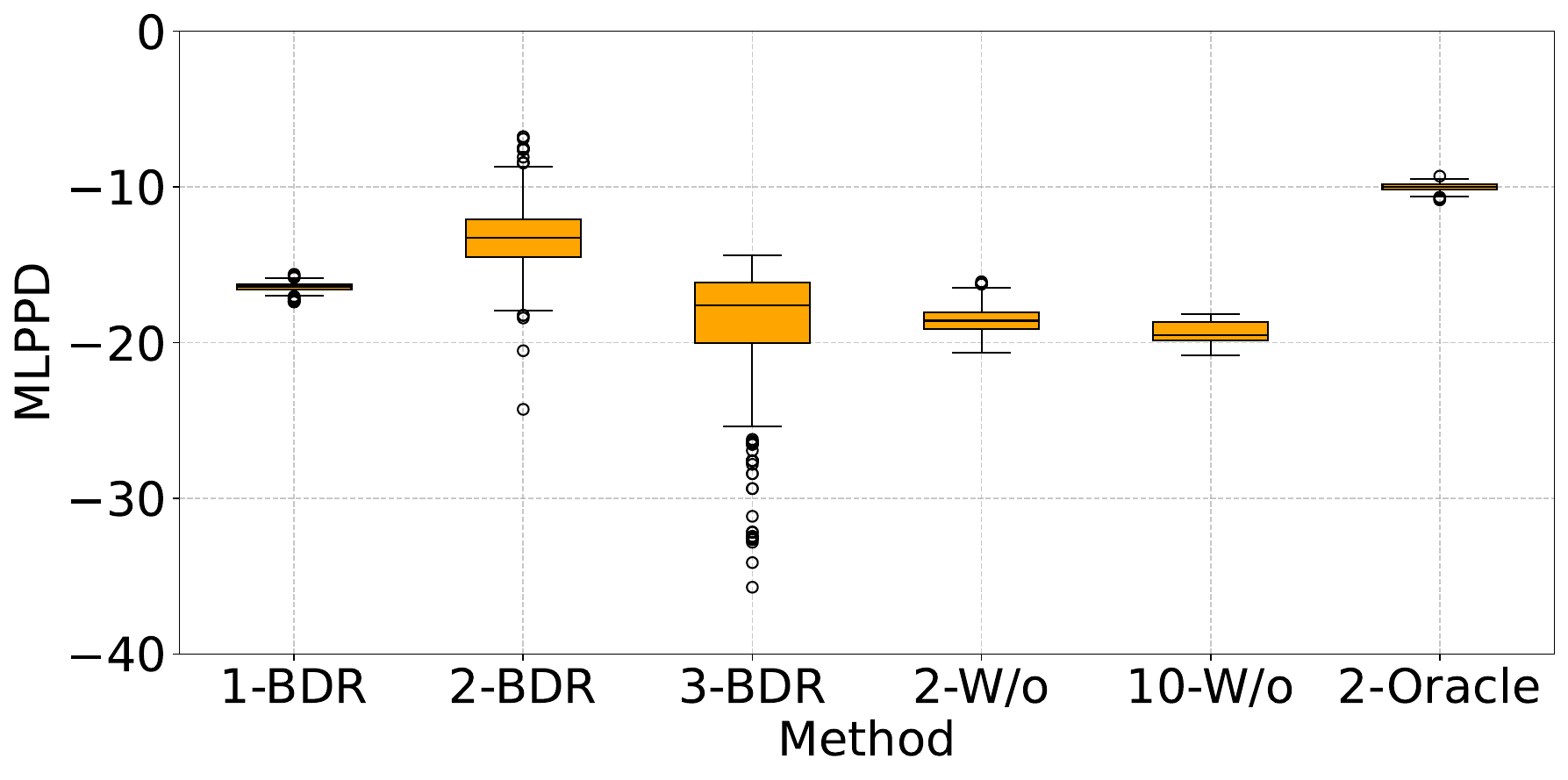}
        \caption{3-layer DGP}
        \label{fig:case1_2d_mlppd_280_l3}
    \end{subfigure}
    \begin{subfigure}{0.49\textwidth}
        \centering
        \includegraphics[width=\linewidth]{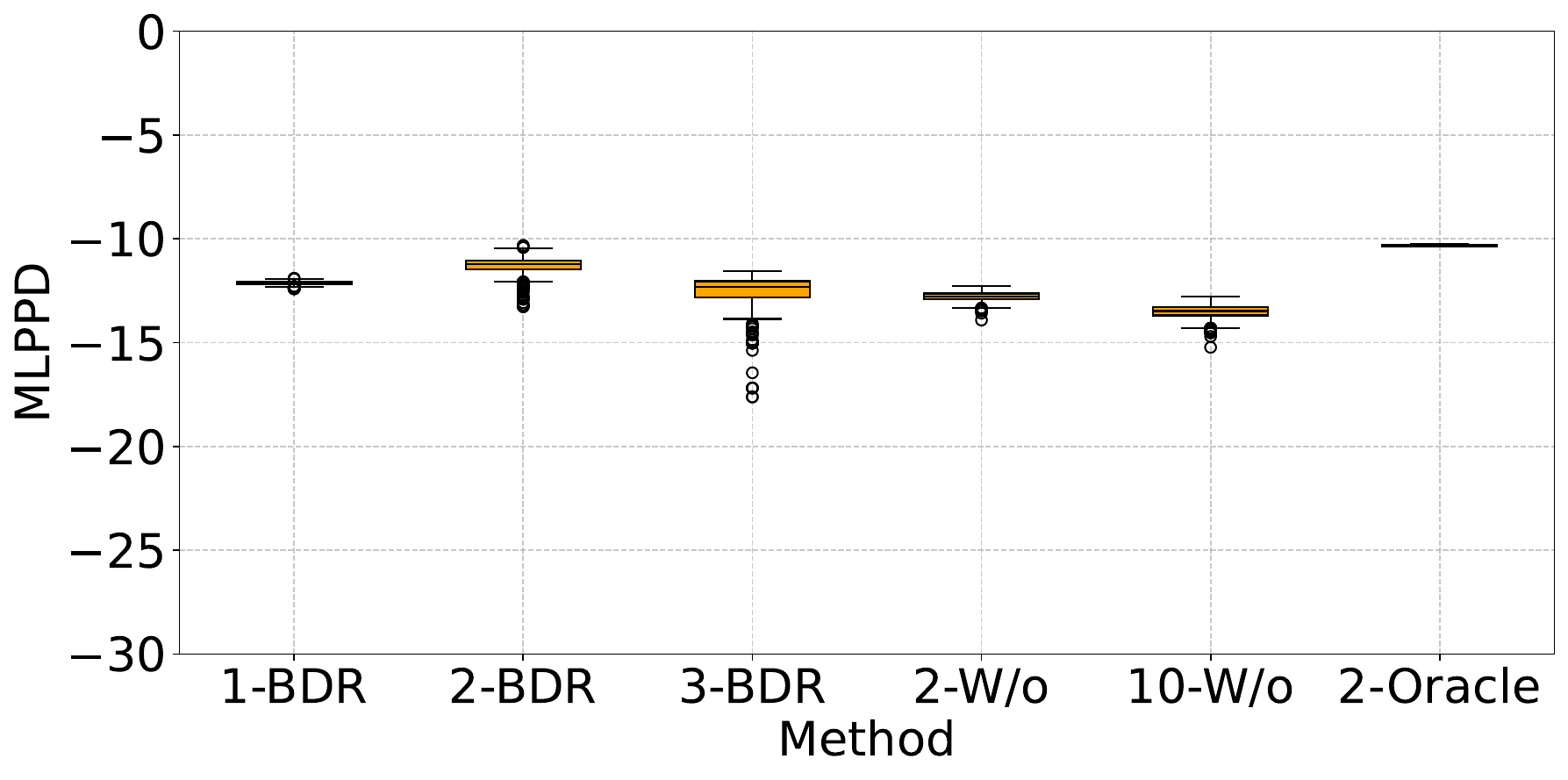}
        \caption{3-layer DGP}
        \label{fig:case1_2d_mlppd_480_l3}
    \end{subfigure}
    \caption{MLPPD for 2D input subspace at $n_{\text{train}}=280$ and $480$ based on the response surface of a polynomial function with known structure.}
    \label{fig:case1_2d_mlppd_280_and_480}
\end{figure}

The results for the two-dimensional case closely parallel the one-dimensional setting, with GP(2)–BDR achieving the strongest overall performance across both training sizes. This model attains the lowest RMSPE and CRPS values and the highest MLPPD, NSME, and BIC, indicating accurate subspace recovery and well-calibrated uncertainty (see \Cref{{tab:case1_2d}}, \Cref{fig:case1_2d_rmspe_280_and_480}, and \Cref{fig:case1_2d_mlppd_280_and_480}.

Performance improves notably when increasing the training size from $n_{\text{train}} = 280$ to $480$. RMSPE decreases, CRPS sharpens, and NSME moves closer to the oracle benchmark. These gains reflect the greater amount of information required to recover a two-dimensional active subspace. The BDR mechanism uses this additional information effectively, yielding pronounced improvements in accuracy and uncertainty quantification.

Two-layer DGP(2)–BDR models exhibit further improvements in uncertainty quantification as the sample size increases, suggesting that modest model depth enhances flexibility once the subspace has been reliably identified. By contrast, models fitted without BDR or under misspecified subspace dimensions show consistently inferior performance, and the gap widens with larger training sets—underscoring the importance of jointly estimating $W$ with other model components.

\textcolor{black}{The CP and ALCI results in \Cref{tab:case1_2dT2} provide additional evidence regarding posterior uncertainty quantification. Across both training sizes, the GP(2)-BDR model achieves coverage probabilities near the nominal 95\% level together with relatively short credible intervals, whereas models fitted without BDR or under incorrect subspace dimension tend to show weaker calibration and, in many cases, substantially less favorable interval efficiency. The close correspondence between the BDR and Oracle results further indicates that the learned projection recovers the true two-dimensional input subspace with high fidelity.}

As in the one-dimensional case, the oracle models provide an upper bound on achievable accuracy. The close alignment between BDR-based fits and their oracle counterparts demonstrates that the proposed framework accurately recovers the true two-dimensional manifold. Taken together, the 1D and 2D results for this polynomial chaos function illustrate that the proposed BDR framework scales well with training size. Larger datasets improve subspace recovery, enhance predictive accuracy, and sharpen uncertainty quantification, with these improvements especially evident in the higher-dimensional setting. These findings provide a strong foundation for extending the methodology to more complex functions where deeper DGP architectures may capture richer latent structure.

\begin{table}
\centering
\begin{tabular}{|l|c|c|c|c|c|c|}
\hline
  \multicolumn{7}{ |c| }{$n=280$ with true W: $10 \times 2$} \\
 \hline
Method (D) & TC(mins) &  RMSPE &  NSME & CRPS & Score  & BIC\\
\hline
GP (1) BDR & 600.04 & 0.1387  & 0.9915 &  0.1484  & 168.4025 & 205.74 \\
GP (2) BDR & 1750.00  & \textbf{0.1097}  & \textbf{0.9953} & $\mathbf{0.0767}$  & \textit{\textbf{545.0266}} & \textbf{207.39}\\
GP (3) BDR & 4002.01   &  0.1736 & 0.9804  & 0.2780  & 458.2853 & 203.95 \\
DGP 2-layer (1) BDR & 3500.09 & 0.1584   & 0.9830  &  0.1141  & 442.9296 & 204.02\\
DGP 2-layer (2) BDR & 5020.70  & 0.1105  &  0.9940 & 0.1645  & 156.7907 & 207.04\\
DGP 2-layer (3) BDR & 6000.21   & 0.2134  & 0.9719  & 0.5667  & 215.8817 & 206.80\\
DGP 3-layer (1) BDR & 6850.00   &  0.2084 & 0.9735  & 0.8402  & 99.3556 & 200.61\\
DGP 3-layer (2) BDR &  7390.04  & 0.2023  & 0.9768   & 0.7613  & 108.4518 & 202.05 \\
DGP 3-layer (3) BDR &  9342.50  &  0.2289  & 0.9715  & 0.4960  & 269.0292 & 199.20 \\
\hline
GP (10) W/o & 864.03 & 0.2540  & 0.9645 & 0.2674  & 176.0696 & 199.06 \\
DGP 2-layer (2) W/o & 4028.63 & 0.2479  & 0.9650  & 0.7399  & 104.0099  & 199.52 \\
DGP 2-layer (10) W/o & 12004.17   & 0.2739  & 0.9636 & 0.8730  &  83.6629 & 197.01 \\
DGP 3-layer (2) W/o & 6066.56 & 0.2360  & 0.9675 & 0.5927 &  74.5334 & 199.79 \\
DGP 3-layer (10) W/o &  14960.35  & 0.2874  &  0.9507 & 0.8715 & 89.2398  & 198.44\\
\hline
GP (2) Oracle &  100.20  & \textit{\textbf{0.0731}}  &  \textit{\textbf{0.9989}}  &   \textit{\textbf{0.0164}}  & 83.5188 & \textbf{210.22}\\
DGP 2-layer (2) Oracle &  1040.08  &  0.0995 &  0.9977  & 0.2135  & \textbf{104.8423} & 208.03\\
DGP 3-layer (2) Oracle &  2491.50  & 0.1921  & 0.9799  & 0.3682  & 129.2567 & 203.29\\
\hline
  \multicolumn{7}{ |c| }{$n=480$ with true W: $10 \times 2$} \\
\hline
GP (1) BDR & 901.00 & $0.1055$  & 0.9972  & 0.1023  & 260.5579 & 248.01 \\
GP (2) BDR & 3475.81 & \textbf{0.0815}  & \textbf{0.9982}   & $\mathbf{0.0162}$  & \textit{\textbf{715.9930}} & \textit{\textbf{249.20}} \\ 
GP (3) BDR & 6945.03   & 0.1272  & 0.9940    & 0.2499  & 712.2032 & 247.52 \\
DGP 2-layer (1) BDR & 4205.09 & 0.1437  & 0.9929  &  0.1046 & 643.7125 & 247.00\\
DGP 2-layer (2) BDR & 6843.75 & 0.1100  & 0.9945  &  0.1442 & 266.6124  & 248.33 \\
DGP 2-layer (3) BDR & 7005.10    & 0.1592  &  0.9923   & 0.5855  & 540.6212 & 246.07 \\
DGP 3-layer (1) BDR & 7419.00 & 0.1937  & 0.9873  & 0.9335  & 206.9284 & 244.02\\
DGP 3-layer (2) BDR & 9685.28   & 0.1821 & 0.9914  & 0.9890  & 146.8134 & 245.06\\
DGP 3-layer (3) BDR & 11064.01   &  0.1941 &  0.9850 & 0.5073  & 694.1730  &  243.90\\
\hline
GP (10) W/o &  1043.90  & 0.2194 & 0.9821  & 0.2376  & 220.6803 & 240.00\\
DGP 2-layer (2) W/o & 5035.49   & 0.2003  &  0.9842  &  0.7889 & 325.3600 & 243.80\\
DGP 2-layer (10) W/o & 14790.14   & 0.2175  & 0.9836   & 0.7510  & 317.6212 &243.45\\
DGP 3 -layer (2) W/o & 9029.35  & 0.2329  &  0.9715  & 0.6274  & 320.5221 & 241.36 \\
DGP 3-layer (10) W/o &  16700.03   &  0.2470 & 0.9702   &  0.6450 & 104.6729 & 243.98\\
\hline
GP (2) Oracle & 400.02   & \textit{\textbf{0.0706}}  & \textit{\textbf{0.9993}}   & \textit{\textbf{0.0118}}  & \textbf{324.2333} &  \textit{\textbf{250.08}}\\
DGP 2-layer (2) Oracle &  1930.58  &  0.0981  & 0.9941  & 0.1496  & 317.5155 & 246.59 \\
DGP 3-layer (2) Oracle & 3220.08   & 0.1629  & 0.9920  & 0.2646  & 331.7272  & 247.00 \\
\hline
\end{tabular}
\caption{Performance Metrics for 2D input subspace at $n_{train}=280$ and $480$ based on response surface of polynomial function with known structure. Table entries report, for each metric, the median value across posterior samples.}
\label{tab:case1_2d}
\end{table}

\begin{table}[t]
\centering
\begin{tabular}{|l|c|c|c|c|c|c|c|}
\hline
  \multicolumn{3}{ |c| }{$n=280$ with true W: $10 \times 2$} \\
\hline
Method (D) & CP & ALCI (95\%) \\
\hline
GP (1) BDR & 0.9429 & 2.2627\\ 
GP (2) BDR & 0.9567 & 0.0831\\
GP (3) BDR & 0.8714 & 0.0668\\
DGP 2-layer (1) BDR & 0.9414 & 2.1820\\ 
DGP 2-layer (2) BDR & 0.9218 & 1.1725\\
DGP 2-layer (3) BDR & 0.9571 & 3.6656\\
DGP 3-layer (1) BDR & 0.9142 & 1.2863\\
DGP 3-layer (2) BDR & 0.9294 & 2.4322\\
DGP 3-layer (3) BDR & 0.9714 & 4.5061\\
\hline
GP (10) W/o  & 0.5000 & 0.9248 \\
DGP 2-layer (2) W/o & 0.4920 & 1.3419\\
DGP 2-layer (10) W/o & 0.5667 & 2.4178\\
DGP 3-layer (2) W/o & 0.5520 & 2.5390\\
DGP 3-layer (10) W/o & 0.1000 & 1.5463 \\
\hline
GP (2) Oracle & 0.9460 & 0.0082\\
DGP 2-layer (2) Oracle & 0.9251 & 0.2257\\
DGP 3-layer (2) Oracle & 0.9545 & 1.5371\\
\hline
  \multicolumn{3}{ |c| }{$n=480$ with true W: $10 \times 2$} \\
\hline
GP (1) BDR & 0.9448 & 1.0510\\ 
GP (2) BDR & 0.9667 & 0.0618\\
GP (3) BDR & 0.9167 & 0.0413\\
DGP 2-layer (1) BDR & 0.9516 & 1.0540\\ 
DGP 2-layer (2) BDR & 0.9429 & 0.0763\\
DGP 2-layer (3) BDR & 0.9667 & 2.2576 \\
DGP 3-layer (1) BDR & 0.9317 & 1.1130\\
DGP 3-layer (2) BDR & 0.9350 & 2.1656\\
DGP 3-layer (3) BDR & 0.9855 & 3.6111\\
\hline
GP (10) W/o  & 0.5833 & 0.0112\\
DGP 2-layer (2) W/o & 0.5310 & 1.2473\\
DGP 2-layer (10) W/o &  0.5860 & 0.3619\\
DGP 3-layer (2) W/o & 0.5956 & 2.1480\\
DGP 3-layer (10) W/o & 0.5450 & 0.2416\\
\hline
GP (2) Oracle & 0.9517 & 0.0059\\
DGP 2-layer (2) Oracle & 0.9500 & 0.0214\\
DGP 3-layer (2) Oracle & 0.9589 & 0.6652\\
\hline
\end{tabular}
\caption{Performance Metrics for 2D input subspace at $n_{train}=280$ and $480$ based on response surface of polynomial function with known structure. Table entries report, for each metric, the median value across posterior samples.}
\label{tab:case1_2dT2}
\end{table}

\subsection{Case 2: Synthetic Data}
\label{susec:case2}

This set of experiments evaluates the Bayesian dimension reduction framework under more complex response surfaces that include discontinuities, regime shifts, and strong nonlinear interactions. We examine the performance of two- and three-layer DGPs using synthetic datasets generated from two response functions with known latent structure. Design points are sampled using Latin Hypercube Sampling (LHS), and responses are computed using known projection matrices. Details of the data-generation mechanisms follow the setups described in \cite{Sauer02012023}.

\subsubsection{Synthetic Response Surface from a Piecewise Function}
\label{subsubsec:case2_1D}

We consider a univariate piecewise function \( \eta(z) \), where the scalar input
\[
z = W^\top x,
\]
is obtained by projecting high-dimensional vectors \( x \in [0,1]^{10} \) using a known \(10 \times 1\) matrix \( W \) (as defined in \Cref{subsubsec:case1_1D}). The response function is
\begin{equation}
\eta(z) =
\begin{cases}
1.35 \cos(12\pi z), & z < 0.333, \\
1.35, & 0.333 \leq z \leq 0.666, \\
1.35 \cos(6\pi z), & z > 0.666,
\end{cases}
\label{eq:First Data Generation Scenario}
\end{equation}
which introduces oscillatory behavior in two regions separated by a flat plateau.

High-dimensional inputs \( x \in \mathbb{R}^{10} \) are generated using LHS with sample sizes \( n = 300 \) and \( n = 500 \). Each dataset is split into training (80\%) and test (20\%) sets, and Gaussian noise with mean \(0\) and standard deviation \(0.1\) is added to produce noisy outputs \( y \). One-, two-, and three-layer DGP models are then fitted using the observed data, with the analysis focusing on the recovery of the underlying \(10 \times 1\) projection matrix.

\begin{figure}[h!]
    \centering

    \begin{subfigure}{0.49\textwidth}
        \centering
        \includegraphics[width=\linewidth]{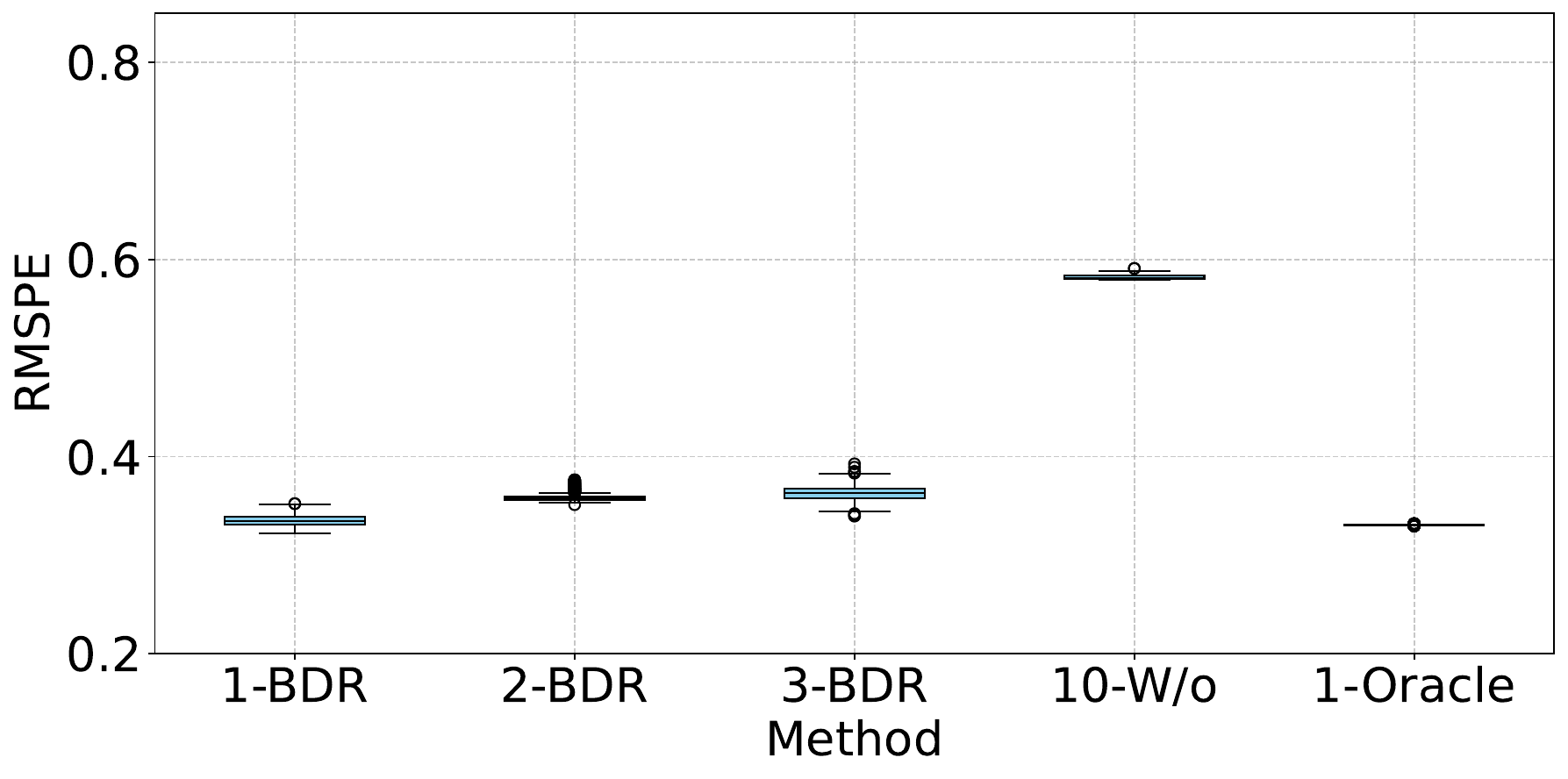}
        \caption{Standard GP at $n_{\text{train}}=240$}
        \label{fig:case2_1d_rmspe_240_l1}
    \end{subfigure}
    \hfill
    \begin{subfigure}{0.49\textwidth}
        \centering
        \includegraphics[width=\linewidth]{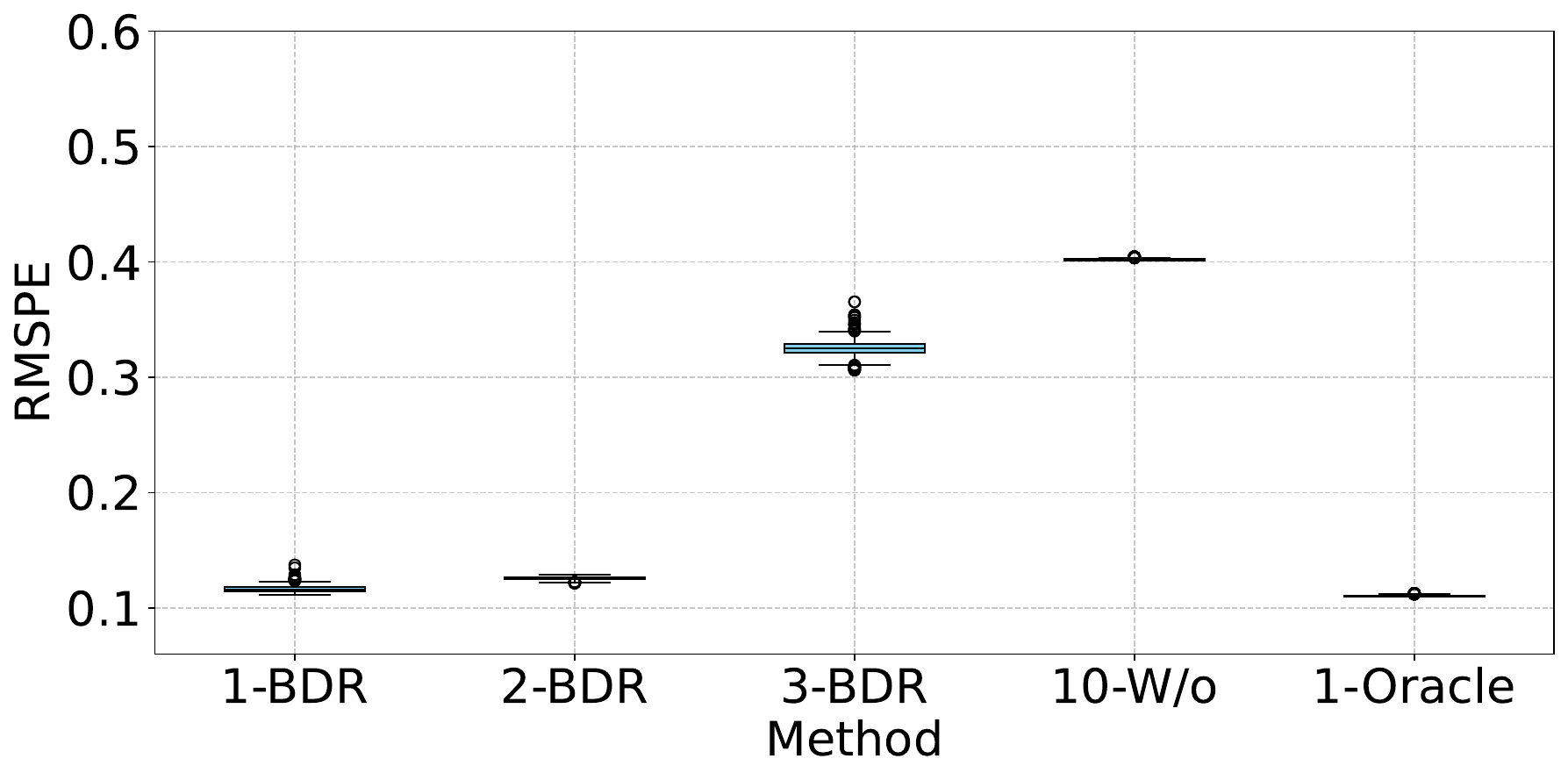}
        \caption{Standard GP at $n_{\text{train}}=400$}
        \label{fig:case2_1d_rmspe_400_l1}
    \end{subfigure}
    \begin{subfigure}{0.49\textwidth}
        \centering
        \includegraphics[width=\linewidth]{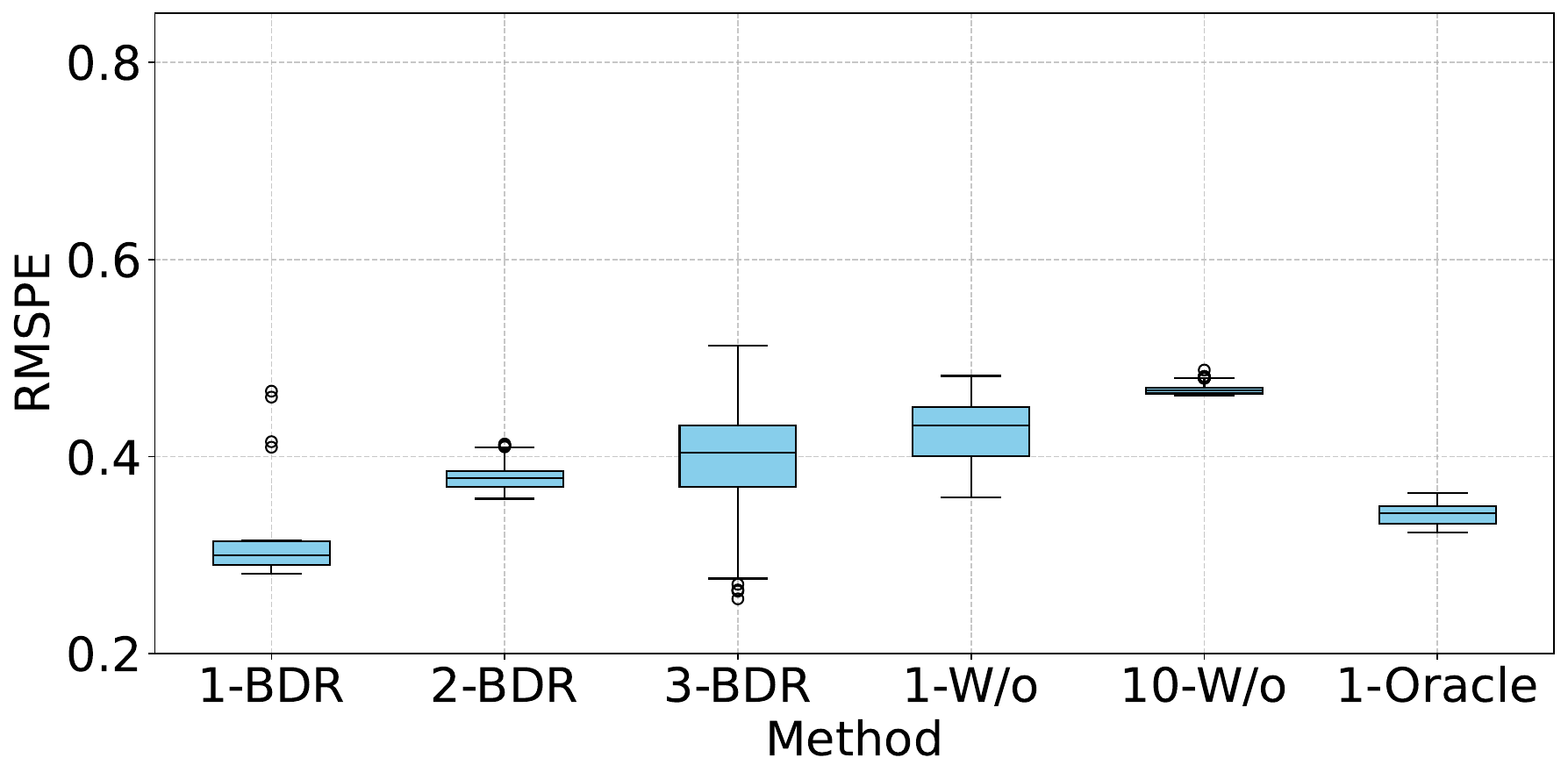}
        \caption{2-layer DGP at $n_{\text{train}}=240$}
        \label{fig:case2_1d_rmspe_240_l2}
    \end{subfigure}
    \hfill
    \begin{subfigure}{0.49\textwidth}
        \centering
        \includegraphics[width=\linewidth]{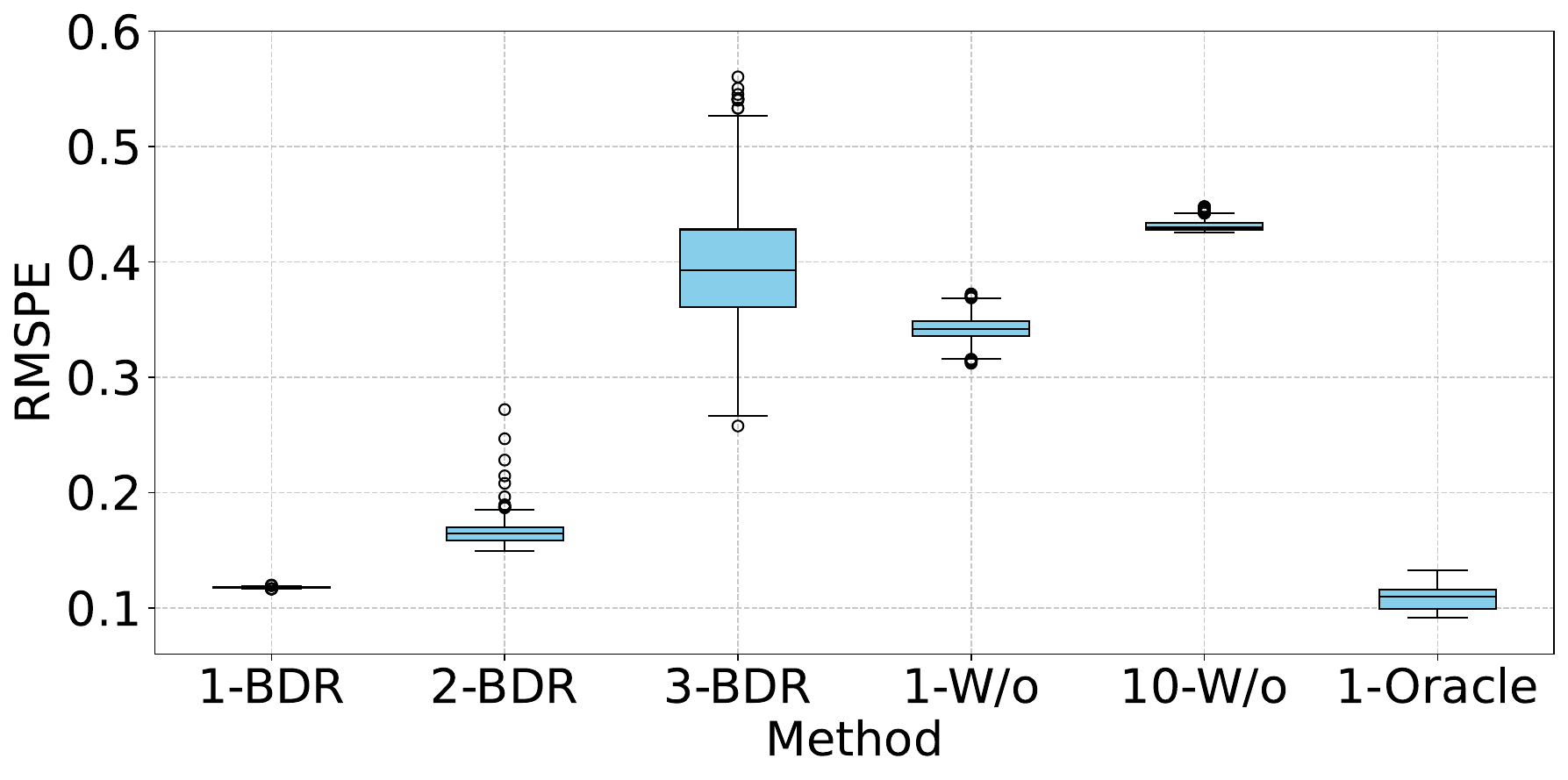}
        \caption{2-layer DGP at $n_{\text{train}}=400$}
        \label{fig:case2_1d_rmspe_400_l2}
    \end{subfigure}
    \begin{subfigure}{0.49\textwidth}
        \centering
        \includegraphics[width=\linewidth]{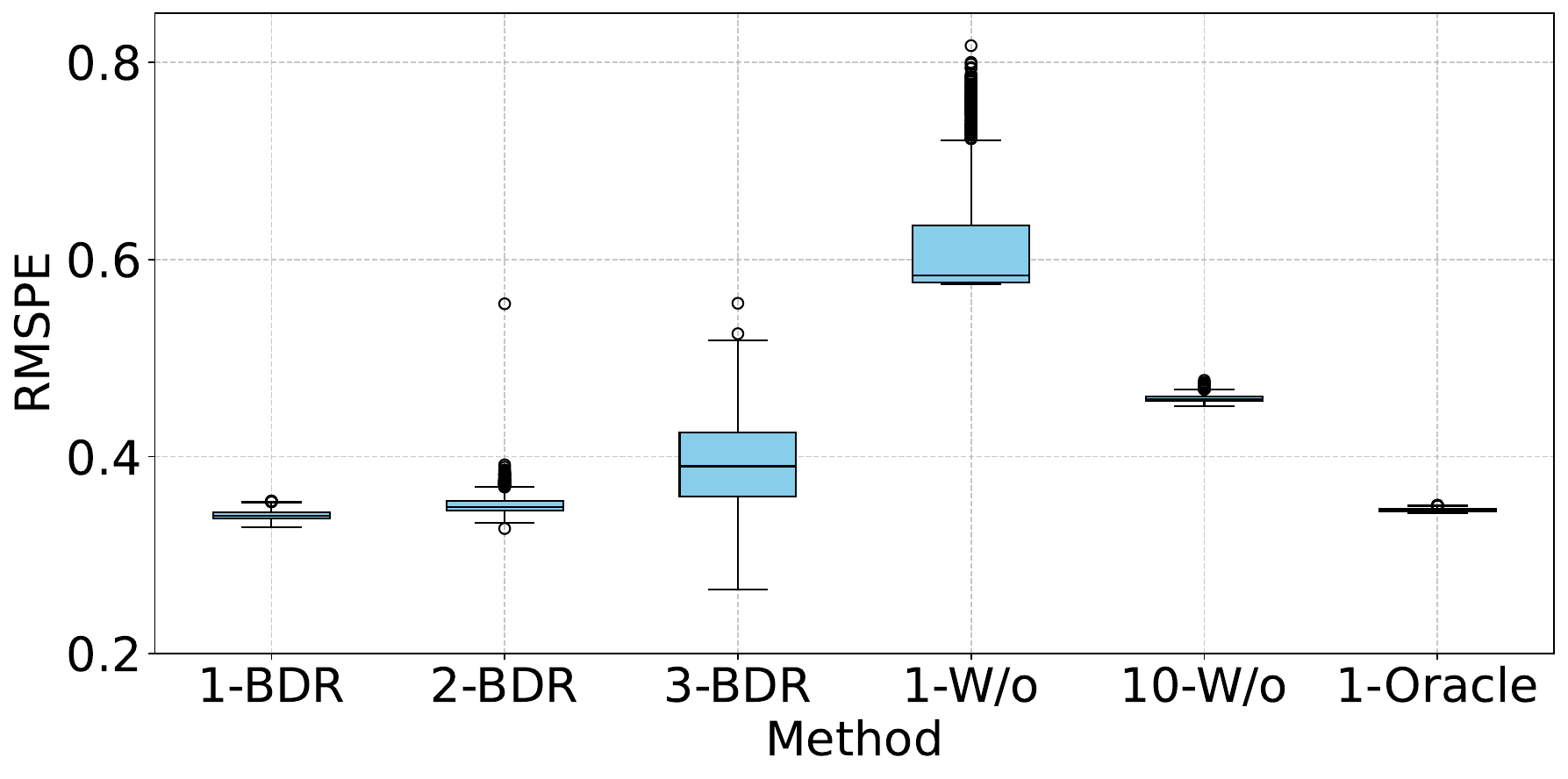}
        \caption{3-layer DGP at $n_{\text{train}}=240$}
        \label{fig:case2_1d_rmspe_240_l3}
    \end{subfigure}
    \begin{subfigure}{0.49\textwidth}
        \centering
        \includegraphics[width=\linewidth]{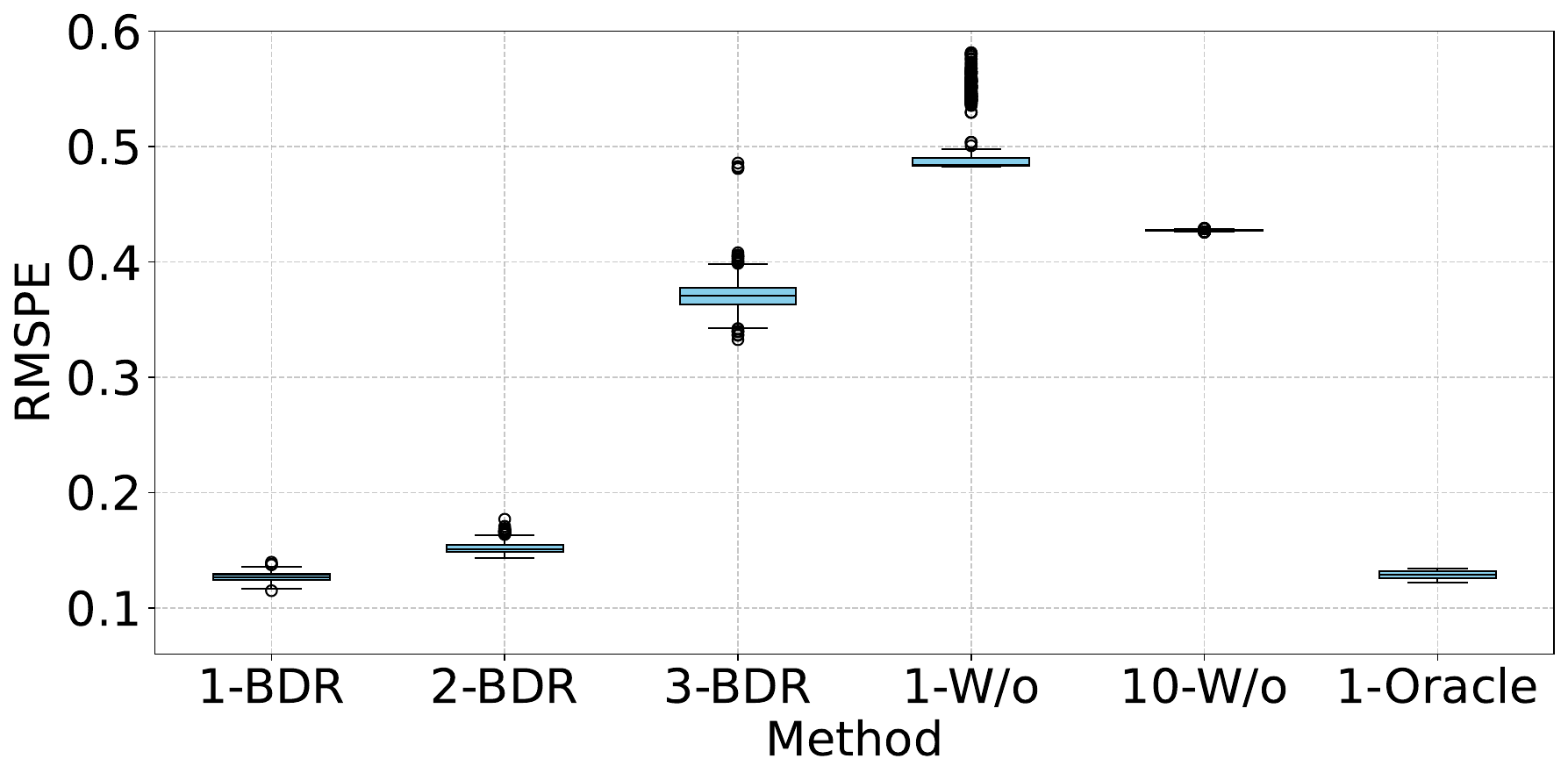}
        \caption{3-layer DGP at $n_{\text{train}}=400$}
        \label{fig:case2_1d_rmspe_400_l3}
    \end{subfigure}
    \caption{RMSPE for 1D input subspace at $n_{\text{train}}=240$ and $400$ based on the response surface of a generated piecewise function.}
    \label{fig:case2_1d_rmspe_240_and_400}
\end{figure}

\begin{figure}[h!]
    \centering

    \begin{subfigure}{0.49\textwidth}
        \centering
        \includegraphics[width=\linewidth]{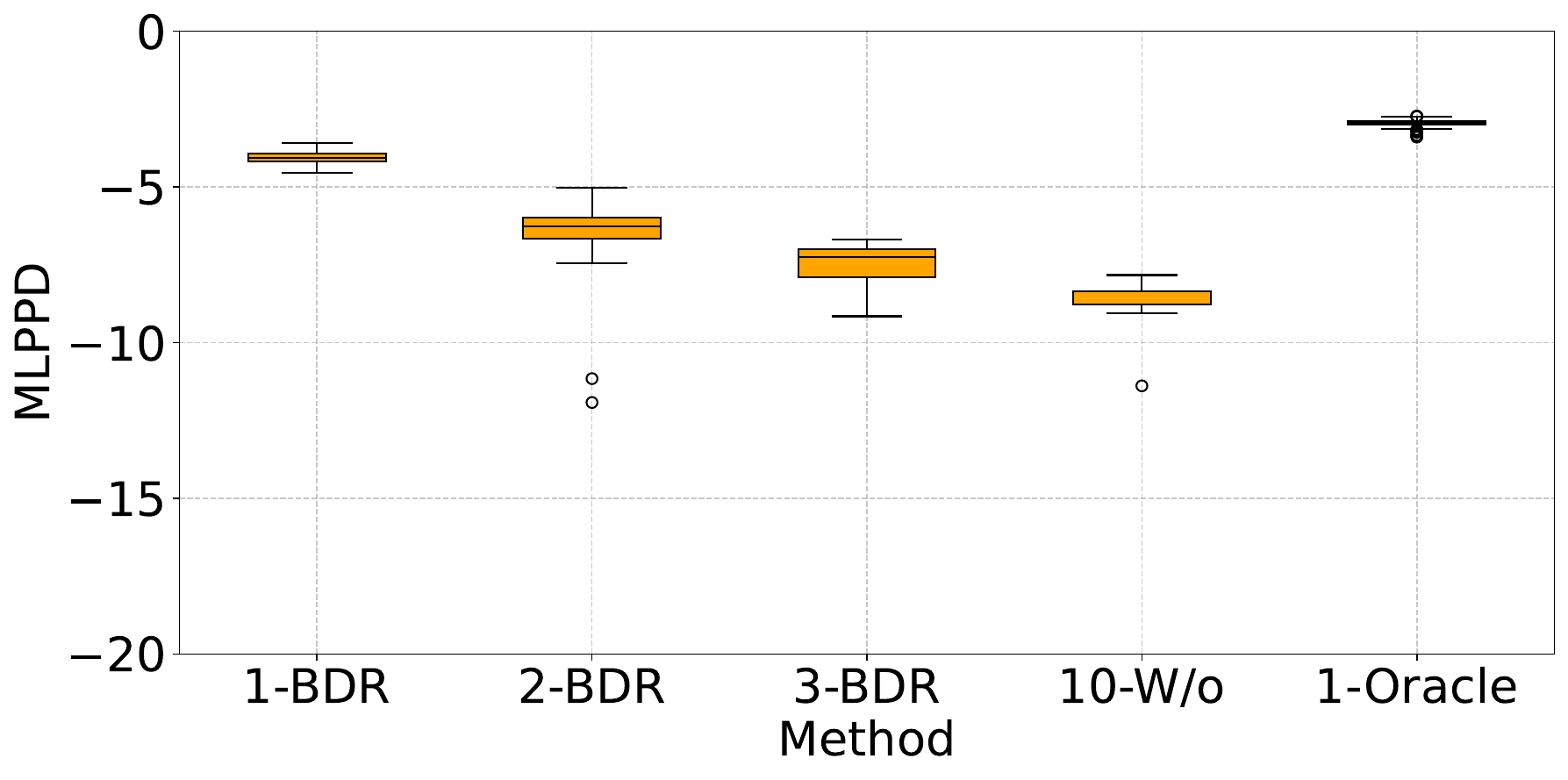}
        \caption{Standard GP at $n_{\text{train}}=240$}
        \label{fig:case2_1d_mlppd_240_l1}
    \end{subfigure}
    \hfill
    \begin{subfigure}{0.49\textwidth}
        \centering
        \includegraphics[width=\linewidth]{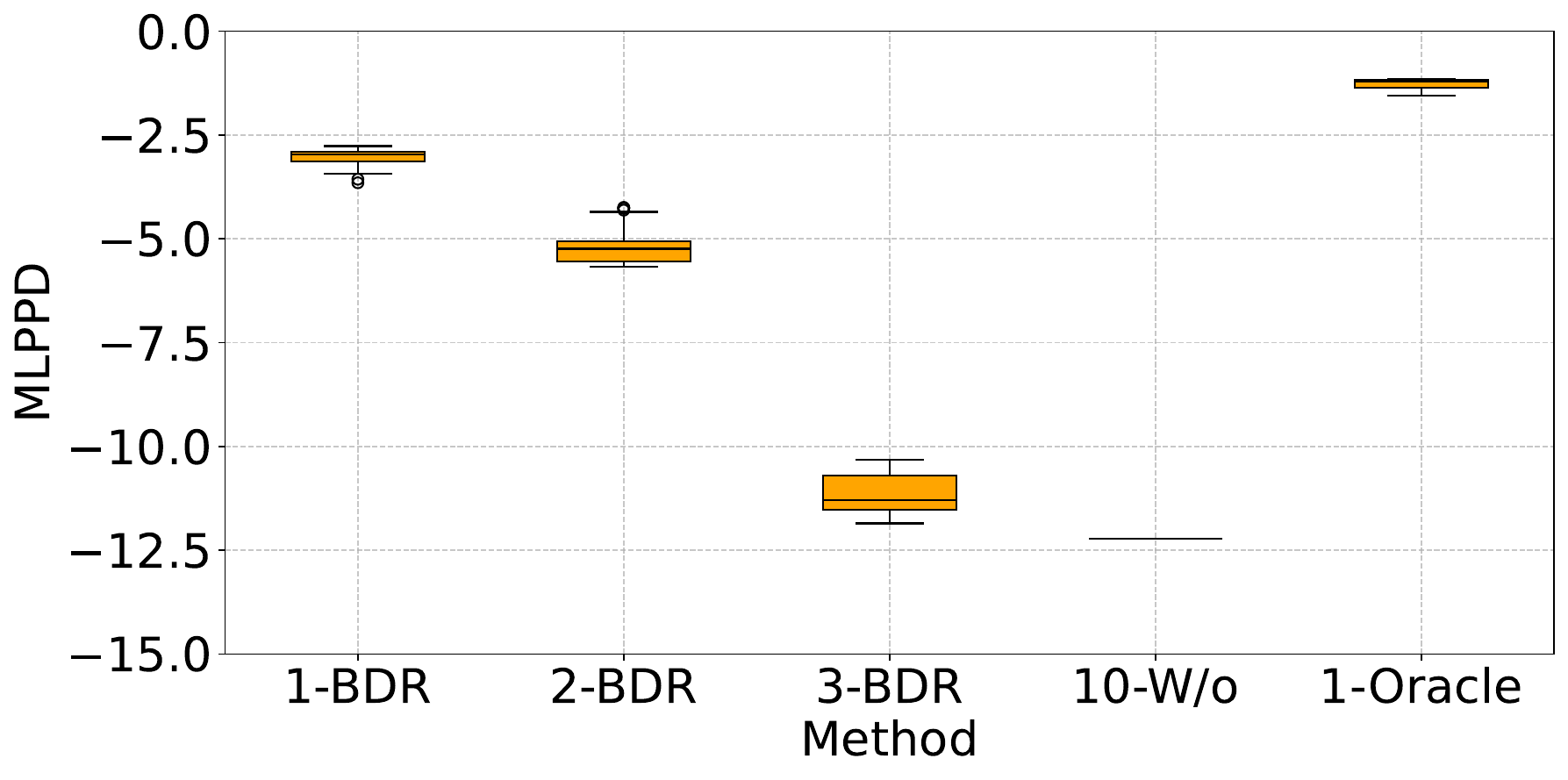}
        \caption{Standard GP at $n_{\text{train}}=400$}
        \label{fig:case2_1d_mlppd_400_l1}
    \end{subfigure}
    \begin{subfigure}{0.49\textwidth}
        \centering
        \includegraphics[width=\linewidth]{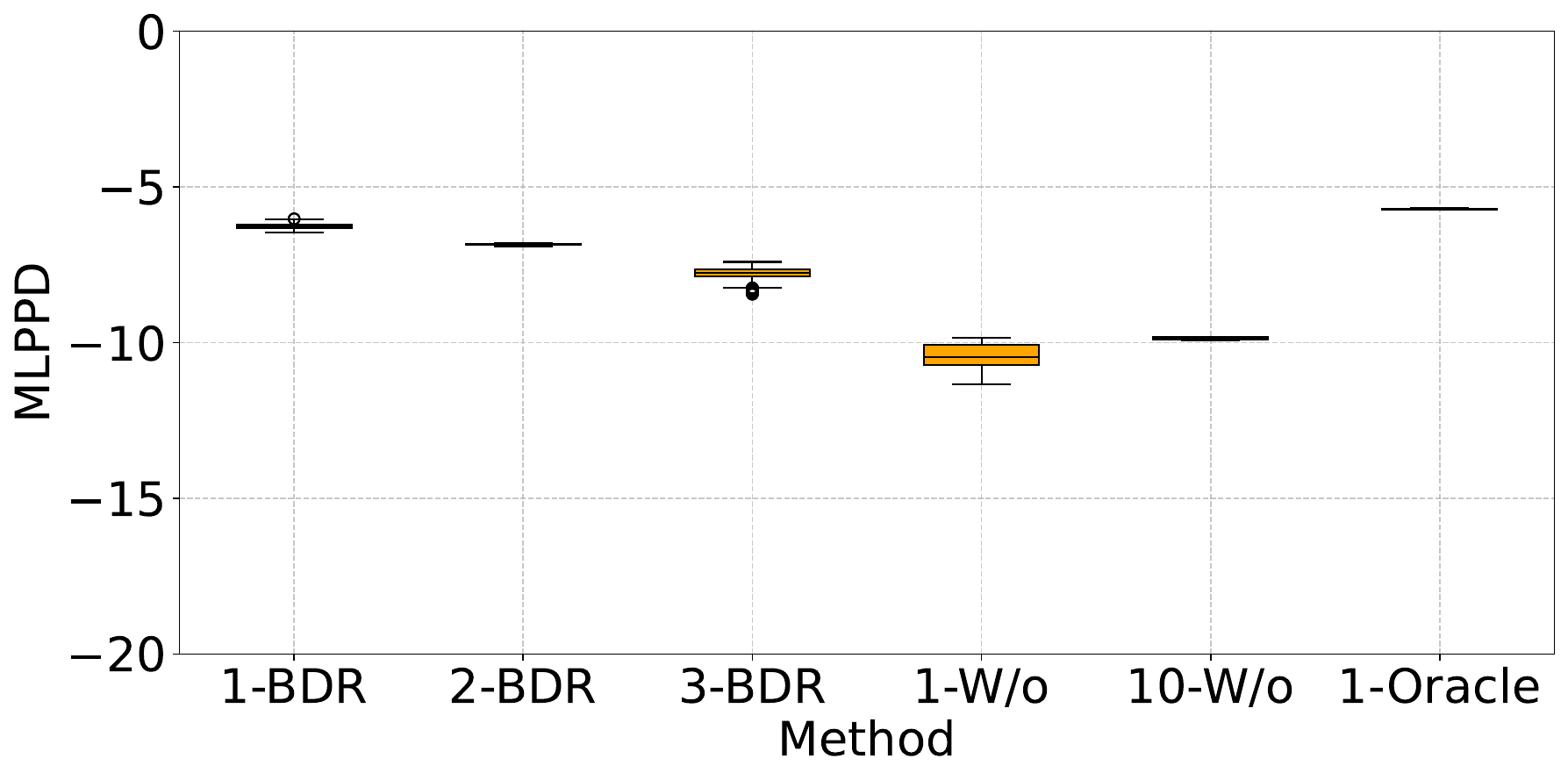}
        \caption{2-layer DGP at $n_{\text{train}}=240$}
        \label{fig:case2_1d_mlppd_240_l2}
    \end{subfigure}
    \begin{subfigure}{0.49\textwidth}
        \centering
        \includegraphics[width=\linewidth]{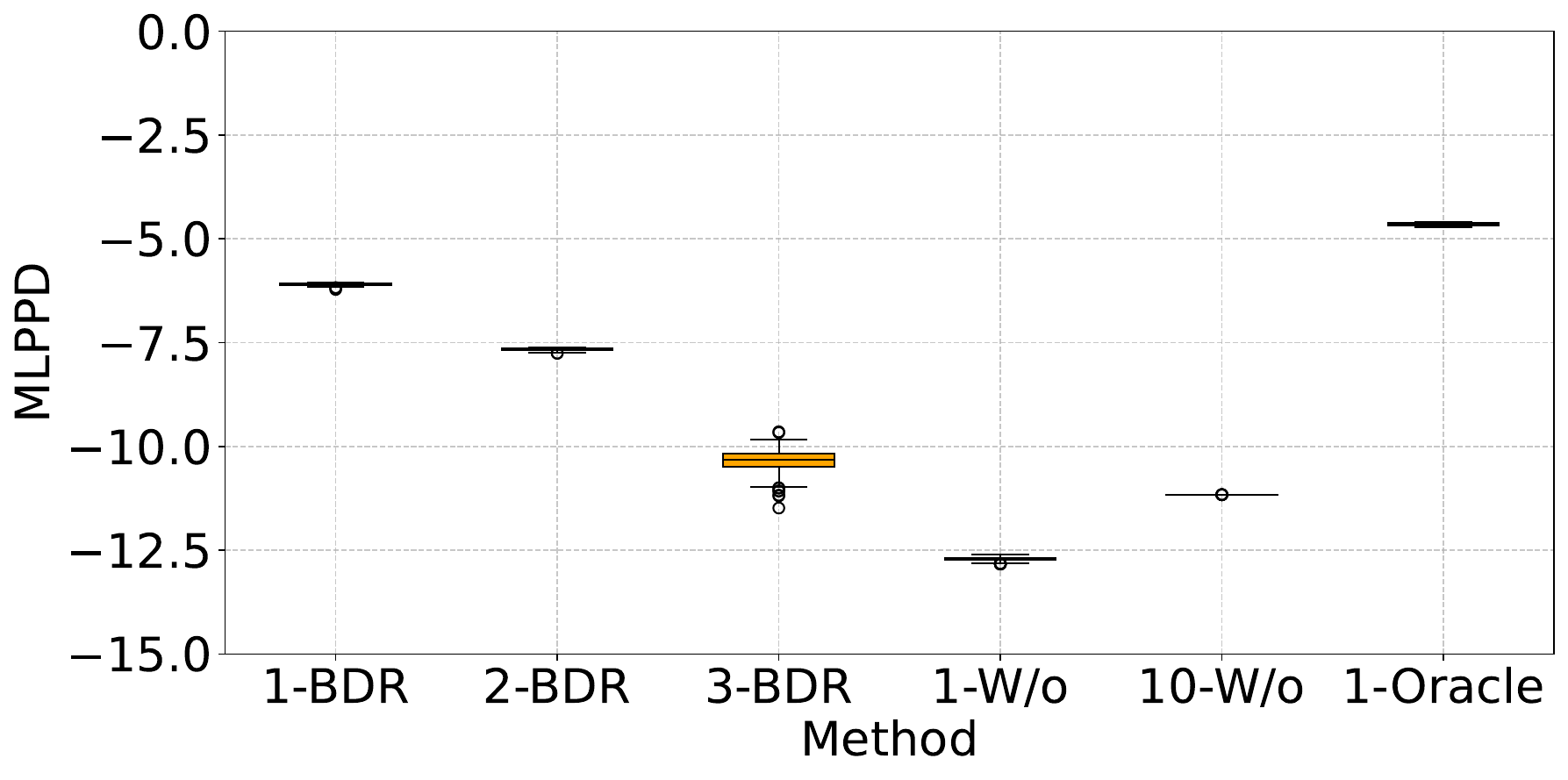}
        \caption{2-layer DGP at $n_{\text{train}}=400$}
        \label{fig:case2_1d_mlppd_400_l2}
    \end{subfigure}
    \hfill
    \begin{subfigure}{0.49\textwidth}
        \centering
        \includegraphics[width=\linewidth]{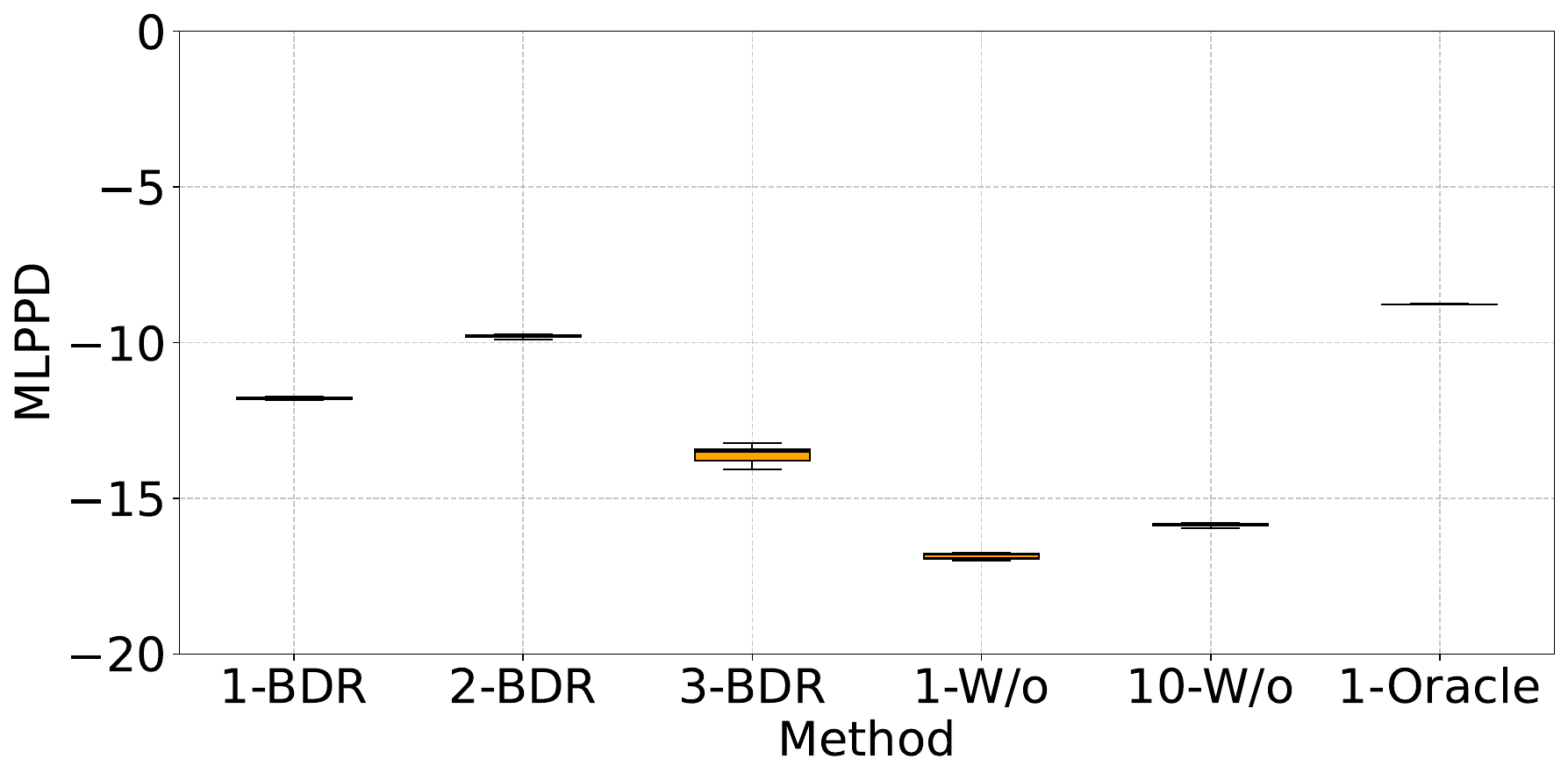}
        \caption{3-layer DGP at $n_{\text{train}}=240$}
        \label{fig:case2_1d_mlppd_240_l3}
    \end{subfigure}
 \begin{subfigure}{0.49\textwidth}
        \centering
        \includegraphics[width=\linewidth]{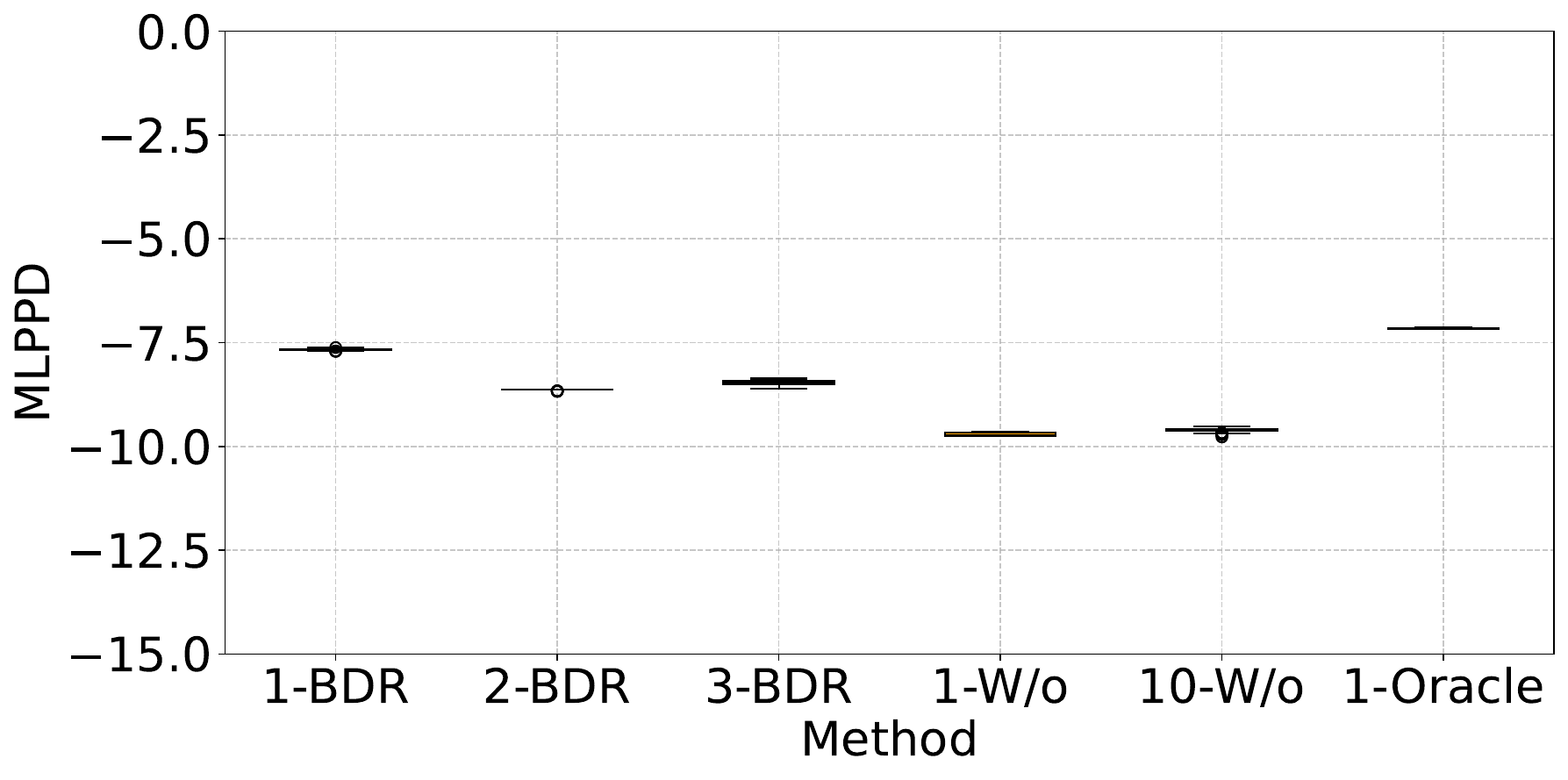}
        \caption{3-layer DGP at $n_{\text{train}}=400$}
        \label{fig:case2_1d_mlppd_400_l3}
    \end{subfigure}
    \caption{MLPPD for 1D input subspace at $n_{\text{train}}=240$ and $400$ based on the response surface of a generated piecewise function.}
    \label{fig:case2_1d_mlppd_240_and_400}
\end{figure}

This piecewise example is particularly challenging because, although the response is driven by a one-dimensional projection, the resulting function exhibits abrupt changes in smoothness and clear regime-dependent behavior. Across both training sizes, the 1D BDR models substantially outperform the models trained in the full 10D space, achieving lower RMSPE and CRPS and higher NSME. At \( n_{\text{train}} = 240 \), the two-layer DGP(1)–BDR model provides the best balance among accuracy, uncertainty quantification, and model complexity, benefiting from the added nonlinear mapping needed to capture non-smooth transitions (see \Cref{tab:case2_1d},  \Cref{fig:case2_1d_rmspe_240_and_400}, and \Cref{fig:case2_1d_mlppd_240_and_400}). When the training size increases to \( n_{\text{train}} = 400 \), GP(1)–BDR becomes competitive with the deeper model, achieving the lowest RMSPE and highest BIC, while the two-layer DGP continues to excel on likelihood-based metrics such as MLPPD. Improvements from \( n_{\text{train}} = 240 \) to \( 400 \) are most pronounced in RMSPE and NSME for the BDR models, reflecting that additional data help resolve both the projection direction and the sharp transitions in the piecewise structure. In contrast, models fit without dimension reduction fail to recover the underlying regime structure even with increased sample size; they retain high RMSPE and low NSME at both training levels. The close alignment between the BDR and oracle results confirms that the proposed framework accurately reconstructs the 1D manifold and remains robust to non-smooth and discontinuous features in the response.

\textcolor{black}{The CP and ALCI results in \Cref{tab:case2_1d2} provide further evidence on posterior uncertainty quantification in this more challenging non-smooth setting. In particular, the 1D BDR models, especially the DGP 2-layer(1)-BDR and GP(1)-BDR fits, attain coverage probabilities near the nominal 95\% level with comparatively shorter credible intervals than many competing alternatives, whereas models without BDR exhibit substantial undercoverage and less reliable interval behavior; the close agreement between the BDR and Oracle fits again indicates accurate recovery of the true one-dimensional subspace.}

\begin{table}[h] 
\centering
\begin{tabular}{|l|c|c|c|c|c|c|}
\hline
  \multicolumn{7}{ |c| }{$n=240$ with true W: $10 \times 1$} \\
 \hline
Method (D) & TC(mins) &  RMSPE &  NSME & CRPS & Score  & BIC\\
\hline
 GP (1) BDR & 1284.51  &  0.3365 &  0.9213  & \textbf{0.3327}   & 109.4969 & 443.52\\
 GP (2) BDR & 1807.90   & 0.3544  &  0.9129  &   0.6040  & 96.4711 & 436.08\\
 GP (3) BDR & 4206.00   & 0.3623  & 0.9104   & 0.6140  & 111.1986 & 435.71 \\
 DGP 2-layer (1) BDR  & 4049.37  &  \textit{\textbf{0.3200}}  &  \textit{\textbf{0.9390}}     &   0.3505  & \textit{\textbf{179.5591}} &  \textit{\textbf{445.83}}\\
 DGP 2-layer (2) BDR  & 4373.82  &  0.3987  & 0.9085  & 0.5152  & 162.8384 & 442.00\\
 DGP 2-layer (3) BDR &  7509.22  & 0.4118  &  0.9000  & 0.6554  & 91.2029 & 440.16\\
 DGP 3-layer (1) BDR &   6038.55    &   0.3428 & 0.9205 &  0.5396  & 100.0682 & 439.30\\ 
 DGP 3-layer (2) BDR & 6616.80 &   0.3532  & 0.9189  &  0.6975 & 160.2100 & 435.72\\
 DGP 3-layer (3) BDR & 8001.34   & 0.3831  & 0.9090  & 0.5015   & 72.5859 & 430.09\\
 \hline
GP (10) W/o &  1785.02   & 0.5819  & 0.8513  & 0.6056  & 153.0783  & 420.59\\
DGP 2-layer (1) W/o & 3157.57   &  0.4452  &  0.8900  & 0.7729 & 102.4485  & 441.67 \\
DGP 2-layer (10) W/o &  11094.00  &  0.4575 &  0.8689 & 0.7983  & 105.9849  & 421.04\\
DGP 3-layer (1) W/o & 5434.28   & 0.5870  & 0.8440  & 0.6934 & 70.7221  & 417.48\\
DGP 3-layer (10) W/o & 17584.00   & 0.4650  &  0.8704 &  0.6010 & 134.8137 & 425.62\\
  \hline
GP (1) Oracle  & 65.90 & 0.3309  & 0.9249  &  0.4964 & 62.2320 &  440.34 \\
DGP 2-layer (1) Oracle & 2779.50   &  \textbf{0.3301} &  \textbf{0.9255} & \textit{\textbf{0.3075}}   &  \textbf{173.9056} & \textbf{444.91} \\
DGP 3-layer (1) Oracle &  4155.37  &  0.3483  & 0.9202  & 0.5735  &    125.3930  & 439.21\\
\hline
  \multicolumn{7}{ |c| }{$n=400$ with true W: $10 \times 1$}\\
  \hline
 GP (1) BDR & 4067.31   & \textbf{0.1166} & \textbf{0.9661}  & \textbf{0.3091}  & 135.5500 & \textbf{485.94}\\
 GP (2) BDR & 4981.71   & 0.1182 &  0.9502 & 0.5747  & 138.8517 &  483.40\\
 GP (3) BDR& 6001.91   & 0.3222  & 0.9100  & 0.7737  & 150.8997 &  482.99\\
 DGP 2-layer (1) BDR  & 7161.58   &  0.1179  &  0.9549 &  0.3478  & \textit{\textbf{270.0840}} & 485.22\\
 DGP 2-layer (2) BDR &  9336.50  &  0.1441  & 0.9445  & 0.5330 & 204.5334 & 482.30\\
 DGP 2-layer (3) BDR & 9820.80   &   0.3984  & 0.9002 & 0.5661  &  228.5347 &  477.50\\
 DGP 3-layer (1) BDR & 70487.28   &  0.1246 & 0.9478 & 0.4648  & 157.1503 & 481.73\\
 DGP 3-layer (2) BDR & 13994.64   &  0.1513   & 0.9311 &  0.5121 & 165.9127 & 479.00 \\
 DGP 3-layer (3) BDR & 15590.36   & 0.3480  & 0.9106  & 0.7284  & 216.3719 & 476.90\\
 \hline
GP (10) W/o & 2231.06   &  0.4019 & 0.8750  & 0.5935  & 257.6286 & 474.00 \\
DGP 2-layer (1) W/o &  5768.48 &  0.3420 & 0.9030 &  0.6569  &  139.2502 & 478.04\\
DGP 2-layer (10) W/o & 16821.02  & 0.4304  & 0.8272  &  0.7097  &  129.6682 & 472.16\\
DGP 3-layer (1) W/o &  6016.28 & 0.4816  & 0.8120 &  0.6540   & 122.6498  & 470.01\\
DGP 3-layer (10) W/o & 18750.00   & 0.4273   & 0.8570  & 0.7840  & 143.9562 & 471.67\\
\hline
GP (1) Oracle & 122.32  &  \textit{\textbf{0.1103}}  & \textit{\textbf{0.9741}}  &  0.29 & 127.3673 & \textbf{486.02}\\
DGP 2-layer (1) Oracle & 3095.38  &  0.1161  & 0.9686 & \textit{\textbf{0.2471}}  &    \textbf{258.3528} & 485.21\\
DGP 3-layer (1) Oracle & 4524.45  &  0.1135  &   0.9690 & 0.5032 & 159.6055 &  484.09\\
 \hline
\end{tabular}
\caption{Performance Metrics for the methods at $n_{train}=240$ and $400$ based on response surface of a generated piecewise function  with known structure. Table entries report, for each metric, the median value across posterior samples.}
\label{tab:case2_1d}
\end{table}

\begin{table}[t]
\centering
\begin{tabular}{|l|c|c|c|c|c|c|c|}
\hline
  \multicolumn{3}{ |c| }{$n=240$ with true W: $10 \times 1$} \\
\hline
Method (D) & CP & ALCI (95\%)\\
\hline
GP (1) BDR & 0.9208 & 0.6381 \\ 
GP (2) BDR & 0.9200 & 0.4390 \\
GP (3) BDR & 0.9047 & 0.6842 \\
DGP 2-layer (1) BDR & 0.9455 & 0.5920\\ 
DGP 2-layer (2) BDR & 0.9410 & 1.1425\\
DGP 2-layer (3) BDR & 0.9059 & 4.3551\\
DGP 3-layer (1) BDR & 0.9502 & 0.7605\\
DGP 3-layer (2) BDR & 0.9203 & 0.8392\\
DGP 3-layer (3) BDR & 0.9161 & 4.6283\\
\hline
GP (10) W/o  & 0.6700 & 0.0247\\
DGP 2-layer (1) W/o & 0.6025  & 0.7659 \\
DGP 2-layer (10) W/o & 0.5930 & 0.1645\\
DGP 3-layer (1) W/o & 0.6547 & 0.9225\\
DGP 3-layer (10) W/o & 0.7010 & 0.2160\\
\hline
GP (1) Oracle & 0.9350 & 0.5548\\
DGP 2-layer (1) Oracle & 0.9281 & 0.8323\\
DGP 3-layer (1) Oracle & 0.9434 & 0.7970\\
\hline
  \multicolumn{3}{ |c| }{$n=400$ with true W: $10 \times 1$} \\
\hline
GP (1) BDR & 0.9439 & 0.2641 \\ 
GP (2) BDR & 0.9670 & 0.2910\\
GP (3) BDR & 0.9313 & 0.5313\\
DGP 2-layer (1) BDR & 0.9511 & 0.2546 \\ 
DGP 2-layer (2) BDR & 0.9715 & 0.6515\\
DGP 2-layer (3) BDR & 0.9102 & 0.9495\\
DGP 3-layer (1) BDR & 0.9610 & 0.3738\\
DGP 3-layer (2) BDR & 0.9820 & 0.6930\\
DGP 3-layer (3) BDR & 0.9333 & 1.3323\\
\hline
GP (10) W/o  & 0.7101 & 0.0155\\
DGP 2-layer (1) W/o & 0.7305 & 0.3804\\
DGP 2-layer (10) W/o & 0.7907 & 0.1396\\
DGP 3-layer (1) W/o & 0.7529 & 0.6358\\
DGP 3-layer (10) W/o & 0.6205 & 0.2090\\
\hline
GP (1) Oracle & 0.9630 & 0.2431\\
DGP 2-layer (1) Oracle & 0.9822 & 0.5075\\
DGP 3-layer (1) Oracle & 0.9700 & 0.3400\\
\hline
\end{tabular}
\caption{Performance Metrics for the methods at $n_{train}=240$ and $400$ based on response surface of a generated piecewise function  with known structure. Table entries report, for each metric, the median value across posterior samples.}
\label{tab:case2_1d2}
\end{table}

\subsubsection{Synthetic Response Surface from an Exponential Function}
\label{subsubsec:case2_2D}

We next consider a two-dimensional input subspace for a nonlinear response surface defined by
\begin{equation}
    f(z_1, z_2) = 10 z_1 \exp(-z_1^2 - z_2^2),
    \label{Second Data Generation Scenario}
\end{equation}
where the latent variables \( z = (z_1, z_2)^\top \) are obtained by projecting high-dimensional input vectors \( x \in [0,1]^{10} \) using a known \(10 \times 2\) projection matrix \( W \) (see \Cref{subsubsec:case1_2D}). To enhance variability in the response surface, the transformed components are rescaled to the interval \([1,7]\) via  
\[
    z_j = (z_j - 0.5)\cdot 6 + 1, \qquad j = 1,2.
\]
Gaussian noise with mean zero and standard deviation \(0.1\) is then added to the outputs to simulate measurement error. As in \Cref{subsubsec:case2_1D}, we consider sample sizes \(n = 300\) and \(500\), with an 80/20 training–test split. Models with one, two, and three DGP layers are trained and evaluated assuming a \(10 \times 2\) projection structure to investigate the effect of dimensionality on predictive performance.

\begin{figure}[h!]
    \centering

    \begin{subfigure}{0.49\textwidth}
        \centering
        \includegraphics[width=\linewidth]{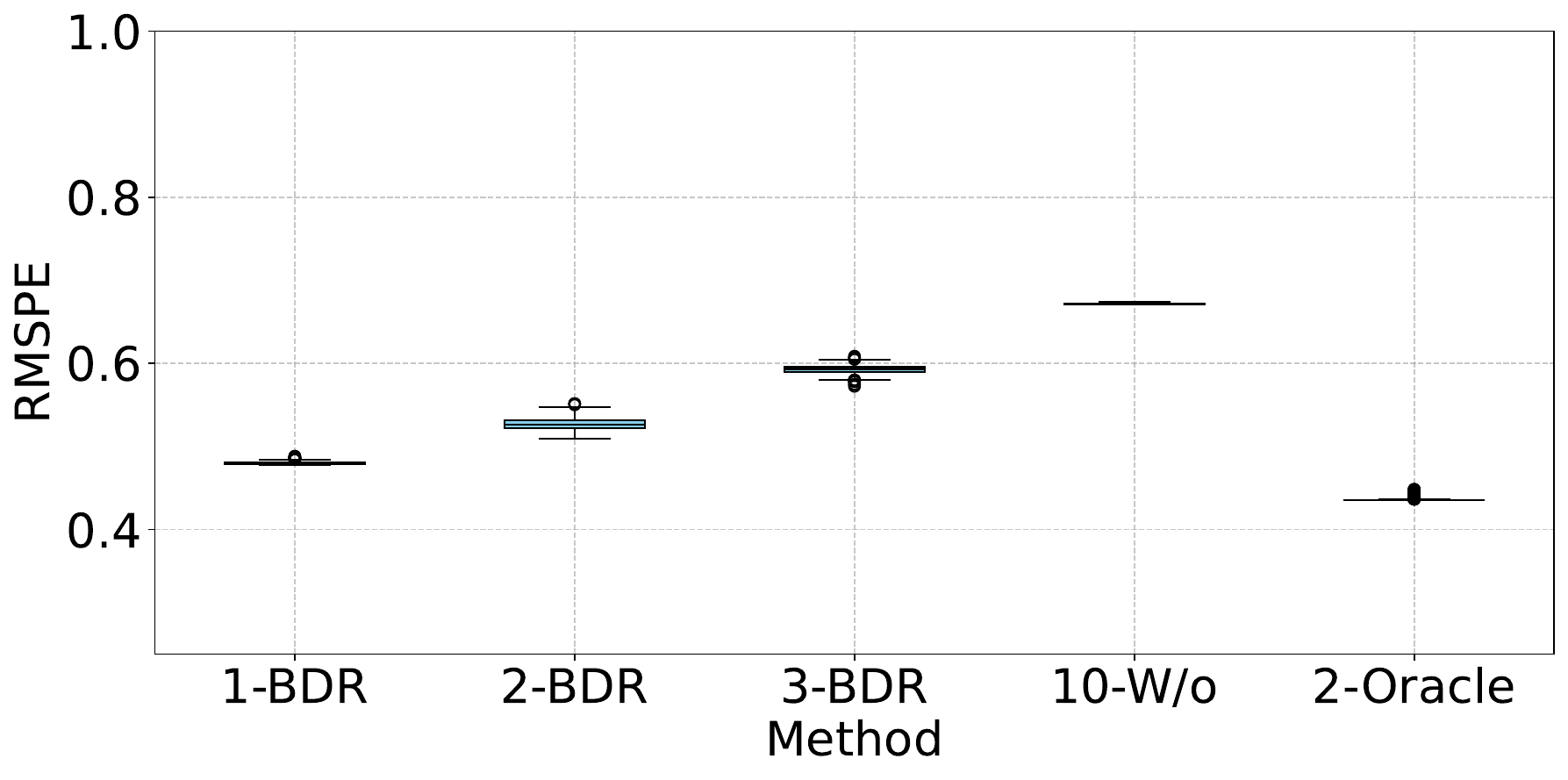}
        \caption{Standard GP at $n_{\text{train}}=240$}
        \label{fig:case2_2d_rmspe_240_l1}
    \end{subfigure}
    \hfill
    \begin{subfigure}{0.49\textwidth}
        \centering
        \includegraphics[width=\linewidth]{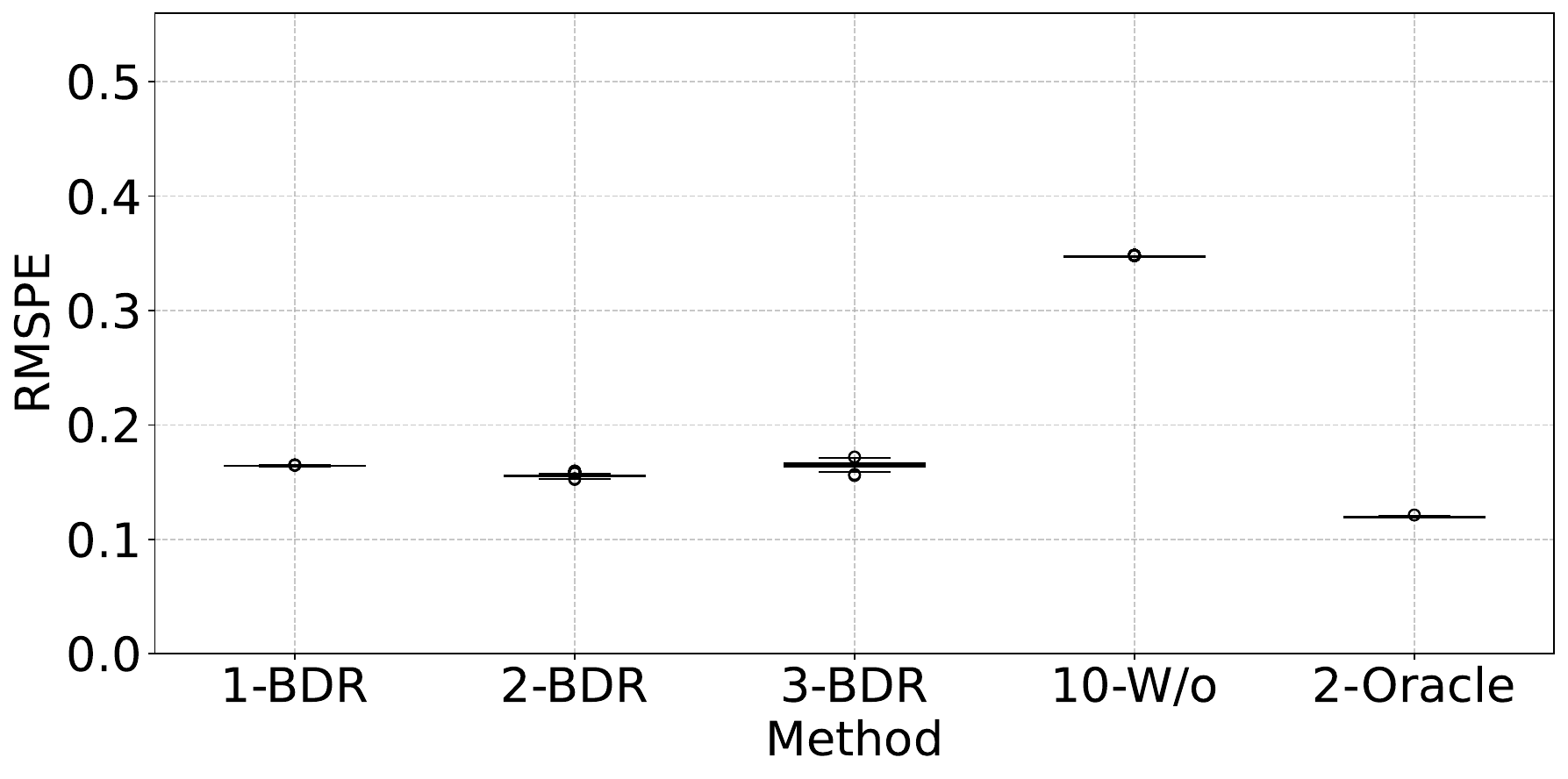}
        \caption{Standard GP at $n_{\text{train}}=400$}
        \label{fig:case2_2d_rmspe_400_l1}
    \end{subfigure}
    \begin{subfigure}{0.49\textwidth}
        \centering
        \includegraphics[width=\linewidth]{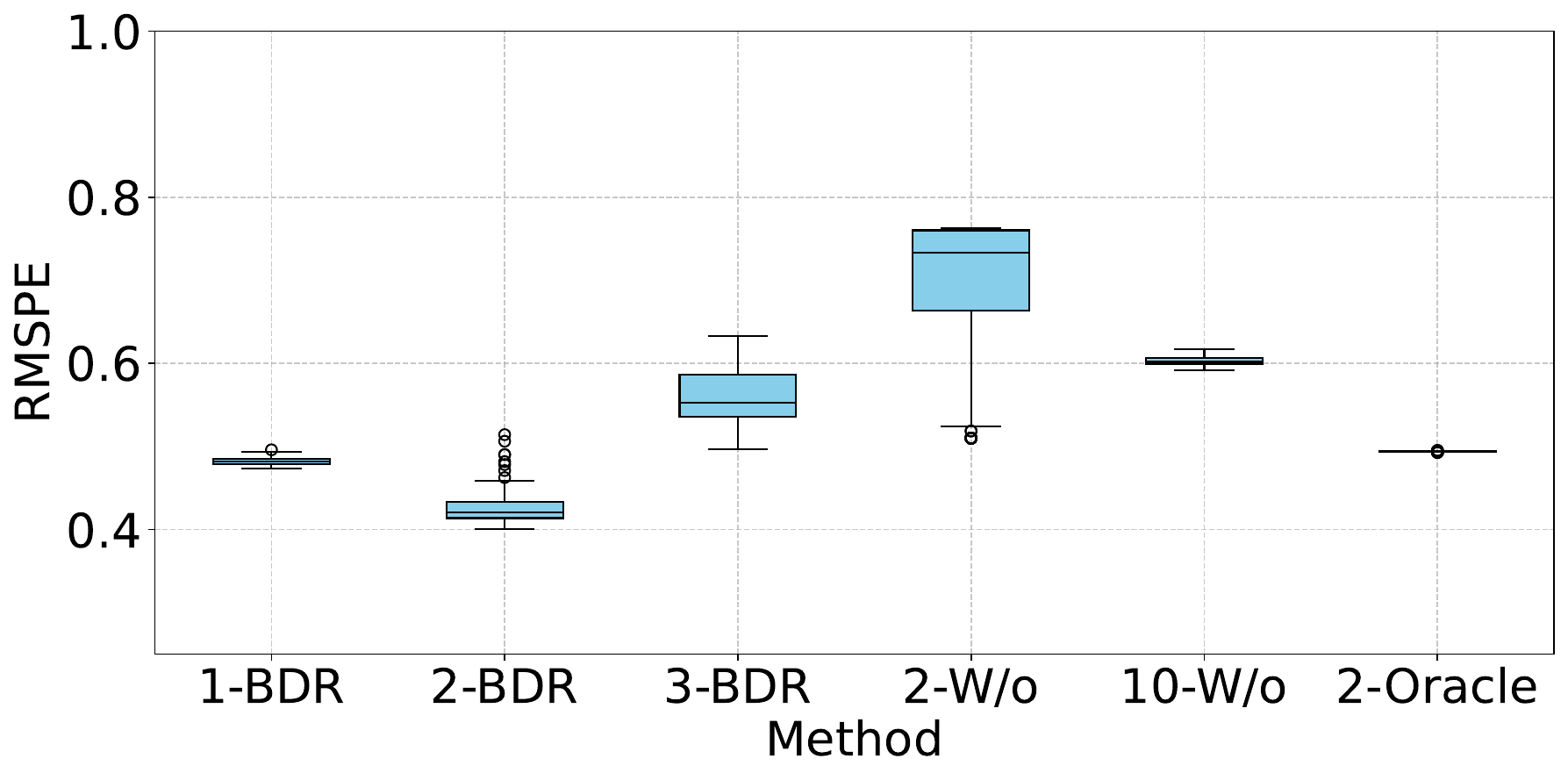}
        \caption{2-layer DGP at $n_{\text{train}}=240$}
        \label{fig:case2_2d_rmspe_240_l2}
    \end{subfigure}
    \hfill
    \begin{subfigure}{0.49\textwidth}
        \centering
        \includegraphics[width=\linewidth]{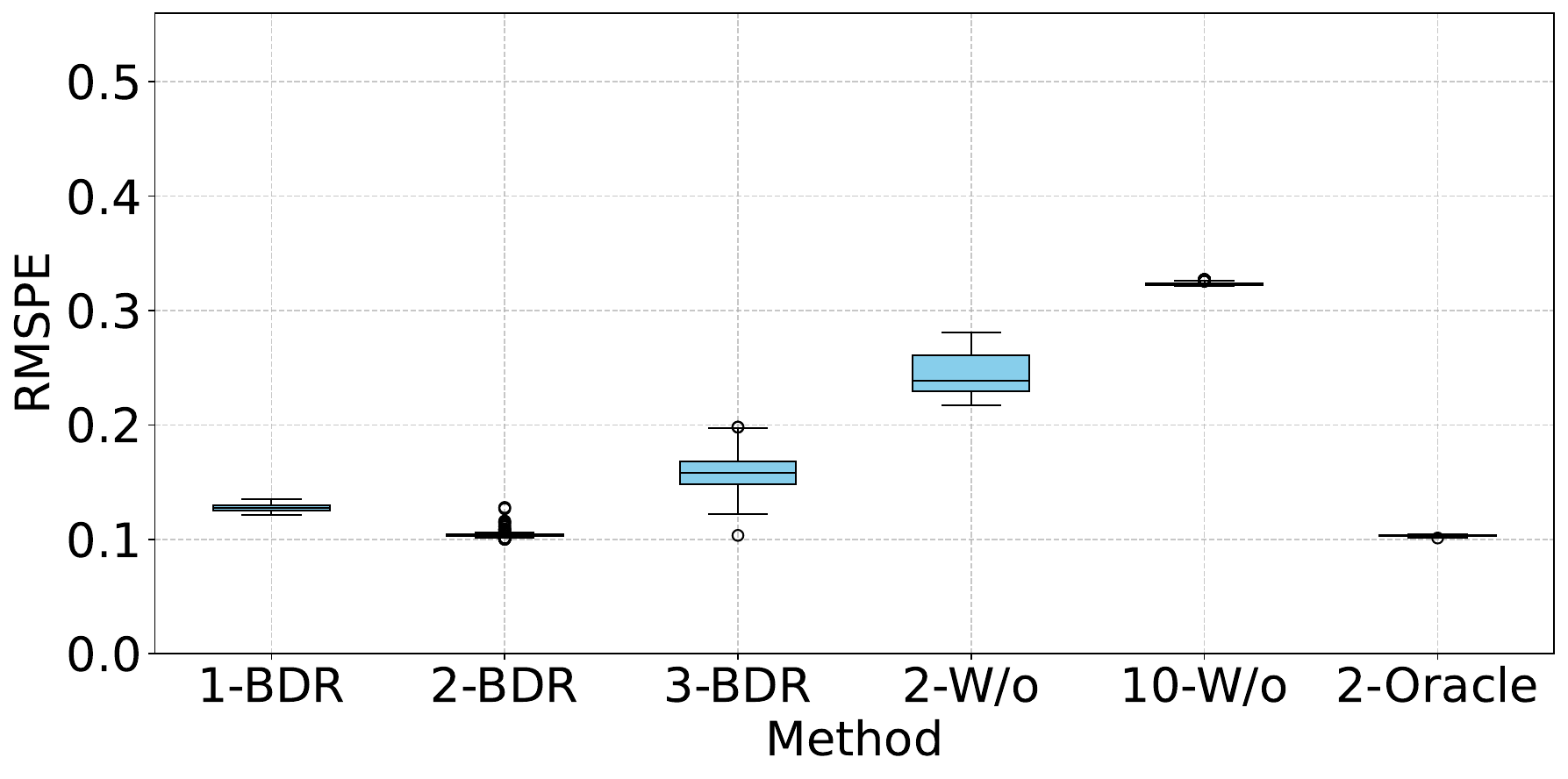}
        \caption{2-layer DGP at $n_{\text{train}}=400$}
        \label{fig:case2_2d_rmspe_400_l2}
    \end{subfigure}
    \begin{subfigure}{0.49\textwidth}
        \centering
        \includegraphics[width=\linewidth]{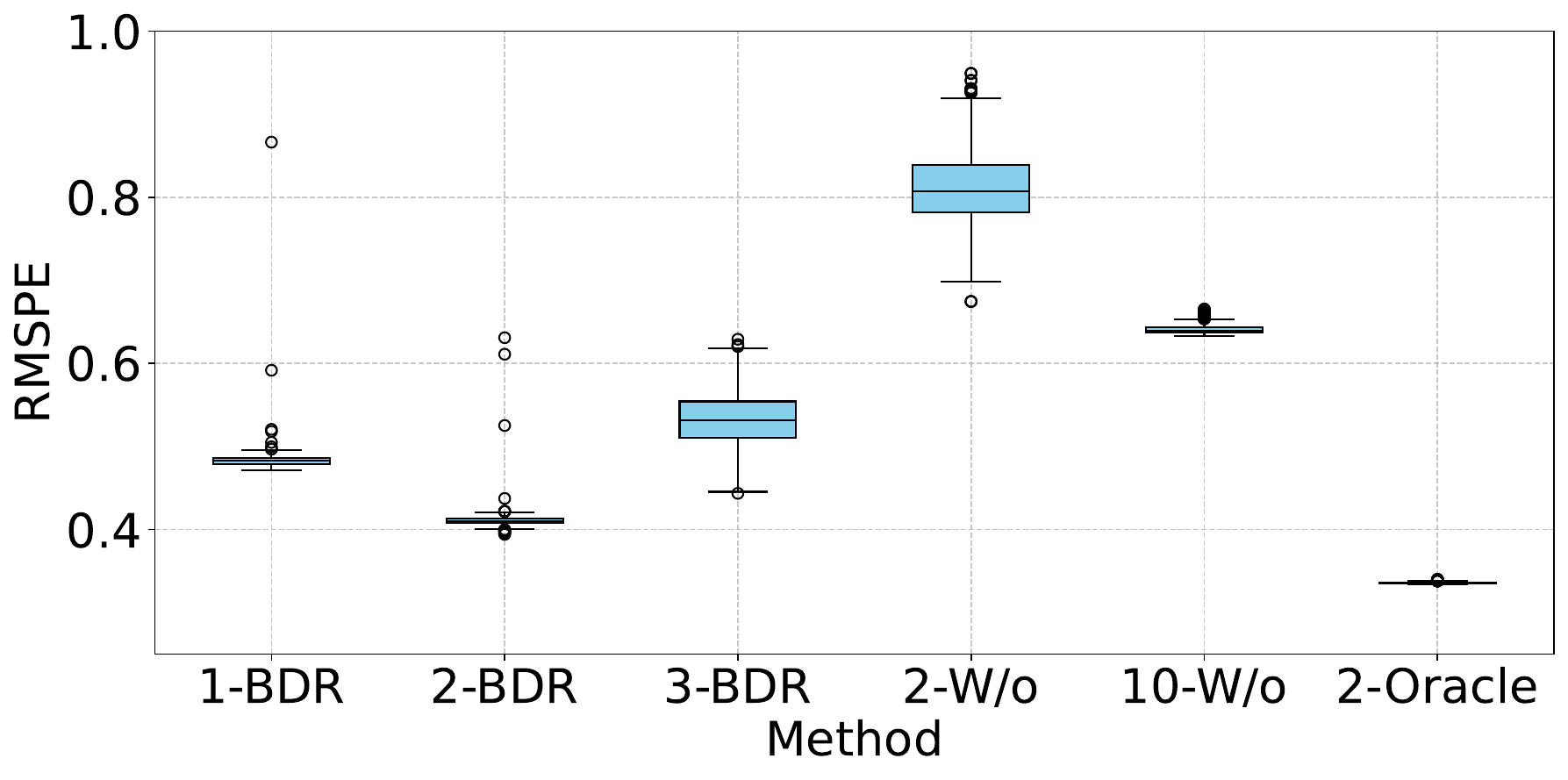}
        \caption{3-layer DGP at $n_{\text{train}}=240$}
        \label{fig:case2_2d_rmspe_240_l3}
    \end{subfigure}
    \begin{subfigure}{0.49\textwidth}
        \centering
        \includegraphics[width=\linewidth]{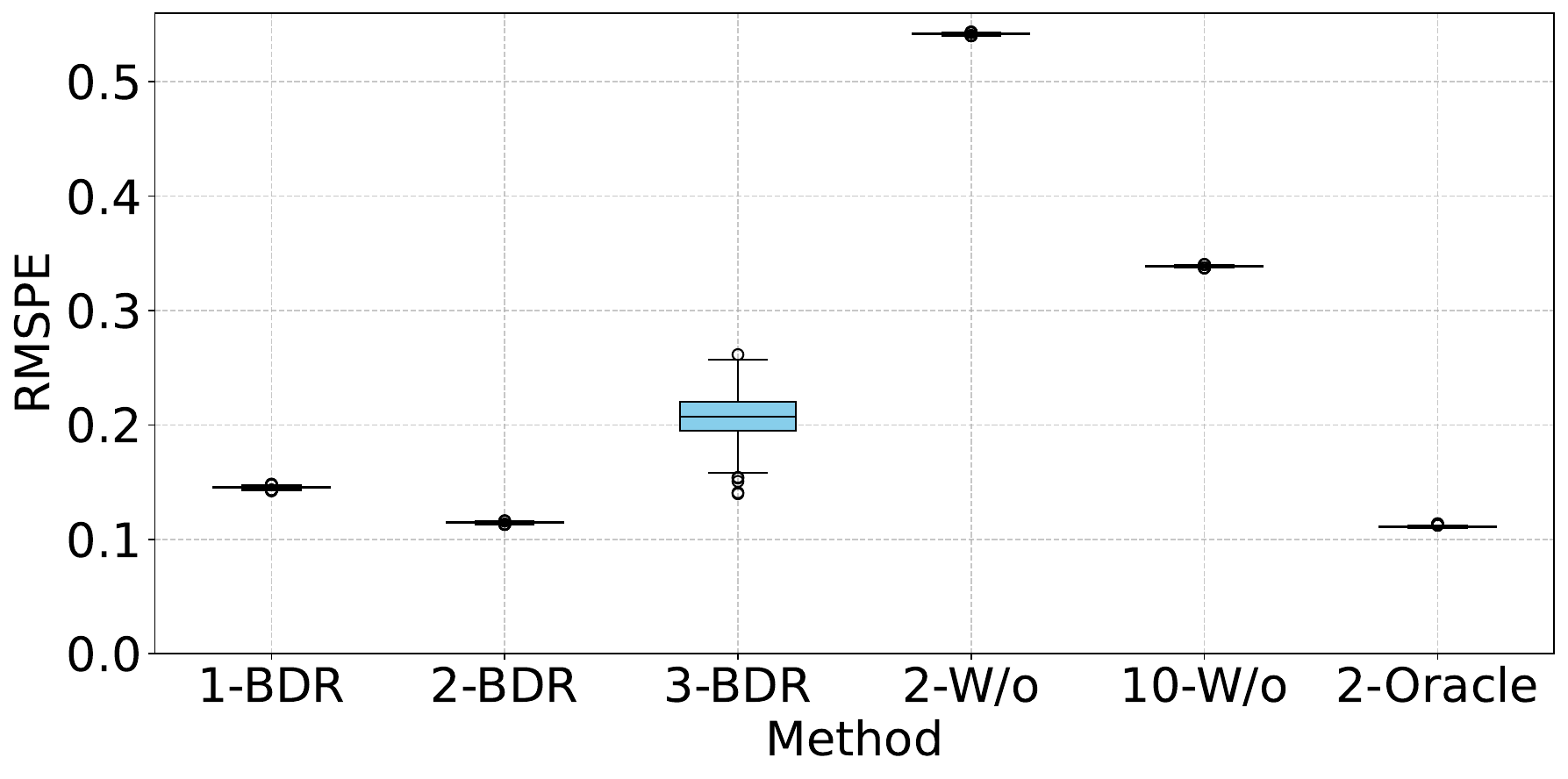}
        \caption{3-layer DGP at $n_{\text{train}}=400$}
        \label{fig:case2_2d_rmspe_400_l3}
    \end{subfigure}
    \caption{RMSPE for 2D input subspace at $n_{\text{train}}=240$ and \(400\) based on the response surface of a generated exponential function.}
    \label{fig:case2_2d_rmspe_240_and_400}
\end{figure}

\begin{figure}[h!]
    \centering

    \begin{subfigure}{0.49\textwidth}
        \centering
        \includegraphics[width=\linewidth]{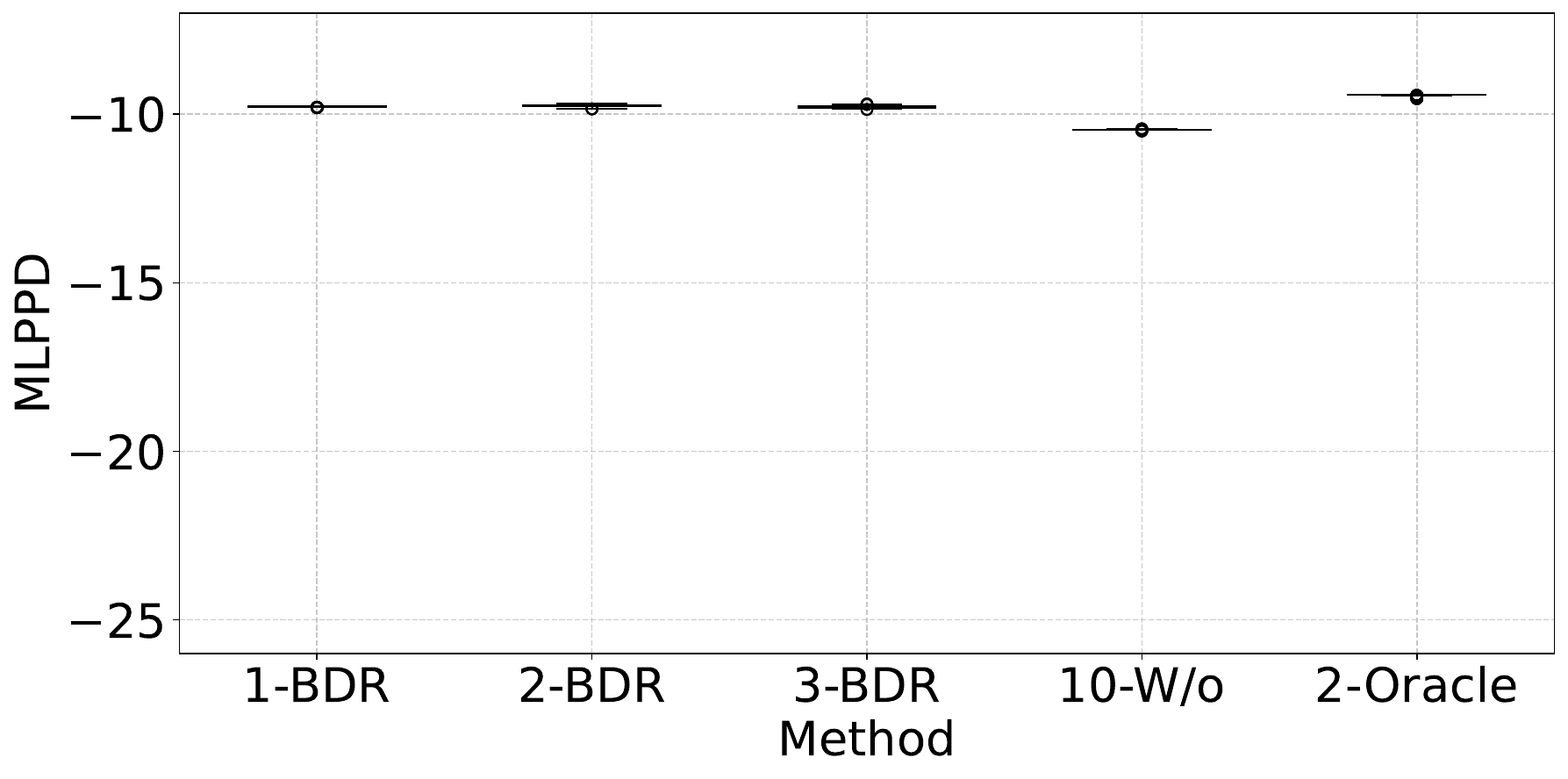}
        \caption{Standard GP at $n_{\text{train}}=240$}
        \label{fig:case2_2d_mlppd_240_l1}
    \end{subfigure}
    \begin{subfigure}{0.49\textwidth}
        \centering
        \includegraphics[width=\linewidth]{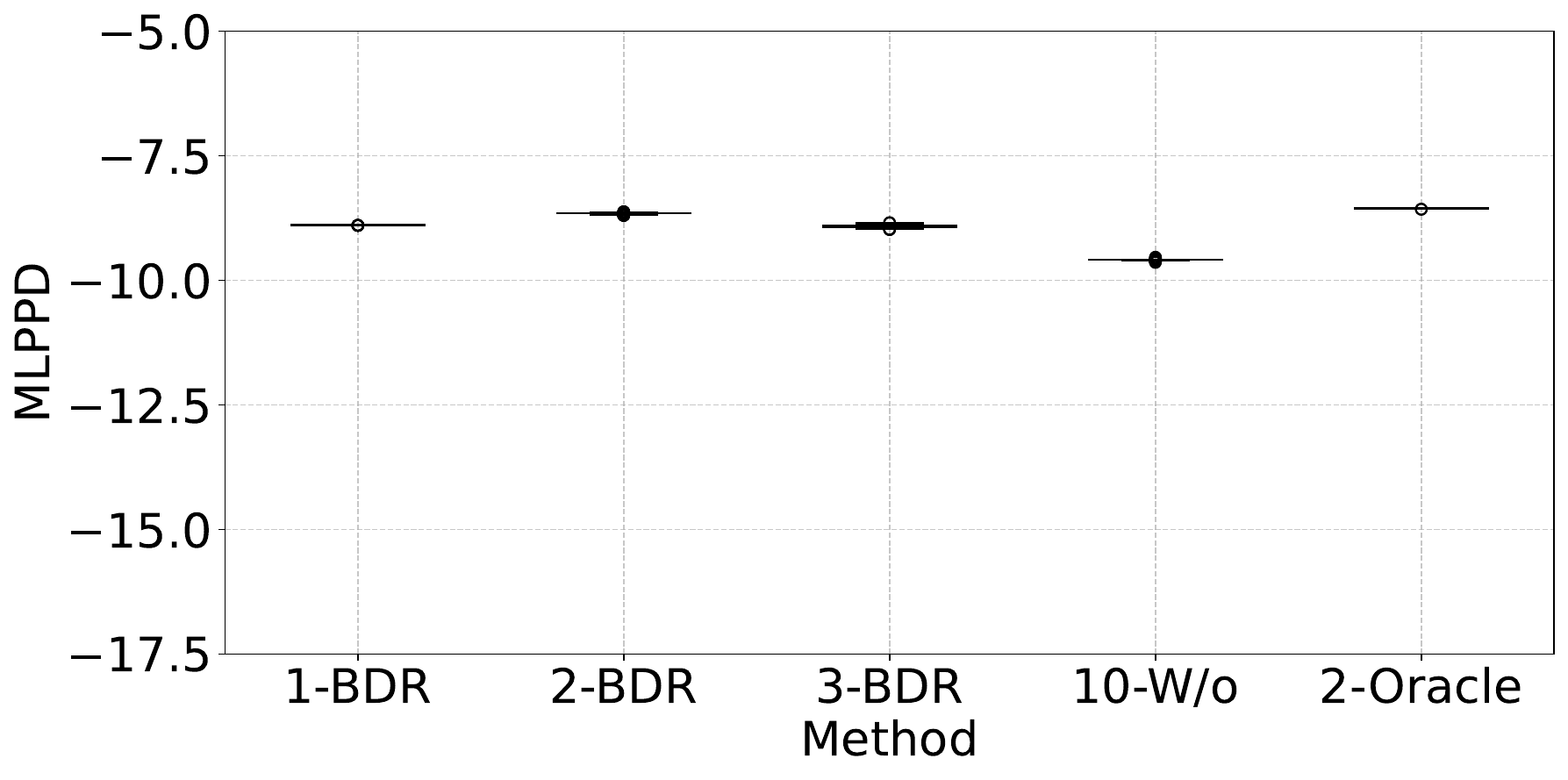}
        \caption{Standard GP at $n_{\text{train}}=400$}
        \label{fig:case2_2d_mlppd_400_l1}
    \end{subfigure}
    \hfill
    \begin{subfigure}{0.49\textwidth}
        \centering
        \includegraphics[width=\linewidth]{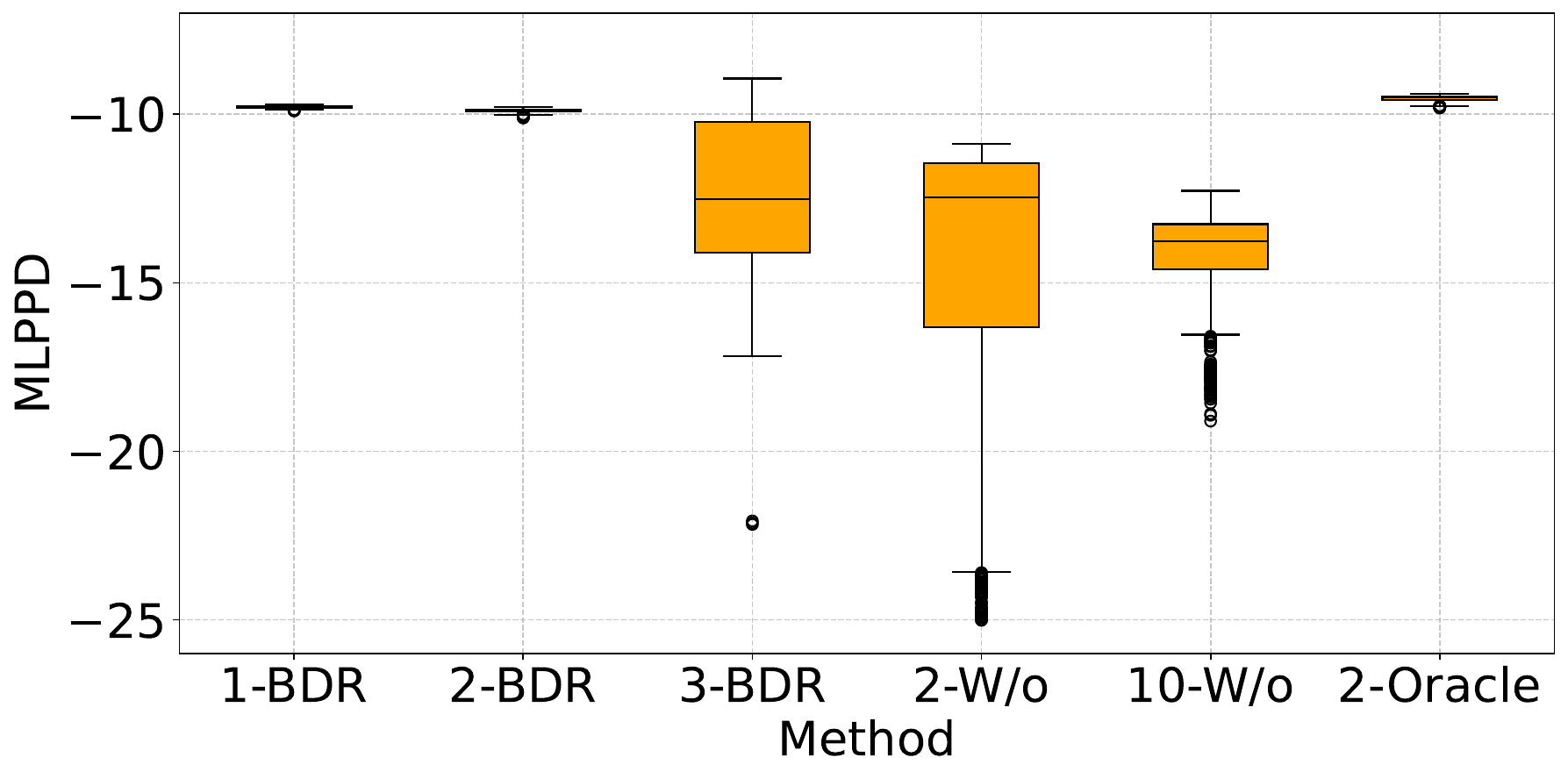}
        \caption{2-layer DGP at $n_{\text{train}}=240$}
        \label{fig:case2_2d_mlppd_240_l2}
    \end{subfigure}
    \hfill
    \begin{subfigure}{0.49\textwidth}
        \centering
        \includegraphics[width=\linewidth]{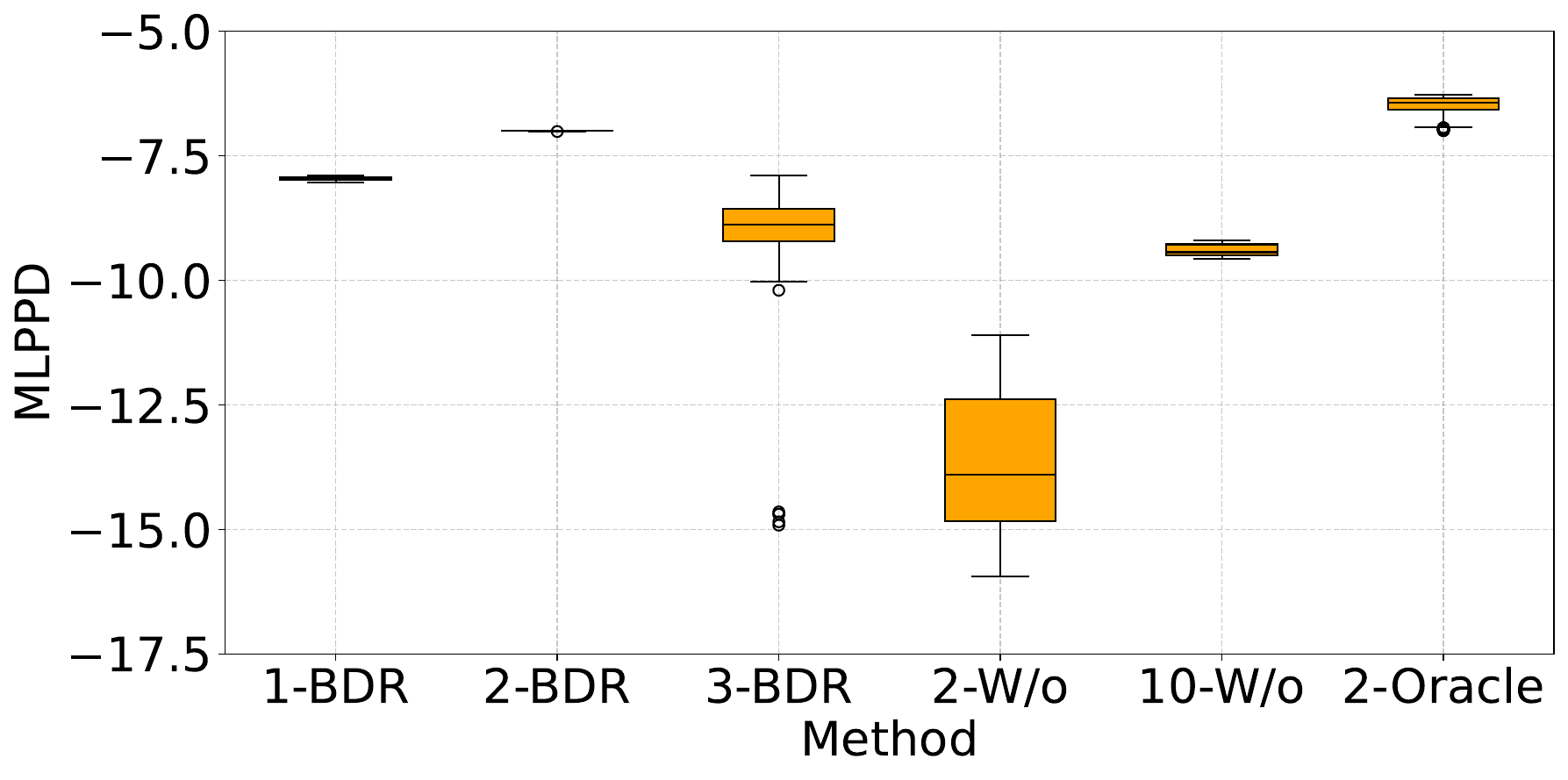}
        \caption{2-layer DGP at $n_{\text{train}}=400$}
        \label{fig:case2_2d_mlppd_400_l2}
    \end{subfigure}
    \begin{subfigure}{0.49\textwidth}
        \centering
        \includegraphics[width=\linewidth]{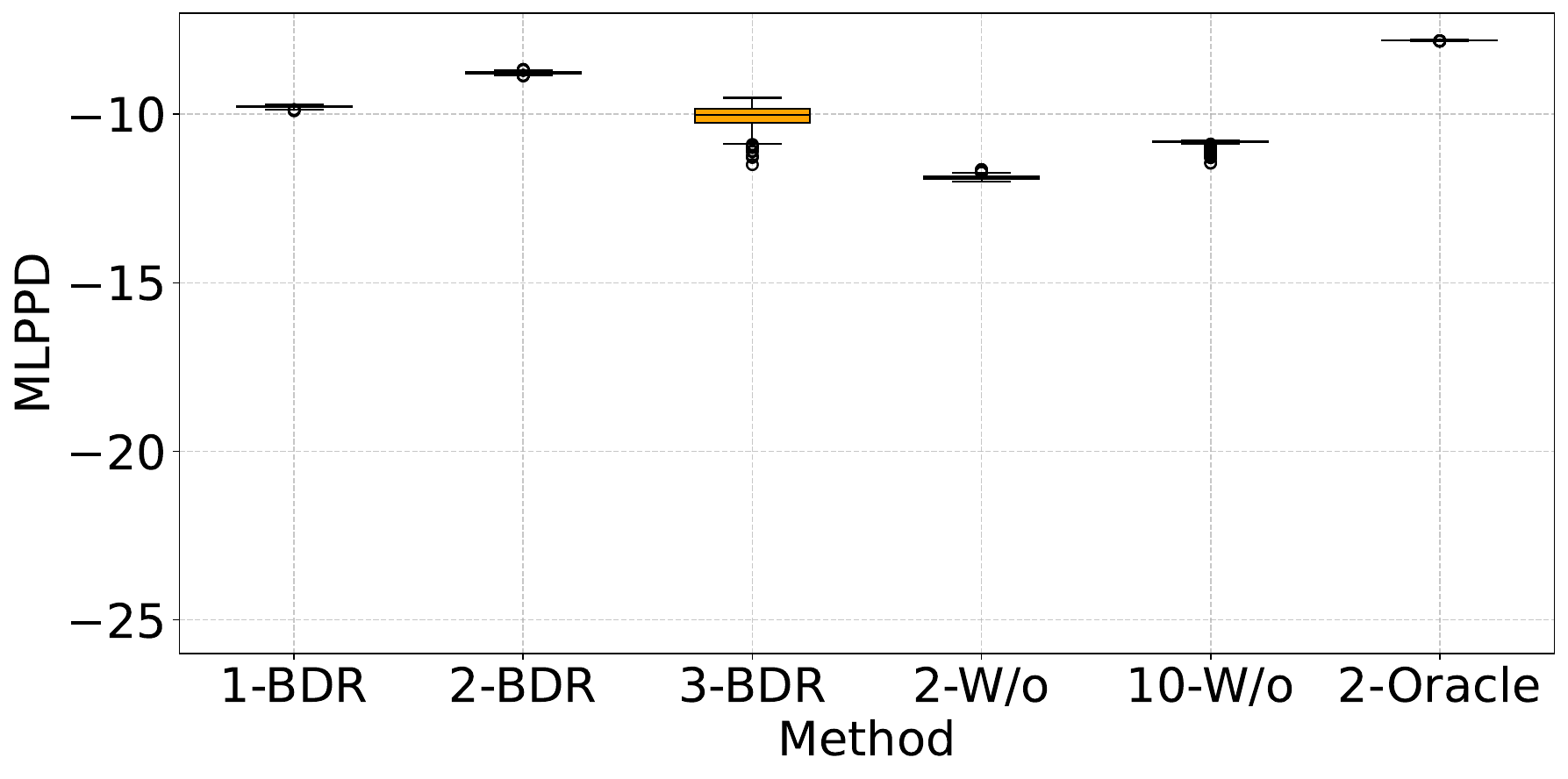}
        \caption{3-layer DGP at $n_{\text{train}}=240$}
        \label{fig:case2_2d_mlppd_240_l3}
    \end{subfigure}
    \begin{subfigure}{0.49\textwidth}
        \centering
        \includegraphics[width=\linewidth]{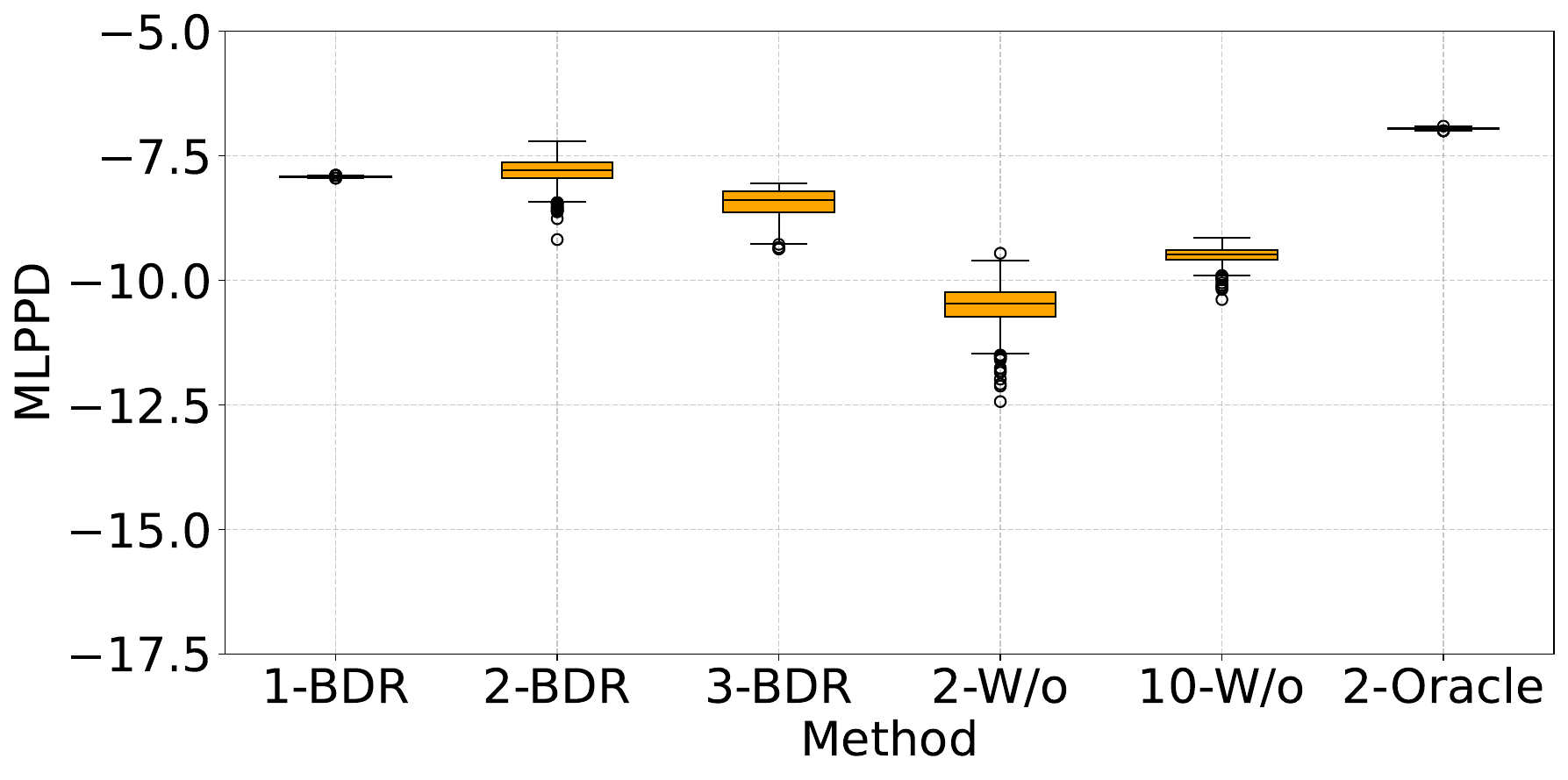}
        \caption{3-layer DGP at $n_{\text{train}}=400$}
        \label{fig:case2_2d_mlppd_400_l3}
    \end{subfigure}
    \caption{MLPPD for 2D input subspace at $n_{\text{train}}=240$ and $400$ based on the response surface of a generated exponential function.}
    \label{fig:case2_2d_mlppd_240_and_400}
\end{figure}

This exponential scenario defines a smooth yet highly nonlinear two-dimensional manifold with strong interactions between inputs. In this setting, the advantages of deeper architectures become more pronounced. Across both training sizes, the three-layer DGP(2)–BDR model achieves the lowest RMSPE, highest NSME, and largest BIC and predictive score, demonstrating its ability to capture complex curvature while maintaining calibrated predictive uncertainty. At \( n_{\text{train}} = 240 \), this model performs comparably to its oracle counterpart, indicating successful recovery of the underlying two-dimensional projection directions despite the substantial nonlinearity (see \Cref{tab:case2_2d}, \cref{fig:case2_2d_rmspe_240_and_400}, and \cref{fig:case2_2d_mlppd_240_and_400}). 

Increasing the training size to \( n_{\text{train}} = 400 \) leads to systematic improvements across all BDR models, particularly in NSME and CRPS. The two-layer DGP(2)–BDR model becomes increasingly competitive as its latent mappings stabilize with additional data. These gains are more pronounced than in the 1D piecewise setting, reflecting the higher intrinsic dimensionality and the smoother but more complex nonlinear structure. In contrast, models fitted without BDR perform poorly at both sample sizes, underscoring the difficulty of implicitly learning a 10D-to-2D mapping and the detrimental effects of unstructured high-dimensional inputs on calibration and generalization. As before, the oracle models mark the upper performance limit, and the close agreement between the BDR fits and these benchmarks confirms that the framework recovers the governing 2D latent manifold even under strong nonlinearity.

\textcolor{black}{The CP and ALCI results in \Cref{tab:case2_2d2} further corroborate the quality of posterior uncertainty quantification under the proposed framework. In particular, the BDR models with correctly specified subspace dimension achieve coverage probabilities near the nominal 95\% level across both training sizes, whereas models fitted without BDR show clear undercoverage and poorer interval behavior. The close correspondence between the BDR-based and Oracle results provides additional evidence that the learned projection accurately recovers the true two-dimensional manifold, even under substantial nonlinearity.}

Across these synthetic studies, the BDR-enhanced GP and DGP models consistently recover the correct low-dimensional structure, achieve \textbf{a well-calibrated} predictive accuracy close to oracle performance, and benefit substantially from larger training sizes. In contrast, models without BDR degrade significantly, highlighting the necessity of learning the projection matrix rather than relying on the ambient high-dimensional representation. These results demonstrate that the proposed framework is robust across nonlinear and discontinuous settings and scales effectively with intrinsic dimension, providing a strong foundation for applications to more complex high-dimensional problems where reliable subspace discovery and uncertainty quantification are essential.

\begin{table}[h!]
\centering
\begin{tabular}{|l|c|c|c|c|c|c|}
\hline
 \multicolumn{7}{ |c| }{$n=240$ with true W: $10 \times 2$}\\
  \hline
Method (D) & TC(mins) &  RMSPE &  NSME & CRPS & Score  & BIC\\
\hline 
 GP (1) BDR & 2641.13 & 0.4795  &  0.9035  & 0.5938  & 104.6915 & 615.28 \\
 GP (2) BDR & 3970.30  & 0.5302  & 0.8873  & 0.4549  & 63.3460  & 612.07 \\
 GP (3) BDR &  5009.01  & 0.5948  & 0.8291 & 0.5863  & 72.6261  & 608.65\\
 DGP 2-layer (1) BDR & 4382.70  & 0.4778  & 0.9079 & 0.4097  & 126.6638 & 616.90\\
 DGP 2-layer (2) BDR & 4570.02   &  0.4217  & 0.9192  & 0.4490  &  100.6576 & 618.34\\
 DGP 2-layer (3) BDR & 8221.47   & 0.5616  &  0.8449  & 0.5677  & 73.6139 & 610.71\\
 DGP 3-layer (1) BDR & 7185.66  &  0.4823  &  0.8937 & \textbf{0.3705}  & 87.8325 & 616.10 \\
  DGP 3-layer (2) BDR &  8536.20  &  \textbf{0.4045}  & \textbf{0.9240}  & 0.4178  & \textit{\textbf{128.5832}} & \textit{\textbf{619.20}}\\
  DGP 3-layer (3) BDR &  9270.45  & 0.5221  & 0.8895  &  0.5377 & 104.7551 & 615.37 \\
\hline
GP (10) W/o  &  2224.01   & 0.6761  & 0.8013  & 0.7710   & 80.3565  & 600.03\\
DGP 2-layer (2) W/o & 4058.97   & 0.7403 & 0.8005 & 0.7180  & 76.0480 & 605.25 \\
DGP 2-layer (10) W/o &  12007.44  & 0.6028  & 0.8159  &  0.6382  & 54.3635 & 603.22\\
DGP 3-layer (2) W/o &  7843.40  & 0.8005 & 0.8000  & 0.5523  & 59.9743 & 606.88\\
DGP 3-layer (10) W/o &  19026.30  & 0.6339  & 0.8096  &  0.6329  & 69.3009  & 601.48\\
\hline
GP (2) Oracle &  526.15  &  0.4344  & 0.9150  &  \textbf{0.4994} & 53.8692 & 617.46\\
DGP 2-layer (2) Oracle & 3113.28 &  0.4917  & 0.8929  & 0.5207  & 90.0178 & 616.95 \\
DGP 3-layer (2) Oracle & 5385.52  &  \textit{\textbf{0.3215}}  & \textit{\textbf{0.9371}}  & 0.5308  & \textbf{113.6072}  & \textit{\textbf{620.88}}\\
\hline
 \multicolumn{7}{ |c| }{$n=400$ with true W: $10 \times 2$}\\
 \hline
 GP (1) BDR  & 3585.62 &  0.1253  &  0.9269 & 0.4032  & 215.8007 & 747.33 \\
 GP (2) BDR  & 4376.22  &  0.1245  & 0.9371  & 0.3942  & 217.0940 & 750.00 \\
 GP (3) BDR & 6245.80   & 0.1529  &  0.9160 & 0.6416  & 232.9859 & 746.81 \\
 DGP 2-layer (1) BDR & 5149.60  &  0.1247  & 0.9403 &  0.3759 & 262.7613 &  748.95\\ 
 DGP 2-layer (2) BDR & 6425.10  &  \textbf{0.1049} & \textbf{0.9531}  &   0.3631  & 226.7972 & 752.67\\
 DGP 2-layer (3) BDR & 8954.07   & 0.1562  & 0.9083  & 0.5153  & 220.3793 & 745.93\\
 DGP 3-layer (1) BDR & 9737.80  &  0.1242  & 0.9407  & \textbf{0.2976}  & 239.7502 & 749.60\\
 DGP 3-layer (2) BDR &  13502.34  &  0.1144  & 0.9453 & 0.4846  & \textit{\textbf{294.6458}}  & \textbf{751.02} \\
 DGP 3-layer (3) BDR &  14930.00  & 0.2045 & 0.9069  & 0.5537  & 187.0360 &  743.48\\
 \hline
GP (10) W/o & 6381.02   & 0.3602  &  0.8500 & 0.5990  & 193.7770 & 743.56 \\
DGP 2-layer (2) W/o &  5143.22  & 0.2500  &  0.8867  & 0.7041  &  207.3369  & 746.73\\
DGP 2-layer (10) W/o & 15360.11    & 0.3232  & 0.8762  & 0.5696  & 154.0303 & 745.03\\
DGP 3-layer (2) W/o & 9305.46   & 0.5447  & 0.8629  & 0.5260  & 150.3455   & 740.58\\
DGP 3-layer (10) W/o &  21452.70   & 0.3387  &  0.8650 & 0.6109  & 133.6470 & 743.49\\
\hline
GP (2) Oracle & 600.53  &  0.1292   & 0.9201  & 0.2932  &  101.9608 & 745.39 \\
DGP 2-layer (2) Oracle & 4334.36  &  \textit{\textbf{0.1011}}  & \textit{\textbf{0.9603}}  &  \textit{\textbf{0.2529}}  &  228.7382  & \textbf{753.18} \\
DGP 3-layer (2) Oracle & 8050.99  &  0.1034  &  0.9550 & 0.4350   & \textbf{273.9225} &  751.04\\
\hline
\end{tabular}
\caption{Performance Metrics for the methods at $n_{train}=240$ and $400$ based on response surface of an exponential function with known structure. Table entries report, for each metric, the median value across posterior samples.}
\label{tab:case2_2d}
\end{table}

\begin{table}[t]
\centering
\begin{tabular}{|l|c|c|c|c|c|c|c|}
\hline
  \multicolumn{3}{ |c| }{$n=240$ with true W: $10 \times 2$} \\
\hline
Method (D) & CP & ALCI (95\%) \\
\hline
GP (1) BDR & 0.9329 & 0.4703 \\ 
GP (2) BDR & 0.9525 & 1.9222 \\
GP (3) BDR & 0.9049 & 0.8226 \\
DGP 2-layer (1) BDR & 0.9559 & 1.6426\\ 
DGP 2-layer (2) BDR & 0.9627 & 2.5319\\
DGP 2-layer (3) BDR & 0.9183 & 5.6908 \\
DGP 3-layer (1) BDR & 0.9541 & 0.7884\\
DGP 3-layer (2) BDR & 0.9423 & 1.0974\\
DGP 3-layer (3) BDR & 0.9118 & 5.3347\\
\hline
GP (10) W/o  & 0.5029 & 0.0161\\
DGP 2-layer (2) W/o & 0.5310 & 0.4722\\
DGP 2-layer (10) W/o & 0.5152 & 0.3196\\
DGP 3-layer (2) W/o & 0.5745 & 0.3467\\
DGP 3-layer (10) W/o & 0.5288 & 0.3995\\
\hline
GP (2) Oracle & 0.9423 & 0.4247 \\
DGP 2-layer (2) Oracle & 0.9572 & 1.2208\\
DGP 3-layer (2) Oracle & 0.9603 & 1.9750 \\
\hline
  \multicolumn{3}{ |c| }{$n=400$ with true W: $10 \times 2$} \\
\hline
GP (1) BDR & 0.9424 & 0.1792 \\ 
GP (2) BDR & 0.9629 & 0.3092 \\
GP (3) BDR & 0.9554 & 0.5969 \\
DGP 2-layer (1) BDR & 0.9641 & 0.6803\\ 
DGP 2-layer (2) BDR & 0.9655 & 0.3420\\
DGP 2-layer (3) BDR & 0.9290 & 4.5036\\
DGP 3-layer (1) BDR & 0.9560 & 0.5933\\
DGP 3-layer (2) BDR & 0.9657 & 0.9260\\
DGP 3-layer (3) BDR & 0.9485 & 4.5164\\
\hline
GP (10) W/o  & 0.6000 & 0.0119 \\
DGP 2-layer (2) W/o & 0.6923 & 0.2901\\
DGP 2-layer (10) W/o & 0.7808 & 0.1542\\
DGP 3-layer (2) W/o & 0.6667 & 0.1452\\
DGP 3-layer (10) W/o & 0.6548 & 0.1735\\
\hline
GP (2) Oracle & 0.9502 & 0.1050\\
DGP 2-layer (2) Oracle & 0.9616 & 0.3094\\
DGP 3-layer (2) Oracle & 0.9625 & 0.2000\\
\hline
\end{tabular}
\caption{Performance Metrics for the methods at $n_{train}=240$ and $400$ based on response surface of an exponential function with known structure. Table entries report, for each metric, the median value across posterior samples.}
\label{tab:case2_2d2}
\end{table}

\subsection{Forward and Adjoint Problem: Stochastic Elliptic Differential Equation}
\label{susec:application1}

We consider a stochastic elliptic PDE \cite{Constantine2016, gautier2021fullybayesian, TRIPATHY2016191} of the form
\begin{equation}
    \frac{\partial}{\partial \mathbf{x}}
    \left[
        a(\mathbf{x};\epsilon)\,
        \frac{\partial}{\partial \mathbf{x}}u(\mathbf{x};\epsilon)
    \right]
    = 1,
    \qquad \mathbf{x} \in [0,1]^2,
\end{equation}
subject to mixed boundary conditions,
\begin{equation}
\begin{aligned}
    & u(\mathbf{x};\epsilon) = 0, && \mathbf{x} \in \Gamma_1, \\
    & \frac{\partial}{\partial \mathbf{x}}u(\mathbf{x};\epsilon)\cdot \mathbf{n} = 0, && \mathbf{x} \in \Gamma_2,
\end{aligned}
\end{equation}
where $\Gamma_1$ denotes the left, top, and bottom boundaries (homogeneous Dirichlet), and $\Gamma_2$ denotes the right boundary (homogeneous Neumann).  

The log-conductivity field is modeled using a Karhunen–Loève (KL) expansion,
\begin{equation}
    a(\mathbf{x};\epsilon)
    = a_0 + \exp\!\left(
        \sigma \sum_{i=1}^L \epsilon_i \sqrt{\lambda_i}\, \phi_i(\mathbf{x})
    \right),
\end{equation}
where $\epsilon = (\epsilon_1,\dots,\epsilon_L)$ are independent uniform random variables on $[-1,1]$, and $\{(\lambda_i,\phi_i)\}$ are the eigenpairs of the covariance kernel. We represent the log-conductivity field \(\log a(\mathbf{x};\epsilon)\) using a Gaussian process with exponential correlation,
\begin{equation}
    C(\mathbf{x},\mathbf{x}';\beta)
    = \sigma^2
      \exp\!\left[-\frac{|x_1 - x_1'| + |x_2 - x_2'|}{\beta}\right],
      \qquad \beta > 0,
\end{equation}
where $\beta$ denotes the correlation length. We evaluate input-subspace recovery under two regimes: a long correlation length ($\beta = 1$) and a short one ($\beta = 0.01$). A truncation level of $L=100$ is used, yielding a $100$-dimensional parameter space.

The quantity of interest (QoI) is the average solution over the Neumann boundary,
\begin{equation}
    \text{QoI}(\epsilon)
    = \frac{1}{|\Gamma_2|}
      \int_{\Gamma_2} u(\mathbf{x};\epsilon)\, d\mathbf{x},
\end{equation}
where the closed-form solution of the PDE can be derived by differentiating twice:
\begin{equation}
    u(x;\epsilon)
    = u(0) 
    + \int_0^x
        \frac{a(0)u'(0) - s - 1}{a(s;\epsilon)}\, ds,
\end{equation}
with
\begin{equation}
    a(0)u'(0)
    =
    \frac{
        \displaystyle \int_0^1 \frac{s+1}{a(s;\epsilon)}\, ds
    }{
        \displaystyle \int_0^1 \frac{1}{a(s;\epsilon)}\, ds
    }.
\end{equation}

For this study, we set \( u(0)=0 \), \( a_0 = 0 \), and \( \sigma=1 \), and examine both correlation-length scenarios ($\beta = 1$ and $\beta = 0.01$). We use a training set of 270 noiseless observations and a test set of 30 points. The surrogate models employ a Matérn–$3/2$ separable covariance function.

\begin{figure}[h!]
    \centering

    \begin{subfigure}{0.49\textwidth}
        \centering
        \includegraphics[width=\linewidth]{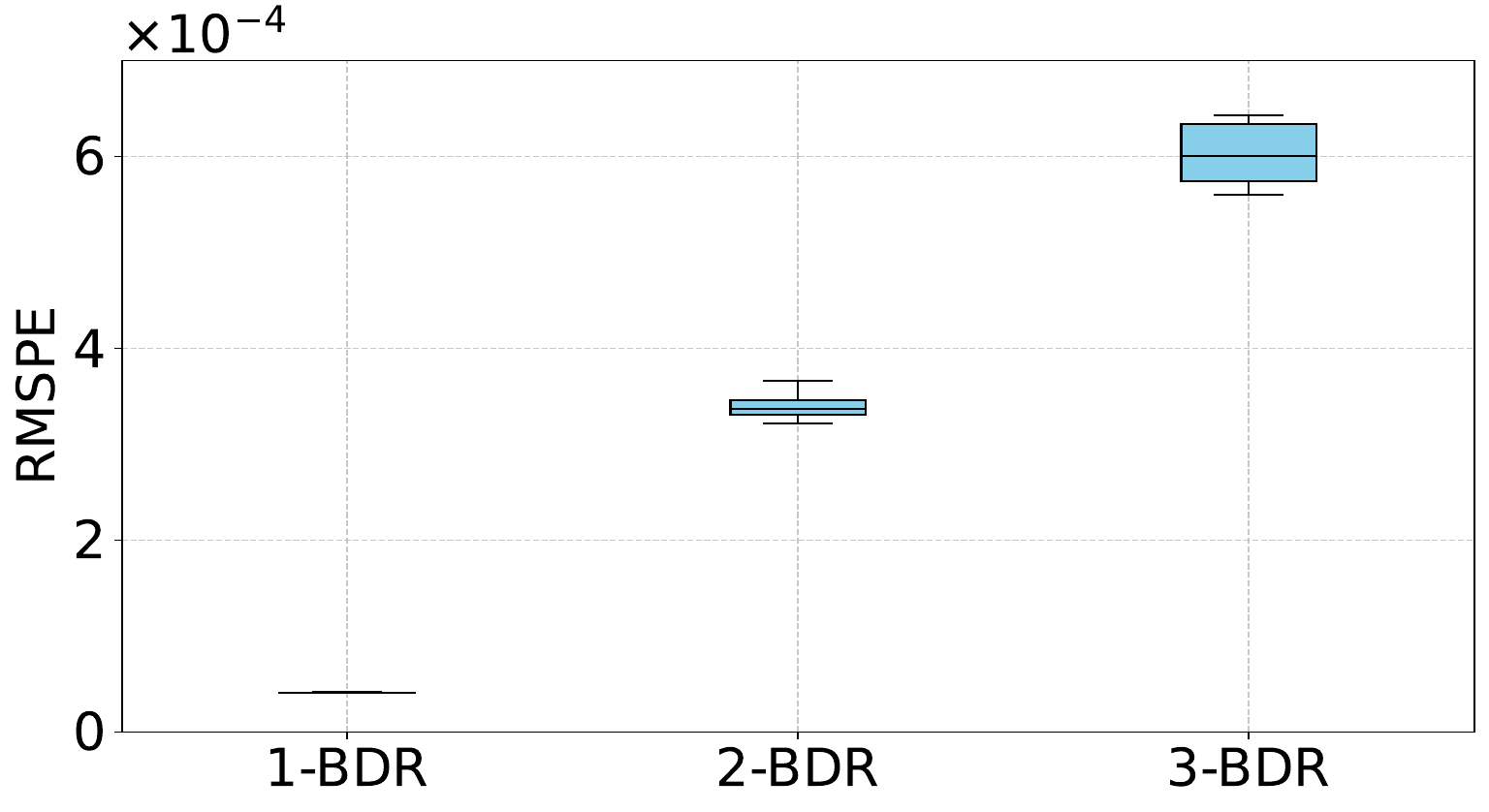}
        \caption{Standard GP at $\beta = 1$}
        \label{fig:1ld_matern_PDE_RMSPE_beta1}
    \end{subfigure}
    \hfill
    \begin{subfigure}{0.49\textwidth}
        \centering
        \includegraphics[width=\linewidth]{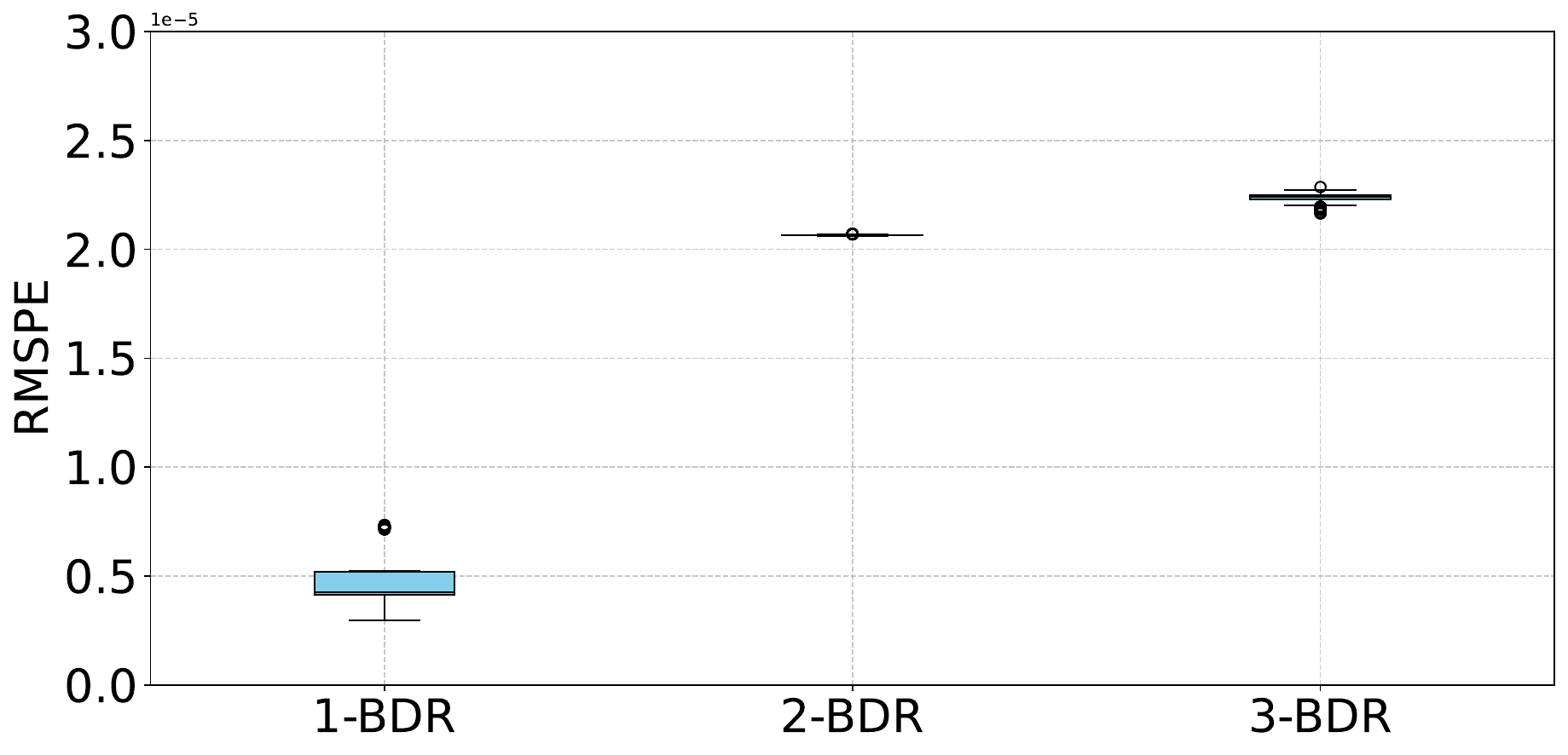}
        \caption{Standard GP at $\beta = 0.01$}
        \label{fig:1ld_matern_PDE_RMSPE_beta2}
    \end{subfigure}
    \begin{subfigure}{0.49\textwidth}
        \centering
        \includegraphics[width=\linewidth]{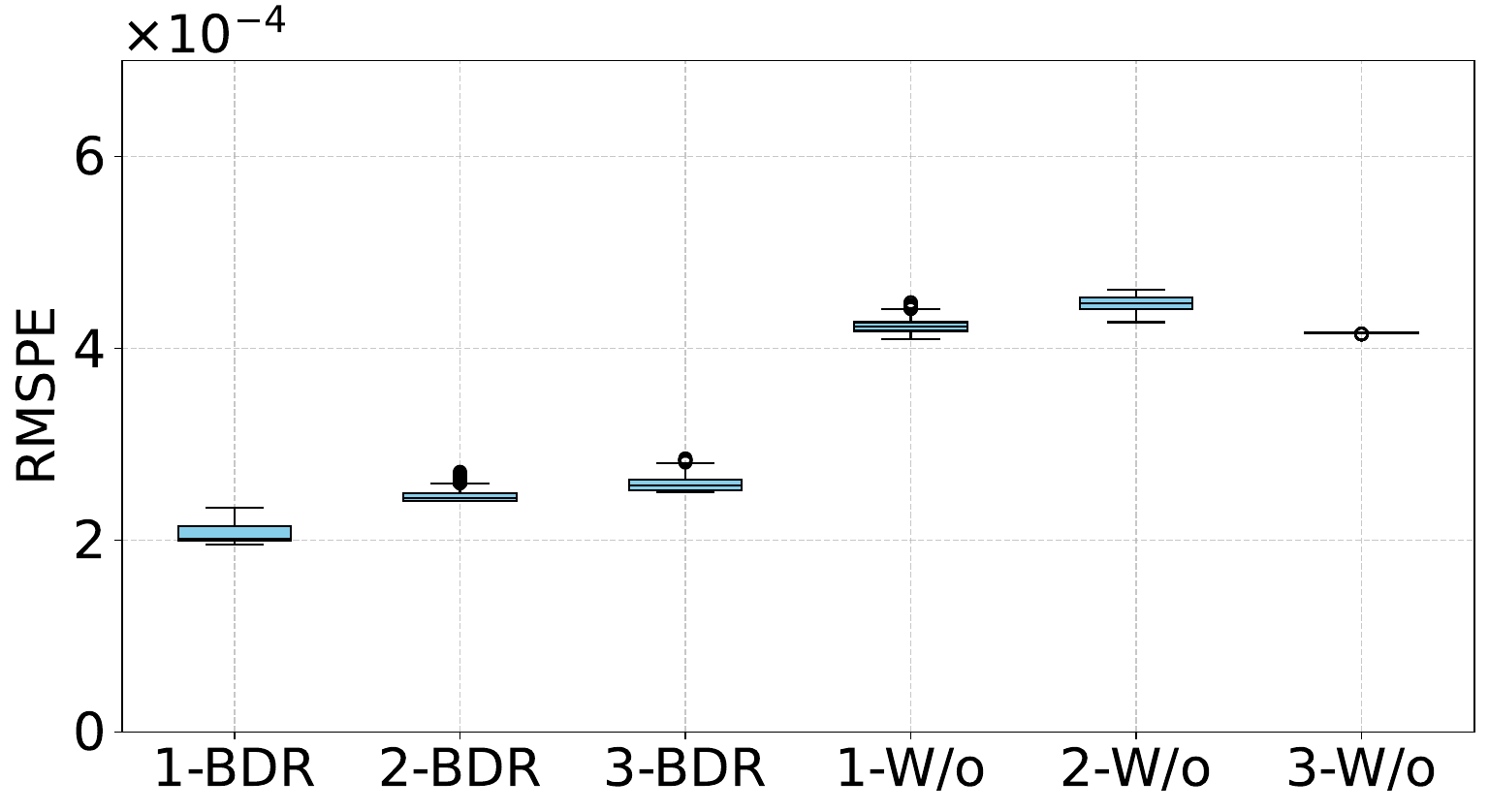}
        \caption{2-layer DGP at $\beta = 1$}
        \label{fig:2ld_matern_PDE_RMSPE_beta1}
    \end{subfigure}
    \hfill
    \begin{subfigure}{0.49\textwidth}
        \centering
        \includegraphics[width=\linewidth]{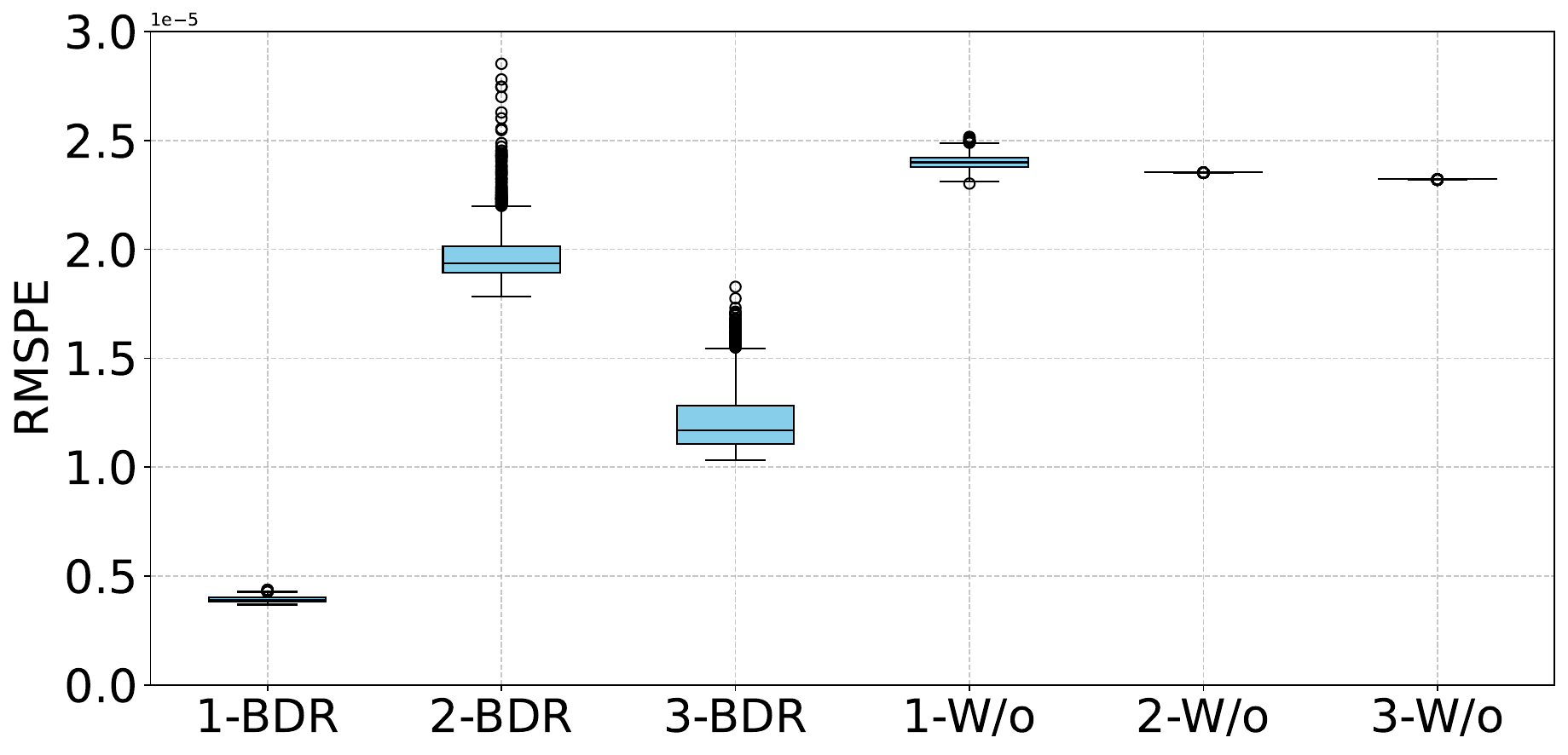}
        \caption{2-layer DGP at $\beta = 0.01$}
        \label{fig:2ld_matern_PDE_RMSPE_beta2}
    \end{subfigure}
    \begin{subfigure}{0.49\textwidth}
        \centering
        \includegraphics[width=\linewidth]{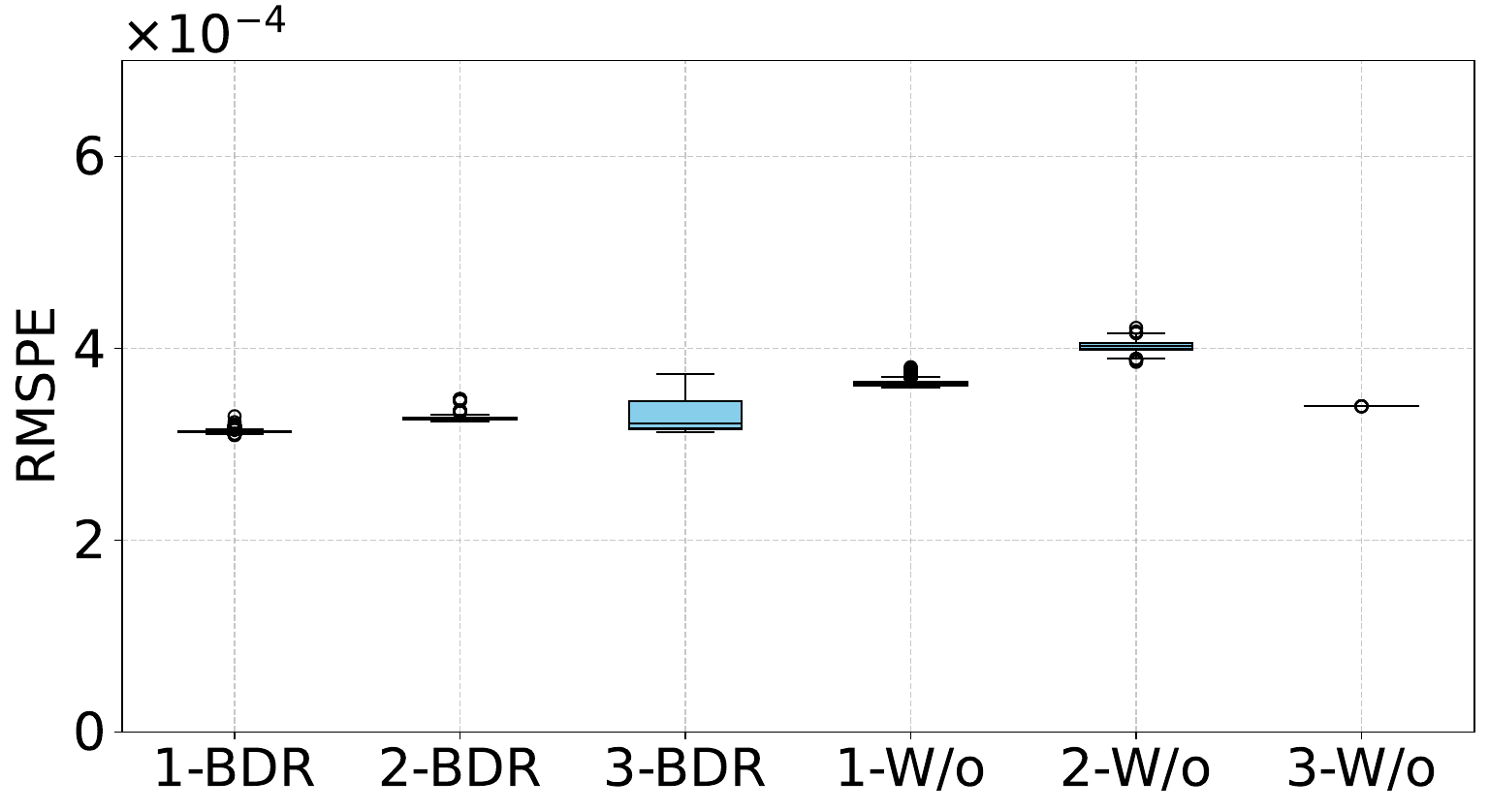}
        \caption{3-layer DGP at $\beta = 1$}
        \label{fig:3ld_matern_PDE_RMSPE_beta1}
    \end{subfigure}
    \begin{subfigure}{0.49\textwidth}
        \centering
        \includegraphics[width=\linewidth]{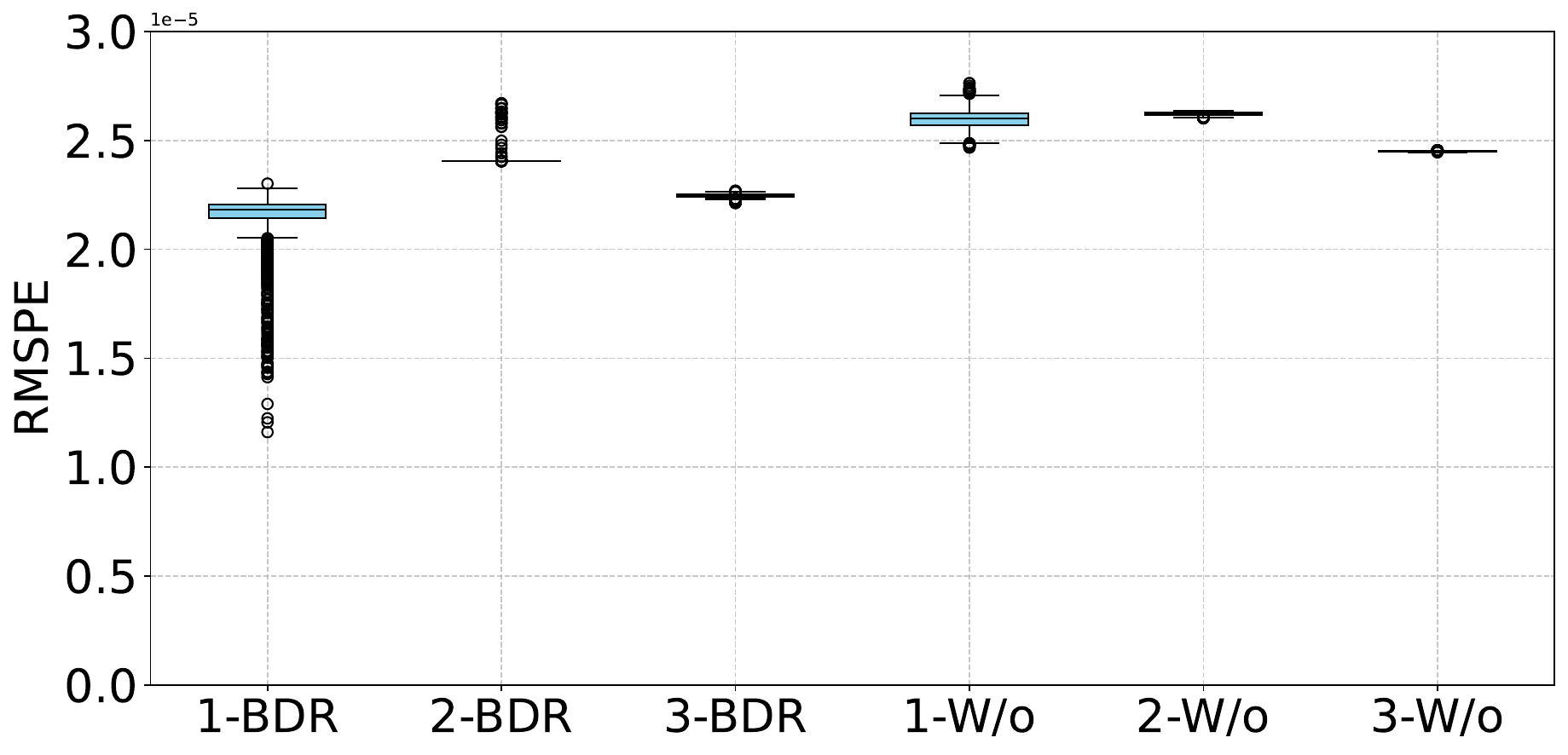}
        \caption{3-layer DGP at $\beta = 0.01$}
        \label{fig:3ld_matern_PDE_RMSPE_beta2}
    \end{subfigure}
    \caption{Elliptic PDE: RMSPE for BDR and W/o across different input subspaces and DGP layers, based on $\beta = 1$ and $\beta =0.01$.}
    \label{fig:matern_PDE_RMSPE_beta1_and_beta2}
\end{figure}

\begin{figure}[h!]
    \centering

    \begin{subfigure}{0.49\textwidth}
        \centering
        \includegraphics[width=\linewidth]{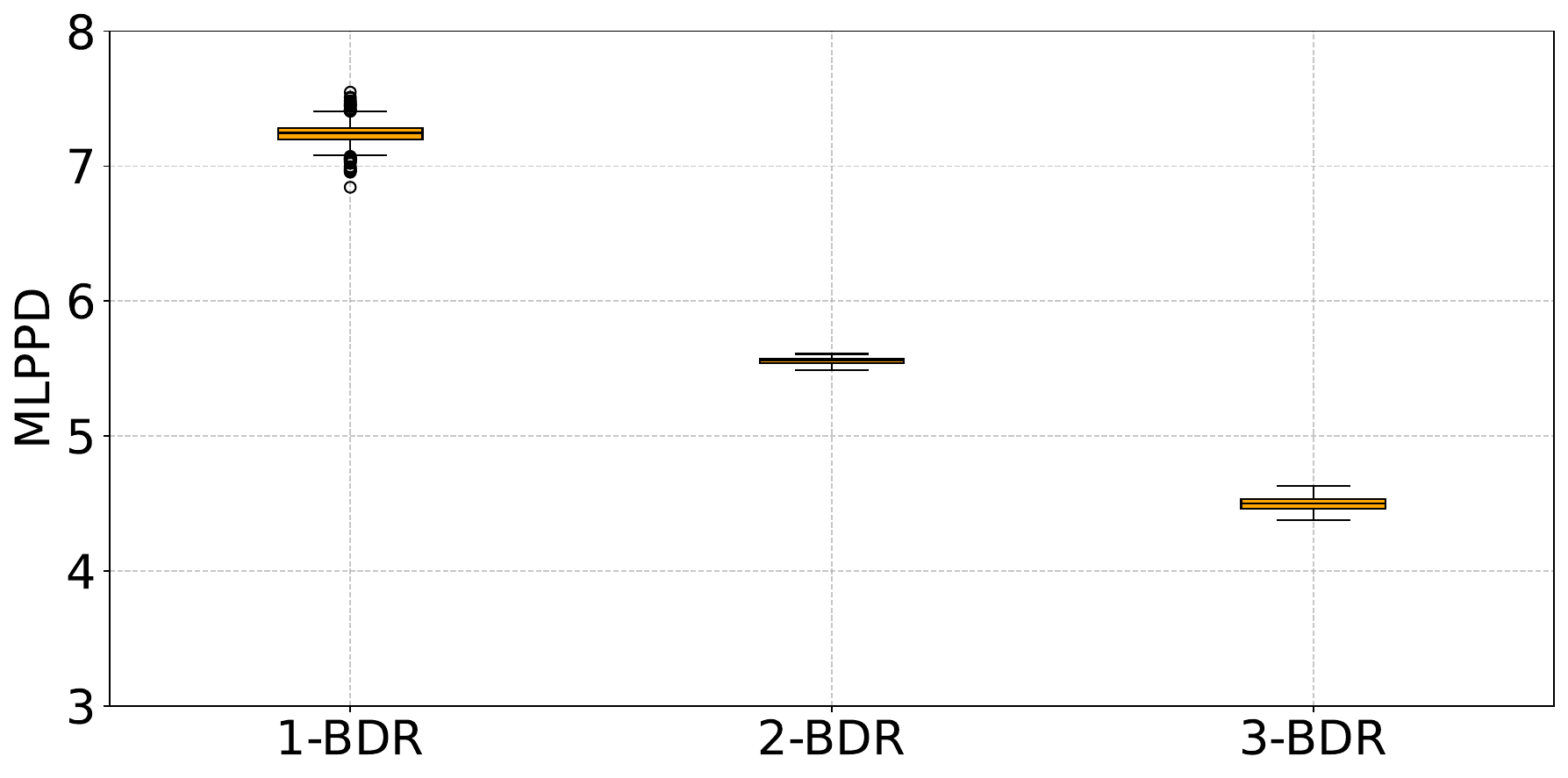}
        \caption{Standard GP at $\beta = 1$}
        \label{fig:1ld_matern_PDE_MLPPD_beta1}
    \end{subfigure}
    \hfill
    \begin{subfigure}{0.49\textwidth}
        \centering
        \includegraphics[width=\linewidth]{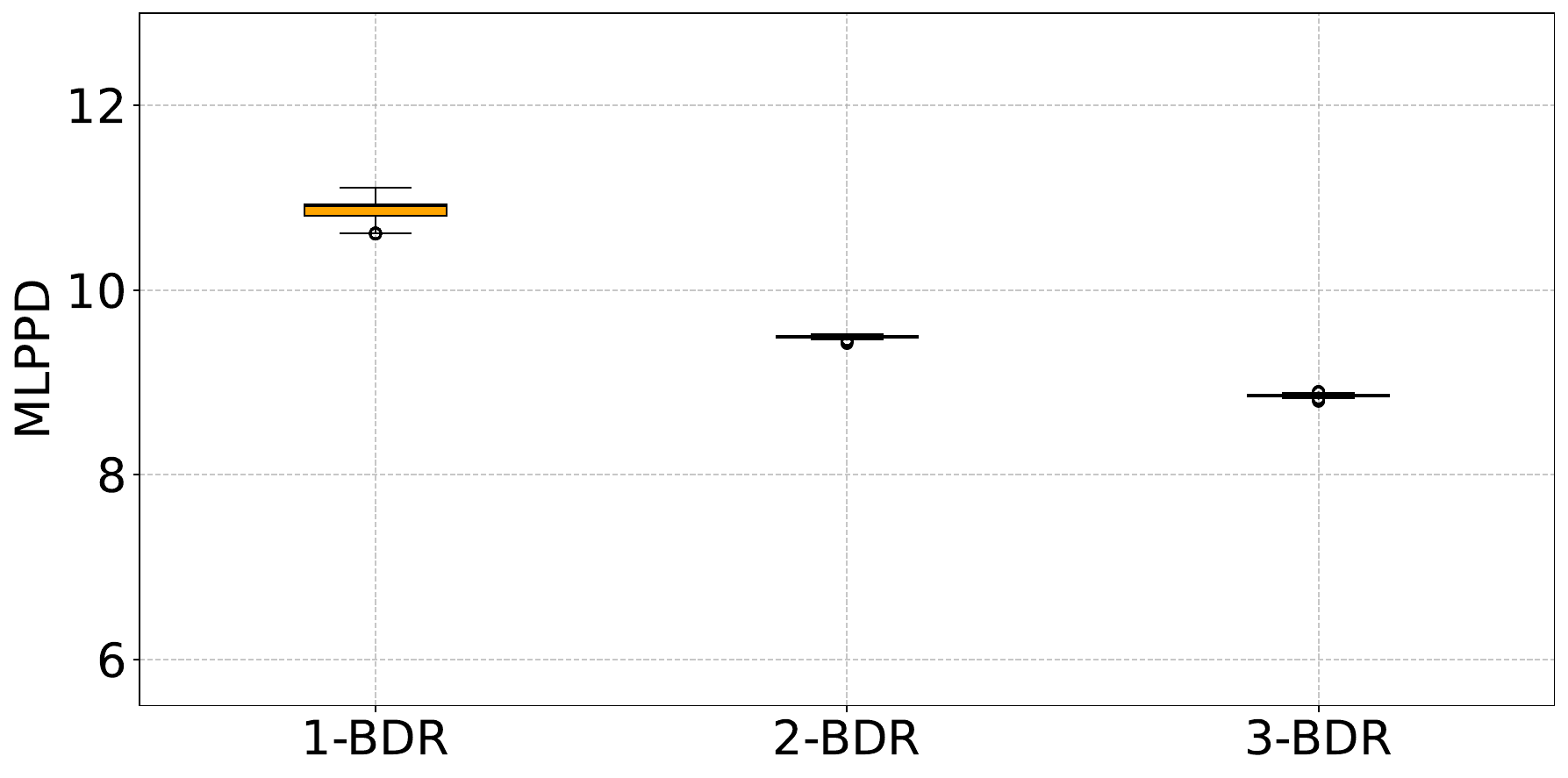}
        \caption{Standard GP at $\beta = 0.01$}
        \label{fig:1ld_matern_PDE_MLPPD_beta2}
    \end{subfigure}
    \begin{subfigure}{0.49\textwidth}
        \centering
        \includegraphics[width=\linewidth]{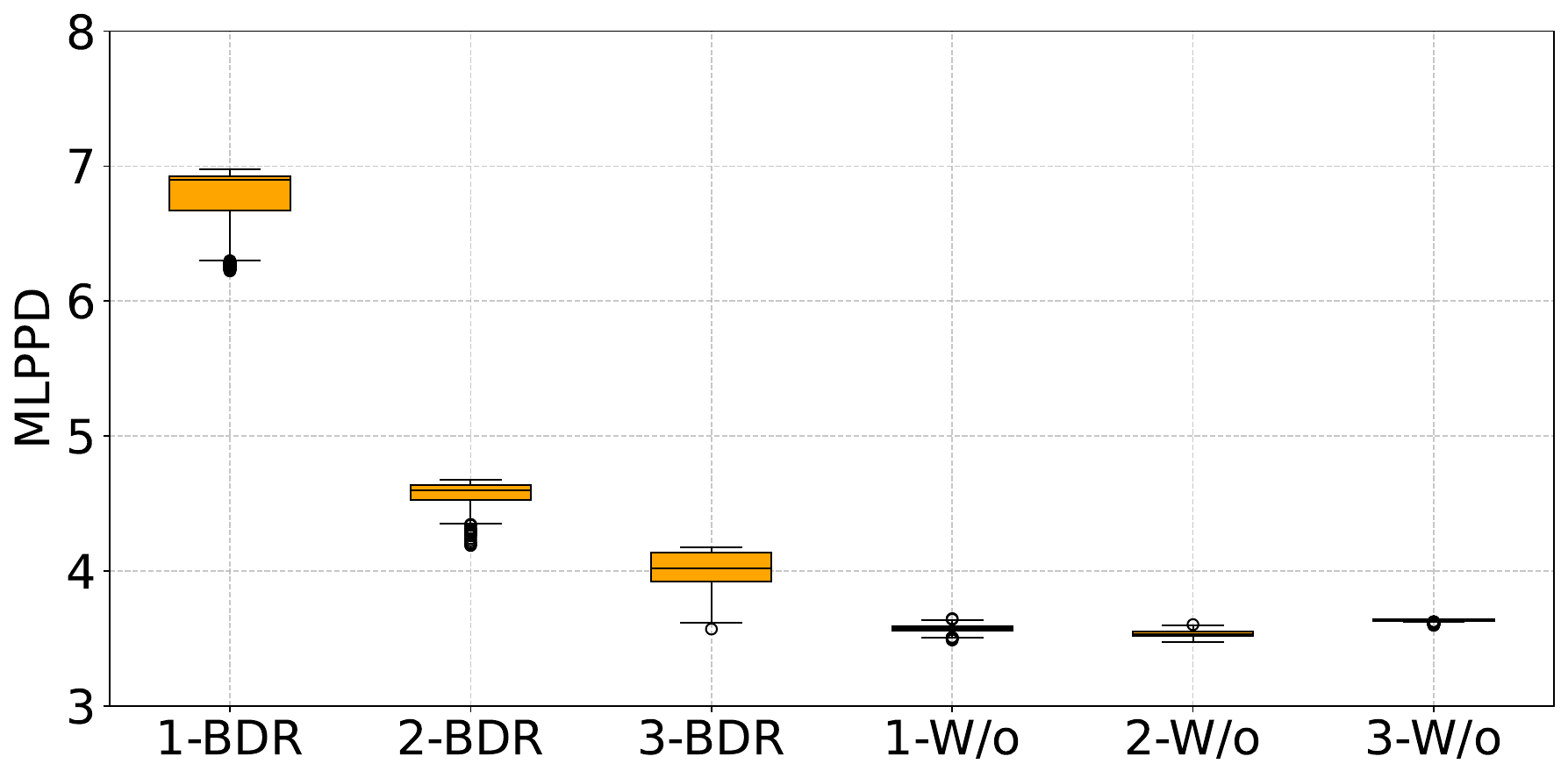}
        \caption{2-layer DGP at $\beta = 1$}
        \label{fig:2ld_matern_PDE_MLPPD_beta1}
    \end{subfigure}
    \hfill
    \begin{subfigure}{0.49\textwidth}
        \centering
        \includegraphics[width=\linewidth]{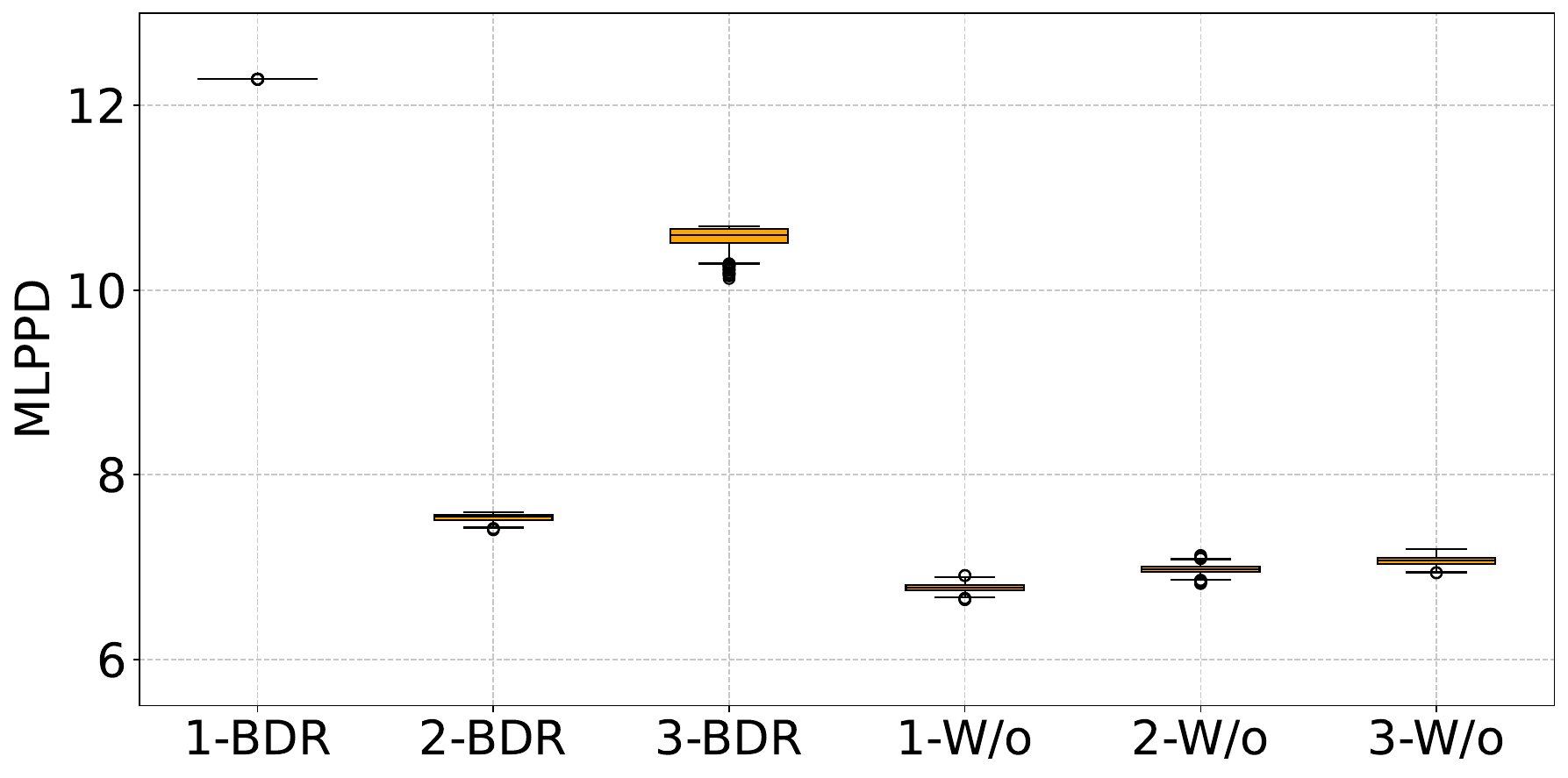}
        \caption{2-layer DGP at $\beta = 0.01$}
        \label{fig:2ld_matern_PDE_MLPPD_beta2}
    \end{subfigure}
    \begin{subfigure}{0.49\textwidth}
        \centering
        \includegraphics[width=\linewidth]{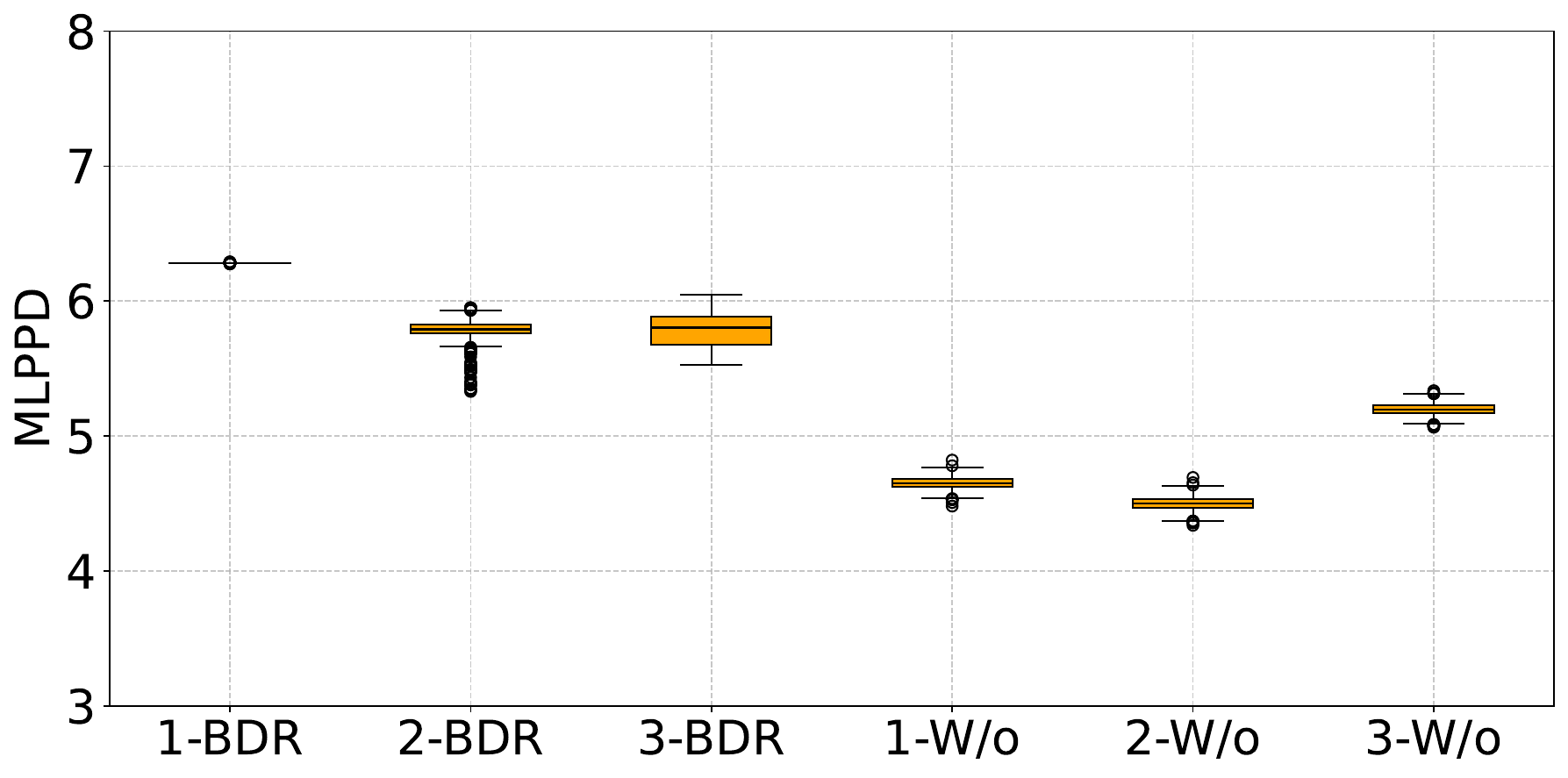}
        \caption{3-layer DGP at $\beta = 1$}
        \label{fig:3ld_matern_PDE_MLPPD_beta1}
    \end{subfigure}
    \begin{subfigure}{0.49\textwidth}
        \centering
        \includegraphics[width=\linewidth]{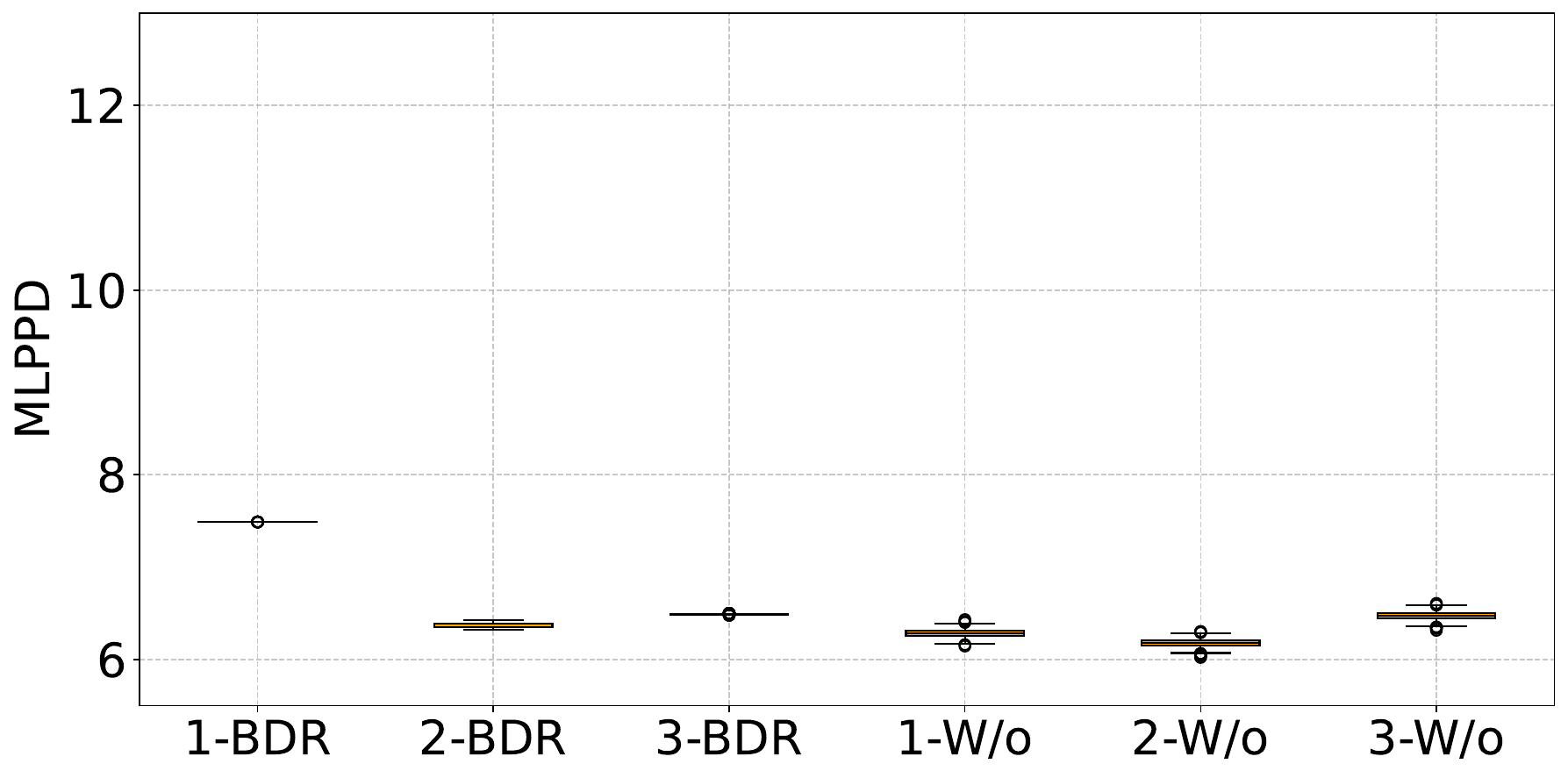}
        \caption{3-layer DGP at $\beta = 0.01$}
        \label{fig:3ld_matern_PDE_MLPPD_beta2}
    \end{subfigure}
    \caption{Elliptic PDE: MLPPD for BDR and W/o across different input subspaces and DGP layers, based on $\beta = 1$ and $\beta = 0.01$.}
    \label{fig:matern_PDE_MLPPD_beta1_and_beta2}
\end{figure}

The results, shown in \Cref{tab:matern_EPDE}, \Cref{fig:matern_PDE_RMSPE_beta1_and_beta2}, and \Cref{fig:matern_PDE_MLPPD_beta1_and_beta2}, demonstrate the effectiveness of the Bayesian GP and DGP BDR frameworks for solving stochastic elliptic PDEs with random coefficients. Across both correlation-length settings, the BDR models consistently outperform the W/o BDR variants in RMSPE, CRPS, and BIC.  

For the long correlation length ($\beta = 1$), the GP(1)–BDR achieves the strongest predictive performance, indicating that a single-layer surrogate with an estimated subspace is sufficient to capture the dominant input directions. For the short correlation length ($\beta = 0.01$), which induces stronger spatial fluctuations in the QoI, the two-layer DGP(1)–BDR attains the lowest RMSPE and CRPS, showing that additional depth is beneficial for modeling high-frequency variability. Models without BDR exhibit larger prediction errors and reduced NSME, confirming that learning the projection matrix $W$ is crucial for accurate dimension reduction and reliable uncertainty quantification. \textcolor{black}{The CP and ALCI results in \Cref{tab:matern_EPDE_T2} provide additional evidence regarding of posterior uncertainty quantification. For both $\beta=1$ and $\beta=0.01$, the BDR-based GP and DGP models yield coverage probabilities close to the nominal 95\% level together with relatively short credible intervals, while the corresponding models without BDR show clear undercoverage and less satisfactory interval behavior.}

Overall, the results show that the proposed Bayesian surrogates with BDR remain stable across both correlation regimes, with deeper architectures providing additional flexibility for more irregular fields. These findings highlight that the BDR-enhanced GP and DGP surrogates offer an accurate and computationally efficient approach for high-dimensional stochastic PDE problems, delivering strong predictive performance and well-calibrated uncertainty even under limited training data.

\begin{table}[h]
\centering
\begin{tabular}{|l|c|c|c|c|c|l|c|}
\hline
 \multicolumn{7}{ |c| }{$\beta = 1$} \\
 \hline
Method (D) &  TC(mins) & RMSPE & NSME  & CRPS & Score & BIC \\
\hline
 GP (1) BDR   & 423.50 & $\mathbf{4.08 \times 10^{-5}}$ & $\mathbf{0.9673}$  & $\mathbf{2.34 \times 10^{-5}}$  & 609.82 & $\mathbf{2549.53}$  \\
 GP (2) BDR   & 813.40 & 3.36$\times 10^{-4}$ & 0.8801  & 1.83$\times 10^{-4}$ & 494.87  &  1166.07 \\
 GP (3) BDR   & 1056.19 & 6.01$\times 10^{-4}$ & 0.8360  & 2.46$\times 10^{-4}$ & $\mathbf{677.244}$ &  937.30 \\
 DGP  2-layer (1) BDR   & 1468.12 & 2.01 $\times 10^{-4}$ & 0.8957 & 1.13$\times 10^{-4}$  &  598.70  &  2329.19  \\
 DGP  2-layer (2) BDR   & 2335.80  & 2.49$\times 10^{-4}$ &  0.8922 &  1.92 $\times 10^{-4}$&  495.01 &  1532.89   \\
 DGP  2-layer (3) BDR   & 2873.45 &2.73$\times 10^{-4}$  & 0.8903  & 3.42 $\times 10^{-4}$ &  525.02 &   652.48 \\
 DGP  3-layer (1) BDR   & 3554.24 & 3.13 $\times 10^{-4}$ & 0.8820  & 2.85$\times 10^{-4}$  & 413.14  & 1046.05  \\
 DGP  3-layer (2) BDR   & 5804.18 & 3.24 $\times 10^{-4}$ & 0.8805   & 3.5 $\times 10^{-4}$ & 417.98 &  663.80  \\ 
 DGP  3-layer (3) BDR   &  7966.48 & 3.20$\times 10^{-4}$ & 0.8811 & 1.07$\times 10^{-4}$ & 407.19 & 601.54  \\
 \hline
 DGP  2-layer (1) W/o   & 1063.04 & 4.22$\times 10^{-4}$ & 0.8508 & 1.78$\times 10^{-4}$ & 487.49 & 1930.27  \\
 DGP  2-layer (2) W/o   & 2307.08 & 4.47$\times 10^{-4}$ & 0.8443  & 1.89$\times 10^{-4}$ & 493.45 & 1130.02   \\
 DGP  2-layer (3) W/o   & 2840.31  & 4.16$\times 10^{-4}$  & 0.8547 & 1.71$\times 10^{-4}$ & 514.35 &  588.20   \\
 DGP  3-layer (1) W/o   & 3349.28 & 3.68$\times 10^{-4}$ & 0.8734 & 3.47$\times 10^{-4}$  & 397.47 &  457.72  \\
 DGP  3-layer (2) W/o   & 4849.56 & 4.06$\times 10^{-4}$  & 0.8609 & 3.56$\times 10^{-4}$ & 415.36 &  580.22  \\
 DGP  3-layer (3) W/o    & 5447.53 & 3.45$\times 10^{-4}$  & 0.8784  & 3.33$\times 10^{-4}$  & 400.62 &  578.69  \\
\hline
 \multicolumn{7}{ |c| }{$\beta = 0.01$} \\
 \hline
 GP (1) BDR   & 439.80  & 4.24 $\times 10^{-6}$  & 0.9993  & 4.68 $\times 10^{-6}$ &  $\mathbf{751.88}$ &  2715.69 \\
 GP  (2) BDR   & 647.14 & 2.15$\times 10^{-5}$ & 0.8892  & 1.25$\times 10^{-5}$ & 683.97 & 1955.60    \\
 GP (3) BDR   & 992.70 & 2.24 $\times 10^{-5}$ & 0.8836  & 1.29 $\times 10^{-5}$ &  694.31 &  1671.21   \\
 DGP  2-layer (1) BDR   & 1437.29 & $\mathbf{3.91 \times 10^{-6}}$ &  $\mathbf{0.9991}$ & $\mathbf{2.37 \times 10^{-6}}$ & 552.75 &  $\mathbf{2848.68}$\\
 DGP  2-layer (2) BDR   & 2540.56  & 1.94 $\times 10^{-5}$  & 0.8961  & 2.87 $\times 10^{-5}$ & 509.48 &  2534.31  \\
DGP  2-layer (3) BDR   & 3194.34 & 1.17 $\times 10^{-5}$  & 0.8985  & 2.78 $\times 10^{-5}$ & 513.10  &  2510.90   \\
DGP  3-layer (1) BDR   & 3606.64 & 2.18 $\times 10^{-5}$ & 0.8873  & 2.34 $\times 10^{-5}$ & 514.48  & 1586.93  \\
DGP  3-layer (2) BDR   & 4326.32 & 2.41$\times 10^{-5}$ & 0.8790  & 7.07$\times 10^{-5}$ & 506.22 &  570.84   \\
DGP  3-layer (3) BDR   & 5873.33  & 2.05$\times 10^{-5}$ &  0.8910  & 2.34 $\times 10^{-5}$ & 504.47 &  719.88  \\
\hline
DGP  2-layer (1) W/o   & 1000.59  & 2.40$\times 10^{-5}$ & 0.8796 & 6.48$\times 10^{-5}$ & 492.29 & 2215.53   \\
DGP  2-layer (2) W/o   & 2134.00  & 2.35$\times 10^{-5}$ & 0.8801 & 6.52$\times 10^{-5}$ & 491.93 & 2472.54  \\
DGP  2-layer (3) W/o    & 2780.21 & 2.32$\times 10^{-5}$ & 0.8821 & 7.22$\times 10^{-5}$ & 489.97 & 2471.72\\
DGP  3-layer (1) W/o   & 3195.27 & 2.60$\times 10^{-5}$ & 0.8682 & 6.46$\times 10^{-5}$ & 492.46 &  305.95  \\
DGP  3-layer (2) W/o   & 4220.45 & 2.62$\times 10^{-5}$ & 0.8676 & 6.50$\times 10^{-5}$ & 492.19 &  300.87  \\
DGP  3-layer (3) W/o   & 5759.80 & 2.45$\times 10^{-5}$ & 0.8724 & 6.52$\times 10^{-5}$ & 491.97 &  397.24  \\
 \hline
\end{tabular}
\caption{Elliptic PDE:Performance metrics for different BDR and W/o  layers with training and testing sizes $n_{train}=270$ and $n_{test}=30$,  respectively. Table entries report, for each metric, the median value across posterior samples.}
\label{tab:matern_EPDE}
\end{table}

\begin{table}[t]
\centering
\begin{tabular}{|l|c|c|c|c|c|c|c|}
\hline
  \multicolumn{3}{ |c| }{$\beta = 1$} \\
\hline
Method (D) & CP & ALCI (95\%)\\
\hline
GP (1) BDR & 0.9506 & \(4.09 \times 10^{-4}\)\\ 
GP (2) BDR & 0.9411 & \(5.13 \times 10^{-4}\)\\
GP (3) BDR & 0.9589 & \(2.27 \times 10^{-4}\)\\
DGP 2-layer (1) BDR & 0.9547 & \(5.25 \times 10^{-4}\)\\ 
DGP 2-layer (2) BDR & 0.9468 & \(6.40 \times 10^{-4}\)\\
DGP 2-layer (3) BDR & 0.9447 & \(3.45 \times 10^{-4}\) \\
DGP 3-layer (1) BDR & 0.9516 & \(8.54 \times 10^{-4}\)\\
DGP 3-layer (2) BDR & 0.9203 & \(7.13 \times 10^{-4}\)\\
DGP 3-layer (3) BDR & 0.9642 & \(9.64 \times 10^{-4}\)\\
\hline
DGP 2-layer (1) W/o & 0.5260 & \(8.29 \times 10^{-4}\)\\
DGP 2-layer (2) W/o & 0.4053 & \(7.21 \times 10^{-4}\)\\
DGP 2-layer (3) W/o & 0.3158 & \(9.30 \times 10^{-4}\)\\
DGP 3-layer (1) W/o & 0.6053 & \(6.76 \times 10^{-4}\)\\
DGP 3-layer (2) W/o & 0.6432 & \(5.45 \times 10^{-4}\)\\
DGP 3-layer (3) W/o & 0.6211 & \(8.80 \times 10^{-4}\)\\
\hline
  \multicolumn{3}{ |c| }{$\beta = 0.01$} \\
\hline
GP (1) BDR & 0.9615 & \(2.12 \times 10^{-5}\)\\ 
GP (2) BDR & 0.9539 & \(1.16 \times 10^{-5}\)\\
GP (3) BDR & 0.9684 & \(1.62 \times 10^{-5}\)\\
DGP 2-layer (1) BDR & 0.9608 & \(1.66 \times 10^{-5}\)\\ 
DGP 2-layer (2) BDR & 0.9542 & \(4.35 \times 10^{-5}\)\\
DGP 2-layer (3) BDR & 0.9895 & \(3.99 \times 10^{-5}\) \\
DGP 3-layer (1) BDR & 0.9614 & \(1.92 \times 10^{-5}\)\\
DGP 3-layer (2) BDR & 0.9200 & \(2.43 \times 10^{-5}\)\\
DGP 3-layer (3) BDR & 0.9842 & \(2.93 \times 10^{-5}\)\\
\hline
DGP 2-layer (1) W/o & 0.5667 & \(3.51 \times 10^{-5}\)\\
DGP 2-layer (2) W/o & 0.4800 & \(4.80 \times 10^{-5}\)\\
DGP 2-layer (3) W/o & 0.6579 & \(4.89 \times 10^{-5}\)\\
DGP 3-layer (1) W/o & 0.6207 & \(2.60 \times 10^{-5}\)\\
DGP 3-layer (2) W/o & 0.6667 & \(1.69 \times 10^{-5}\)\\
DGP 3-layer (3) W/o & 0.7245 & \(2.41 \times 10^{-5}\)\\
\hline
\end{tabular}
\caption{Elliptic PDE:Performance metrics for different BDR and W/o  layers with training and testing sizes $n_{train}=270$ and $n_{test}=30$,  respectively. Table entries report, for each metric, the median value across posterior samples.}
\label{tab:matern_EPDE_T2}
\end{table}

\subsection{ONERA M6 (Lift and Drag)}
\label{susec:application2}

We apply the proposed methodology to the ONERA–M6 wing design problem \cite{Lukaczyk_ONERA, gautier2021fullybayesian}, using lift and drag coefficients as outputs. The wing geometry is parameterized using a Free-Form Deformation (FFD) framework with \(m = 50\) design variables representing control-point displacements. An initial Bézier volume (FFD box) enclosing the wing is constructed, and points inside this volume are mapped via a trivariate Bernstein polynomial representation:
\begin{equation}
    X(u,v,w)
    = \sum_{i=0}^{L} \sum_{j=0}^{M} \sum_{k=0}^{N}
        P_{i,j,k}\, B_i^{L}(u) B_j^{M}(v) B_k^{N}(w),
\end{equation}
where \(0 \le u,v,w \le 1\) are parametric coordinates,  
\(B_i^{L}, B_j^{M}, B_k^{N}\) are Bernstein polynomials,  
and \(P_{i,j,k}\) denote the FFD control points.  
Deforming the FFD control points smoothly warps both the wing surface and the surrounding volume mesh.

The 50 design variables define perturbations to selected control points, enabling changes in wing thickness, twist, sweep, and other geometric features. The design space is the scaled hypercube 
\[
    \mathcal{X} = [-0.05,0.05]^{50},
\]
with bounds selected to ensure that most CFD runs converge. For each new geometry, mesh morphing is automated: the baseline half-wing mesh is partitioned for parallel execution, deformed according to the FFD mapping, and passed to the CFD solver. The workflow of \cite{Kenway2014} reads a matrix of design variables, applies the deformation, executes the CFD solve, and returns flow quantities for post-processing.

All aerodynamic analyses use the Euler solver from \cite{Kenway2014}. The flow is treated as inviscid and compressible, governed by the 3D Euler equations, with free-stream conditions Mach \(0.8395\) and angle of attack \(3.06^\circ\). These conditions produce transonic flow with a characteristic \(\lambda\)-shock on the upper surface, a dominant contributor to wave drag. For each geometry, the solver provides steady-state pressure fields and adjoint sensitivities. The lift and drag coefficients, \(C_L\) and \(C_D\), are computed by integrating pressure forces over the wing surface. The baseline wing yields \(C_L \approx 0.2864\) and \(C_D \approx 0.0118\), satisfying the lift requirement and serving as a reference for optimization.

To construct the surrogate modeling dataset, 297 wing geometries were sampled using Latin Hypercube Sampling in the 50-dimensional space. For each sample, the Euler solver evaluated the lift and drag, and the adjoint solver computed sensitivities with respect to the design variables. A subset of 250 samples was used for training and 47 for testing. For additional background on the ONERA–M6 benchmark, see \cite{Lukaczyk_ONERA, Kenway2014, gautier2021fullybayesian}.

\begin{figure}[h!]
    \centering

    \begin{subfigure}{0.49\textwidth}
        \centering
        \includegraphics[width=\linewidth]{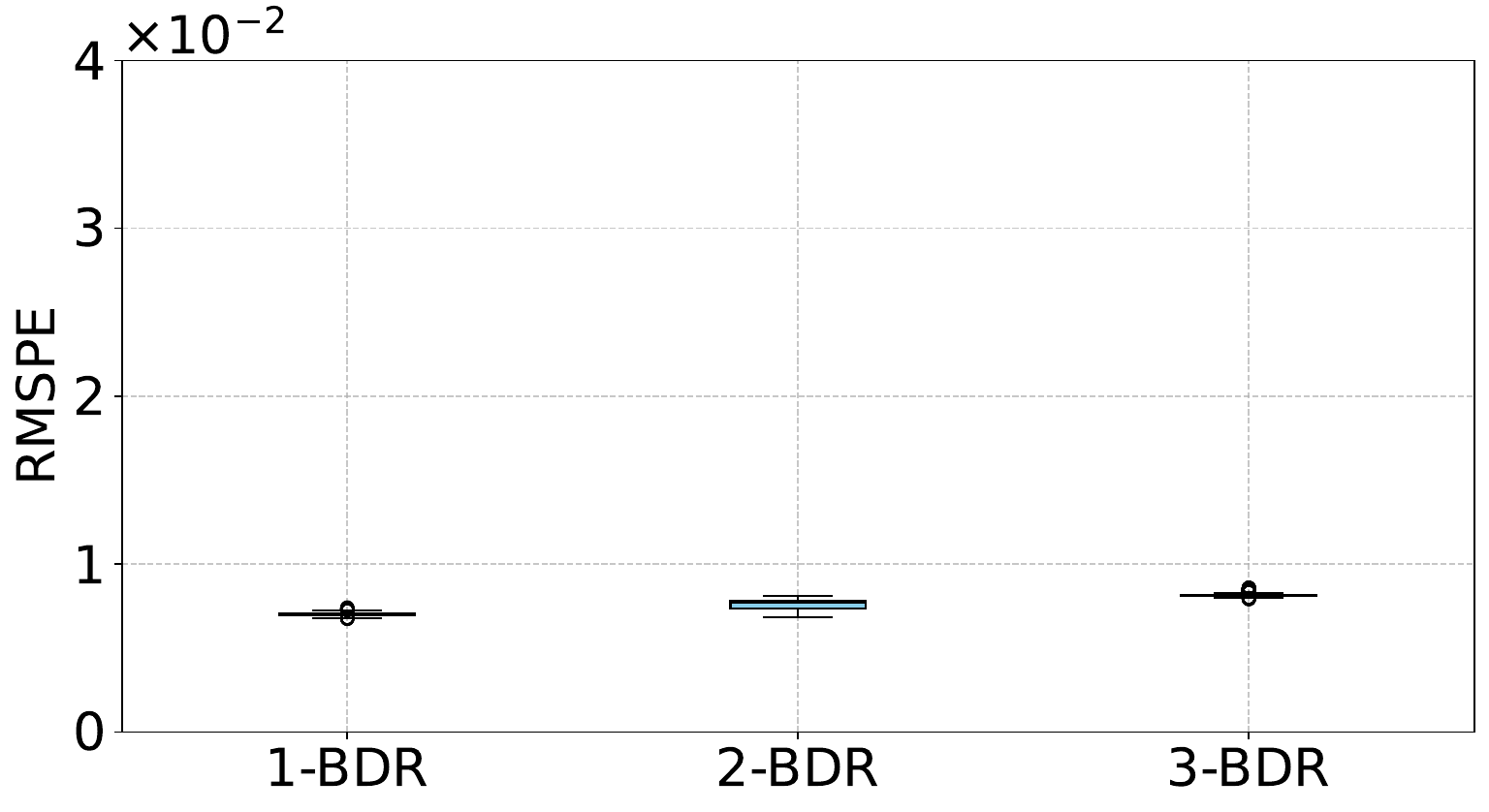}
        \caption{Standard GP for ONERA M6 (drag)}
        \label{fig:1ld_drag_RMSPE}
    \end{subfigure}
    \hfill
    \begin{subfigure}{0.49\textwidth}
        \centering
        \includegraphics[width=\linewidth]{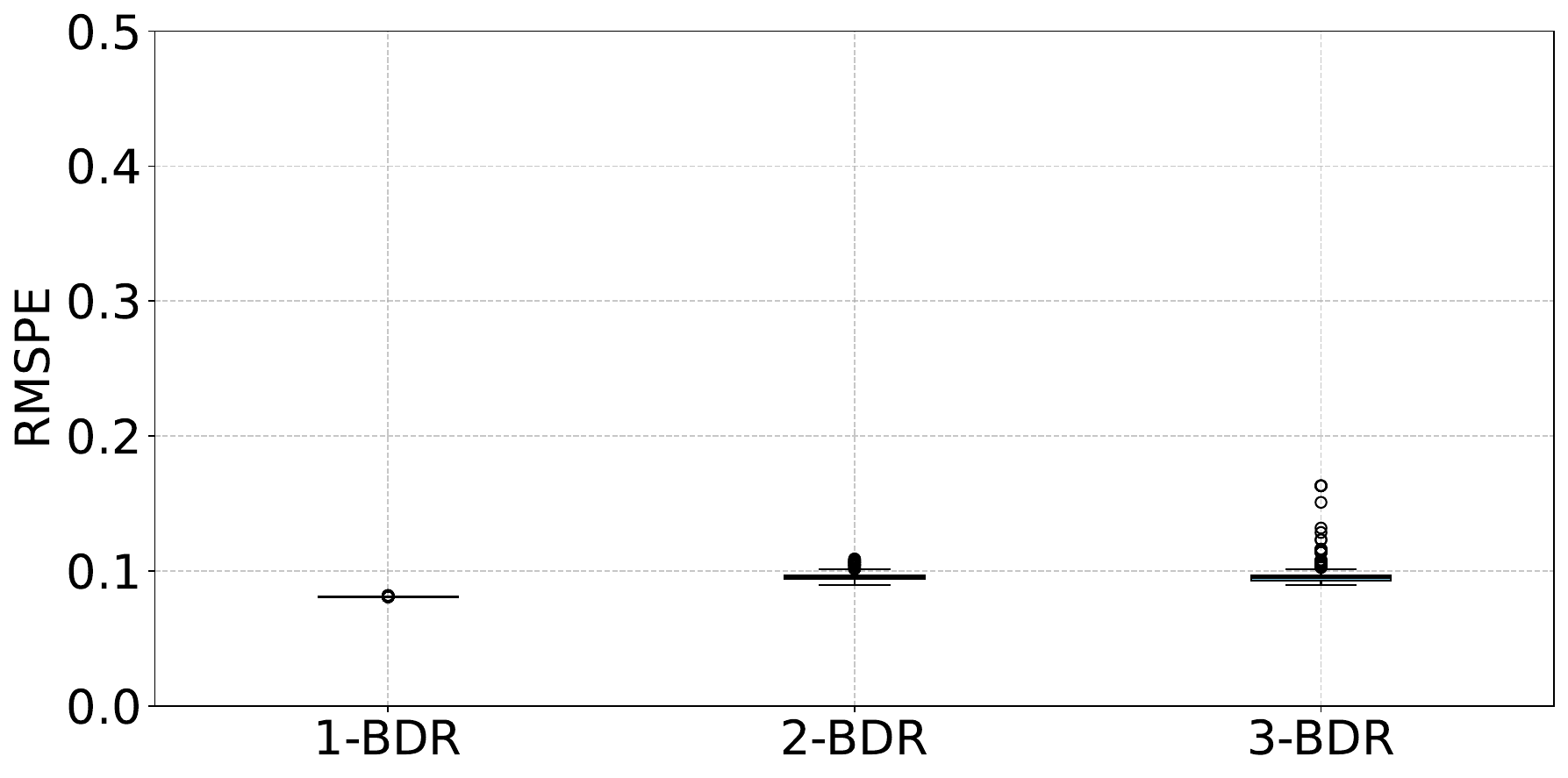}
        \caption{Standard GP for ONERA M6 (lift) }
        \label{fig:1ld_lift_RMSPE}
    \end{subfigure}
    \begin{subfigure}{0.49\textwidth}
        \centering
        \includegraphics[width=\linewidth]{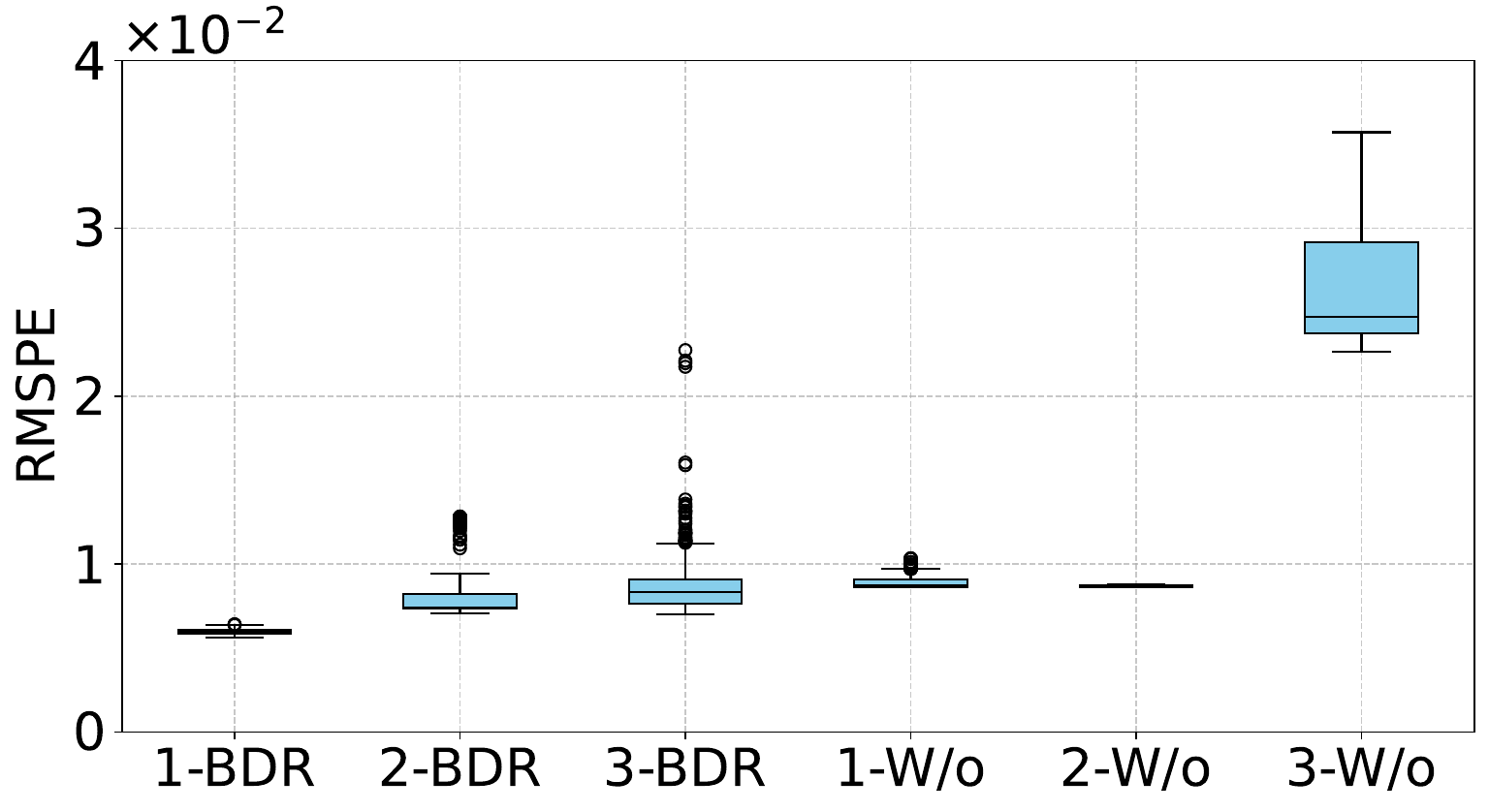}
        \caption{2-layer DGP for ONERA M6 (drag) }
        \label{fig:2ld_drag_RMSPE}
    \end{subfigure}
    \hfill
    \begin{subfigure}{0.49\textwidth}
        \centering
        \includegraphics[width=\linewidth]{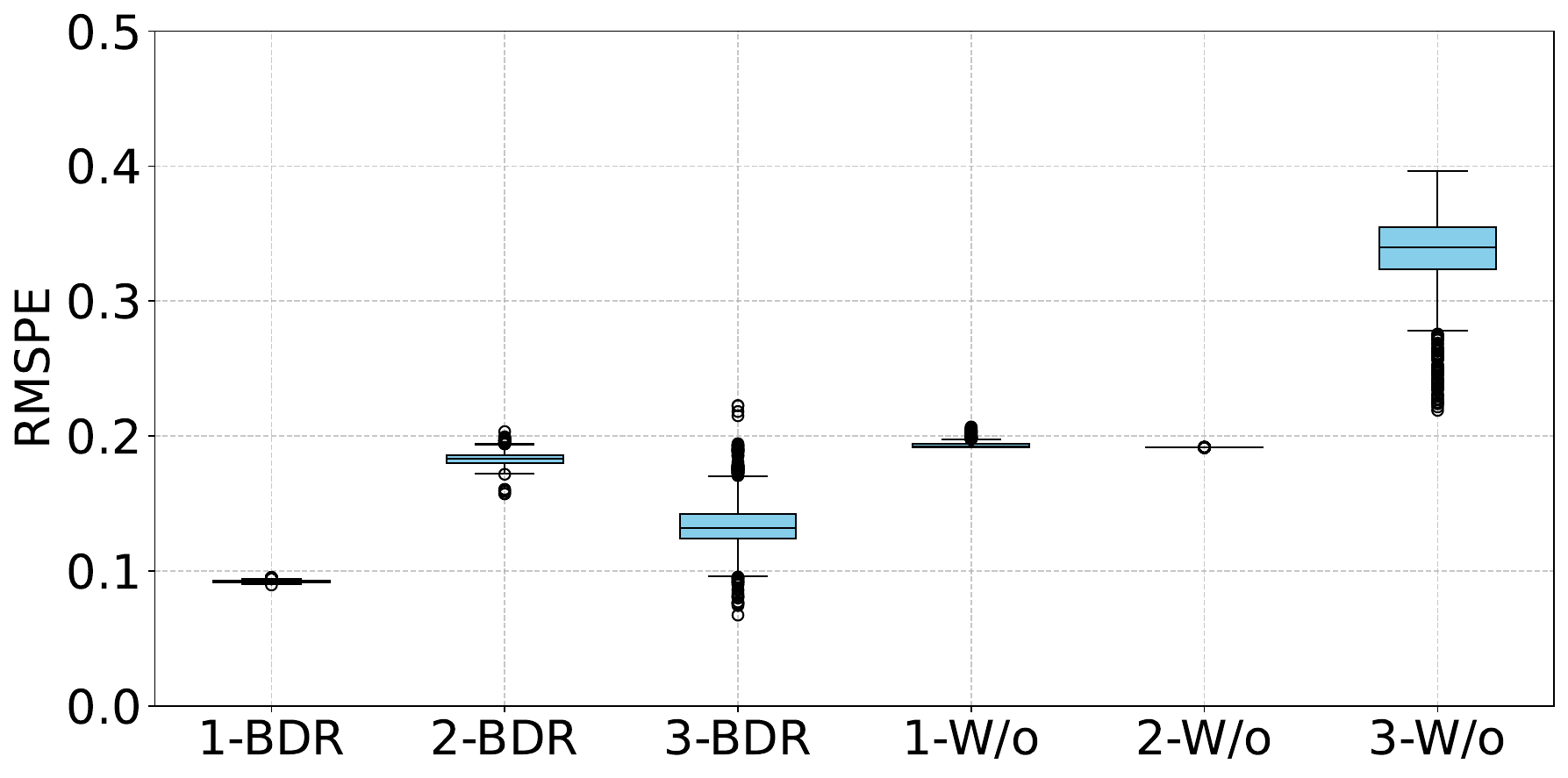}
        \caption{2-layer DGP for ONERA M6 (lift) }
        \label{fig:2ld_lift_RMSPE}
    \end{subfigure}
    \begin{subfigure}{0.49\textwidth}
        \centering
        \includegraphics[width=\linewidth]{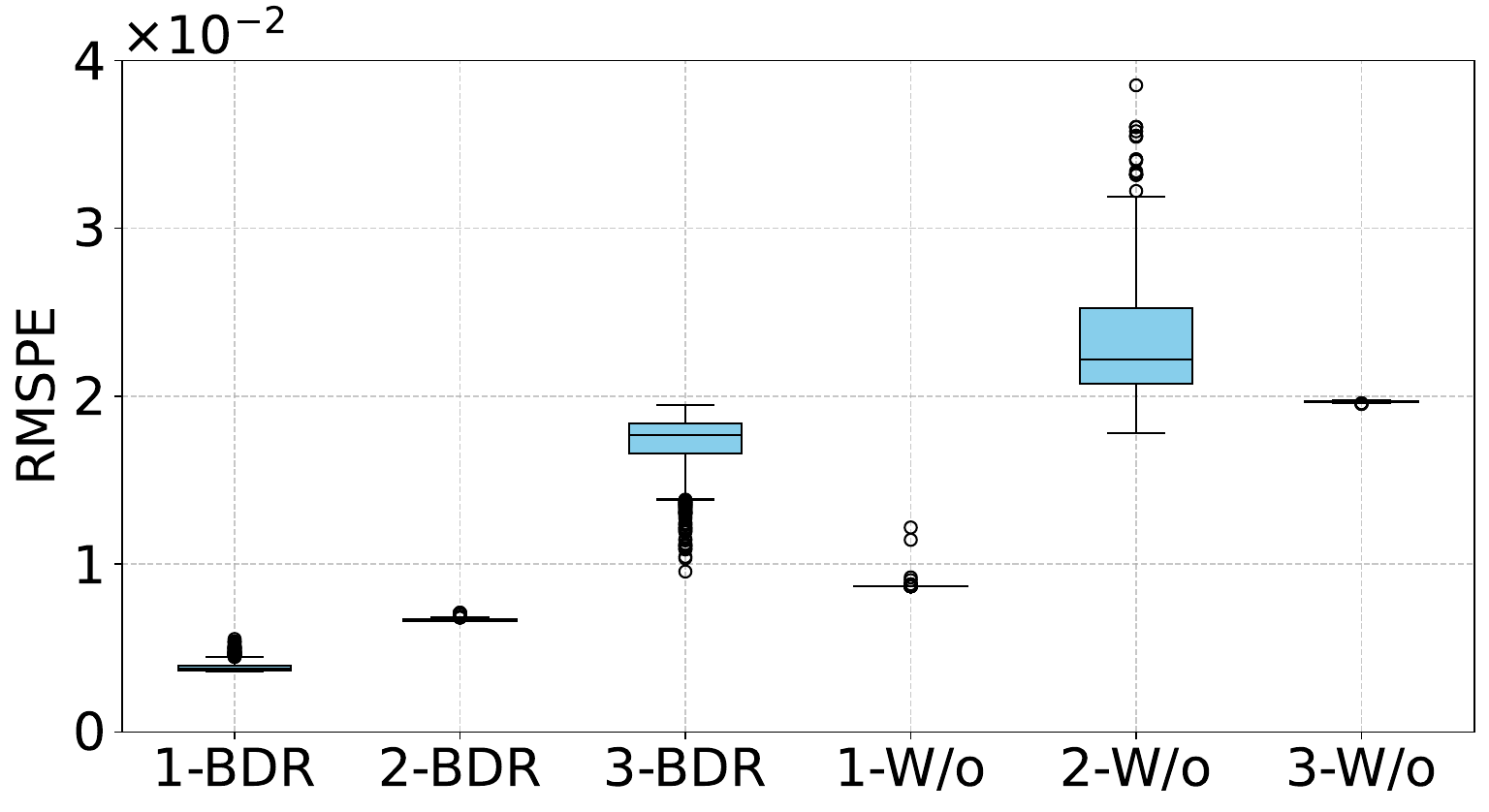}
        \caption{3-layer DGP for ONERA M6 (drag) }
        \label{fig:3ld_drag_RMSPE}
    \end{subfigure}
    \begin{subfigure}{0.49\textwidth}
        \centering
        \includegraphics[width=\linewidth]{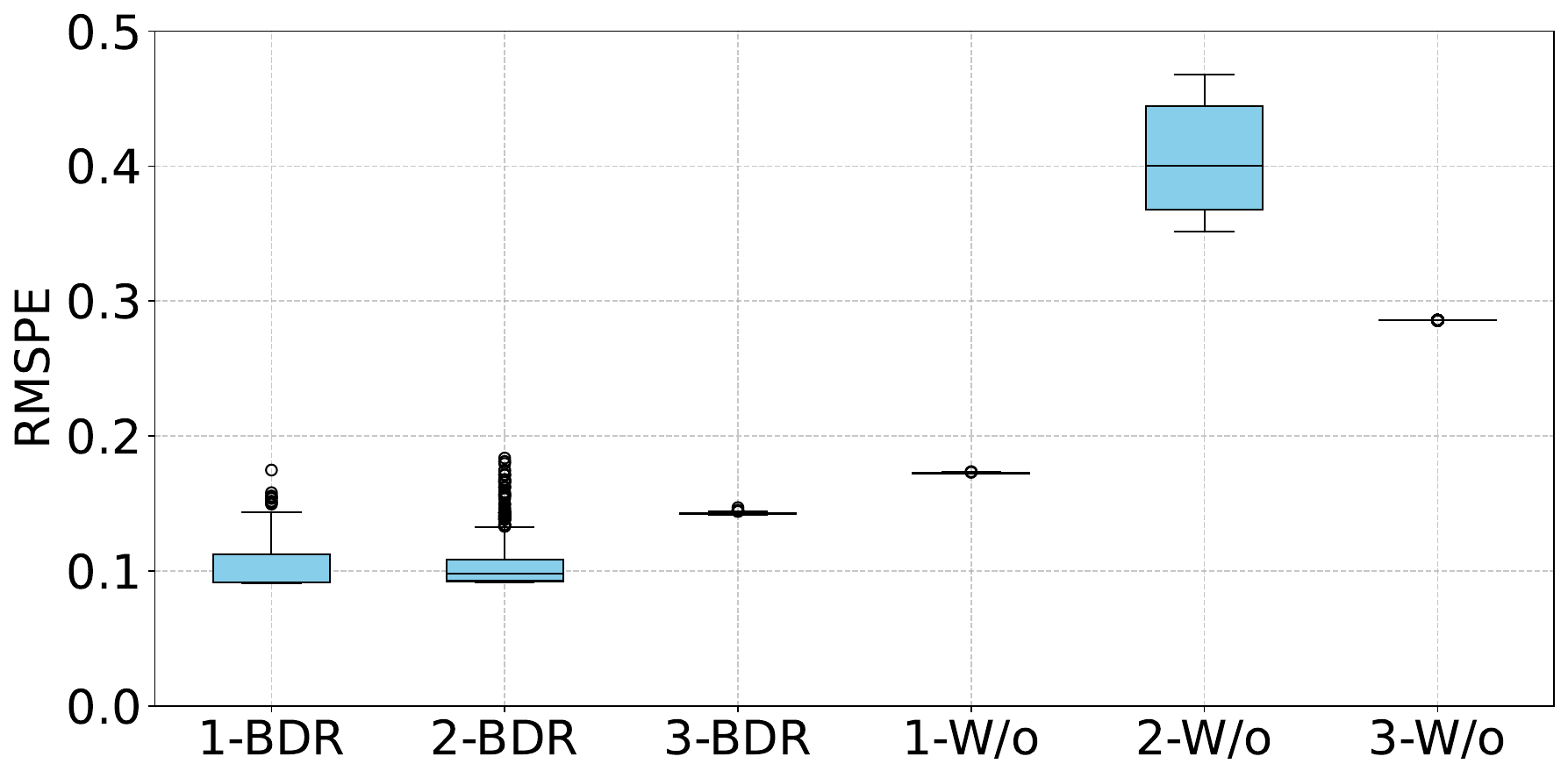}
        \caption{3-layer DGP for ONERA M6 (lift) }
        \label{fig:3ld_lift_RMSPE}
    \end{subfigure}
    \caption{ONERA M6 (drag and lift): RMSPE for BDR and W/o across different input subspaces and DGP layers.}
    \label{fig:ONERA_drag_and_lift_RMSPE}
\end{figure}

\begin{figure}[h!]
    \centering

    \begin{subfigure}{0.49\textwidth}
        \centering
        \includegraphics[width=\linewidth]{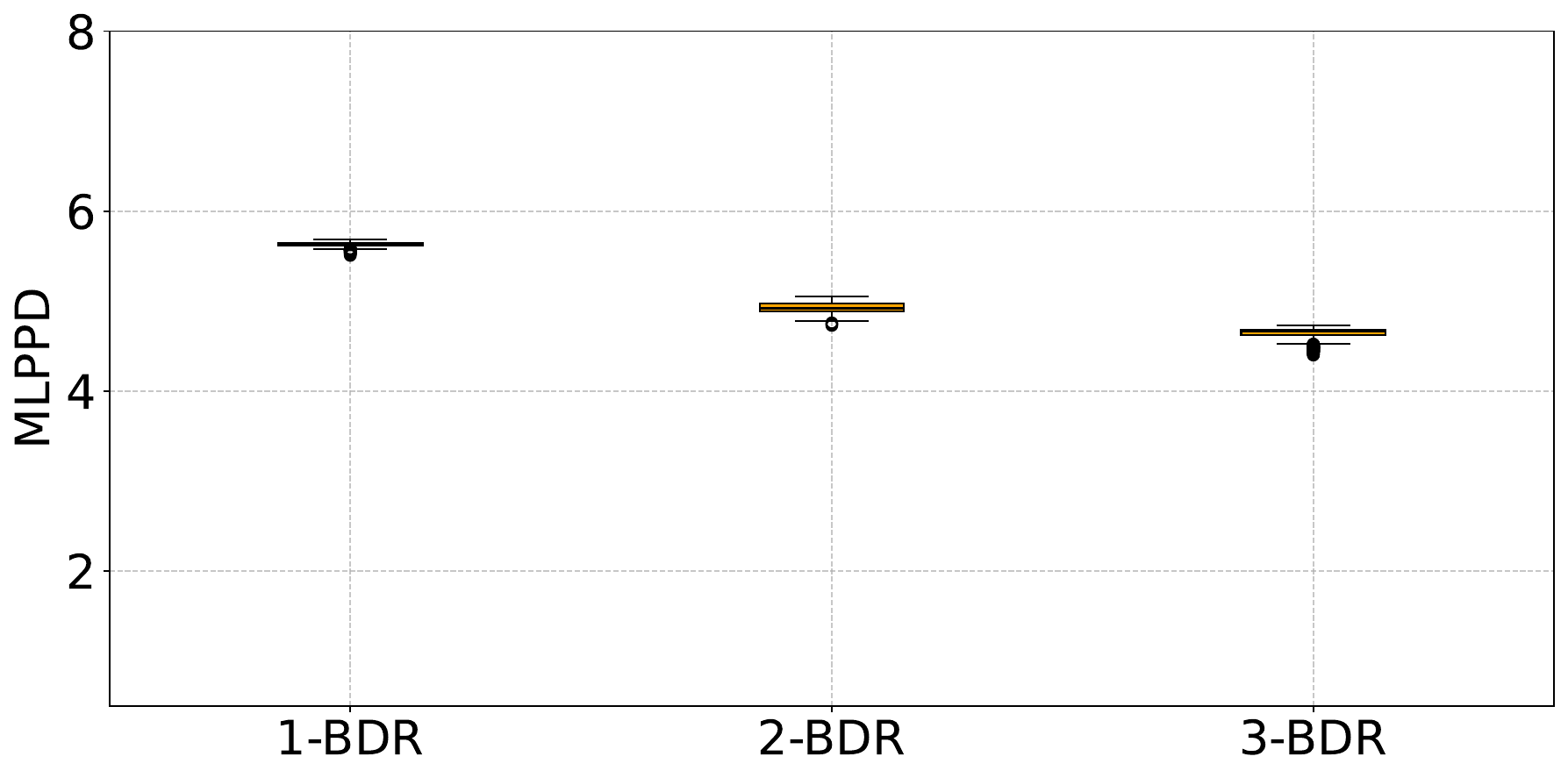}
        \caption{Standard GP for ONERA M6 (drag)}
        \label{fig:1ld_drag_MLPPD}
    \end{subfigure}
    \hfill
    \begin{subfigure}{0.49\textwidth}
        \centering
        \includegraphics[width=\linewidth]{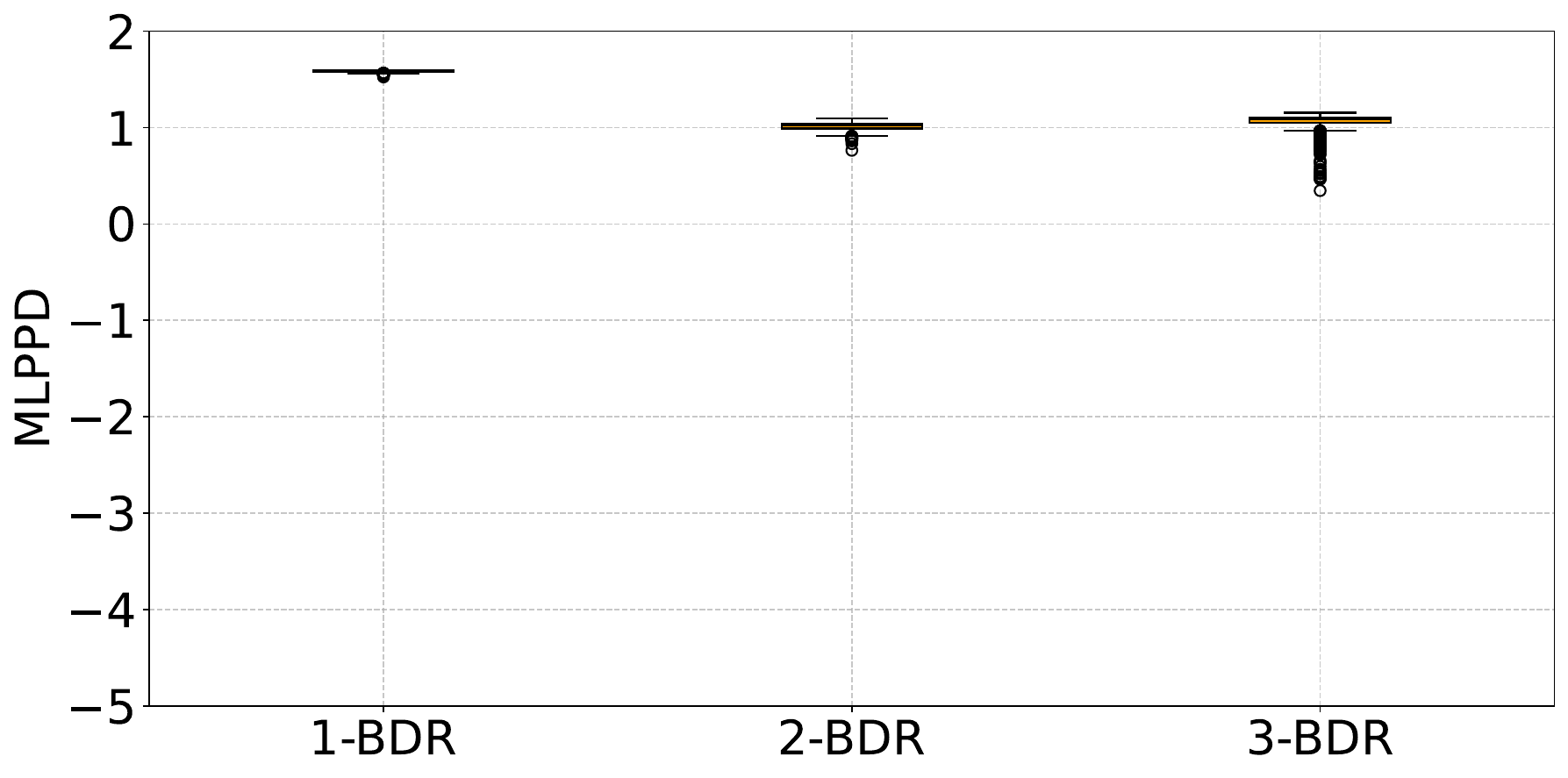}
        \caption{Standard GP for ONERA M6 (lift)}
        \label{fig:1ld_lift_MLPPD}
    \end{subfigure}
    \begin{subfigure}{0.49\textwidth}
        \centering
        \includegraphics[width=\linewidth]{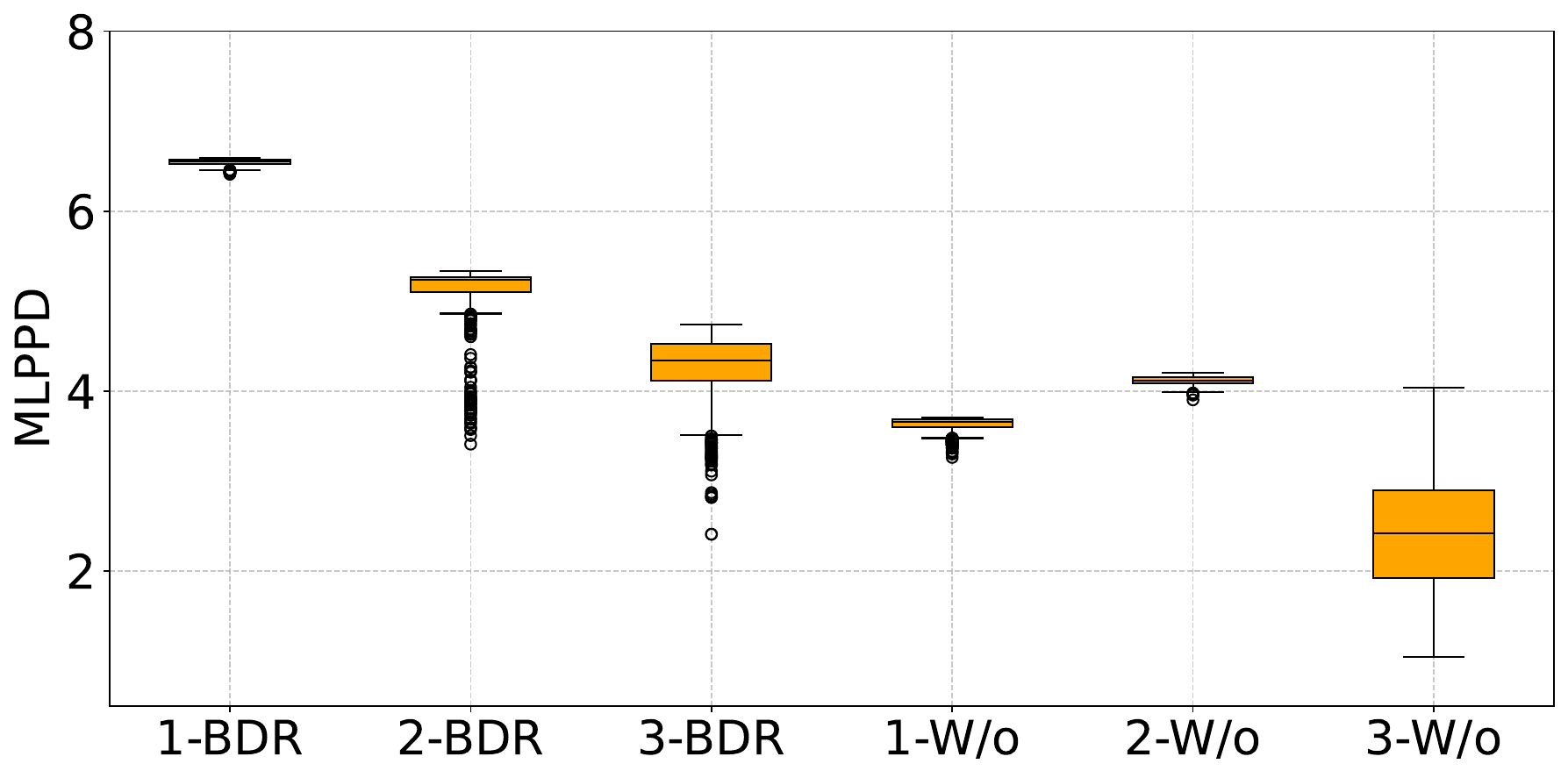}
        \caption{2-layer DGP for ONERA M6 (drag)}
        \label{fig:2ld_drag_MLPPD}
    \end{subfigure}
    \hfill
    \begin{subfigure}{0.49\textwidth}
        \centering
        \includegraphics[width=\linewidth]{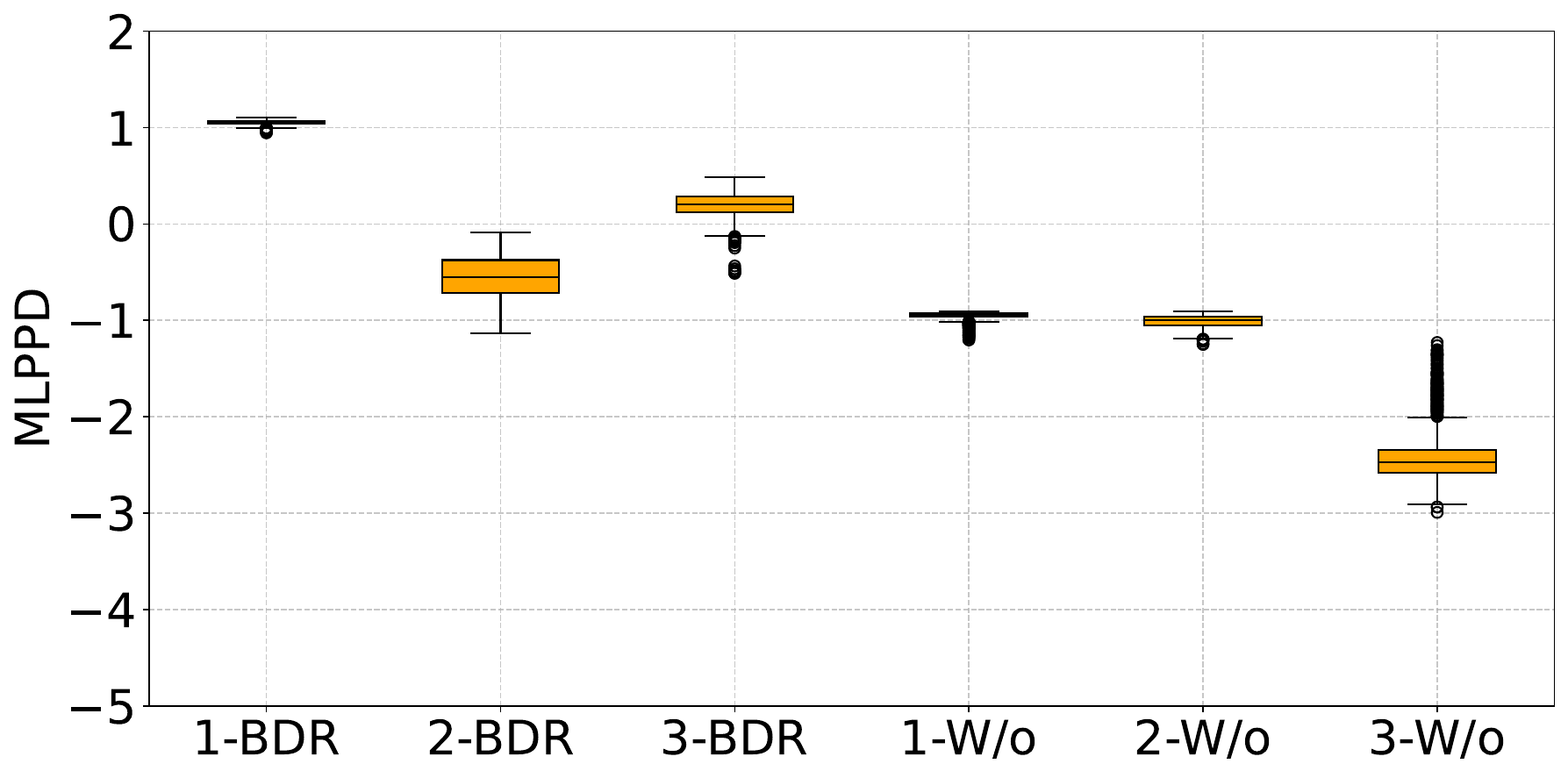}
        \caption{2-layer DGP for ONERA M6 (lift)}
        \label{fig:2ld_lift_MLPPD}
    \end{subfigure}
    \begin{subfigure}{0.49\textwidth}
        \centering
        \includegraphics[width=\linewidth]{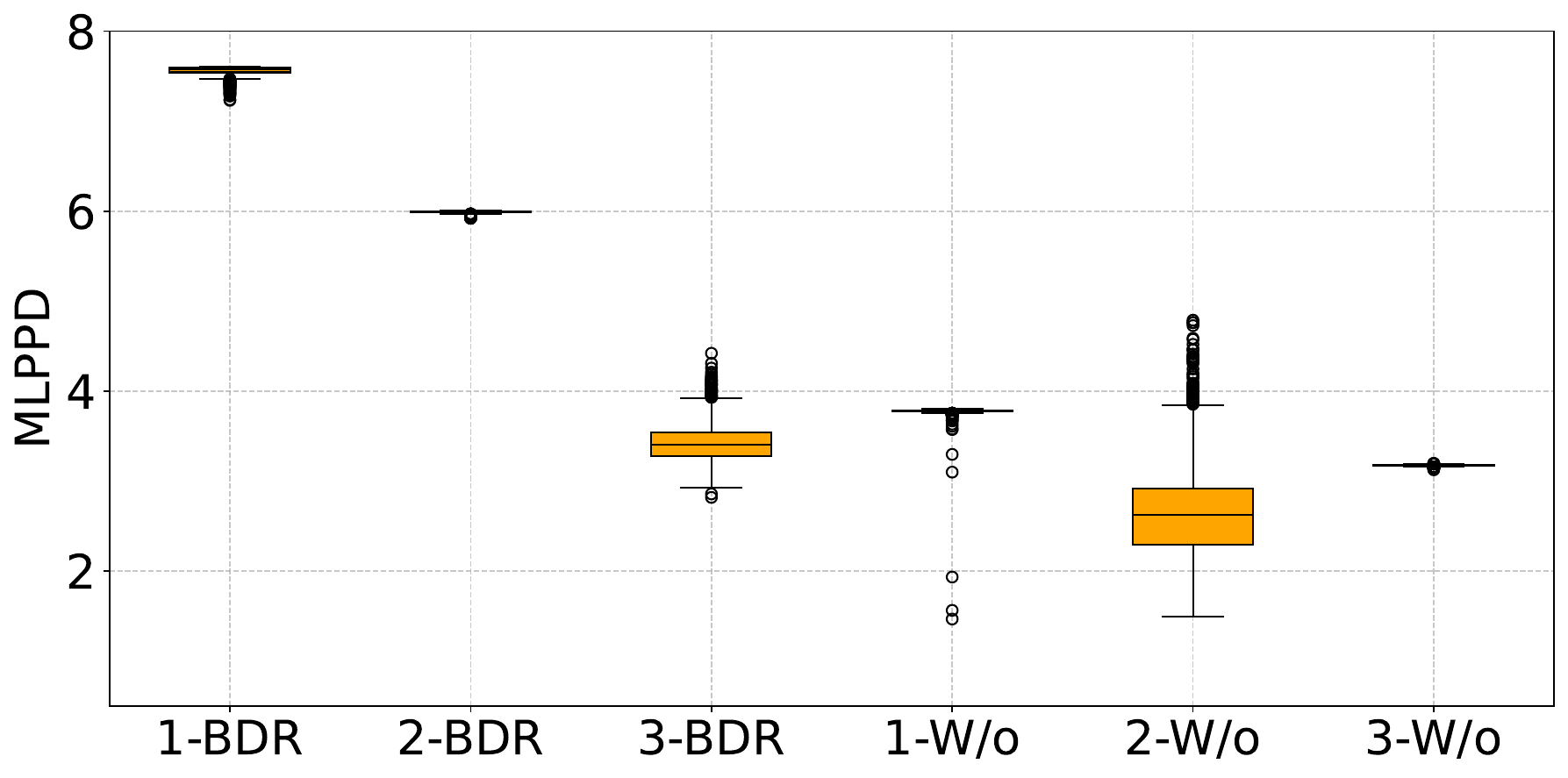}
        \caption{3-layer DGP for ONERA M6 (drag)}
        \label{fig:3ld_drag_MLPPD}
    \end{subfigure}
    \begin{subfigure}{0.49\textwidth}
        \centering
        \includegraphics[width=\linewidth]{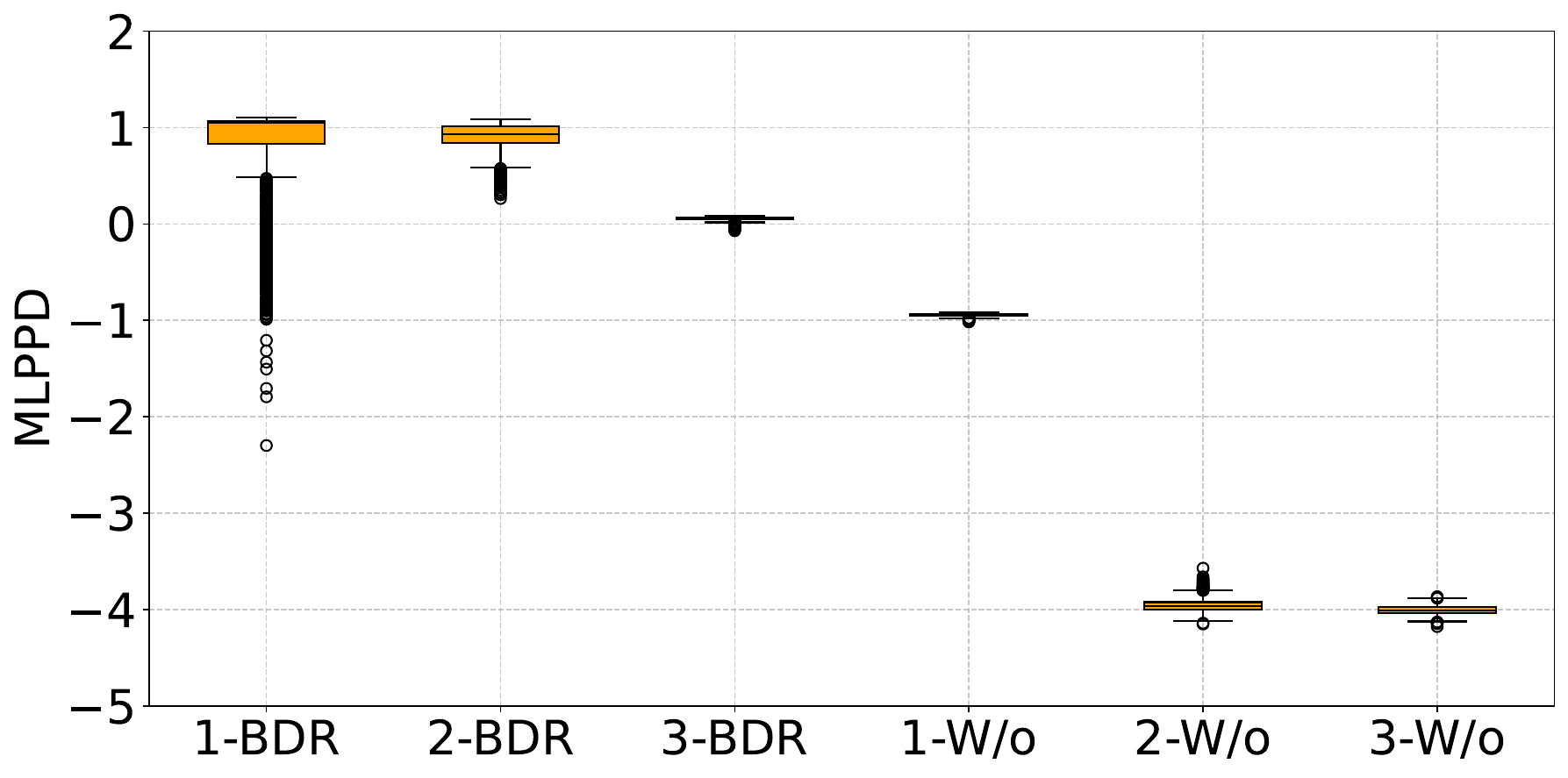}
        \caption{3-layer DGP for ONERA M6 (lift)}
        \label{fig:3ld_lift_MLPPD}
    \end{subfigure}

    \caption{ONERA M6 (drag and lift): MLPPD for BDR and W/o across different input subspaces and DGP layers.}
    \label{fig:ONERA_drag_and_lift_MLPPD}
\end{figure}

The results in \Cref{tab:ONERA_M6}, \Cref{fig:ONERA_drag_and_lift_RMSPE}, and \Cref{fig:ONERA_drag_and_lift_MLPPD}  demonstrate the effectiveness of the proposed BDR-based GP and DGP models on this high-dimensional aerodynamic problem.  For the drag coefficient, the DGP 3-layer(1)–BDR model provides the best overall performance, achieving the lowest RMSPE, highest NSME, and the largest BIC and predictive score. This indicates that deeper surrogates paired with an estimated low-dimensional subspace effectively capture the nonlinear drag behavior. Other BDR variants, such as GP(5)–BDR and DGP 2-layer(1)–BDR, remain competitive. In contrast, models without BDR exhibit substantially larger errors and weaker uncertainty calibration, underscoring the importance of learning the projection matrix rather than relying on the raw 50-dimensional inputs. For lift, GP(1)–BDR achieves the lowest RMSPE and the strongest NSME and BIC, indicating that a single-layer GP with built-in dimension reduction is sufficient for this smoother quantity of interest. While deeper BDR models (e.g., DGP 2-layer(1)–BDR) provide comparable predictive performance, they offer no clear advantage in this case. Without BDR, model performance degrades noticeably, especially for larger subspace sizes and deeper DGPs. \textcolor{black}{The CP and ALCI results in \Cref{tab:ONERA_M6_T2} provide further evidence on posterior uncertainty quantification for both drag and lift. Across both quantities of interest, the BDR-based models attain coverage probabilities close to the nominal 95\% level, whereas the corresponding without BDR models show clear undercoverage, indicating that learning the projection matrix is essential not only for predictive accuracy but also for well-calibrated and reliable uncertainty quantification in this high-dimensional aerodynamic setting.}

Overall, the findings show that Bayesian surrogates with BDR deliver accurate and stable predictions for both lift and drag across a challenging 50-dimensional design space. They achieve a favorable balance between predictive accuracy, uncertainty quantification, and model complexity, while classical DGPs without dimension reduction struggle in this high-dimensional, small-sample regime.

\begin{table}[h]
\centering
\begin{tabular}{|l|c|c|c|c|c|l|c|}
\hline
 \multicolumn{7}{ |c| }{$drag$} \\
 \hline
Method (D) &  TC(mins) & RMSPE & NSME  & CRPS & Score & BIC \\
\hline
 GP (1) BDR   & 90.42 & 7.00$\times 10^{-3}$ & 0.9511  & 4.05$\times 10^{-3}$  & 528.06 & 780.52 \\
 GP (2) BDR   & 124.62 & 7.68$\times 10^{-3}$  & 0.9349 & 4.45$\times 10^{-3}$ & 528.31 & 765.79\\
 GP (3) BDR   & 147.72 & 8.16$\times 10^{-3}$ & 0.9270  & 4.22$\times 10^{-3}$ & 533.00  & 751.93\\
 DGP  2-layer (1) BDR   & 884.43 & 5.89$\times 10^{-3}$ & 0.9692  & 3.97$\times 10^{-3}$  & 530.92 & 792.56 \\
  DGP  2-layer (2) BDR   & 920.25 & 7.38$\times 10^{-3}$ & 0.9481  & 4.27$\times 10^{-3}$ & 548.91  & 770.29\\
 DGP  2-layer (3) BDR   & 2522.07 & 8.19$\times 10^{-3}$ & 0.9206  & 6.41 $\times 10^{-3}$ & 476.01 & 746.02\\
 DGP  3-layer (1) BDR   & 1189.17 & $\mathbf{3.67 \times 10^{-3}}$ & $\mathbf{0.9813}$  & $\mathbf{2.88 \times 10^{-3}}$ & $\mathbf{553.08}$ & $\mathbf{805.37}$\\
 DGP  3-layer (2) BDR   & 2520.43 & 6.63$\times 10^{-3}$ & 0.9578  & 3.82$\times 10^{-3}$ & 525.85 & 774.60\\
 DGP  3-layer (3) BDR   & 3724.59 & 1.77$\times 10^{-2}$ & 0.8171  & 1.16$\times 10^{-2}$ & 528.28 & 650.22 \\
 \hline 
 DGP  2-layer (1) W/o   & 1167.30 & 8.77 $\times 10^{-3}$ & 0.9013  & 3.92$\times 10^{-3}$ & 541.12 & 708.15 \\
 DGP  2-layer (2)  W/o  & 1740.15 & 8.64$\times 10^{-3}$ & 0.9147  & 3.92$\times 10^{-3}$ & 515.45 & 720.39 \\
 DGP  2-layer (3)  W/o  & 2657.70 & 2.48$\times 10^{-2}$ & 0.8004  & 2.47$\times 10^{-2}$ & 516.59  & 571.42\\
 DGP  3-layer (1)  W/o  & 1659.96 & 8.67 $\times 10^{-3}$ & 0.9125  & 3.87$\times 10^{-3}$ & 524.77 & 714.16\\
 DGP  3-layer (2)  W/o  & 1801.71 & 2.22 $\times 10^{-2}$ & 0.8045  & 3.39$\times 10^{-2}$ & 514.88 & 590.64\\
 DGP  3-layer (3) W/o   & 3240.60 & 1.98$\times 10^{-2}$ &  0.8104 & 3.90$\times 10^{-2}$ & 525.45 & 610.21\\
\hline
 \multicolumn{7}{ |c| }{$lift$} \\
 \hline
 GP (1) BDR   & 108.78 & $\mathbf{0.0713}$  & $\mathbf{0.9603}$  & $\mathbf{0.0465}$ & 304.98 & $\mathbf{148.39}$\\
 GP (2) BDR   & 153.30 & 0.0956 &  0.9413 & 0.0496 & 304.43 & 130.90 \\
 GP (3) BDR   & 214.64 & 0.0945 &  0.9421 & 0.0467 & 300.36 & 132.45\\
 DGP  2-layer (1) BDR   & 1012.74 & 0.0819 & 0.9547  & 0.0474 & $\mathbf{313.35}$ & 146.57\\
  DGP  2-layer (2) BDR   & 1269.42& 0.1821 & 0.8519  & 0.1963 & 309.75 & 125.03\\
DGP  2-layer (3) BDR   & 1680.69 & 0.1318 & 0.9167  & 0.0948 & 236.89 & 134.20\\
DGP  3-layer (1) BDR   & 1816.20& 0.0931 & 0.9452  & 0.0477 & 300.67 & 143.89\\
DGP  3-layer (2) BDR   & 2919.18 & 0.0974 & 0.9358  & 0.0585 & 304.78 & 140.27\\
DGP  3-layer (3) BDR   & 5493.87 & 0.1429  & 0.9060 & 0.1477 & 303.93 & 131.67\\
\hline
 DGP  2-layer (1) W/o  & 768.75 & 0.1920 & 0.8371  & 0.3474 & 310.42 & 119.14\\
 DGP  2-layer (2) W/o  & 1413.54 & 0.1914 & 0.8391  & 0.3049 & 305.16 & 108.74\\
 DGP  2-layer (3) W/o  & 2481.30 & 0.3398 & 0.6624  & 0.5125 & 299.32 & 98.01\\ 
 DGP  3-layer (1) W/o  & 2823.53 & 0.1724 & 0.8617  & 0.1474 & 308.59 & 129.60\\
 DGP  3-layer (2) W/o  & 2473.15 & 0.4002 & 0.5618  &  0.9522& 295.21  & 73.05\\
 DGP  3-layer (3)  W/o  & 4665.30 & 0.2860  & 0.7411  & 0.6130 & 298.82 & 115.23\\
 \hline
\end{tabular}
\caption{ONERA M6 (lift and drag):Performance metrics for different BDR and W/o layers with training and testing sizes $n_{train}=250$ and $n_{test}=47$,  respectively. Table entries report, for each metric, the median value across posterior samples.}
\label{tab:ONERA_M6}
\end{table}

\begin{table}[t]
\centering
\begin{tabular}{|l|c|c|c|c|c|c|c|}
\hline
  \multicolumn{3}{ |c| }{$drag$} \\
\hline
Method (D) & CP & ALCI (95\%) \\
\hline
GP (1) BDR & 0.9588 & 0.0906  \\ 
GP (2) BDR & 0.9633 & 0.0544\\
GP (3) BDR & 0.9605 & 0.0161 \\
DGP 2-layer (1) BDR & 0.9513 & 0.0168\\ 
DGP 2-layer (2) BDR & 0.9756 & 0.0153\\
DGP 2-layer (3) BDR & 0.9833 & 0.0300 \\
DGP 3-layer (1) BDR & 0.9667 & 0.0181\\
DGP 3-layer (2) BDR & 0.9400 & 0.0185\\
DGP 3-layer (3) BDR & 0.9333 & 0.0078\\
\hline
DGP 2-layer (1) W/o & 0.7952 & 0.0167\\
DGP 2-layer (2) W/o & 0.8000 & 0.0195\\
DGP 2-layer (3) W/o & 0.8367 & 0.1354\\
DGP 3-layer (1) W/o & 0.7800 & 0.0169\\
DGP 3-layer (2) W/o & 0.8467 & 0.0217\\
DGP 3-layer (3) W/o & 0.7015 & 0.0170\\
\hline
  \multicolumn{3}{ |c| }{$lift$} \\
\hline
GP (1) BDR & 0.9503 & 0.2133\\ 
GP (2) BDR & 0.9452 & 0.3688 \\
GP (3) BDR & 0.9700 & 0.3840 \\
DGP 2-layer (1) BDR & 0.9520 & 0.3442\\ 
DGP 2-layer (2) BDR & 0.9667 & 0.5947\\
DGP 2-layer (3) BDR & 0.9400 & 0.5150\\
DGP 3-layer (1) BDR & 0.9645 & 0.5709\\
DGP 3-layer (2) BDR & 0.9453 & 0.4960\\
DGP 3-layer (3) BDR & 0.9733 & 0.3118\\
\hline
DGP 2-layer (1) W/o & 0.6001 & 0.2142\\
DGP 2-layer (2) W/o & 0.6527 & 0.3721\\
DGP 2-layer (3) W/o & 0.8089 & 0.5460\\
DGP 3-layer (1) W/o & 0.6800 & 0.4818\\
DGP 3-layer (2) W/o & 0.8201 & 0.2353\\
DGP 3-layer (3) W/o & 0.8556 & 0.4667\\
\hline
\end{tabular}
\caption{ONERA M6 (lift and drag):Performance metrics for different BDR and W/o layers with training and testing sizes $n_{train}=250$ and $n_{test}=47$,  respectively. Table entries report, for each metric, the median value across posterior samples.}
\label{tab:ONERA_M6_T2}
\end{table}

\section{Conclusions and Discussion}
\label{sec:Con_Discus}

We developed a Bayesian framework for Gaussian process (GP) surrogate modeling with built-in dimension reduction (BDR). By placing a matrix Langevin prior on the projection matrix and performing posterior inference using HMC, the method jointly learns a low-dimensional input structure and the GP surrogate while propagating uncertainty through all components of the hierarchy. The framework extends naturally to deep Gaussian processes (DGPs), enabling flexible and nonstationary surrogates appropriate for complex computer experiments. Across a suite of synthetic benchmarks and two scientific applications, the Bayesian GP and DGP models equipped with BDR delivered substantially more accurate predictions and better-calibrated uncertainties than approaches lacking dimension reduction.

These results underscore the advantages of integrating dimension reduction directly into the probabilistic surrogate. Treating the projection matrix \(W\) as an inferential quantity allows uncertainty in the latent input subspace to be quantified jointly with predictive uncertainty. This yields more interpretable representations of high-dimensional inputs and improves robustness in small-data regimes where structural uncertainty can be comparable to or greater than predictive noise.

\textcolor{black}{The proposed framework also applies naturally to deterministic functions, including high-dimensional deterministic simulators commonly encountered in computer experiments \cite{Jerome1989,Wenjia2021}. In that setting, the observation model is modified by removing the nugget term associated with observational noise, that is, setting $g=0$ in theory, or replacing it with a very small fixed jitter in practice for numerical stability \cite{Gramacy2012}. The main challenge of high-dimensional inputs remains unchanged, since direct GP modeling still becomes inefficient as the ambient dimension increases \cite{Chen2023,Nathan2022}. Our built-in dimension reduction mechanism continues to address this issue by learning the projection matrix $W$ jointly with the GP or DGP surrogate, thereby identifying the dominant low-dimensional subspace that governs the response. Thus, the methodology remains directly relevant in the deterministic setting. In particular, for deterministic high-dimensional functions, the proposed method continues to identify the dominant low-dimensional subspace governing the response, while providing coherent surrogate-based uncertainty quantification for prediction and emulation.}

Several limitations remain. Full MCMC inference is computationally intensive, particularly for deeper architectures or large training sets, because sampling must be performed both over GP hyperparameters and over the Stiefel-manifold–valued projection matrix. Although HMC provides an efficient sampler for \(W\), the overall cost per iteration can still be high, especially in DGP settings where evaluating latent layers requires repeated GP computations. Variational Bayesian (VB) inference offers a promising complementary alternative: by replacing full posterior sampling with tractable variational approximations, one can obtain substantial reductions in runtime at the expense of some loss in posterior fidelity. Hybrid strategies—such as using VB for latent layers and HMC for \(W\) or kernel parameters—may further accelerate computation while preserving accuracy.

\textcolor{black}{Additionally, this work treated $D$ as fixed throughout the analysis. While this simplifies computation and implementation, it does not account for uncertainty in the choice of the input subspace. A fully Bayesian treatment could be achieved by assigning a discrete prior to $D$ and performing trans-dimensional inference, such as reversible jump MCMC or marginal likelihood-based model averaging. However, these approaches introduce significant computational complexity and are beyond the scope of the present study. Developing efficient and scalable methods for Bayesian inference on $D$ remains an important direction for future research.}

The current formulation focuses on scalar outputs and adopts specific kernel structures and prior choices; exploring alternative kernels, hierarchical priors, and sensitivity analyses would further strengthen the framework. While the method scales well to the moderate dataset sizes typical in computer experiments, additional approximations will be required for very large \(n\).

Future extensions include incorporating \textcolor{black}{randomizing the input subspace ($D$)}, Vecchia-type scalable likelihood approximations for DGPs \cite{SauerVecchia2023}, embedding the framework in sequential design and active learning settings, and developing multivariate-output or multi-fidelity versions. Alternative priors on the Stiefel manifold and more advanced sampling algorithms may also provide computational gains. Together, these directions offer a path toward making Bayesian dimension-reducing GP surrogates broadly applicable to modern large-scale scientific simulation and uncertainty quantification tasks.


\bibliographystyle{siamplain}
\bibliography{references}
\end{document}